\documentclass[10pt,journal,compsoc]{IEEEtran}

\usepackage{
amsfonts,
cite,
pifont,
graphicx,
multirow,
booktabs,
amsmath,
amssymb,
hhline,
url,
hyperref}
\usepackage[ruled]{algorithm2e}
\usepackage[table,xcdraw]{xcolor}
\newcommand\shline{\specialrule{0.8pt}{0pt}{0pt}}
\begin{document}

\title{Practical Compact Deep Compressed Sensing}

\author{Bin Chen and Jian Zhang, \IEEEmembership{Member, IEEE}
\IEEEcompsocitemizethanks{\IEEEcompsocthanksitem Manuscript received January 15, 2023; revised June 13, 2024; accepted November 10, 2024.
\IEEEcompsocthanksitem This work was supported by the Key Program of the National Natural Science Foundation of China (No. 62331011).
\IEEEcompsocthanksitem Bin Chen and Jian Zhang are with the School of Electronic and Computer Engineering, Peking University, Shenzhen 518055, China.
\IEEEcompsocthanksitem Corresponding author: Jian Zhang.
\IEEEcompsocthanksitem
E-mail: chenbin@stu.pku.edu.cn; zhangjian.sz@pku.edu.cn}}

\markboth{Journal of \LaTeX\ Class Files,~2023}%
{Shell \MakeLowercase{\textit{et al.}}: Bare Demo of IEEEtran.cls for Computer Society Journals}

\IEEEtitleabstractindextext{
\begin{abstract}
Recent years have witnessed the success of deep networks in compressed sensing (CS), which allows for a significant reduction in sampling cost and has gained growing attention since its inception. In this paper, we propose a new practical and compact network dubbed PCNet for general image CS. Specifically, in PCNet, a novel collaborative sampling operator is designed, which consists of a deep conditional filtering step and a dual-branch fast sampling step. The former learns an implicit representation of a linear transformation matrix into a few convolutions and first performs adaptive local filtering on the input image, while the latter then uses a discrete cosine transform and a scrambled block-diagonal Gaussian matrix to generate under-sampled measurements. Our PCNet is equipped with an enhanced proximal gradient descent algorithm-unrolled network for reconstruction. It offers flexibility, interpretability, and strong recovery performance for arbitrary sampling rates once trained. Additionally, we provide a deployment-oriented extraction scheme for single-pixel CS imaging systems, which allows for the convenient conversion of any linear sampling operator to its matrix form to be loaded onto hardware like digital micro-mirror devices. Extensive experiments on natural image CS, quantized CS, and self-supervised CS demonstrate the superior reconstruction accuracy and generalization ability of PCNet compared to existing state-of-the-art methods, particularly for high-resolution images. Code is available at \url{https://github.com/Guaishou74851/PCNet}.
\end{abstract}

}

\maketitle
\IEEEdisplaynontitleabstractindextext
\IEEEpeerreviewmaketitle

\vspace{-10pt}
\IEEEraisesectionheading{\section{Introduction}\label{sec:introduction}}

\IEEEPARstart{A}{s} a pioneering framework in signal processing, compressed sensing (CS) \cite{donoho2006compressed,candes2006compressive} establishes a foundation to challenge the Nyquist-Shannon sampling theorem \cite{shannon1949communication} and significantly reduce signal acquisition costs. It enables the accurate reconstruction of high-dimensional images from a small number of linear measurements and has led to a wide range of promising applications. These include single-pixel imaging \cite{duarte2008single}, magnetic resonance imaging (MRI) \cite{lustig2008compressed, sun2016deep}, computational tomography (CT) \cite{szczykutowicz2010dual}, and snapshot compressive imaging (SCI) \cite{cheng2022recurrent,liu2024deep}. Additionally, CS has been applied to the source coding \cite{fowler2012block,liu2016compressive} and encryption \cite{zhang2020robust} of images and videos. In this work, we focus on general image CS problems, aiming to keep our contributions practical and extensible for the deployment of real-world CS systems.

\begin{figure*}
\centering
\vspace{-5pt}
\includegraphics[width=0.95\textwidth]{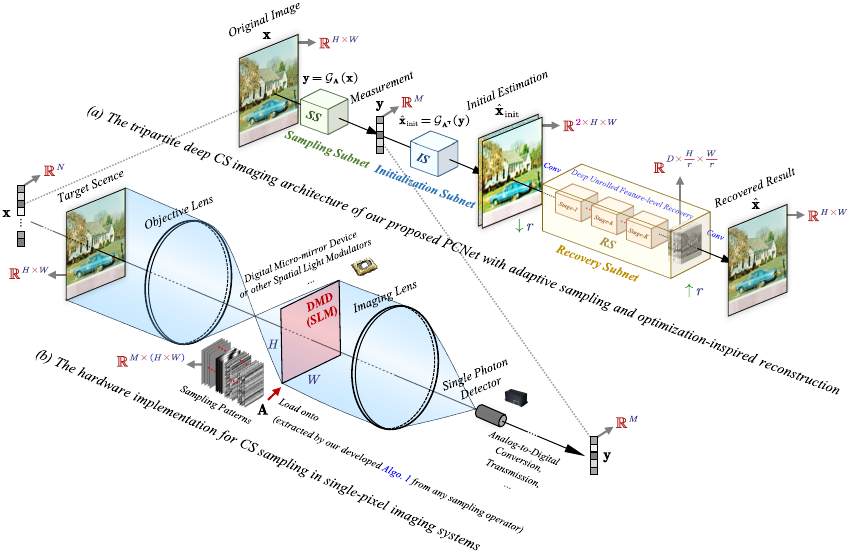}
\vspace{-13pt}
\caption{Illustrations of the proposed PCNet architecture and a typical single-pixel hardware implementation \cite{duarte2008single} for CS acquisition. \textcolor{blue}{\textbf{(a)}} Following \cite{zhang2020optimization,chen2022content}, our PCNet consists of a sampling subnet (SS), an initialization subnet (IS), and a recovery subnet (RS). The SS simulates the sampling process, the IS performs the transformation from measurement to image, while the RS, which is algorithm-unrolled and transmission-augmented, further refines the initialization from the IS to provide a final reconstruction. \textcolor{blue}{\textbf{(b)}} In practice, linear sampling matrices (operators) are converted into a global matrix form as shown in Fig.~\ref{fig:sampling_operator_comparison} (a), of size $M\times N$. This matrix is then split and re-organized into $M$ separate modulating patterns (each corresponding to a matrix row of size $H\times W$) to be individually loaded onto specific spatial light modulators (SLMs) like the digital micro-mirror devices (DMDs), with a spatial resolution of $N=H\times W$, for sequential sampling.}
\label{fig:arch}
\vspace{-10pt}
\end{figure*}

\subsection{Basic Concepts and Task Statement}
Mathematically, CS imaging is to reconstruct a latent clean image $\mathbf{x} \in \mathbb{R}^N$ from its observed low-dimensional measurement $\mathbf{y} \in \mathbb{R}^M$ under a noisy acquisition process described by $\mathbf{y} = \mathbf{Ax} + \mathbf{n}$. Here, $\mathbf{x}$ and $\mathbf{y}$ respectively represent the vector forms of the image $\mathbf{X} \in \mathbb{R}^{H \times W}$ and its measurement $\mathbf{Y}$. The linear projection is achieved via a sampling matrix (or operator) $\mathbf{A} \in \mathbb{R}^{M \times N}$ (or $\mathcal{G}_\mathbf{A}: \mathbb{R}^{N} \rightarrow \mathbb{R}^{M}$). The noise $\mathbf{n}$ is often assumed to be additive white Gaussian noise (AWGN) \cite{zhang2022plug}, characterized by a standard deviation (or noise level) $\sigma$. The CS ratio (or sampling rate) is defined as $\gamma = M/N$. Fig.~\ref{fig:arch} (b) illustrates a CS sampling implementation in single-pixel imaging systems \cite{duarte2008single}. Following \cite{chen2021equivariant,chen2022robust}, the set of all information necessary for CS reconstruction—including the sampling matrix (operator) $\mathbf{A}$ ($\mathcal{G}_\mathbf{A}$), measurement $\mathbf{y}$, CS ratio $\gamma$, and the linear acquisition model itself—is termed \textit{sampling physics knowledge}, or simply \textit{``physics"} in this paper.

In practice, extremely low $\gamma$ values, where $M \ll N$, offer benefits such as sampling acceleration \cite{sun2016deep}, energy savings \cite{duarte2008single}, and reduced X-ray radiation \cite{szczykutowicz2010dual}. However, these also introduce challenges, since inferring $\mathbf{x}$ from only $\mathbf{y}$ and $\mathbf{A}$ is ill-posed without additional information.

\vspace{-5pt}
\subsection{Progress of Image CS Methods: From Hand-Crafted Optimization Framework to Deep Algorithm Unrolling}

In CS research, the two fundamental issues are the design of sampling operators and the development of reconstruction algorithms \cite{gan2007block,shi2019image}. Since the advent of CS theory, sampling matrices such as Gaussian \cite{candes2006compressive}, binary/bipolar/ternary \cite{amini2011deterministic}, structurally random \cite{do2008fast,gan2008fast}, and Toeplitz \cite{haupt2010toeplitz} matrices have been extensively studied. Early-stage model-driven reconstruction methods have exploited image priors like sparsity \cite{zhang2014group} and low-rankness \cite{dong2014compressive} to build regularizers for optimization. Robust data-driven methods have also been developed, including dictionary learning \cite{aharon2006k}, tight-frame learning \cite{cai2014data}, and convolutional operator learning \cite{chun2019convolutional}.

Over the past decade, the development of neural networks (NNs) has significantly improved image reconstruction in terms of both quality and speed. Leveraging the representational capabilities of deep NNs, numerous network-driven approaches directly learn an inverse mapping from the measurement domain to the original image signal domain \cite{zhang2018ista}. Representative end-to-end CS reconstruction networks include ReconNet \cite{kulkarni2016reconnet}, CSNet \cite{shi2019image}, SCSNet \cite{shi2019scalable}, DPA-Net \cite{sun2020dual}, RK-CCSNet \cite{zheng2020sequential}, and MR-CCSNet \cite{fan2022global}. These focus on accuracy and scalability\footnote{In the context of image CS, following \cite{shi2019scalable,you2021coast,chen2022content}, we define a CS network as ``scalable'' when it is able to use a single set of learned network parameters to handle the CS tasks of multiple sampling rates.}, and have popularized many effective network designs, such as hierarchical and non-local structures \cite{shi2019scalable,zheng2020sequential,fan2022global,chen2023deep}. Numerous optimization-inspired methods organically integrate traditional optimization algorithms with advanced NN designs \cite{zhang2023physics,song2023dynamic,chen2023self,song2023deep,li2024d}. Examples include ADMM-Net \cite{sun2016deep}, ISTA-Net \cite{zhang2018ista}, OPINE-Net \cite{zhang2020optimization}, AMP-Net \cite{zhang2021amp}, MAC-Net \cite{chen2020learning}, COAST \cite{you2021coast}, ISTA-Net$^{++}$ \cite{you2021ista}, MADUN \cite{song2021memory}, FSOINet \cite{chen2022fsoinet}, and CASNet \cite{chen2022content}. This paradigm, which unrolls the truncated inference of an optimization algorithm into NN stages, has attracted attention and become a distinctive stream in image CS.

\subsection{Challenges in Existing Deep CS Networks: Obtaining Practical and Efficient Sampling Operator}
\label{subsec:challenges_sampling}

The success of deep NNs in image restoration \cite{dong2015image,ulyanov2018deep,pang2020self,chen2021equivariant,chen2022robust,quan2022dual,cao2023ntire,li2024resvr} has led researchers to explore learning the sampling matrix alongside recovery NN components in an adaptive, end-to-end manner from image datasets. Due to computational and memory/storage limitations, most CS networks adopt a block-based (or block-diagonal) sampling scheme \cite{gan2007block} and collaboratively learn the linear sampling matrix and recovery NN. These networks divide the high-dimensional image of shape $N=H\times W$ into non-overlapping blocks of shape $n=B\times B$ and then obtain measurements block-by-block using a small-sized sampling matrix $\mathbf{A}_B\in \mathbb{R}^{(\gamma n)\times n}$, as Fig.~\ref{fig:sampling_operator_comparison} (b) illustrates. Some recent works \cite{zheng2020sequential,fan2022global} employ a linear convolutional NN (CNN) to reduce the size of image, simulating the under-sampling process. Despite these advances, there are at least three problems in existing sampling operators that limit their effectiveness in practice:

\begin{figure}
\centering
\vspace{-5pt}
\includegraphics[width=0.44\textwidth]{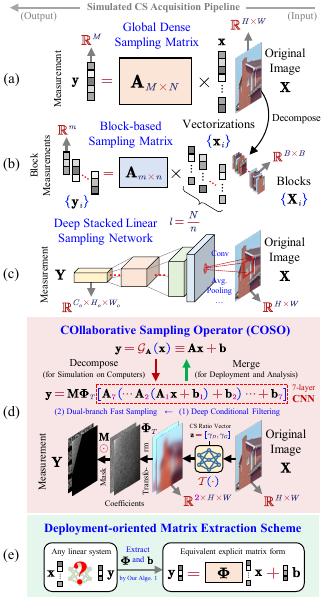}
\vspace{-12pt}
\caption{Illustrations of different CS sampling operators. \textcolor{blue}{\textbf{(a)}} Global dense matrix linearly aggregates all the $N$ pixel values of image $M$ times. \textcolor{blue}{\textbf{(b)}} Block-diagonal matrix divides image into non-overlapping blocks and performs sampling block-by-block. \textcolor{blue}{\textbf{(c)}} Deep sampling NN \cite{zheng2020sequential,fan2022global} uses stacked convolutions to gradually reduce the data volume and obtain compressed measurements. \textcolor{blue}{\textbf{(d)}} Our collaborative sampling operator (COSO) consists of a deep filtering step and a fast global sampling step. The former (denoted by $\mathcal{T}$) contains seven convolution layers and is modulated by the CS ratio. It learns an implicit compact representation of matrix into lightweight $\mathcal{T}$ from data. The latter is achieved through a dual-branch transformation and masking. More details are illustrated in Fig.~\ref{fig:sampling_operator} (a)-(c). Note that in CS imaging systems like the one shown in Fig.~\ref{fig:arch} (b), any sampling operator must be converted to its global matrix form as in (a) for hardware deployment. \textcolor{blue}{\textbf{(e)}} We further provide a practical scheme (see Algo.~\ref{alg:matrix_extraction}) to extract the explicit matrix from any given (black-box) linear system for sampling matrix deployment and analysis.}
\label{fig:sampling_operator_comparison}
\vspace{-10pt}
\end{figure}

\textbf{(1) Failing to efficiently capture the image information into a few measurements.} To reduce time and memory complexity in obtaining measurements, many existing methods employ block-diagonal matrices and linear NN-based operators that aggregate only the signal intensities (or pixel values) of a small local image area. However, we demonstrate that these sampling schemes have substantial room for improvement in capturing the image information due to their limited receptive fields. This issue becomes more significant as the image size increases.

\textbf{(2) Being uninterpretable and mysterious for research.} Existing data-driven CS sampling operators, which learn dimension reduction directly without incorporating specific knowledge about natural signals, are often trained as a \textit{black box}, making it unclear what they have learned and how the structure contributes to their sampling ability. For example, whether the receptive field of a sampling operator is effective for improving the quality of reconstructed images? We lack studies and research tools for this open question.

\textbf{(3) Becoming inflexible and disconnected from practical deployments and hardware advancements.} Most deep CS methods are designed to learn fixed sampling matrices tailored for specific tasks, which often lack flexibility for different sampling rates and image sizes. Previous works involving NN-based sampling operators \cite{zheng2020sequential,fan2022global} do not address their deployment on hardware platforms such as digital micro-mirror devices (DMDs) used as spatial light modulators (SLMs) \cite{duarte2008single}, as shown in Fig.~\ref{fig:arch} (b). Additionally, the commonly adopted image block sizes $B\in\{8,16,32,33\}$ \cite{shi2019image,zhang2020optimization} are far smaller than the sizes of real-world sampling devices and can hardly satisfy the increasing demands for high-resolution imaging required in real applications. In 2016, \cite{kulkarni2016reconnet} used a DMD with 1920$\times$1080 micro-mirror elements to capture real measurements, whereas from 2020 to 2022, Texas Instruments developed 4K ultra-high-definition (UHD) DMDs with a resolution of 3840$\times$2160 \cite{ti}. We will show that existing deep-learned sampling operators are not sufficiently effective at sampling large images of such sizes. Recent content-aware methods \cite{chen2022content,zhang2023progressive} necessitate auxiliary devices or a two-step interactive sampling process to implement dynamic and non-uniform block-diagonal sampling, inevitably complicating hardware implementation.

\vspace{-5pt}
\subsection{Challenges in Existing Deep CS Networks: Building Comprehensive and Robust Reconstruction Backbone}

Recently, the design of recovery networks has become the primary focus of deep NN-based CS research. For example, at the level of \textit{optimization framework}, \cite{cui2022fast} introduce momentum mechanisms to accelerate convergence and reduce the number of unrolled stages. At the level of \textit{network structure}, there are advancements in hierarchical architectures \cite{shi2019scalable} for enhancing multi-scale context perception, and innovations in controllable, flexible structures \cite{shi2019scalable,you2021coast,you2021ista,chen2022content}, dual-path attention modules \cite{sun2020dual}, and memory-augmented mechanisms \cite{chen2020learning,song2021memory}. Despite this significant progress, two major challenges hinder further improvement in CS NNs:

\textbf{(1) The marginal performance benefit from increased capacity diminishes rapidly.} Existing CS NNs require large capacities to learn abundant image priors for high-quality reconstruction, which leads to an increased burden in parameters and computation. Due to an imbalanced development of sampling operators and reconstruction networks, their performance tends to plateau and becomes difficult to further enhance as the capacity of NN grows.

\textbf{(2) The dated component designs and training configurations limit the potential for improvement.} The choice of structure, datasets, and training strategies is critical for training an effective recovery NN \cite{liu2022convnet,zhang2023practical}. Despite many developments in image processing, such as advancements in model backbones and parameter optimization technology \cite{deng2020deep,liang2021swinir}, most CS networks \cite{you2021coast,you2021ista,song2021memory} rely on basic, less effective modules (e.g., the classic residual block), use small datasets (e.g., T91 \cite{dong2015image} and Train400 \cite{zhang2017beyond}), and employ dated training strategies. There is a lack of a flexible CS network adaptable to different sampling ratios and image sizes.

\vspace{-5pt}
\subsection{Contributions of This Work}

This work aims to resolve the above challenges by developing an effective and efficient deep NN-based CS sampling and reconstruction pipeline. Our contributions are four-fold:

\vspace{3pt}
\noindent \ding{113}~\textbf{(1) We propose a novel collaborative sampling operator (COSO) for linear CS acquisition.} This sampling operator disentangles the data-driven learning of image priors from the dimension reduction process, exhibiting more effective information preservation and robustness compared to existing methods. Specifically, we employ a CNN for fast filtering on the input image, followed by a discrete cosine transform (DCT) \cite{ahmed1974discrete} and a scrambled block-diagonal Gaussian matrix \cite{do2008fast} to capture image structure and details, respectively. The proposed sampling operator is efficient, has low memory and parameter overhead, and can be applied to existing CS NNs to consistently achieve performance improvements.

\vspace{3pt}
\noindent \ding{113}~\textbf{(2) We provide a collection of sampling operator design, deployment-oriented matrix extraction, and analysis methods.} Based on pioneering works \cite{gan2007block,gan2008fast} and the current context of deep learning, we present several guiding principles for the design of CS sampling operators. We empirically demonstrate the shortcomings of classic block-diagonal matrix, as well as the superiority of our introduced scrambled block-diagonal matrix from three perspectives of restricted isometry property (RIP) \cite{candes2005decoding}, frequency-domain representation, and effective receptive field (ERF) \cite{luo2016understanding}. We also provide a structural reparameterization-inspired \cite{ding2022scaling} extraction scheme to obtain the matrix form of any linear sampling operator for deployment and analysis. We hope that it can become an auxiliary tool for CS research.

\vspace{3pt}
\noindent \ding{113}~\textbf{(3) We construct a new deep unrolled recovery network, equipped with ten enhancement strategies.} Inspired by the recent advances in general computer vision \cite{liu2022convnet} and image restoration \cite{liang2021swinir}, we construct a proximal gradient descent (PGD) \cite{parikh2014proximal} algorithm-unrolled recovery NN. Concretely, we develop a flexible CS network dubbed PCNet (see Fig.~\ref{fig:arch} (a)) that benefits from efficient window-based Transformer blocks, a large-scale dataset, and improved training strategies. It can serve as a new baseline for future studies.

\vspace{3pt}
\noindent \ding{113}~\textbf{(4) We conduct experiments on three typical tasks: natural image CS, quantized CS (QCS), and self-supervised CS.} Extensive results demonstrate the superiority and generalization ability of our method and the impact of different designs in PCNets. Corresponding analyses are presented to provide an understanding of their working principles.

\vspace{-5pt}
\subsection{Organization of This Paper}

The remainder of this paper is organized as follows. Sec.~\ref{sec:designing_a_deep_sampling_operator}, describes the design of our proposed collaborative sampling operator (COSO), which is motivated by existing sampling methods. This section also provides an algorithm for converting any linear systems to their explicit matrix forms. Sec.~\ref{sec:modernizing_a_reconstruction_backbone} explores a flexible recovery NN with ten enhancement strategies, constructing our PCNets. Extensive experimental results are reported in Sec.~\ref{sec:experiments}. Sec.~\ref{sec:conclusion} summarizes this paper.

\begin{figure}
\centering
\vspace{-8pt}
\includegraphics[width=0.46\textwidth]{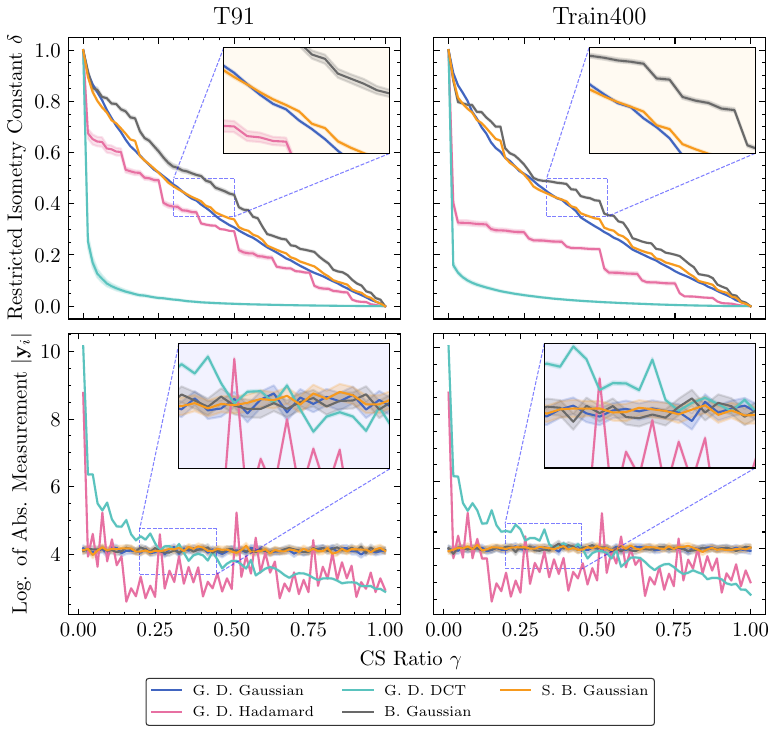}
\vspace{-12pt}
\caption{Our pilot experiments evaluating five sampling matrices on T91 \cite{dong2015image} (left) and Train400 \cite{zhang2017beyond} (right). These matrices include global dense (G. D.) Gaussian, DCT, Hadamard, block-diagonal (B.) Gaussian, and scrambled block-diagonal (S. B.) Gaussian matrices. All images are center-cropped to 256$\times$256, and the block size $B$ is set to 32. \textcolor{blue}{\textbf{(1) Top:}} For an image set $\mathcal{S}=\left\{\mathbf{x}_i\right\}$, the restricted isometry constant of sampling matrix $\mathbf{A}$ is estimated by $\delta \approx \max_i \left\{abs({\lVert \mathbf{Ax}_i \rVert_2^2}/{\lVert \mathbf{x}_i \rVert_2^2}-1)\right\}$. Small $\delta$ values are preferred as they indicate that $\mathbf{A}$ nearly preserves orthogonality within the domain of $\mathcal{S}$. \textcolor{blue}{\textbf{(2) Bottom:}} The average absolute measurement value of $\lvert \mathbf{y}_i \rvert=\lvert \mathbf{A}_i\mathbf{x}\rvert$ in logarithmic scale is calculated by $\frac{1}{j}\Sigma_j\log \lvert \mathbf{A}_i\mathbf{x}_j \rvert$ at the point of $i=\gamma N$ for all possible CS sampling rates $\gamma\in[0,1]$. A large $\lvert \mathbf{y}_i \rvert$ value suggests a high correlation between the $i$-th matrix row $\mathbf{A}_i$ and the images in $\mathcal{S}$.}
\label{fig:ric_amv}
\vspace{-10pt}
\end{figure}

\vspace{-5pt}
\section{Designing a Deep Sampling Operator}
\label{sec:designing_a_deep_sampling_operator}

\subsection{Motivation}
\label{sec:motivation}
\subsubsection{A Wish-List for Designing \texorpdfstring{~$\mathcal{G}_\mathbf{A}$}{A~} in Deep Learning Era}

To address the challenges outlined in Sec.~\ref{subsec:challenges_sampling}, and following \cite{gan2007block,gan2008fast}, we compile a wish list of five ideal characteristics for $\mathcal{G}_\mathbf{A}$ that are instructive for its design in modern CS NNs:

\begin{itemize}
  \item \textbf{Optimal Sampling Performance:} Capturing as much important information as possible for accurately reconstructing both the image structure and details.
  \item \textbf{Universality and Robustness:} Keeping insensitive to noise and not limited to several specific scenarios.
  \item \textbf{Low Memory Expense:} Compact, requiring a small set of parameters to represent a latent (large) matrix, thus reducing storage and transmission burdens.
  \item \textbf{Fast Computation:} Enjoying real-time sensing speed for large-scale training and inferences of CS NNs.
  \item \textbf{Friendliness for Soft-/Hard-ware and Optical Implementation:} Supporting implementation and deployment in commonly adopted modern deep learning frameworks, GPUs, and SLMs such as DMDs.
\end{itemize}

\subsubsection{Rethinking Existing CS Sampling Operators}

\begin{figure*}
\centering
\vspace{-8pt}
\includegraphics[width=0.96\textwidth]{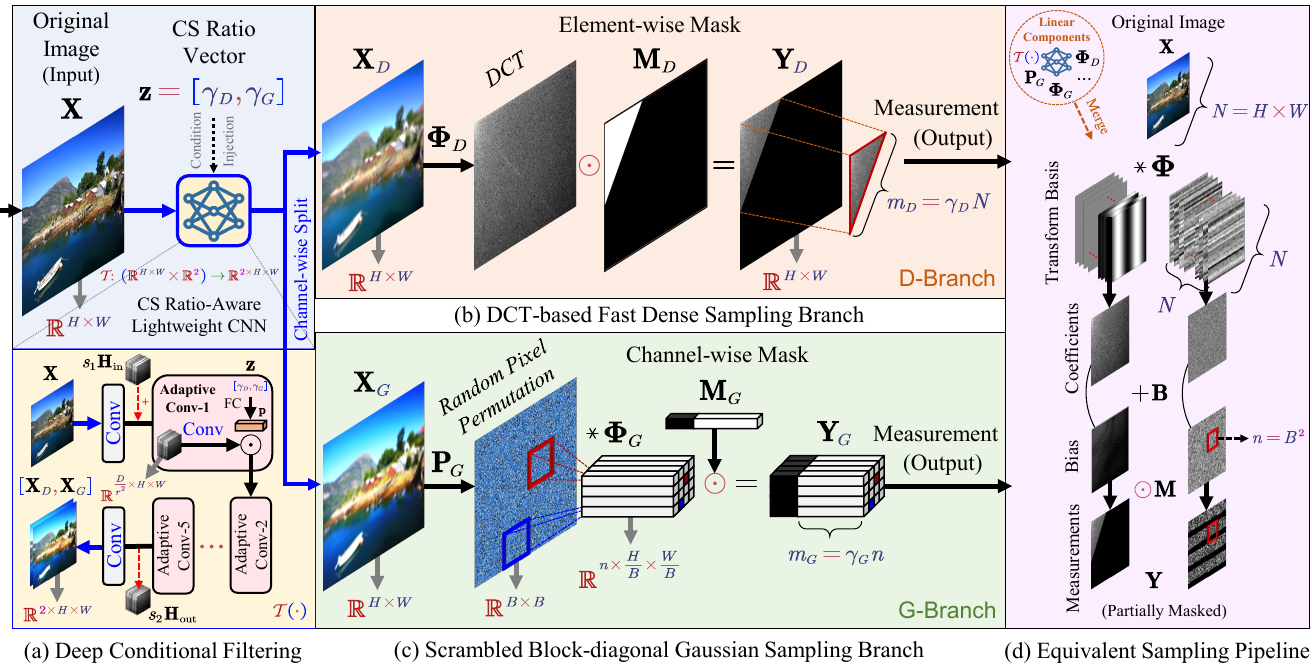}
\vspace{-10pt}
\caption{Illustration of our proposed collaborative sampling operator (COSO) $\mathcal{G}_\mathbf{A}$ for sampling an image with size $N=H\times W$ and rate $\gamma$. \textcolor{blue}{\textbf{(a)}} The input $\mathbf{X}$ is first adaptively filtered by a CS ratio-aware lightweight CNN $\mathcal{T}$, which consists of seven convolution layers and is conditioned by the rate vector $\mathbf{z}=\left[\gamma_D,\gamma_G\right]$ satisfying $\gamma=\gamma_D+\gamma_G$, to generate two smoothed results $\mathbf{X}_D$ and $\mathbf{X}_G$. When the whole sampling operator is utilized in the reconstruction network, we introduce two supplementary paths into $\mathcal{T}$ (indicated by the red dotted lines with arrows) to enable a high-throughput feature-level information flow. \textcolor{blue}{\textbf{(b)}} The DCT-based fast dense sampling branch (D-branch) transforms $\mathbf{X}_D$ into the DCT domain and discards the last $(1-\gamma_D)N$ coefficients in Zig-Zag order to obtain the measurement $\mathbf{Y}_D$. \textcolor{blue}{\textbf{(c)}} The scrambled block-diagonal Gaussian sampling branch (G-branch) randomly permutes all the pixels of $\mathbf{X}_G$ and performs a channel-wise masked convolution to simulate block-diagonal sampling and produce measurement $\mathbf{Y}_G$ at the ratio of $\gamma_G$. \textcolor{blue}{\textbf{(d)}} For practical deployment and analysis, all components of our operator, composed of (a)-(c), can be merged into a simple equivalent linear pipeline, $\mathbf{y=M(\Phi x+b^\prime)=(M\Phi) x+(Mb^\prime)}$, by applying the matrix distributive and associative laws.}
\label{fig:sampling_operator}
\vspace{-10pt}
\end{figure*}

We begin by examining three existing sampling operators: global dense matrix, block-diagonal matrix, and linear sampling NN \cite{zheng2020sequential,fan2022global}. As illustrated in Fig.~\ref{fig:sampling_operator_comparison} (a)-(c), the first conducts $M$ linear aggregations across all $N$ image pixels, the second independently projects each decomposed $B\times B$ image block into $m$ measurements, and the last alternates between convolutions and average-poolings to gradually reduce data volume and obtain the compressed measurement. We initially select three classic $N\times N$ complete bases: DCT \cite{ahmed1974discrete}, Hadamard \cite{gan2008fast}, and orthonormalized i.i.d. Gaussian matrix, simulating the CS sampling process by discarding the last $(N-M)$ (or $n-m$) coefficients of each full image (or block) within the transform domain. In particular, the DCT coefficients are sorted in Zig-Zag order \cite{marcellin2000overview} (see Fig.~\ref{fig:sampling_operator} (b)) by default to retain only the most significant ones.

\begin{figure}
\centering
\hspace{-2pt}
\vspace{-10pt}
\includegraphics[width=0.46\textwidth]{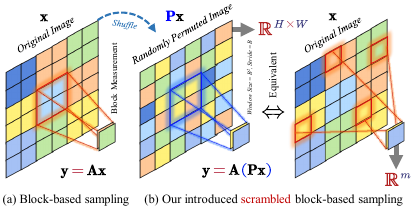}
\vspace{-2pt}
\caption{Illustration of two block-based CS sampling schemes for an image of size $N=H\times W$, block of size $n=B^2$, and rate $\gamma=m/n$. For simplicity, we only present the simulated sampling process for one block, setting $H=W=6$ and $B=m=2$ in our toy example. \textcolor{blue}{\textbf{(a)}} The block-diagonal matrix \cite{gan2007block} divides the image into non-overlapping $B\times B$ blocks and obtains their measurements one-by-one. This operation can be efficiently simulated by a convolution layer (left). \textcolor{blue}{\textbf{(b)}} The introduced scrambled block-diagonal matrix \cite{do2008fast} randomly shuffles all pixels before its block-by-block sampling process (middle). This approach is equivalent to non-overlapping sampling performed on the whole image (right), thus benefiting from a global receptive field and low complexity.}
\label{fig:scrambled_matrix}
\vspace{-10pt}
\end{figure}

To evaluate the characteristics of different matrices, we employ the RIP condition \cite{candes2005decoding} to measure their sampling effectiveness. Our pilot experiments in the top two subgraphs of Fig.~\ref{fig:ric_amv} show that in the domain of two natural image sets, T91 \cite{dong2015image} and Train400 \cite{zhang2017beyond}, the global DCT matrix outperforms others under the indicator of restricted isometry constant $\delta$. Moreover, the global Gaussian matrix is preferred over the block-diagonal one in nearly all cases for $\gamma\in[0,1]$. To bridge the gap between these two Gaussian matrices, we follow \cite{do2008fast} and construct a globally scrambled matrix that shuffles all pixels before block-based sampling to uniformly preserve global information into block measurements (see Fig.~\ref{fig:scrambled_matrix} (b)). We observe that by expanding the receptive field to the entire image, the scrambled block-diagonal sampling matrix (see the orange curves) consistently achieves $\delta$ values close to those of global dense matrix. These results suggest that the operation of global pixel permutation may enhance the sampling effectiveness of block-diagonal matrix.

\begin{table}
\caption{Time complexity of different CS matrices (operators) in sampling an $N$-dimensional image signal. The DCT-/Hadamard-based matrices benefit from fast algorithms \cite{makhoul1980fast,fino1976unified} in practice. Given that the CS ratio $\gamma$ ranges from $[0,1]$, the number of measurements $M$ is approximately $\mathcal{O}(N)$. When the block size $n$ and network depth (number of layers) $L$ are pre-defined constant numbers, both the block-diagonal and network-based operators achieve a linear sampling time of $\mathcal{O}(N)$.}
\label{tab:basis_time_complexity}
\vspace{-8pt}
\centering
\hspace{-2pt}\resizebox{0.49\textwidth}{!}{
\begin{tabular}{ccc}
\shline
\rowcolor[HTML]{EFEFEF} 
Category & Sampling Scheme & Time Complexity\\ \hline \hline
DCT & Global Dense   & $\mathcal{O}(N\log N)$ \\
/Hadamard & Block-Diagonal & $\mathcal{O}((N/n)\times (n\log n))=\mathcal{O}(N\log n)$ \\ \hline
& Global Dense & $\mathcal{O}(N^2)$ \\
\multirow{-2}{*}{Gaussian} & Block-Diagonal & $\mathcal{O}((N/n)\times n^2)=\mathcal{O}(Nn)$ \\ \hline
Network & - & $\mathcal{O}(NL)$ \\ \shline
\end{tabular}}
\vspace{-12pt}
\end{table}

In the bottom two subgraphs of Fig.~\ref{fig:ric_amv}, we evaluate the measurement power \cite{proakis2001digital} of different sampling matrix rows $\mathbf{A}_i$. We observe that the former rows of DCT and Hadamard matrices exhibit higher responses to images than the latter rows, while the measurement magnitudes of Gaussian rows remain stable with only minor fluctuations. We attribute this phenomenon to the fact that the former DCT/Hadamard rows are low-frequency components dominant in natural images, whereas the fixed random Gaussian matrices perform isotropic projection without an ordered structure.

Based on the practical implementation in Fig.~\ref{fig:arch} (b) and above empirical results, we introduce three auxiliary principles or clarifications below to guide our method design:

First, in real applications, linear sampling operators must be converted to global matrix forms (see Fig.~\ref{fig:sampling_operator_comparison} (a)), which can be deployed on hardware like DMDs. Therefore, a matrix extraction method is necessary for sampling operators.

Second, we focus on the complexity of sampling operators in simulation, not the real physical process, because their equivalent $M\times N$ global floating-point matrices do not bring significant differences in time or memory cost on the same optical platform. Tab.~\ref{tab:basis_time_complexity} provides a comparison of the time complexity for simulation among different operators. Global dense matrices like Gaussian, which lack fast algorithms, are too costly compared to others since their simulation essentially involves large-scale matrix multiplication with complexity quadratic to the image size. Consequently, such matrices are not widely used in most recent CS NNs.

Third, inspired by the comparisons in Fig.~\ref{fig:ric_amv} and Tab.~\ref{tab:basis_time_complexity}, we propose to combine the advantages of DCT, Gaussian matrix, and network to develop an effective and efficient deep linear sampling operator. Notably, the global DCT basis performs well in capturing low-frequency image components, the scrambled block-diagonal Gaussian matrix maintains stability across mid-/high-frequency bands, while the deep network can learn adaptive parameters from data. All these linear operators are compatible with arbitrary image sizes and enjoy efficient implementations.

\vspace{-5pt}
\subsection{Proposed Two-Step Sampling Architecture}
\label{subsec:sampling_operator_architecture}

Instead of directly learning all the sampling matrix elements \cite{shi2019image,zhang2020optimization} or a stacked linear compressing network \cite{zheng2020sequential,fan2022global} for one-step sampling, as Figs.~\ref{fig:sampling_operator_comparison} (d) and \ref{fig:sampling_operator} (a)-(c) exhibit, our proposed collaborative sampling operator (COSO) accomplishes the acquisition process in two sequential steps: a \textit{deep conditional filtering} step and a \textit{dual-branch fast sampling} step, in which the former first generates two linearly smoothed versions $\mathbf{X}_D$ and $\mathbf{X}_G$ of the input image $\mathbf{X}$, while the latter then employs a global DCT basis and a scrambled block-diagonal Gaussian matrix to under-sample these two results at rates of $\gamma_D$ and $\gamma_G$ satisfying $\gamma=\gamma_D+\gamma_G$, respectively. In the following, we elaborate on both steps in detail.

\subsubsection{Deep Conditional Filtering}
Before the under-sampling process, we first employ seven filters to enhance the dominant structural components of input images. As Fig.~\ref{fig:sampling_operator} (a) shows, these filters are integrated into a seven-layer linear CNN $\mathcal{T}$. Given an image $\mathbf{X}$, the first layer extracts shallow features and transitions them into a multi-channel space. To flexibly handle the requirements of different CS ratios, each of the middle five convolution layers is individually equipped with a fully connected layer that takes the CS ratio vector $\mathbf{z}=\left[\gamma_D,\gamma_G\right]$ as its input and generates a modulating vector $\mathbf{p}$ to scale output channels\footnote{It is important to note that in practice, the channel-by-channel feature modulating in $\mathcal{T}$ maintains the linearity of our sampling operator, since when $\gamma_D$ and $\gamma_G$ are predefined and fixed for a specific CS task, the scaling factors in vector $\mathbf{p}$ become constant numbers and can be absorbed into their corresponding convolution layers \cite{ding2022scaling}.}. The last layer transforms the final feature to two images $\left[\mathbf{X}_D,\mathbf{X}_G\right]$ to drive the subsequent under-sampling step.

\subsubsection{Dual-Branch Fast Sampling}
To exploit the merits of DCT and scrambled block-diagonal Gaussian matrix across different frequency bands, we propose a dual-branch sampling step. This consists of a DCT-based fast dense sampling branch (D-branch) and a scrambled block-diagonal Gaussian sampling branch (G-branch) operating in parallel to sample two measurements $\mathbf{Y}_D$ and $\mathbf{Y}_G$, from the smoothed $\mathbf{X}_D$ and $\mathbf{X}_G$ generated by $\mathcal{T}$, respectively. Specifically, as Fig.~\ref{fig:sampling_operator} (b) illustrates, the D-branch transforms $\mathbf{X}_D$ into the DCT domain using the $N\times N$ complete DCT basis $\mathbf{\Phi}_D$, and then discards the last insignificant $(1-\gamma_D)N$ coefficients in Zig-Zag order \cite{marcellin2000overview} using a binary ($\{0,1\}$-) element-wise mask $\mathbf{M}_D$ to obtain the measurement $\mathbf{Y}_D$. Fig.~\ref{fig:sampling_operator} (c) exhibits the G-branch pipeline, which first performs a global random pixel permutation $\mathbf{P}_G$ on $\mathbf{X}_G$ and then uses a convolution layer to execute block division and transformation simultaneously. In our implementation, we re-organize the complete Gaussian matrix $\mathbf{\Phi}_G$ as $n$ filters (each of size $1\times B\times B$ corresponds to a matrix row) and set the moving strides to $B$. Finally, the measurement $\mathbf{Y}_G$ is obtained by discarding the last $(1-\gamma_G)n$ coefficients of each transformed block through a channel-wise mask $\mathbf{M}_G$.

Compared to existing methods, as discussed in Sec.~\ref{sec:motivation}, our sampling operator offers three advantages as follows:

First, rather than relying on fixed matrices or networks, it utilizes a lightweight CNN as a compact, implicit representation of a transformation matrix for arbitrary CS ratios and image sizes. This benefits from the powerful learning capabilities of NNs while minimizing parameter number.

Second, our two-step design leads to an interpretable structure with well-studied DCT and Gaussian matrices. All its operations have efficient implementations on modern GPUs. According to Tab.~\ref{tab:basis_time_complexity}, with an $L$-layer $\mathcal{T}$, our $\mathcal{G}_\mathbf{A}$ has time complexity $\mathcal{O}(N\times (\log N + n+L))= \mathcal{O}(N\log N)$.

Third, it enables both local and global image perceptions in two cascaded steps to enhance the preservation of information. The introduced random permutation mitigates the potential generation risk of blocking artifacts \cite{kulkarni2016reconnet,zhang2021amp}.

\vspace{-5pt}
\subsection{Deployment-Oriented Matrix Extraction}

Inspired by the success of the recent structural reparameterization technique \cite{ding2022scaling} in general computer vision tasks, this section provides a viable method to extract the matrix form of our sampling operator for deployment and analysis.

Given a linear operator $\mathcal{F}:\mathbb{R}^N$~$\rightarrow\mathbb{R}^M$, there uniquely exist a transformation matrix $\mathbf{\Phi}\in\mathbb{R}^{M\times N}$ and a bias vector $\mathbf{b}\in\mathbb{R}^M$ satisfying that $\forall \mathbf{x}\in \mathbb{R}^N,\mathcal{F}(\mathbf{x})\equiv \mathbf{\Phi x+b}$ \cite{gentle2017matrix}. We extract the explicit forms of $\mathbf{\Phi}$ and $\mathbf{b}$ from $\mathcal{F}$ through $\mathbf{b}=\mathcal{F}(\mathbf{0}_N)$ and $\mathbf{\Phi}=\left[\mathcal{F}(\mathbf{e}_1)-\mathbf{b},\cdots,\mathcal{F}(\mathbf{e}_N)-\mathbf{b}\right]$, where $\mathbf{e}_i$ is the $i$-th column of the identity matrix $\mathbf{I}_N$. Thus, for a linear network $\mathcal{G}$ (as Figs.~\ref{fig:sampling_operator_comparison} (a)-(e) show) that takes an image of size $N=H\times W$ as the input and outputs a measurement of size $M=C_o\times H_o\times W_o$, we extract its bias and $i$-th column of transformation matrix by $\mathcal{G} (\mathbf{O})$ and $\left[\mathcal{G} (\mathbf{E}_i)-\mathcal{G} (\mathbf{O})\right]$, where the input tensors $\mathbf{O}$ and $\mathbf{E}_i$ of size $H\times W$ are the reshaped forms of $\mathbf{0}_N$ and $\mathbf{e}_i$, respectively. Algo.~\ref{alg:matrix_extraction} presents an implementation of our matrix extraction scheme, which can be efficiently executed on modern GPUs.

In our proposed COSO $\mathcal{G}_\mathbf{A}$, the network $\mathcal{T}$, DCT basis $\mathbf{\Phi}_D$, permutation $\mathbf{P}_G$, and block-diagonal Gaussian matrix $\mathbf{\Phi}_G$ can be merged into a linear transformation matrix $\mathbf{\Phi}\in\mathbb{R}^{(2N)\times N}$ and a bias vector $\mathbf{b}^\prime\in\mathbb{R}^{2N}$. Our two binary masks $\mathbf{M}_D$ and $\mathbf{M}_G$ can be merged into $\mathbf{M}\in \left\{ 0,1 \right\}^{(2N)\times (2N)}$, with only $M=\gamma N$ diagonal entries set to $1$. As illustrated in Fig.~\ref{fig:sampling_operator} (d), the entire operator is equivalent to a simple pipeline $\mathbf{y}$~$=$~$\mathbf{M(\Phi x + b^\prime)=(M\Phi)x+(Mb^\prime)}$. Given parameters of $\gamma_D$, $\gamma_G$, $H$, and $W$, we first extract $\mathbf{A=M\Phi}$ and $\mathbf{b=Mb}^\prime$ from our $\mathcal{G}_\mathbf{A}$ via Algo.~\ref{alg:matrix_extraction} and retain only their unmasked rows to obtain the final explicit global matrix of shape $M\times N$ and the $M$-dimensional bias vector, respectively. In application, the extracted matrix can be loaded onto SLMs (see Fig.~\ref{fig:arch} (b)) at the sampling end, while the additive bias injection can be equivalently performed by the measurement receiver at the reconstruction end.

It is pertinent to mention that the introduction of Algo.~\ref{alg:matrix_extraction} is based on fundamental concepts from linear algebra in a straightforward and natural manner. While the approach is grounded in basic principles, it potentially offers several advantages and connections to existing research:

First, it facilitates a connection between NNs and practical system implementations, enabling the use of various $\mathcal{G}_\mathbf{A}$ configurations. Its integration with our sampling operator reduces the storage cost, allowing only the parameters of the learned filtering network $\mathcal{T}$ and a few random seeds to be transmitted across devices, and handle various image (DMD) sizes and sampling rates.

Second, it is agnostic to NN structures and hard-/soft-ware platforms, making it adaptable for integration with existing CS models and applicable to a range of CS-based inverse problems, such as MRI, CT, and SCI.

Third, in contrast to the existing structural reparameterization techniques \cite{ding2022scaling} that typically merge simple linear operations in the inference stage of NNs, it can be applied to any black-box linear system, not just NNs, and is capable of handling arbitrary internal topologies.

\vspace{-5pt}
\newcommand\mycommfont[1]{\footnotesize\ttfamily{#1}}
\SetCommentSty{mycommfont}
\begin{algorithm}
\label{alg:matrix_extraction}
\caption{Our deployment-oriented extraction scheme for CS sampling matrix and additive bias.}
\LinesNumbered
\KwIn{Linear sampling network or any black-box linear system $\mathcal{G}:\mathbb{R}^{H\times W}\rightarrow\mathbb{R}^{C_o\times H_o\times W_o}$ with $N=H\times W$ and $M=C_o\times H_o\times W_o$.}
\KwOut{The equivalent transformation matrix $\mathbf{\Phi}\in\mathbb{R}^{M\times N}$ and additive bias $\mathbf{b}\in\mathbb{R}^{M}$.}
$\mathbf{O}:=\text{zeros}([H\times W])$\textcolor{blue}{\tcp*{Extract the bias}}
$\mathbf{B}:=\mathcal{G}(\mathbf{O})$\;
$\mathbf{b}:=\text{reshape}(\mathbf{B},~[M])$\;
$\mathbf{\Phi}:=\text{zeros}([M\times N])$\textcolor{blue}{\tcp*[f]{Extract the matrix}}\\
\For{$i\in\{1,\cdots,N\}$}{
    Set $\mathbf{e}_i$ as the $i$-th column of $\mathbf{I}_N$\;
    $\mathbf{E}_i:=\text{reshape}(\mathbf{e}_i,~[H\times W])$\;
    $\mathbf{T}_i:=\mathcal{G}(\mathbf{E}_i)$\;
    ~$\mathbf{t}_i:=\text{reshape}(\mathbf{T}_i,~[M])$\;
    Set the $i$-th column of $\mathbf{\Phi}$ as $(\mathbf{t}_i-\mathbf{b})$\;    
}(\textcolor{blue}{\tcp*[f]{The loop can be parallelized on GPUs}})
\Return $\left\{\mathbf{\Phi},\mathbf{b}\right\}$\;
\end{algorithm}
\vspace{-8pt}

\begin{figure}[!t]
\centering
\vspace{-5pt}
\hspace{-2pt}\includegraphics[width=0.49\textwidth]{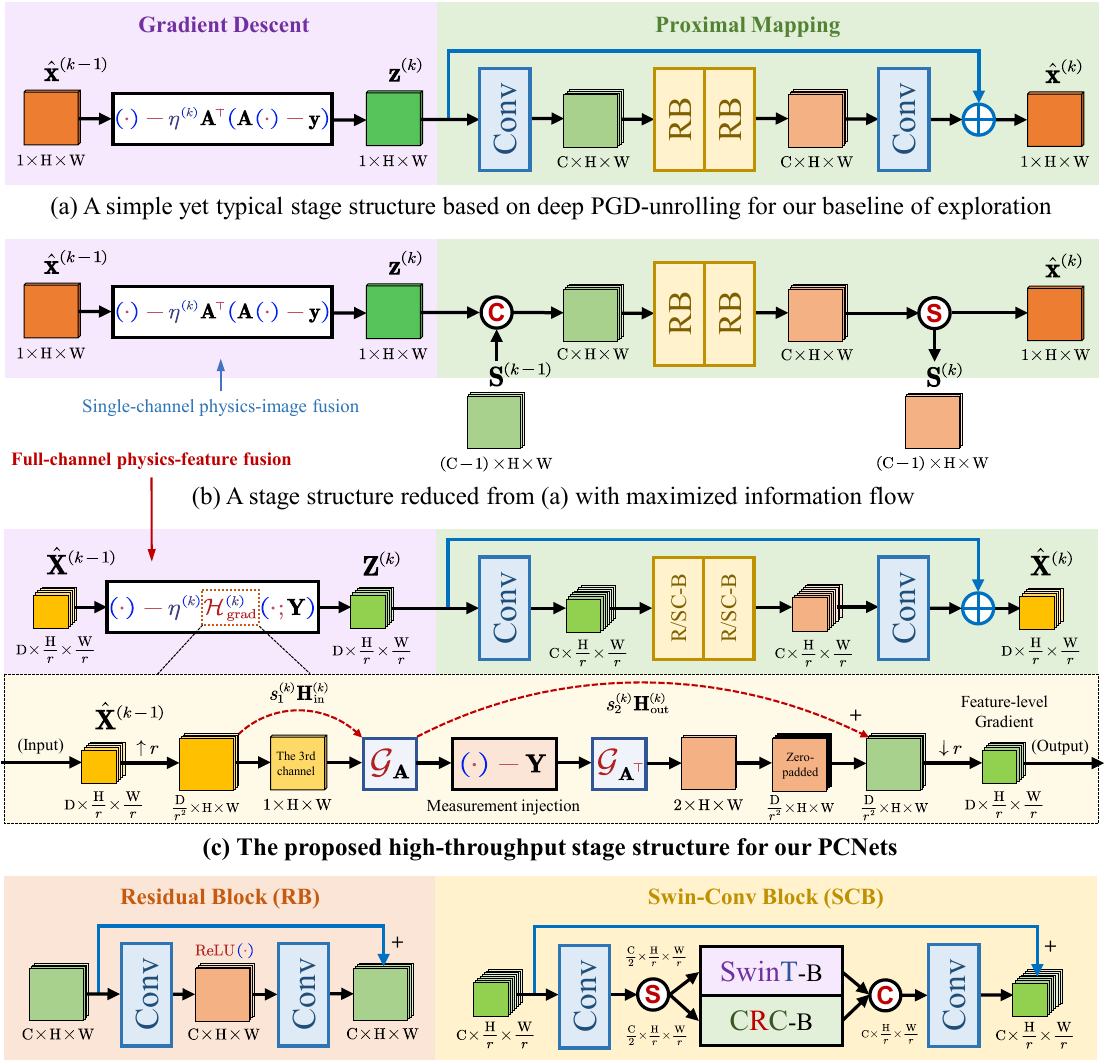}
    \vspace{-20pt}
\caption{Illustrations of three different deep PGD-unrolled stage structure designs. \textcolor{blue}{\textbf{(a)}} In each unrolled stage of our constructed baseline network \cite{zhang2018ista,you2021ista}, the image is first passed through an analytic gradient descent module and then refined by a proximal mapping network. \textcolor{blue}{\textbf{(b)}} We simplify the baseline design shown in (a) and then enable a full-channel feature-level transmission to improve performance. \textcolor{blue}{\textbf{(c)}} Each stage of our PCNet generalizes the optimization process from the original image space to a spatially downscaled feature space, further enhancing recovery efficiency. Here, notations ``C", ``S", ``+" and ``$\uparrow$$/$$\downarrow$~$r$" represent channel-wise concatenation, split, element-wise addition, and up-/down-scaling by Pixel-Shuffle/-Unshuffle layers, respectively.}
\label{fig:stage}
\vspace{-10pt}
\end{figure}

\begin{figure*}[!t]
\centering
\vspace{-8pt}
\includegraphics[width=0.96\textwidth]{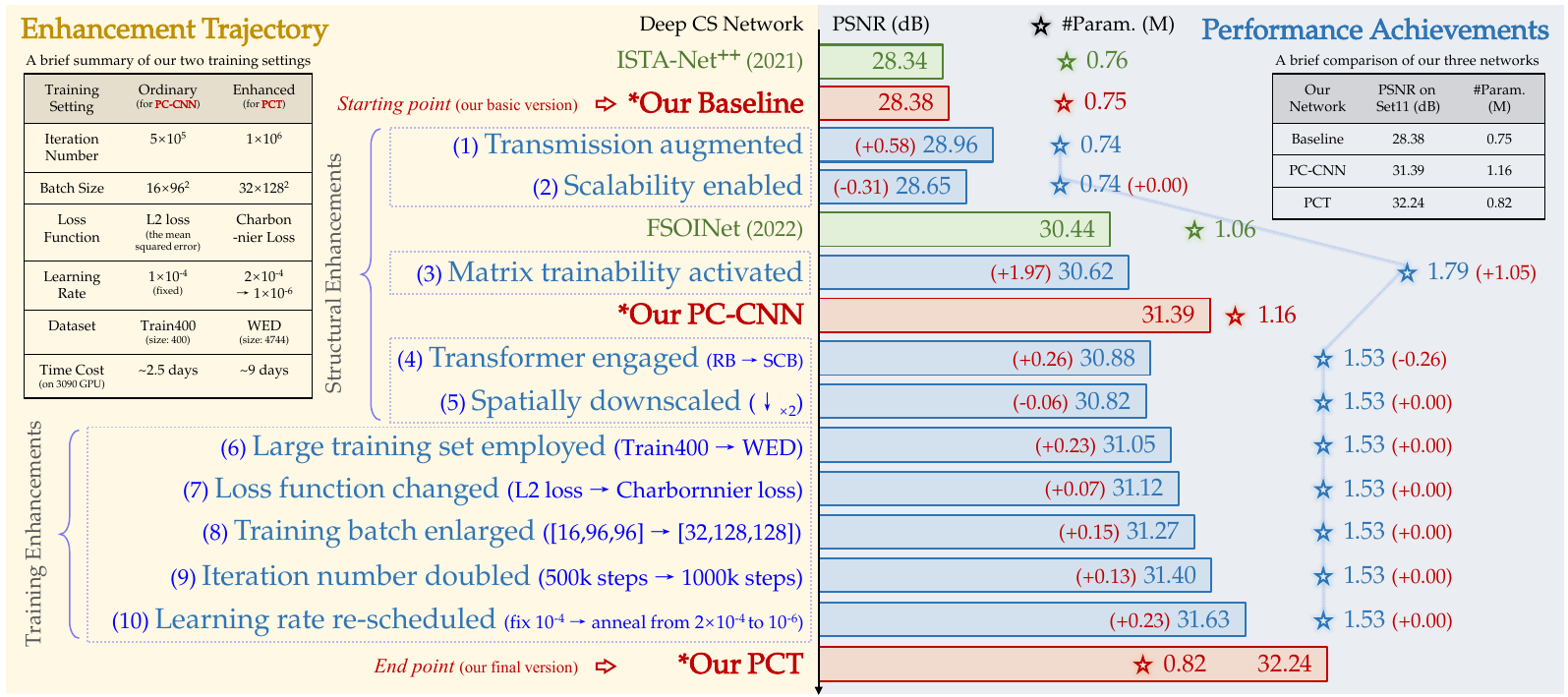}
\vspace{-10pt}
\caption{Illustration of the ``modernization'' process from the baseline network to our proposed PC-CNN and PCT. We represent the average PSNR (dB) on Set11 \cite{kulkarni2016reconnet} at $\gamma =10\%$ and the number of learnable parameters using the height of bars and stars, respectively. The enhancement involves ten sequential steps (marked by blue bars). Two reference networks and the three milestones of our exploration are highlighted in green and red, respectively. The two tables at the top left and right briefly summarize the main differences in training settings and recovery performance among our three networks. Consistent with most deep NN-based CS methods, all results are evaluated by PSNR and SSIM values on the Y channel in YCrCb space (i.e., the luminance component). All our training is conducted on a workstation equipped with an NVIDIA RTX 3090 GPU.}
\label{fig:roadmap}
\vspace{-10pt}
\end{figure*}

\vspace{-5pt}
\section{Modernizing a Reconstruction Backbone}
\label{sec:modernizing_a_reconstruction_backbone}

\subsection{PGD-Unrolled CS Architecture}
Generally, given sampling matrix $\mathbf{A}\in\mathbb{R}^{M\times N}$ and measurement $\mathbf{y}\in\mathbb{R}^M$, CS recovery problem can be formulated as:
\begin{align}
\label{eq:optimization_target}
\underset{{\mathbf{x}}}{\min}~\frac{1}{2}\lVert \mathbf{A{x}-y}\rVert_2^2 + \lambda \mathcal{R}({\mathbf{x}}),
\end{align}
where $\mathcal{R}:\mathbb{R}^N$$\rightarrow$~$\mathbb{R}$ is a regularization function assumed to be convex, with $\lambda \in\mathbb{R}^+$. As a popular first-order algorithm, the traditional proximal gradient descent (PGD) \cite{parikh2014proximal} solves Eq.~(\ref{eq:optimization_target}) by iterating between the following two update steps:
\begin{align}
\label{eq:pgd_1}
\mathbf{z}^{(k)}=&~\mathbf{\hat{x}}^{(k-1)}-\eta \mathbf{A}^\top(\mathbf{A\hat{x}}^{(k-1)}-\mathbf{y}),\\
&~\mathbf{\hat{x}}^{(k)}=\mathtt{prox}_{\lambda \mathcal{R}}(\mathbf{z}^{(k)}),
\label{eq:pgd_2}
\end{align}
where $k$ and $\eta$ denote the iteration index and step size, and Eqs.~(\ref{eq:pgd_1}) and (\ref{eq:pgd_2}) conduct the gradient descent and proximal mapping processes, respectively. Following \cite{zhang2020optimization}, as Fig.~\ref{fig:arch} (a) presents, we establish a tripartite CS architecture with PGD-unrolled reconstruction. It is composed of a sampling subnet (SS), an initialization subnet (IS), and a recovery subnet (RS) in series. SS simulates the CS acquisition process using a sampling operator $\mathcal{G}_\mathbf{A}$, IS transforms the compressed measurement $\mathbf{y}$ into the image domain by operator $\mathcal{G}_\mathbf{A^\top}$ and generates an initialization $\mathbf{\hat{x}}_\text{init}$, while RS maps $K$ iterations of PGD algorithm to cascaded stages and refines $\mathbf{\hat{x}}_\text{init}$ step-by-step to produce the recovered image $\mathbf{\hat{x}}$. In the following, instead of designing new network architectures, we explore modern, effective NN designs and enhancement strategies to obtain a comprehensive reconstruction framework.

\vspace{-5pt}
\subsection{Starting Point of Exploration: The Baseline}
\label{subsec:baseline_setup}

Motivated by the effective design in \cite{you2021ista}, we first develop a simple yet representative baseline for study. To be concrete, the sampling and initialization operators $\mathcal{G}_\mathbf{A}:\mathbb{R}^{H\times W}\rightarrow\mathbb{R}^{m\times \frac{H}{B}\times \frac{W}{B}}$ and $\mathcal{G}_\mathbf{A^\top}:\mathbb{R}^{m\times \frac{H}{B}\times \frac{W}{B}}\rightarrow\mathbb{R}^{H\times W}$ in the SS and IS are implemented by a convolution layer (see Fig.~\ref{fig:scrambled_matrix} (a)) and a transposed convolution layer that share $m$ kernels of size $1\times B\times B$ with stride $B$, to simulate $\mathbf{A}$ and $\mathbf{A}^\top$ of the block-diagonal CS sampling-initialization scheme \cite{zhang2020optimization}. Their kernel weights are obtained by reshaping a fixed random Gaussian matrix $\mathbf{\Phi}_G\in\mathbb{R}^{m\times n}$. In RS, as Fig.~\ref{fig:stage} (a) exhibits, the two update steps of Eqs.~(\ref{eq:pgd_1}) and (\ref{eq:pgd_2}) in each PGD iteration are mapped to two cascaded parts in each unrolled stage. The first gradient descent part performs an analytic calculation, while the second deep proximal mapping part is composed of a convolution layer, two classic residual blocks (RBs), and another convolution layer to predict a residual image. The data is refined in a $C$-channel feature space. The step size $\eta^{(k)}$ of each stage is learnable and initialized to 1. Following \cite{zhang2018ista,zhang2021amp,you2021coast}, we set the stage number $K=20$, block size $B=32$, and feature channel number $C=32$.

Our above baseline NN is implemented in PyTorch and trained on the Train400 \cite{zhang2017beyond} dataset (containing 400 images of size 180$\times$180) by an $\ell_2$ loss function (\textit{i.e.}, the mean squared error), using the AdamW optimizer. Mini-batches of $16$ randomly cropped $96\times 96$ patches and data augmentations of eight geometric transforms including rotations, flippings, and their combinations are utilized for training. The total iteration number and learning rate are set to $5\times 10^5$ and $1\times 10^{-4}$, respectively. As Fig.~\ref{fig:roadmap} exhibits, our constructed baseline network achieves an average PSNR of 28.38dB and parameter number of 0.75M, comparable with ISTA-Net$^{++}$ \cite{you2021ista}, on the Set11 \cite{kulkarni2016reconnet} benchmark at CS ratio $\gamma=10\%$.

\begin{table*}[!t]
\vspace{-5pt}
\caption{A high-level comparison of functional features, parameter number (M), and inference time (ms) of processing a 256$\times$256 image across seventeen CS networks. We evaluate both the sampling operator and the entire CS network separately. Star marks indicate the degree of physical knowledge utilization: \ding{72} (only measurement utilization for initialization), \ding{72}\ding{72} (iterative measurement injection), \ding{72}\ding{72}\ding{72} (iterative single-channel image-level physics utilization), \ding{72}\ding{72}\ding{72}\ding{72} (iterative full-channel feature-level utilization), and \ding{72}\ding{72}\ding{72}\ding{72}\ding{72} (iterative multi-scale full utilization).}
\vspace{-8pt}
\label{tab:compare_sota_high_level}
\centering
\setlength{\tabcolsep}{2pt}
\resizebox{1.0\textwidth}{!}{
\begin{tabular}{l|cccc|cccccccc}
\shline
\rowcolor[HTML]{EFEFEF} 
\cellcolor[HTML]{EFEFEF} &
  \multicolumn{4}{c|}{\cellcolor[HTML]{EFEFEF}CS Sampling Matrix $\mathbf{A}$ (or Operator $\mathcal{G}_\mathbf{A}$)} &
  \multicolumn{8}{c}{\cellcolor[HTML]{EFEFEF}The Entire CS Sampling and Reconstruction Network} \\ \hhline{>{\arrayrulecolor[HTML]{EFEFEF}}->{\arrayrulecolor{black}}|------------} 
\rowcolor[HTML]{EFEFEF}
\multirow{-2}{*}{\cellcolor[HTML]{EFEFEF}Method} &
  Type &
  \begin{tabular}[c]{@{}c@{}}Receptive\\ Field\end{tabular} &
  \begin{tabular}[c]{@{}c@{}}Traina\\-bility\end{tabular} &
  \begin{tabular}[c]{@{}c@{}}\#Param.\\ of $\mathbf{A}$/$\mathcal{G}_\mathbf{A}$\end{tabular} &
  \begin{tabular}[c]{@{}c@{}}Fine Granular\\ Scalability\end{tabular} &
  \begin{tabular}[c]{@{}c@{}}Deblocking\\ Ability\end{tabular} &
  \begin{tabular}[c]{@{}c@{}}High-throughput\\ Transmission\end{tabular} &
  \begin{tabular}[c]{@{}c@{}}Multiscale\\ Perception\end{tabular} &
  \begin{tabular}[c]{@{}c@{}}Physics\\ Utilization\end{tabular} &
  \begin{tabular}[c]{@{}c@{}}Unrolled\\ Algorithm\end{tabular} &
  \begin{tabular}[c]{@{}c@{}}Inferen\\ -ce Time\end{tabular} &
  \begin{tabular}[c]{@{}c@{}}Total\\\#Param.\end{tabular} \\ \hline \hline
ReconNet \cite{kulkarni2016reconnet} &
   &
   &
   &
   &
  $\times$ &
  $\times$ &
  $\checkmark$ &
  $\times$ &
  \ding{72} &
  - &
  1.75 &
  0.62 \\
ISTA-Net$^+$ \cite{zhang2018ista} &
   &
   &
   &
   &
  $\times$ &
  $\times$ &
  $\times$ &
  $\times$ &
  \ding{72}\ding{72}\ding{72} &
  ISTA \cite{blumensath2009iterative} &
  5.29 &
  0.34 \\
DPA-Net \cite{sun2020dual} &
   &
   &
   &
   &
  $\times$ &
  $\checkmark$ &
  $\checkmark$ &
  $\checkmark$ &
  \ding{72} &
  - &
  36.52 &
  9.31 \\
MAC-Net \cite{chen2020learning} &
   &
   &
   &
   &
  $\times$ &
  $\checkmark$ &
  $\times$ &
  $\times$ &
  \ding{72}\ding{72}\ding{72} &
  - &
  71.04 &
  6.12 \\
ISTA-Net$^{++}$ \cite{you2021ista} &
  \multirow{-5}{*}{\begin{tabular}[c]{@{}c@{}}Block\\ -diagonal\\ Gaussian\\ Matrix\end{tabular}} &
  \multirow{-5}{*}{Local} &
  \multirow{-5}{*}{$\times$} &
  \multirow{-5}{*}{0} &
  $\times$ &
  $\checkmark$ &
  $\times$ &
  $\times$ &
  \ding{72}\ding{72}\ding{72} &
  PGD \cite{parikh2014proximal} &
  12.35 &
  0.76 \\ \hline \hline
CSNet$^+$ \cite{shi2019image} &
   &
   &
   &
  1.05 &
  $\times$ &
  $\checkmark$ &
  $\checkmark$ &
  $\times$ &
  \ding{72} &
  - &
  16.77 &
  1.46 \\
SCSNet \cite{shi2019scalable} &
   &
   &
   &
  1.05 &
  $\checkmark$ &
  $\checkmark$ &
  $\checkmark$ &
  $\times$ &
  \ding{72} &
  - &
  30.91 &
  1.64 \\
OPINE-Net$^+$ \cite{zhang2020optimization} &
   &
   &
   &
  1.19 &
  $\times$ &
  $\checkmark$ &
  $\times$ &
  $\times$ &
  \ding{72}\ding{72}\ding{72} &
  ISTA \cite{blumensath2009iterative} &
  17.31 &
  1.10 \\
AMP-Net \cite{zhang2021amp} &
   &
   &
   &
  1.05 &
  $\times$ &
  $\checkmark$ &
  $\times$ &
  $\times$ &
  \ding{72}\ding{72}\ding{72} &
  AMP \cite{donoho2009message} &
  27.38 &
  1.71 \\
COAST \cite{you2021coast} &
   &
   &
   &
  1.19 &
  $\times$ &
  $\checkmark$ &
  $\times$ &
  $\times$ &
  \ding{72}\ding{72}\ding{72} &
  PGD \cite{parikh2014proximal} &
  45.54 &
  1.60 \\
MADUN \cite{song2021memory} &
   &
   &
   &
  1.19 &
  $\times$ &
  $\checkmark$ &
  $\checkmark$ &
  $\times$ &
  \ding{72}\ding{72}\ding{72} &
  PGD \cite{parikh2014proximal} &
  92.15 &
  3.60 \\
FSOINet \cite{chen2022fsoinet} &
   &
   &
   &
  1.05 &
  $\times$ &
  $\checkmark$ &
  $\checkmark$ &
  $\checkmark$ &
  \ding{72}\ding{72}\ding{72} &
  PGD \cite{parikh2014proximal} &
  96.57 &
  1.06 \\
CASNet \cite{chen2022content} &
  \multirow{-8}{*}{\begin{tabular}[c]{@{}c@{}}Block\\ -diagonal\\ Learned\\ Matrix\end{tabular}} &
  \multirow{-8}{*}{Local} &
  \multirow{-8}{*}{$\checkmark$} &
  1.05 &
  $\checkmark$ &
  $\checkmark$ &
  $\times$ &
  $\checkmark$ &
  \ding{72}\ding{72}\ding{72} &
  PGD \cite{parikh2014proximal} &
  103.94 &
  16.90 \\ \hline \hline
RK-CCSNet \cite{zheng2020sequential} &
   &
   &
   &
  0.11 &
  $\times$ &
  $\checkmark$ &
  $\checkmark$ &
  $\checkmark$ &
  \ding{72} &
  - &
  45.96 &
  0.74 \\
MR-CCSNet$^+$ \cite{fan2022global} &
  \multirow{-2}{*}{\begin{tabular}[c]{@{}c@{}}Stacked\\ Network\end{tabular}} &
  \multirow{-2}{*}{Local} &
  \multirow{-2}{*}{$\checkmark$} &
  0.04 &
  $\times$ &
  $\checkmark$ &
  $\checkmark$ &
  $\checkmark$ &
  \ding{72}\ding{72} &
  - &
  79.97 &
  14.25 \\ \hline \hline
\textbf{PC-CNN (Ours)} &
   &
   &
   &
   &
  $\checkmark$ &
  $\checkmark$ &
  $\checkmark$ &
  $\times$ ($r=1$) &
  \ding{72}\ding{72}\ding{72}\ding{72} &
  PGD \cite{parikh2014proximal} &
  128.26 &
  1.16 \\
\textbf{PCT (Ours)} &
  \multirow{-2}{*}{\begin{tabular}[c]{@{}c@{}}Proposed\\ COSO\end{tabular}} &
  \multirow{-2}{*}{Global} &
  \multirow{-2}{*}{$\checkmark$} &
  \multirow{-2}{*}{0.05} &
  $\checkmark$ &
  $\checkmark$ &
  $\checkmark$ &
  $\checkmark$ ($r>1$) &
  \ding{72}\ding{72}\ding{72}\ding{72}\ding{72} &
  PGD \cite{parikh2014proximal} &
  207.83 &
  0.82 \\ \shline
\end{tabular}}
\vspace{-10pt}
\end{table*}

\vspace{-5pt}
\subsection{Enhancement Strategies}

The rest part of Fig.~\ref{fig:roadmap} organizes a ``roadmap" of modernization trajectory. Specifically, we take a total of ten enhancement steps, in which the former five and the latter five are for network structure and training technique, respectively.

\vspace{-3pt}
\subsubsection{Structural Enhancements}

\noindent \textbf{(1) Transmission augmenting.} The first problem of the baseline is that it transmits a single-channel image among its stages, leading to an information bottleneck in the network trunk. Inspired by \cite{song2021memory}, as Fig.~\ref{fig:stage} (b) exhibits, in each unrolled stage, we remove the channel-expanding/-shrinking convolution layers and introduce a 31-channel zero-initialized auxiliary variable $\mathbf{S}^{(k)}$, to convey feature-level data throughout the trunk. This strategy brings a significant 0.58dB PSNR improvement and 0.01M parameter number reduction.

\vspace{3pt}
\noindent \textbf{(2) Scalability enabling.} Following \cite{shi2019scalable,chen2022content}, we introduce a flexible matrix generation scheme to make the network scalable in handling any ratios by one set of learned parameters. To be concrete, for any $\gamma=m/n$, the corresponding sampling matrix is generated by cropping the first $m$ rows of a fixed complete $n\times n$ Gaussian matrix. In each training iteration, $\gamma$ is randomly sampled from $[0,1]$. This strategy enables ratio scalability at the cost of a 0.31dB PSNR drop.

\vspace{3pt}
\noindent \textbf{(3) Matrix learning.} To improve CS sampling performance, following \cite{zhang2020optimization,zhang2021amp, chen2022content, chen2022fsoinet}, we make the above complete matrix a learnable component and jointly optimized with recovery network end-to-end. This strategy brings a 1.97dB PSNR gain at a cost of 1024$^2$~$\approx$~1.05M additional parameters.

\vspace{3pt}
\noindent \textbf{(4) Transformer block employing.} Fueled by the powerful feature refinement ability of Transformer blocks, as Fig.~\ref{fig:stage} shows, we replace RBs in each stage with the recent Swin-Conv blocks (SCBs) \cite{zhang2023practical}, which are validated to be efficient in image denoising tasks. Each SCB passes the input through a convolution, evenly splits the result into two features, and then feeds them into a Swin Transformer (SwinT) block \cite{liang2021swinir} and a convolutional block. Their outputs are concatenated and transformed by another convolution layer to produce a final residual feature. This replacement brings a PSNR gain of 0.26dB and a parameter number reduction of 0.26M.

\vspace{3pt}
\noindent \textbf{(5) Spatial downscaling.} To reduce computation costs while maintaining a large receptive field, we set the token size of SCBs to $r^2$ and keep an embedding size of $C\times 1^2$. We use a PixelUnshuffle layer to rearrange the $(C/r^2)$-channel feature elements and refine them in a downscaled space with a ratio of $r$, resulting in a feature size of $C\times (H/r)\times (W/r)$. In implementation, we set $r=2$. This strategy brings an acceptable 0.06dB PSNR drop but more than $2\times$ acceleration.

\vspace{-3pt}
\subsubsection{Training Enhancements}
\label{subsec:training_enhancements}

\noindent \textbf{(6) Large dataset utilizing.} To enrich the adaptively learned prior of our CS network, following \cite{li2024d,chen2024invertible}, we replace the original training set Train400 \cite{zhang2017beyond} with the larger WED \cite{ma2016waterloo}, which covers a wide space of 4744 high-quality images. This strategy effectively brings a PSNR improvement of 0.23dB.

\vspace{3pt}
\noindent \textbf{(7) Loss function re-selecting.} The choice of loss function is critical for image recovery networks. Following \cite{zhang2022plug,liang2021swinir}, we utilize a variety of loss functions, including the $\ell_1$, $\ell_2$, Charbonnier, perceptual, adversarial, and their combinations, to train our network. We have found that employing the Charbonnier loss, $\mathcal{L}_{char}=({\lVert \hat{\mathbf{X}}-\mathbf{X} \rVert_F^2+1\times 10^{-6}})^{1/2}$, results in stable convergence and enhanced performance, achieving a PSNR gain of 0.07dB in this instance, where $\hat{\mathbf{X}}$ is the reconstructed result of the ground truth image $\mathbf{X}$.

\vspace{3pt}
\noindent \textbf{(8-10) Sufficient NN training.} We find that there is still considerable potential for improvement in the current variant. We increase the size of each training batch to 32 image patches of 128$\times$128, double the number of iterations, and adjust the learning rate from $2\times 10^{-4}$ to $1\times 10^{-6}$ using cosine annealing. These adjustments have collectively resulted in further gains of 0.15dB, 0.13dB, and 0.23dB on PSNR.

\begin{table*}[!t]
\vspace{-5pt}
\caption{A quantitative comparison of average PSNR (dB) among different CS methods, evaluated on four widely adopted benchmarks: Set11~\cite{kulkarni2016reconnet}, CBSD68~\cite{martin2001database}, Urban100~\cite{huang2015single} and DIV2K~\cite{agustsson2017ntire}, with three sampling rates $\gamma\in\{10\%,30\%,50\%\}$. Throughout this paper, the best, second-best, and third-best results of each test case are highlighted in red, blue, and green colors, respectively.}
\vspace{-8pt}
\label{tab:compare_sota_psnr}
\centering
\resizebox{1.0\textwidth}{!}{
\begin{tabular}{lc|ccc|ccc|ccc|ccc}
\shline
\rowcolor[HTML]{EFEFEF} 
\multicolumn{1}{l|}{\cellcolor[HTML]{EFEFEF}} &
  Test Set &
  \multicolumn{3}{c|}{\cellcolor[HTML]{EFEFEF}Set11 \cite{kulkarni2016reconnet}} &
  \multicolumn{3}{c|}{\cellcolor[HTML]{EFEFEF}CBSD68 \cite{martin2001database}} &
  \multicolumn{3}{c|}{\cellcolor[HTML]{EFEFEF}Urban100 \cite{huang2015single}} &
  \multicolumn{3}{c}{\cellcolor[HTML]{EFEFEF}DIV2K \cite{agustsson2017ntire}} \\ \hhline{>{\arrayrulecolor[HTML]{EFEFEF}}->{\arrayrulecolor{black}}|-------------} 
\rowcolor[HTML]{EFEFEF} 
\multicolumn{1}{l|}{\multirow{-2}{*}{\cellcolor[HTML]{EFEFEF}Method}} &
  CS Ratio $\gamma$ &
  10\% &
  30\% &
  50\% &
  10\% &
  30\% &
  50\% &
  10\% &
  30\% &
  50\% &
  10\% &
  30\% &
  50\% \\ \hline \hline
\multicolumn{2}{l|}{ReconNet (CVPR 2016) \cite{kulkarni2016reconnet}}        & 24.08 & 29.46 & 32.76 & 23.92 & 27.97 & 30.79 & 20.71 & 25.15 & 28.15 & 24.41 & 29.09 & 32.15 \\
\multicolumn{2}{l|}{ISTA-Net$^+$ (CVPR 2018) \cite{zhang2018ista}}    & 26.49 & 33.70 & 38.07 & 25.14 & 30.24 & 33.94 & 22.81 & 29.83 & 34.33 & 26.30 & 32.65 & 36.88 \\
\multicolumn{2}{l|}{DPA-Net (TIP 2020) \cite{sun2020dual}}         & 27.66 & 33.60 & -     & 25.33 & 29.58 & -     & 24.00 & 29.04 & -     & 27.09 & 32.37 & -     \\
\multicolumn{2}{l|}{MAC-Net (ECCV 2020) \cite{chen2020learning}}         & 27.92 & 33.87 & 37.76 & 25.70 & 30.10 & 33.37 & 23.71 & 29.03 & 33.10 & 26.72 & 32.23 & 35.40 \\
\multicolumn{2}{l|}{ISTA-Net$^{++}$ (ICME 2021) \cite{you2021ista}} & 28.34 & 34.86 & 38.73 & 26.25 & 31.10 & 34.85 & 24.95 & 31.50 & 35.58 & 27.82 & 33.74 & 37.78 \\ \hline \hline
\multicolumn{2}{l|}{CSNet$^+$ (TIP 2019) \cite{shi2019image}}       & 28.34 & 34.30 & 38.52 & 27.04 & 31.60 & 35.27 & 23.96 & 29.12 & 32.76 & 28.23 & 33.63 & 37.59 \\
\multicolumn{2}{l|}{SCSNet (CVPR 2019) \cite{shi2019scalable}}          & 28.52 & 34.64 & 39.01 & 27.14 & 31.72 & 35.62 & 24.22 & 29.41 & 33.31 & 28.41 & 33.85 & 37.97 \\
\multicolumn{2}{l|}{OPINE-Net$^+$ (JSTSP 2020) \cite{zhang2020optimization}}   & 29.81 & 35.99 & 40.19 & 27.66 & 32.38 & 36.21 & 25.90 & 31.97 & 36.28 & 29.26 & 35.03 & 39.27 \\
\multicolumn{2}{l|}{AMP-Net (TIP 2021) \cite{zhang2021amp}}         & 29.40 & 36.03 & 40.34 & 27.71 & 32.72 & 36.72 & 25.32 & 31.63 & 35.91 & 29.08 & 35.41 & 39.46 \\
\multicolumn{2}{l|}{COAST (TIP 2021) \cite{you2021coast}}           & 30.02 & 36.33 & 40.33 & 27.76 & 32.56 & 36.34 & 26.17 & 32.48 & 36.56 & 29.46 & 35.32 & 39.43 \\
\multicolumn{2}{l|}{MADUN (ACM MM 2021) \cite{song2021memory}}           & 29.89 & 36.90 & 40.75 & 28.04 & 33.07 & 36.99 & 26.23 & 33.00 & 36.69 & 29.62 & \textcolor{green}{36.04} & 40.06 \\
\multicolumn{2}{l|}{FSOINet (ICASSP 2022) \cite{chen2022fsoinet}}         & \textcolor{green}{30.44} & \textcolor{green}{37.00} & \textcolor{green}{41.08} & 28.12 & \textcolor{green}{33.15} & \textcolor{green}{37.18} & \textcolor{green}{26.87} & \textcolor{green}{33.29} & \textcolor{green}{37.25} & \textcolor{green}{30.01} & 36.03 & \textcolor{green}{40.38} \\
\multicolumn{2}{l|}{CASNet (TIP 2022) \cite{chen2022content}}          & 30.31 & 36.91 & 40.93 & \textcolor{green}{28.16} & 33.05 & 36.99 & 26.85 & 32.85 & 36.94 & \textcolor{green}{30.01} & 35.90 & 40.14 \\ \hline \hline
\multicolumn{2}{l|}{RK-CCSNet (ECCV 2020) \cite{zheng2020sequential}}       & -     & -     & 38.01 & -     & -     & 34.69 & -     & -     & 33.39 & -     & -     & 37.32 \\
\multicolumn{2}{l|}{MR-CCSNet$^+$ (CVPR 2022) \cite{fan2022global}}       & -     & -     & 39.27 & -     & -     & 35.45 & -     & -     & 34.40 & -     & -     & 38.16 \\ \hline \hline
\multicolumn{2}{l|}{\textbf{PC-CNN (Ours)}}   & \textcolor{blue}{31.39} & \textcolor{blue}{37.91} & \textcolor{blue}{41.83} & \textcolor{blue}{28.87} & \textcolor{blue}{34.55} & \textcolor{blue}{39.06} & \textcolor{blue}{27.15} & \textcolor{blue}{33.66} & \textcolor{blue}{37.80} & \textcolor{blue}{30.52} & \textcolor{blue}{36.66} & \textcolor{blue}{41.01} \\
\multicolumn{2}{l|}{\textbf{PCT (Ours)}} &  \textcolor{red}{32.24}   &  \textcolor{red}{38.72} &\textcolor{red}{42.71}& \textcolor{red}{29.18} & \textcolor{red}{34.86} & \textcolor{red}{39.57} &  \textcolor{red}{28.50}& \textcolor{red}{34.88} & \textcolor{red}{39.11} & \textcolor{red}{31.12} & \textcolor{red}{37.26} & \textcolor{red}{41.76}        \\ \shline
\end{tabular}}
\vspace{-10pt}
\end{table*}

\begin{figure*}[!t]
\setlength{\tabcolsep}{0.5pt}
\hspace{-4pt}
\resizebox{1.005\textwidth}{!}{
\tiny
\begin{tabular}{cccccccccccccc}
    Ground Truth & ReconNet & ISTA-Net$^\text{+}$ & ISTA-Net$^\text{++}$ & CSNet$^\text{+}$ & SCSNet & OPINE-Net$^\text{+}$ & AMP-Net & COAST & MADUN & FSOINet & CASNet & \textbf{PC-CNN (Ours)} & \textbf{PCT (Ours)}\\
    \includegraphics[width=0.07\textwidth]{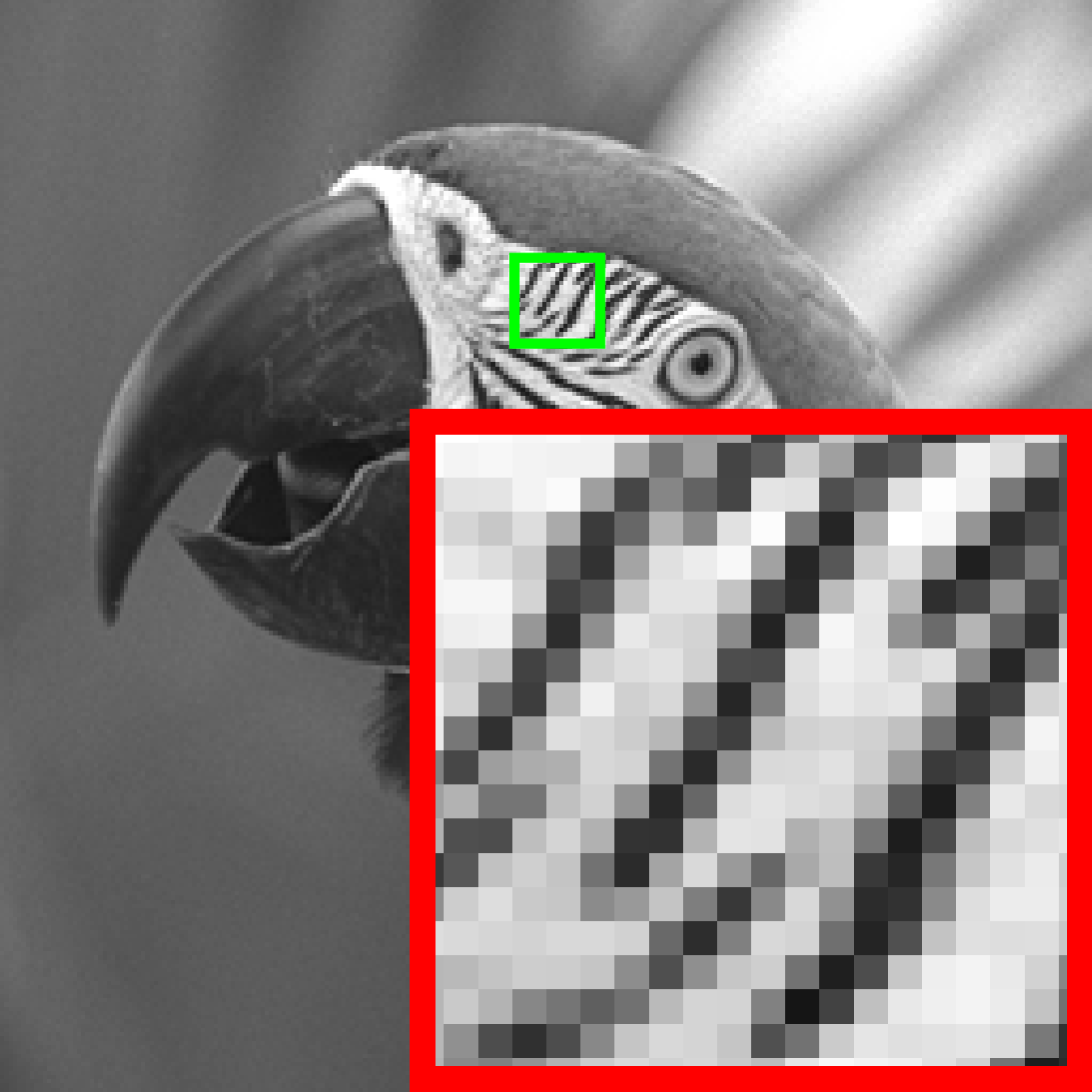}
    &\includegraphics[width=0.07\textwidth]{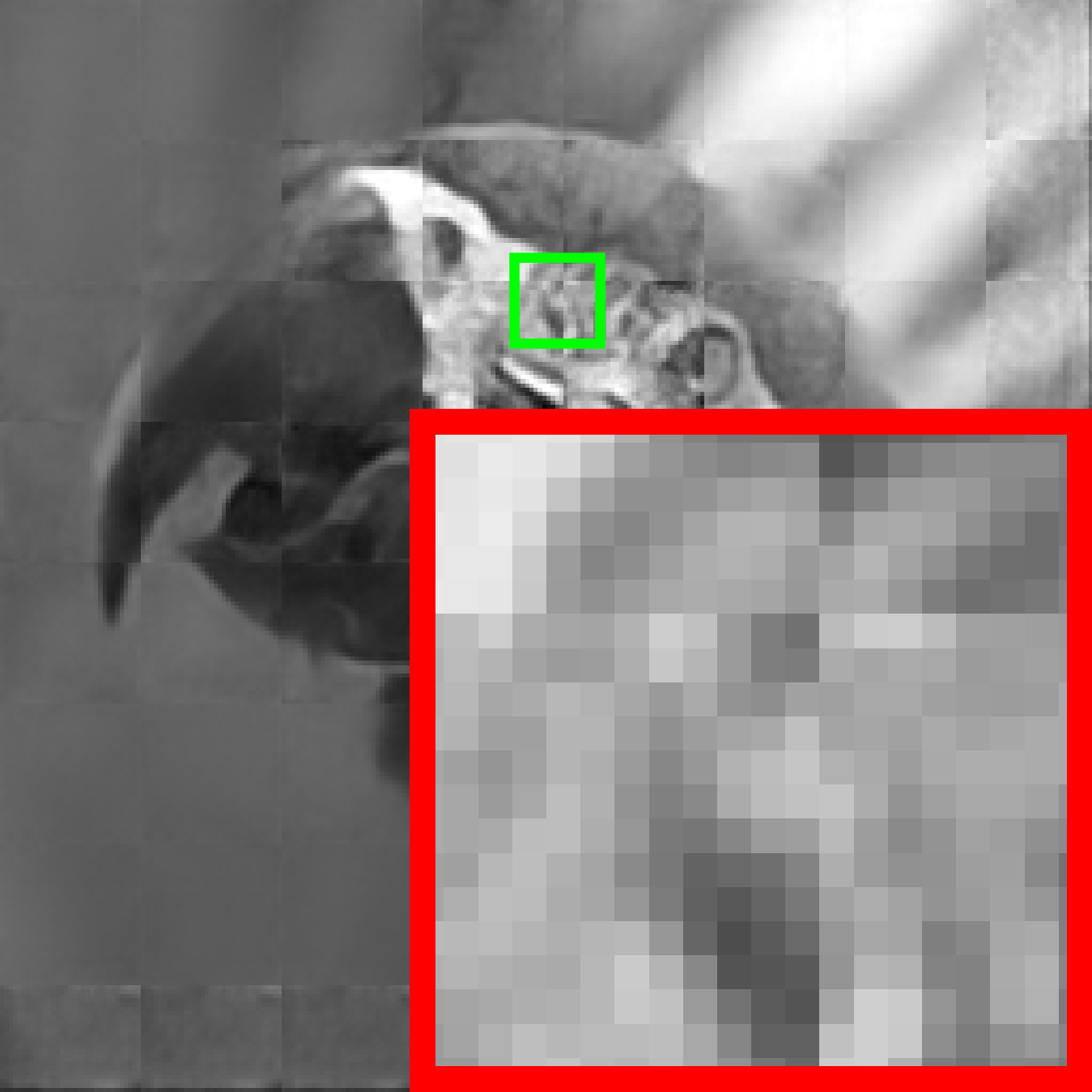}
    &\includegraphics[width=0.07\textwidth]{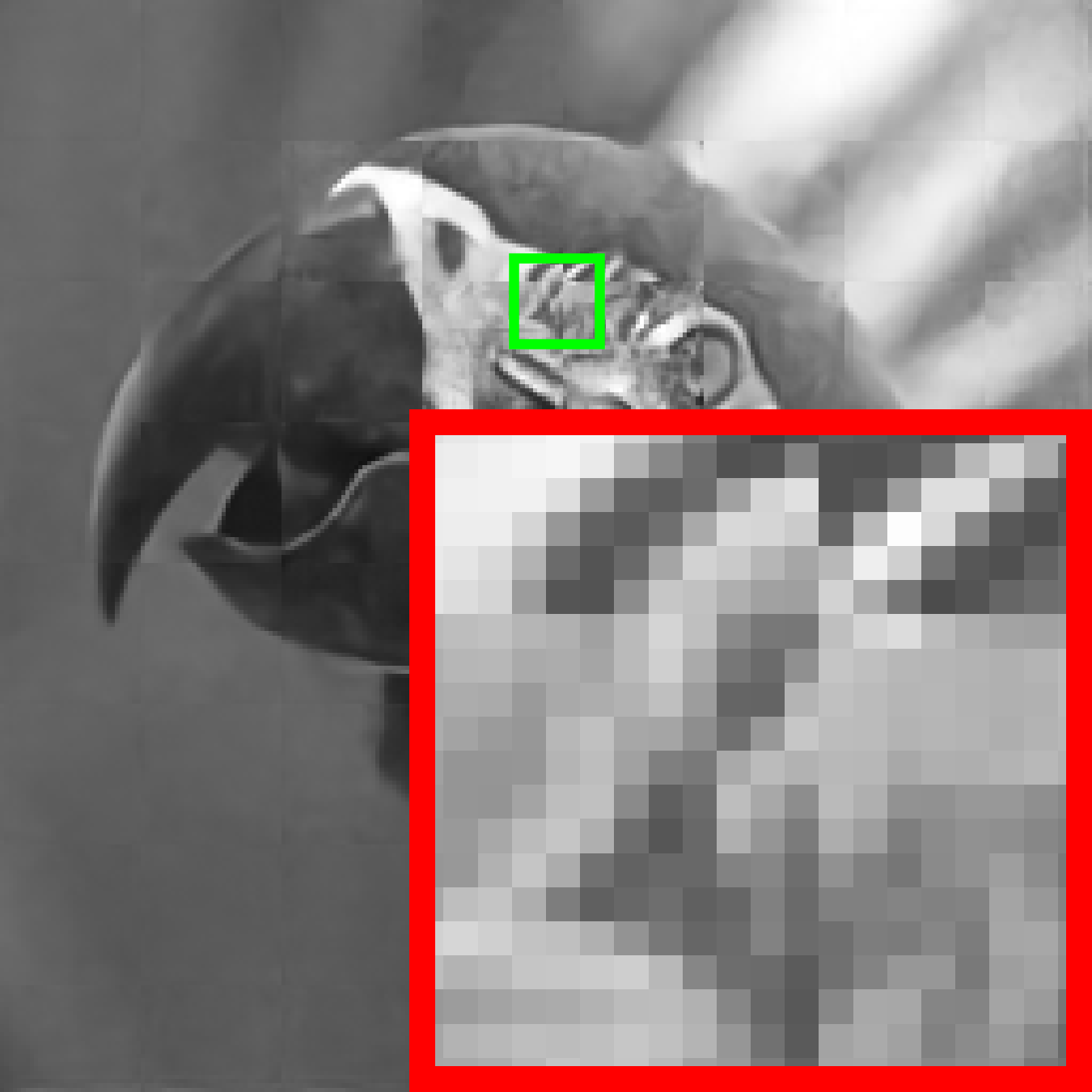}
    &\includegraphics[width=0.07\textwidth]{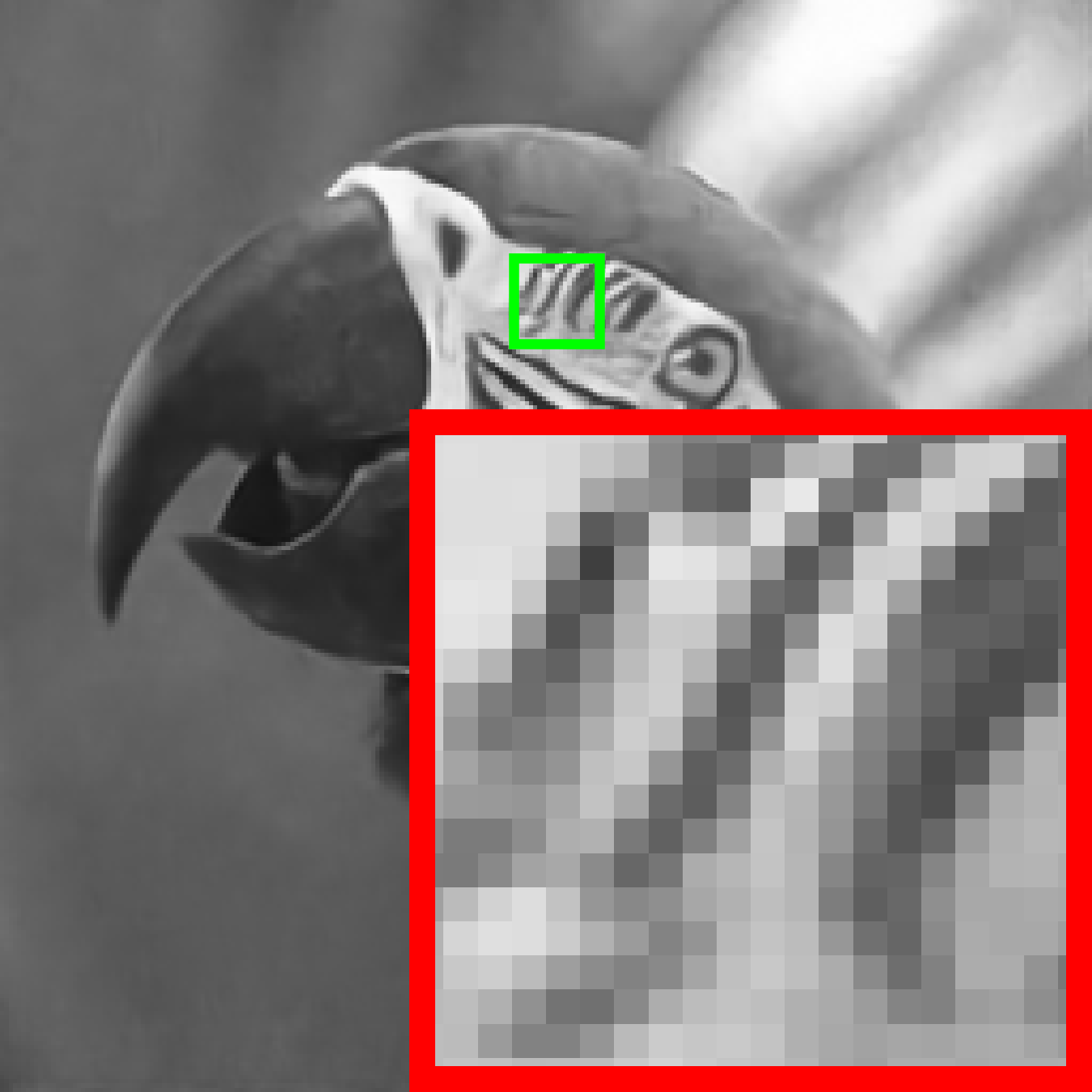}
    &\includegraphics[width=0.07\textwidth]{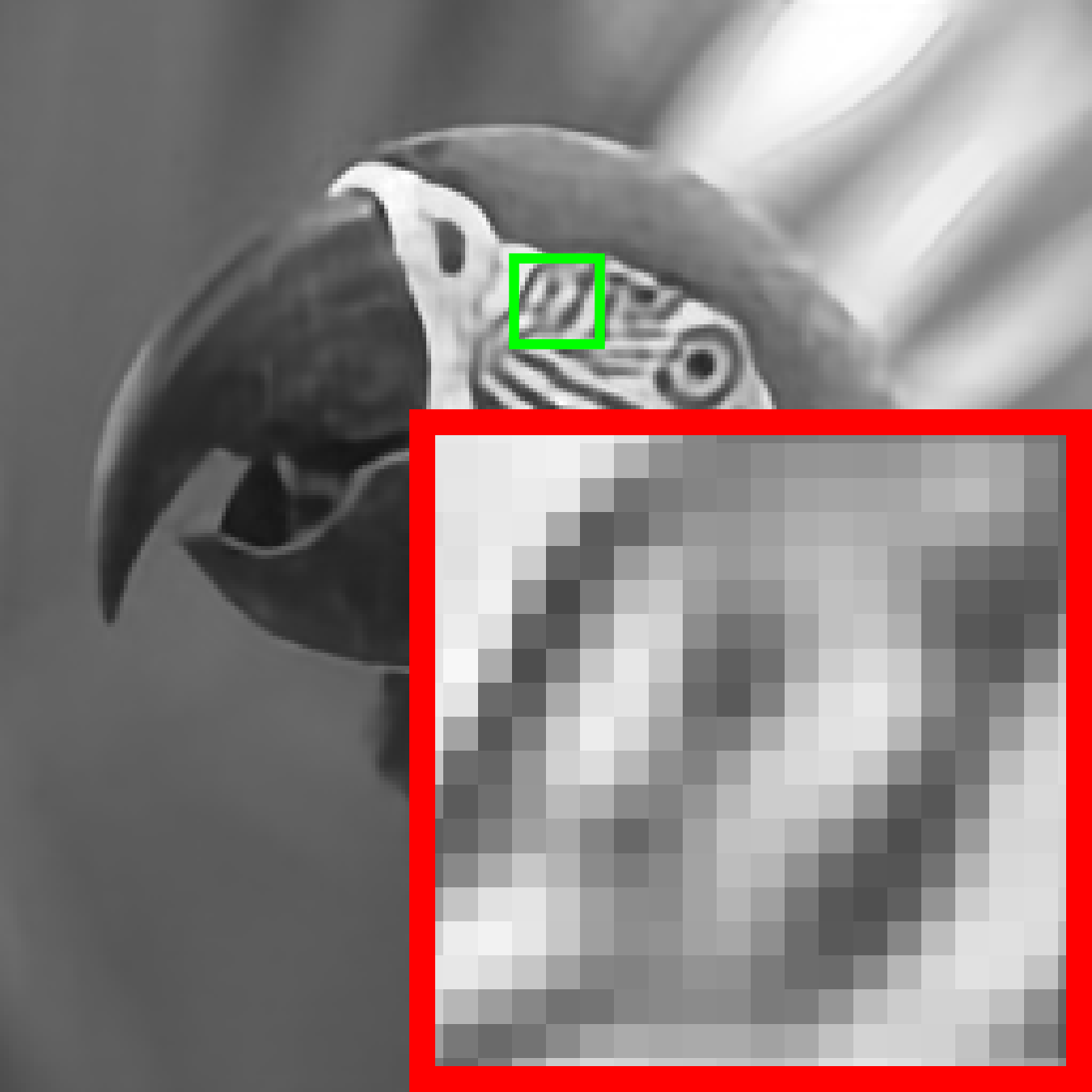}
    &\includegraphics[width=0.07\textwidth]{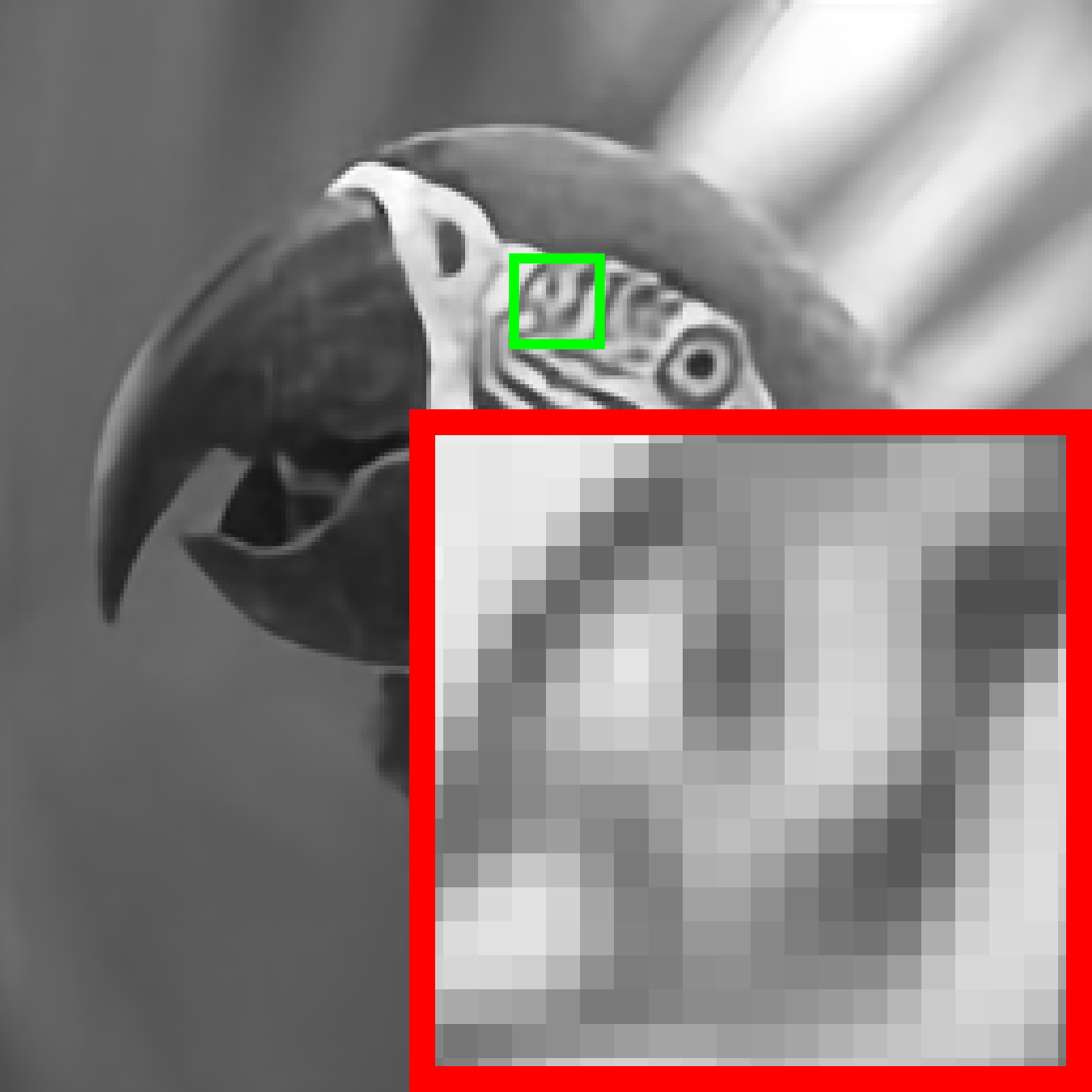}
    &\includegraphics[width=0.07\textwidth]{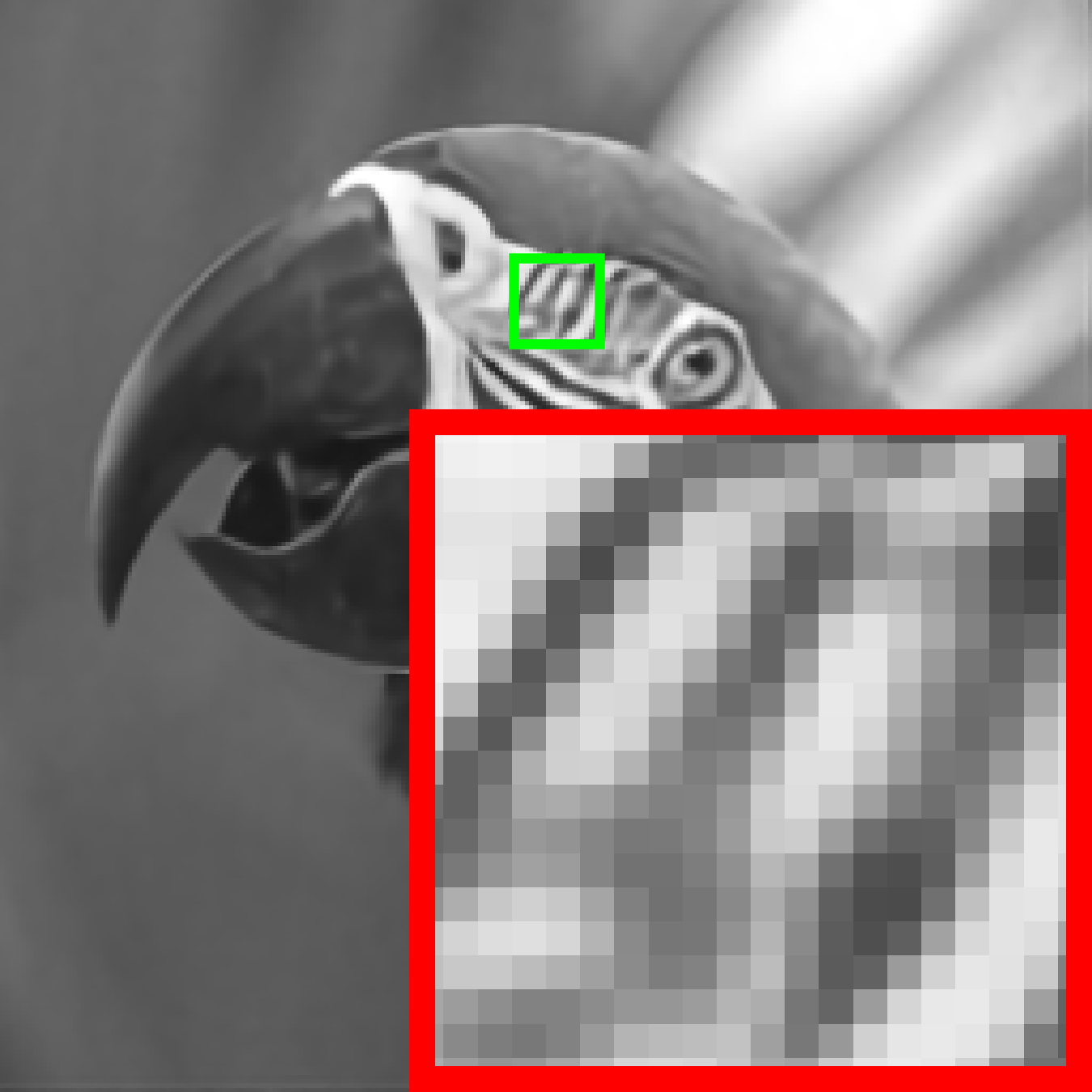}
    &\includegraphics[width=0.07\textwidth]{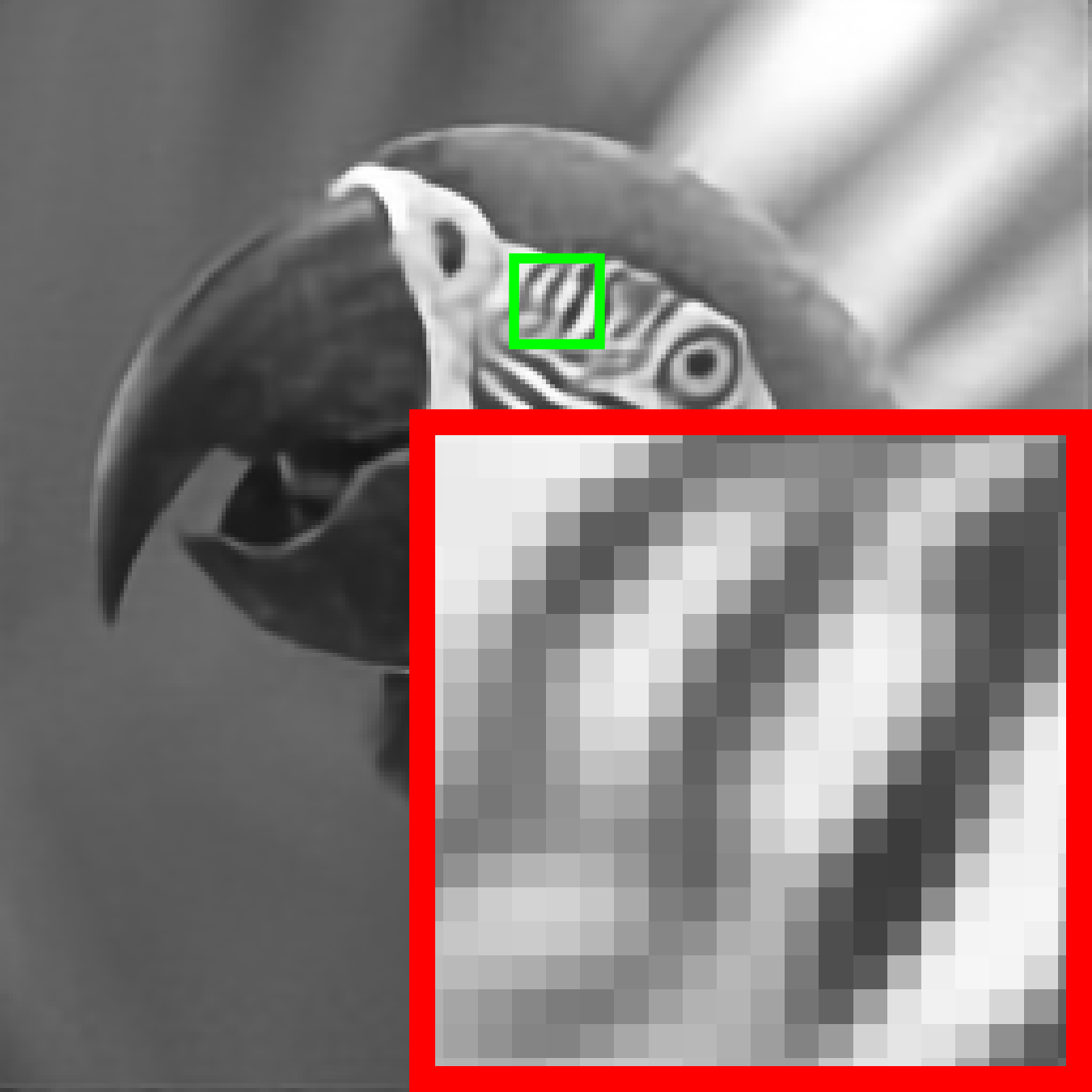}
    &\includegraphics[width=0.07\textwidth]{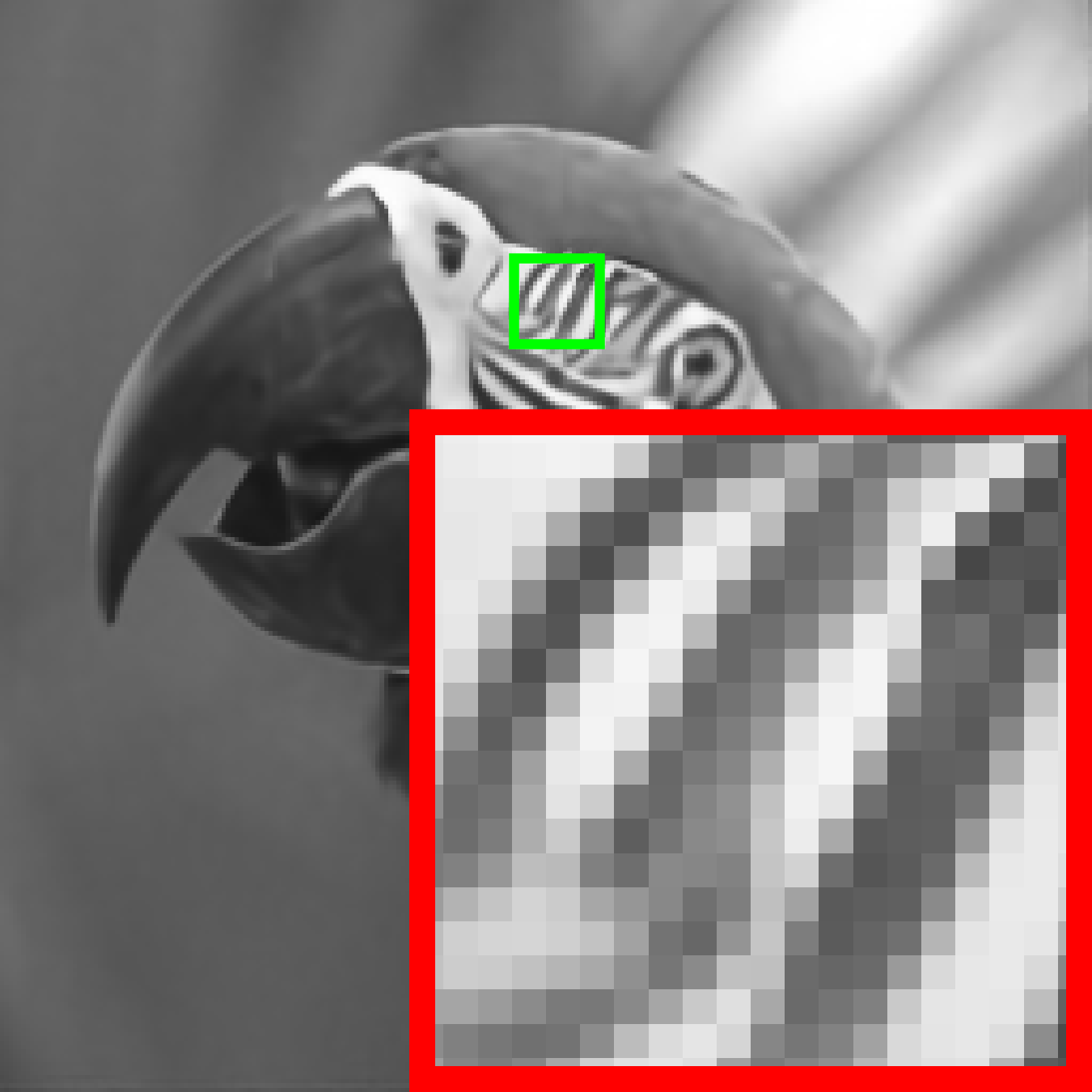}
    &\includegraphics[width=0.07\textwidth]{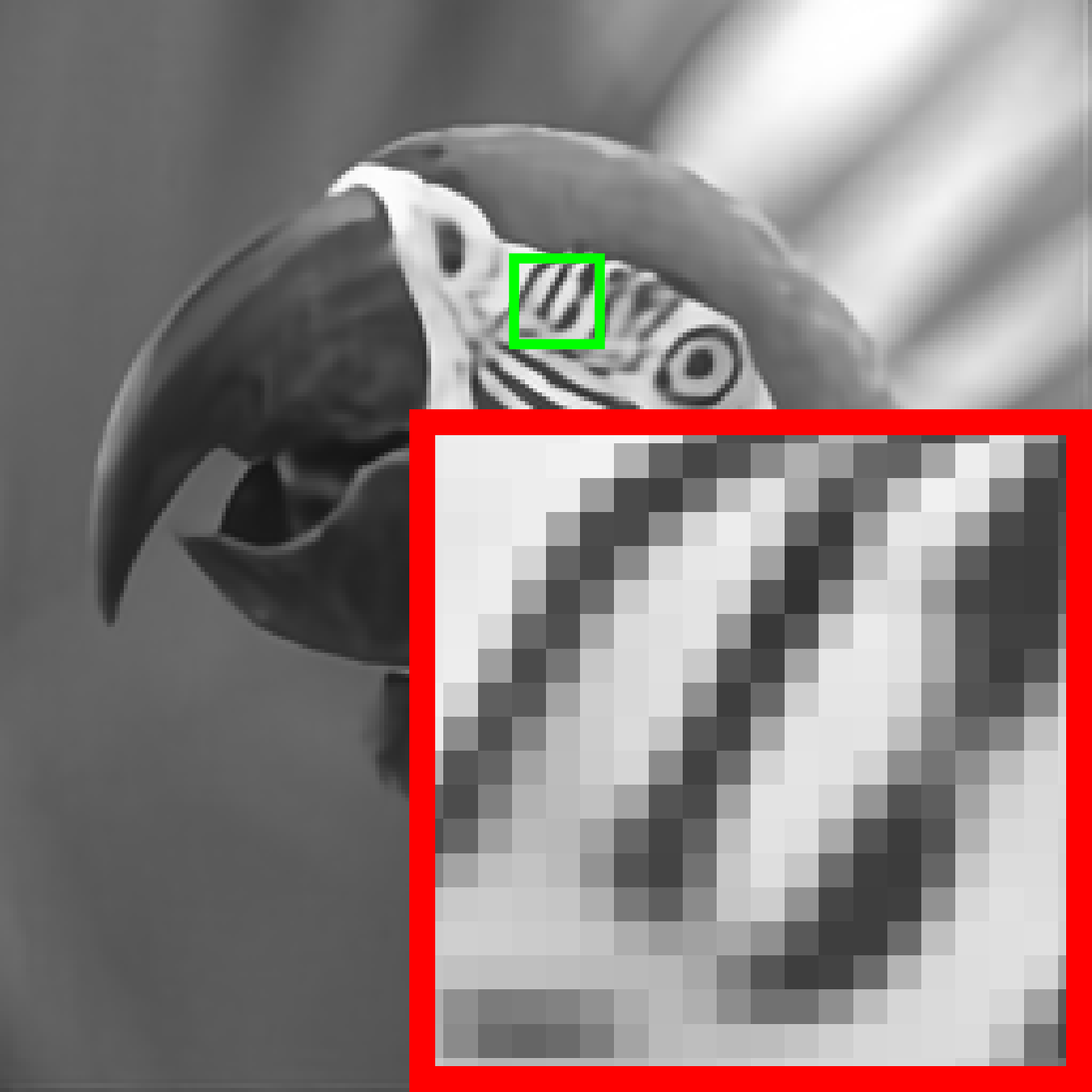}
    &\includegraphics[width=0.07\textwidth]{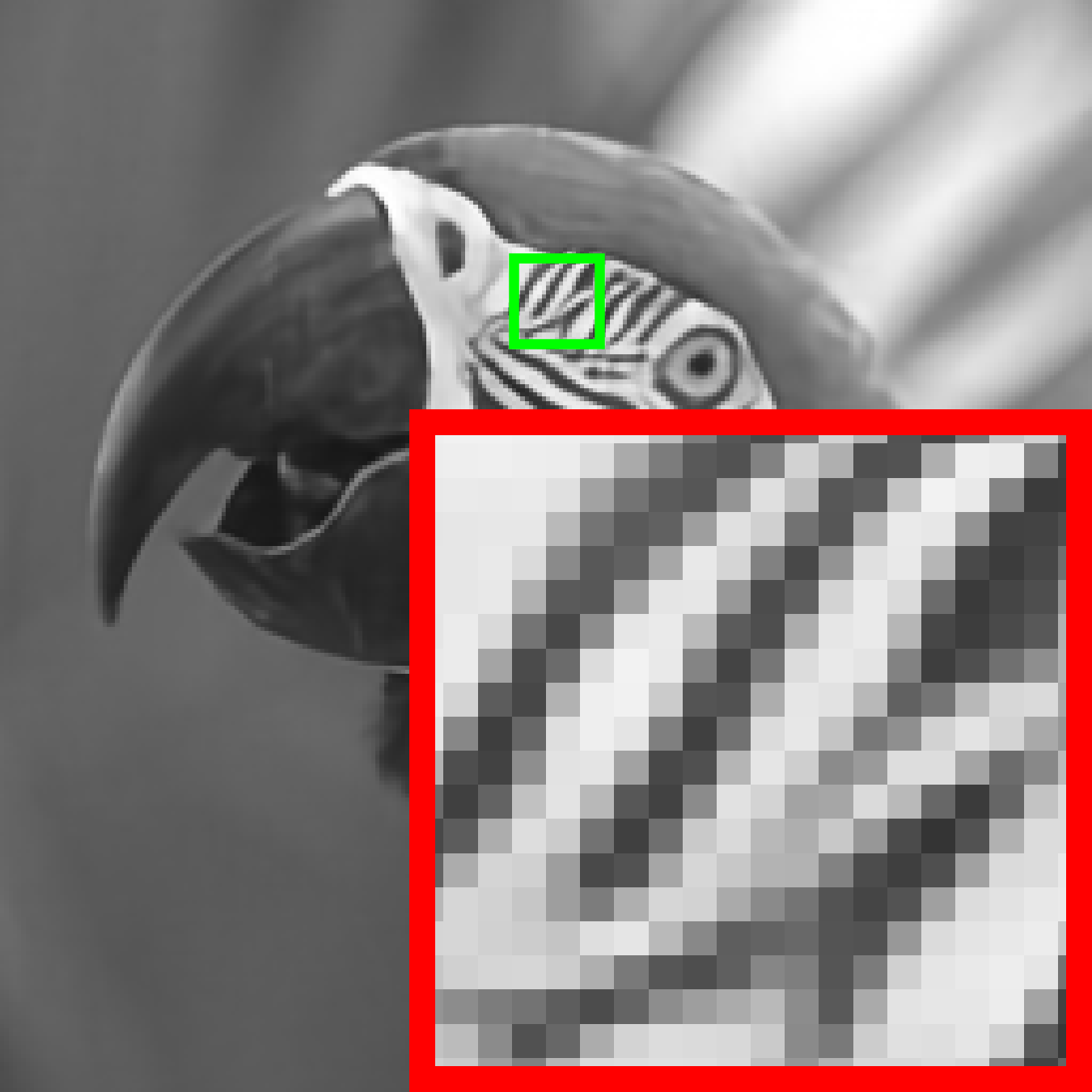}
    &\includegraphics[width=0.07\textwidth]{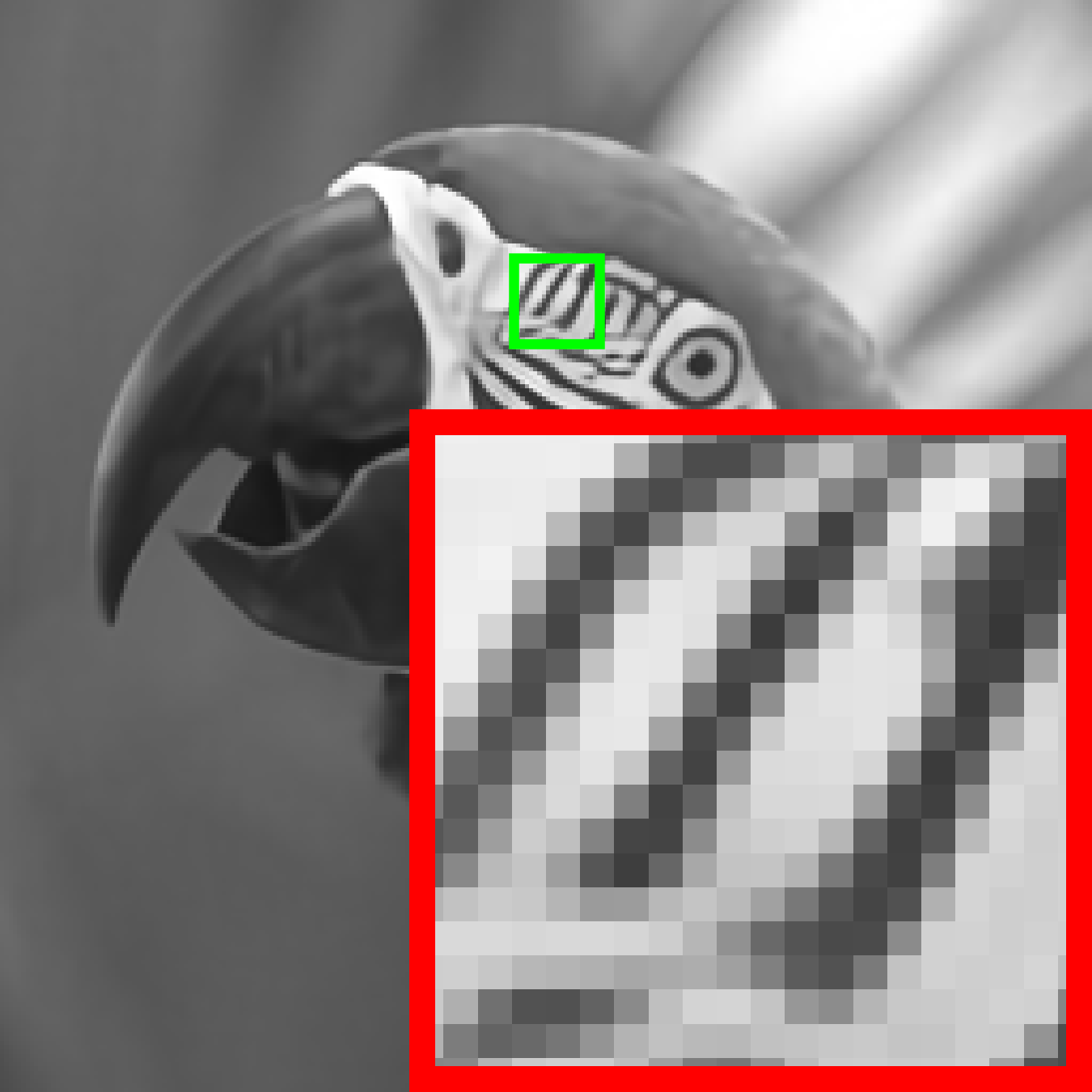}
    &\includegraphics[width=0.07\textwidth]{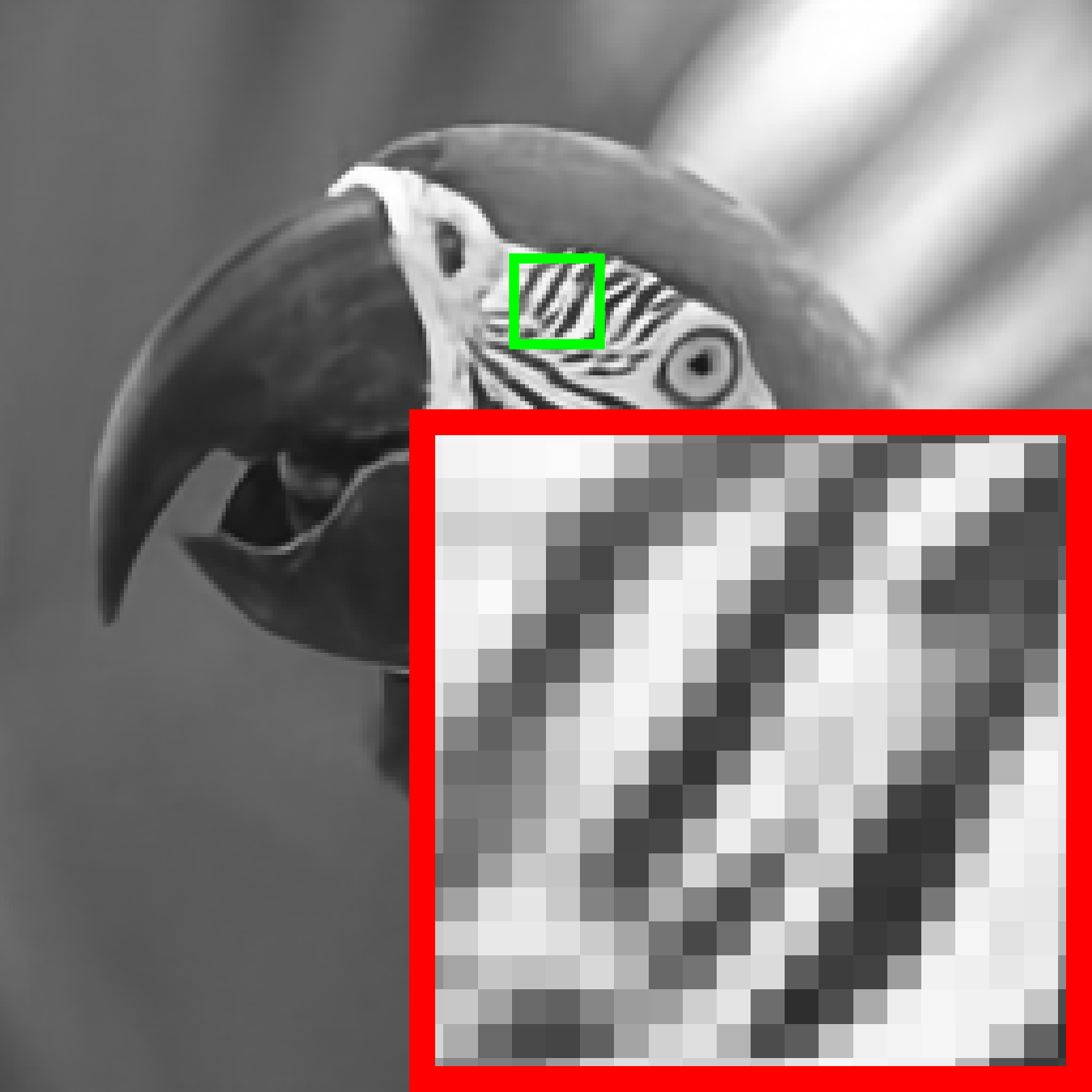}
    &\includegraphics[width=0.07\textwidth]{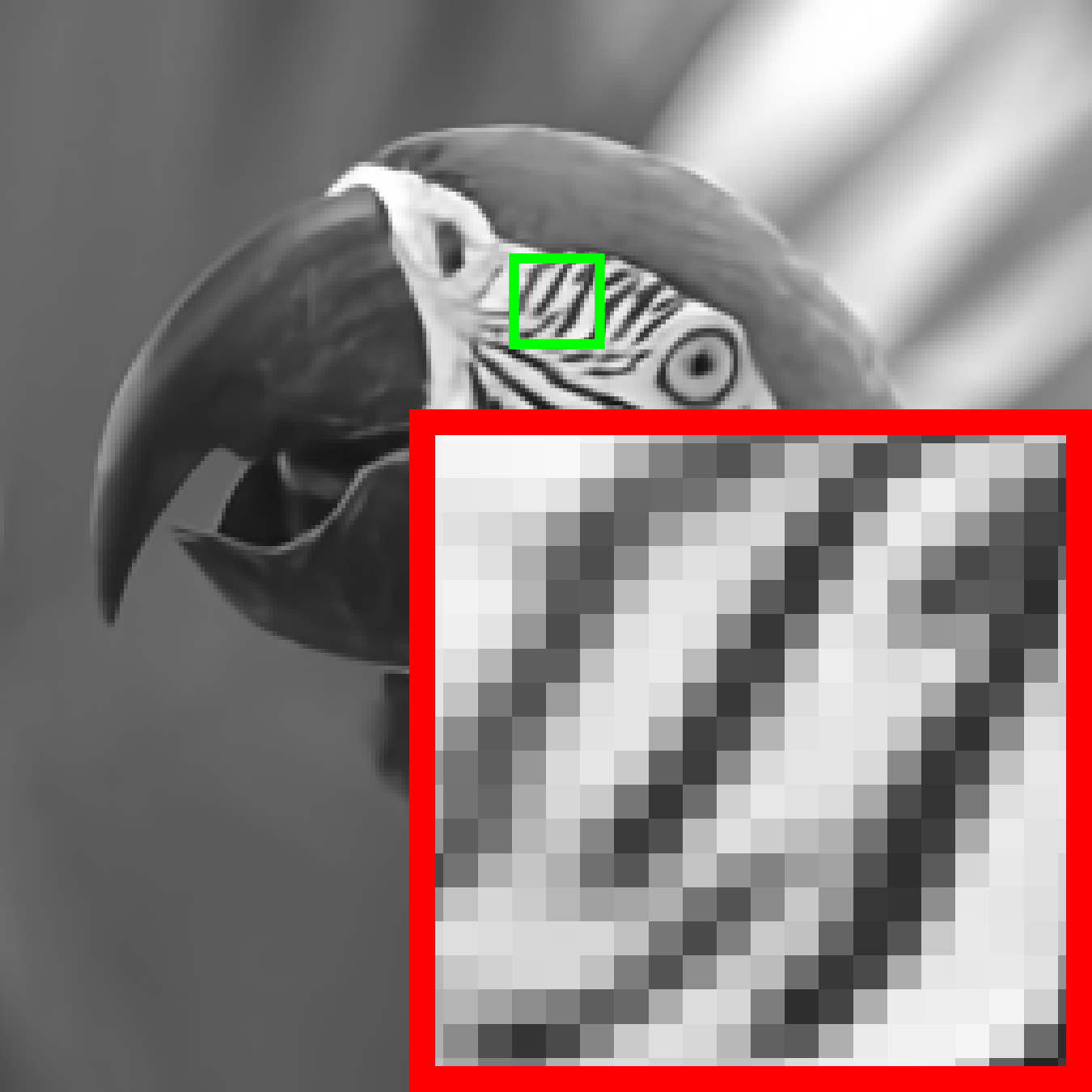}\\
    PSNR/SSIM & 24.21/0.7660 & 26.37/0.8523 & 28.09/0.8864 & 28.11/0.8899 & 28.10/0.8921 & 29.34/0.9155 & 29.20/0.9055 & 29.30/0.9206 & 29.46/0.9207 & \textcolor{green}{29.81}/\textcolor{green}{0.9253} & 29.71/\textcolor{green}{0.9253} & \textcolor{blue}{33.00}/\textcolor{blue}{0.9344} & \textcolor{red}{33.69}/\textcolor{red}{0.9413}\\
    \includegraphics[width=0.07\textwidth]{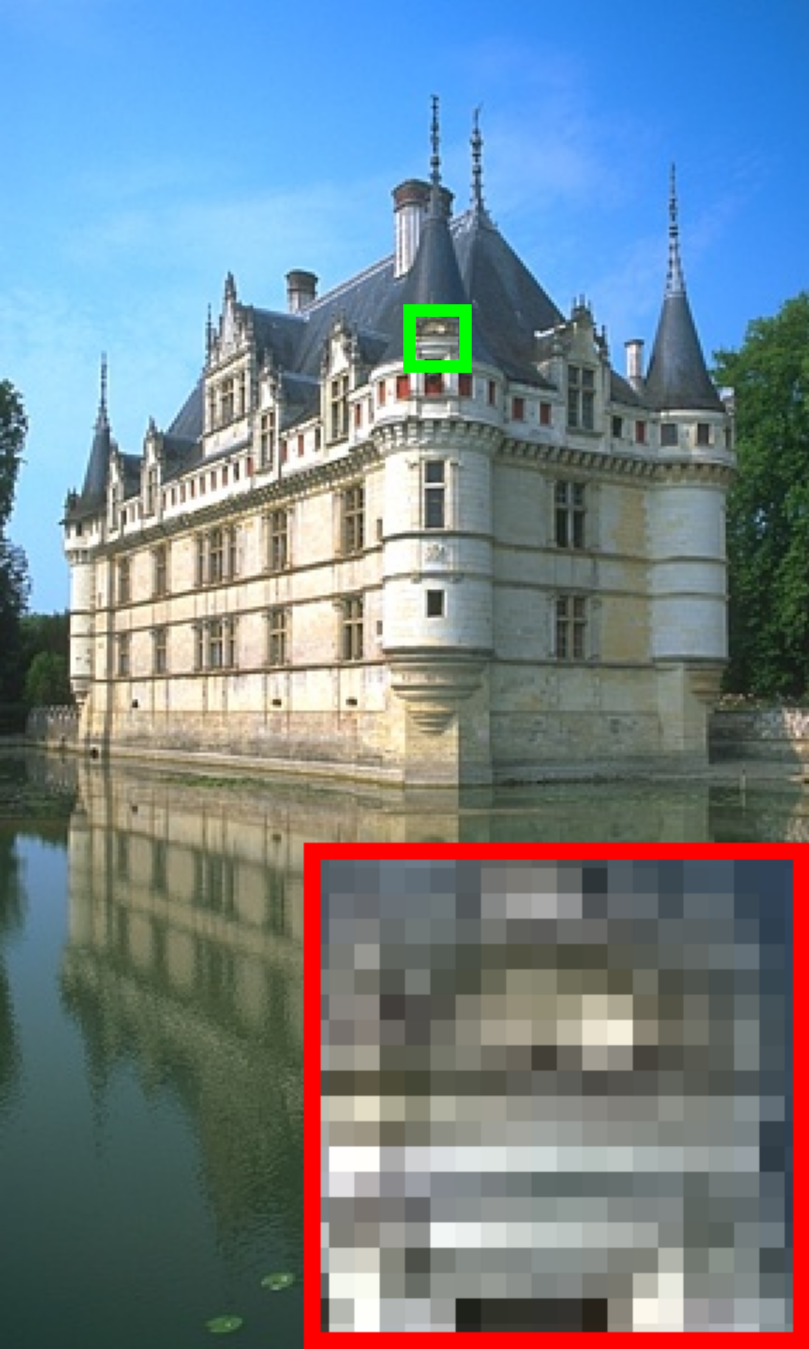}
    &\includegraphics[width=0.07\textwidth]{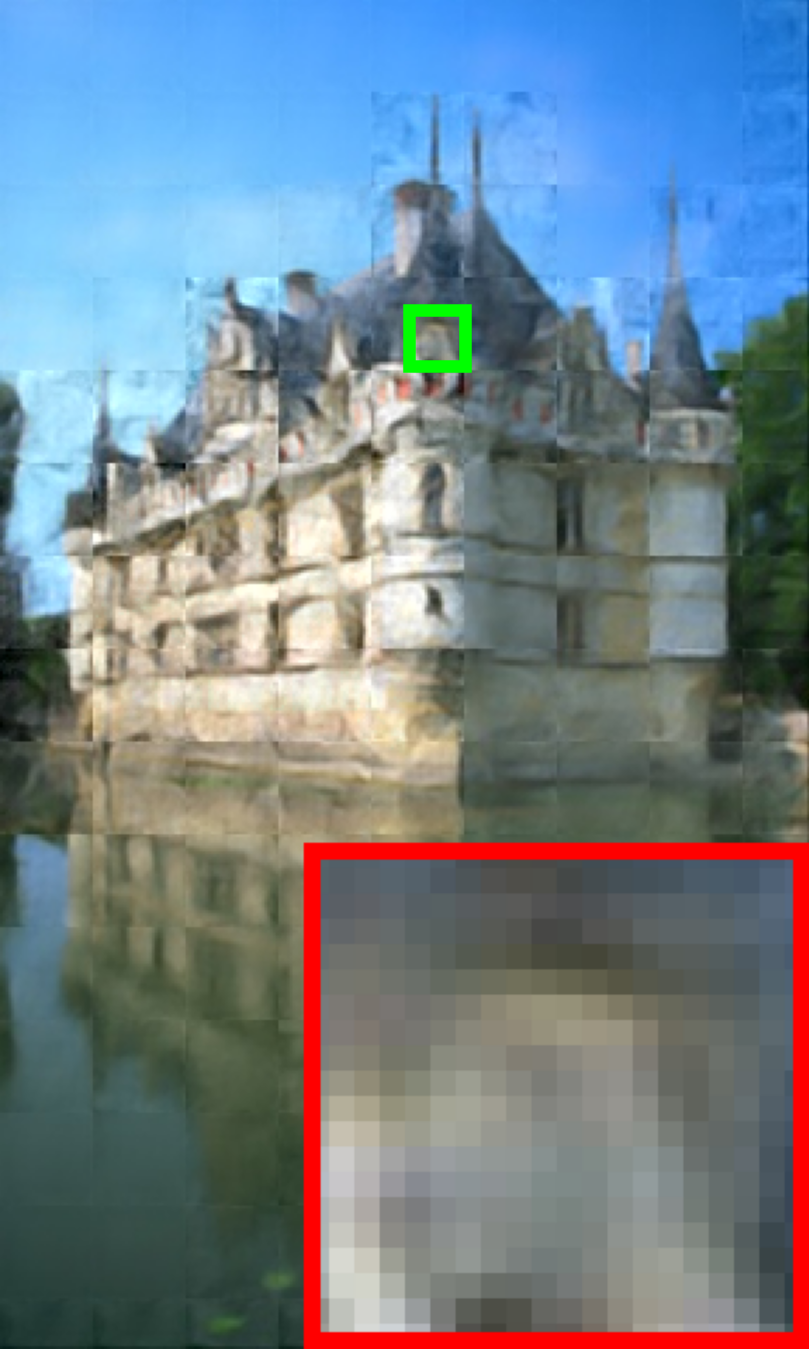}
    &\includegraphics[width=0.07\textwidth]{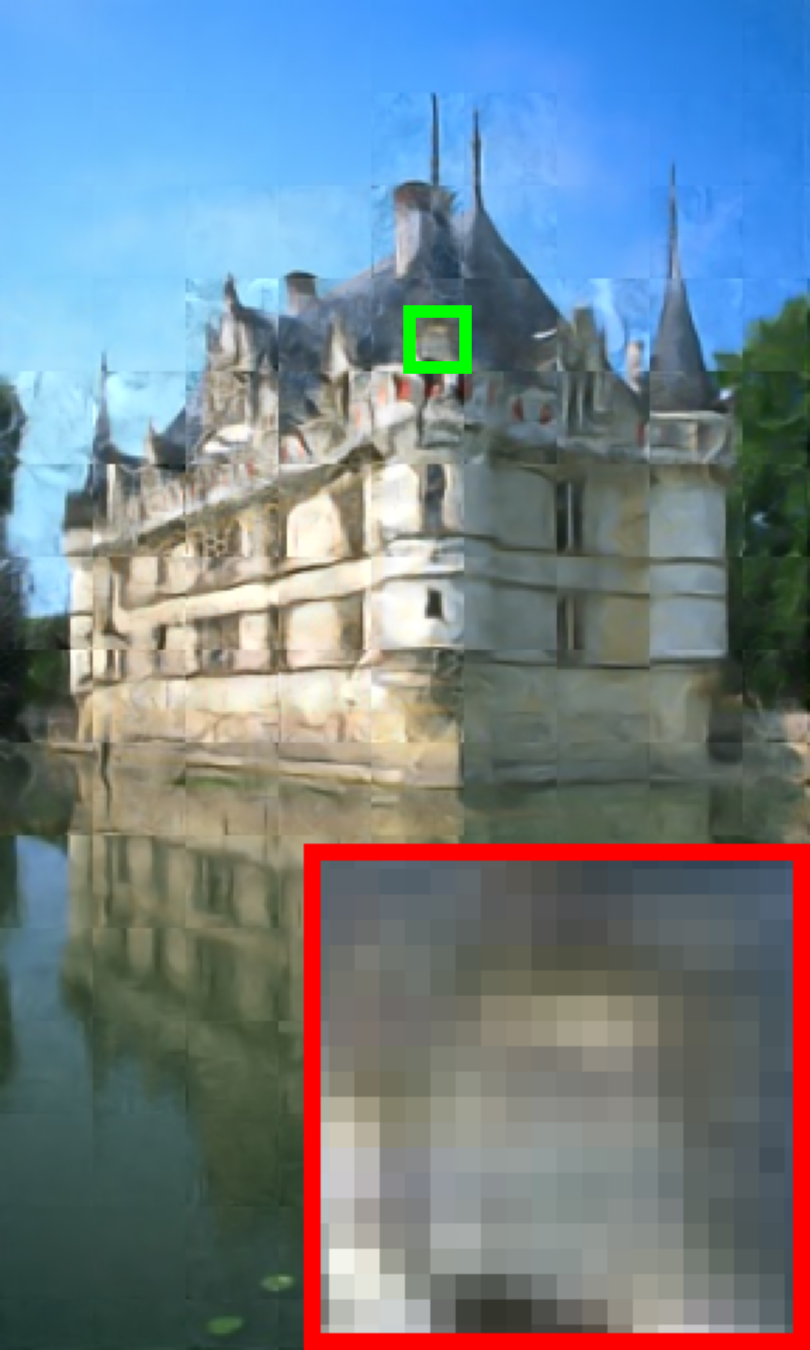}
    &\includegraphics[width=0.07\textwidth]{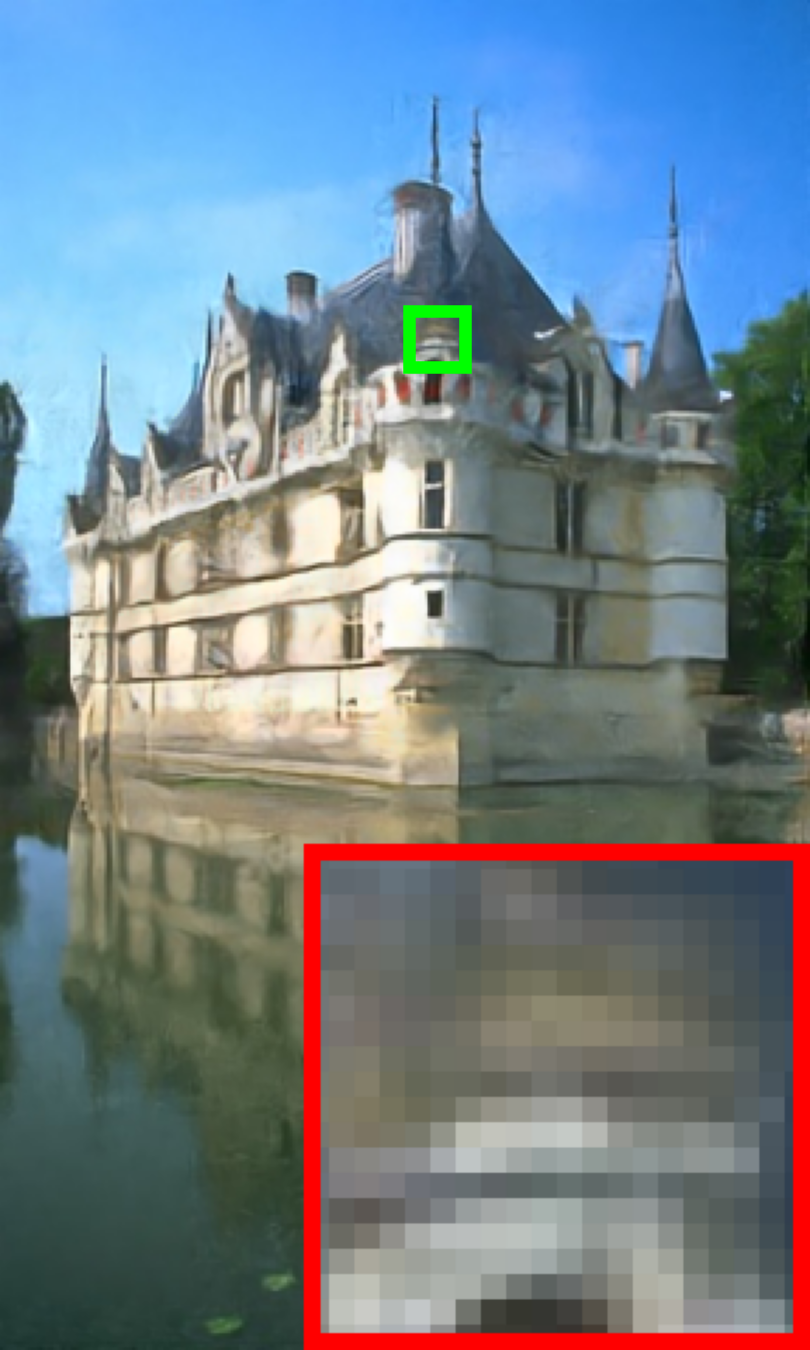}
    &\includegraphics[width=0.07\textwidth]{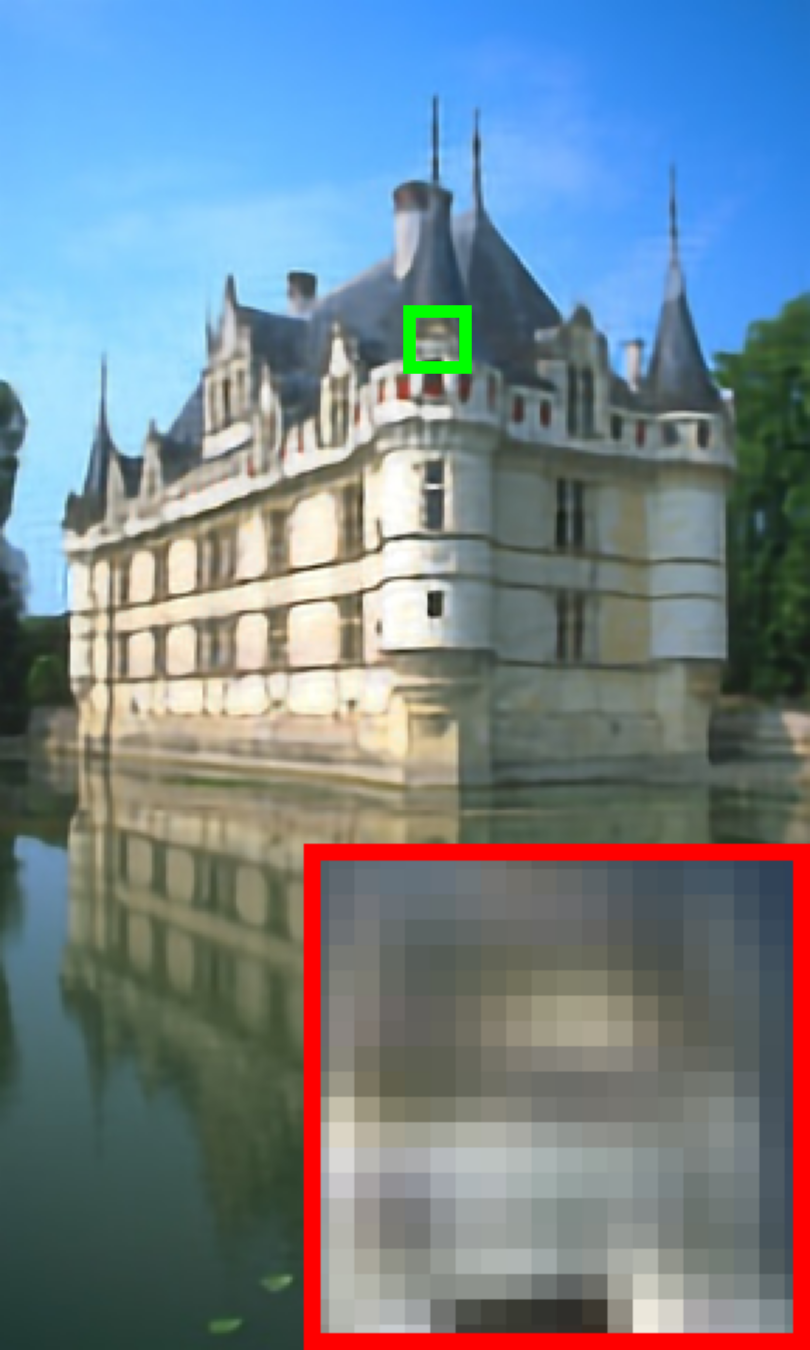}
    &\includegraphics[width=0.07\textwidth]{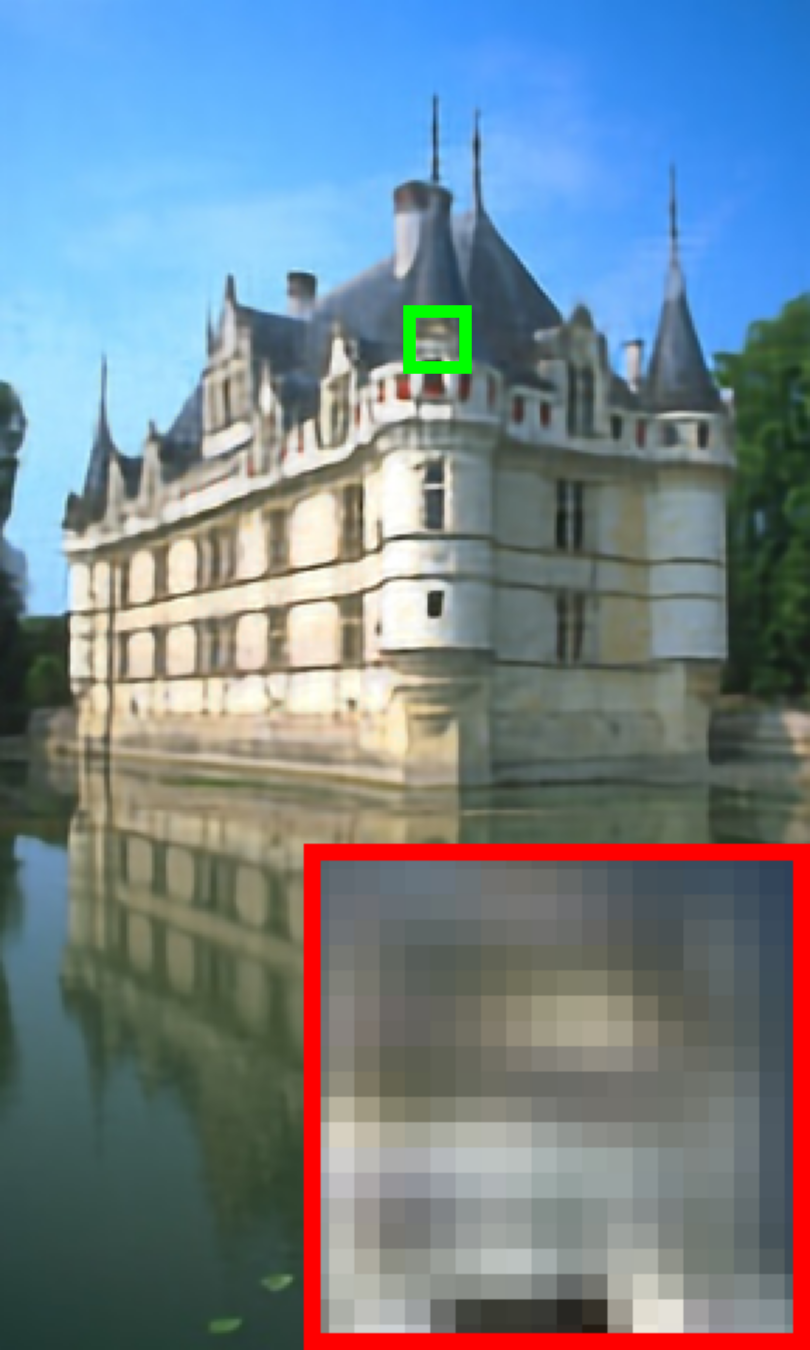}
    &\includegraphics[width=0.07\textwidth]{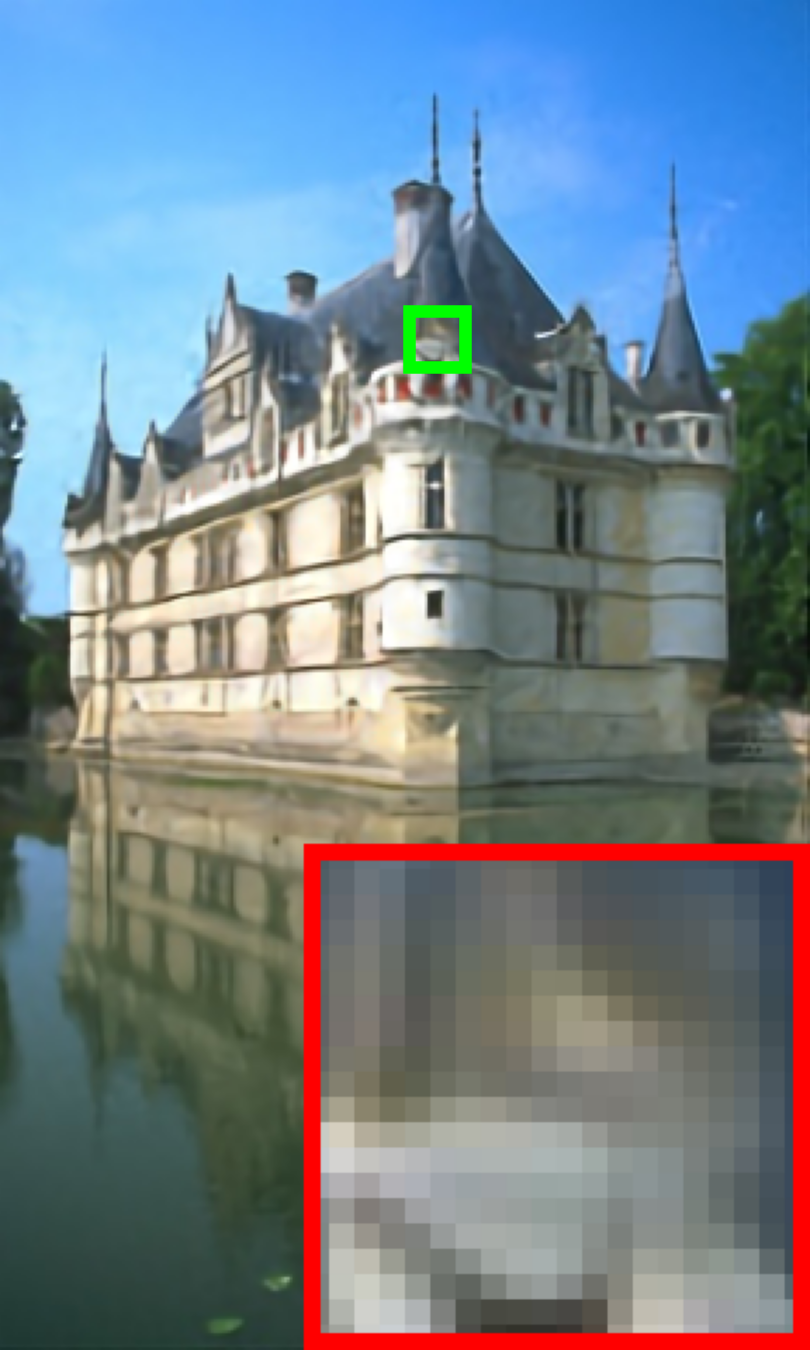}
    &\includegraphics[width=0.07\textwidth]{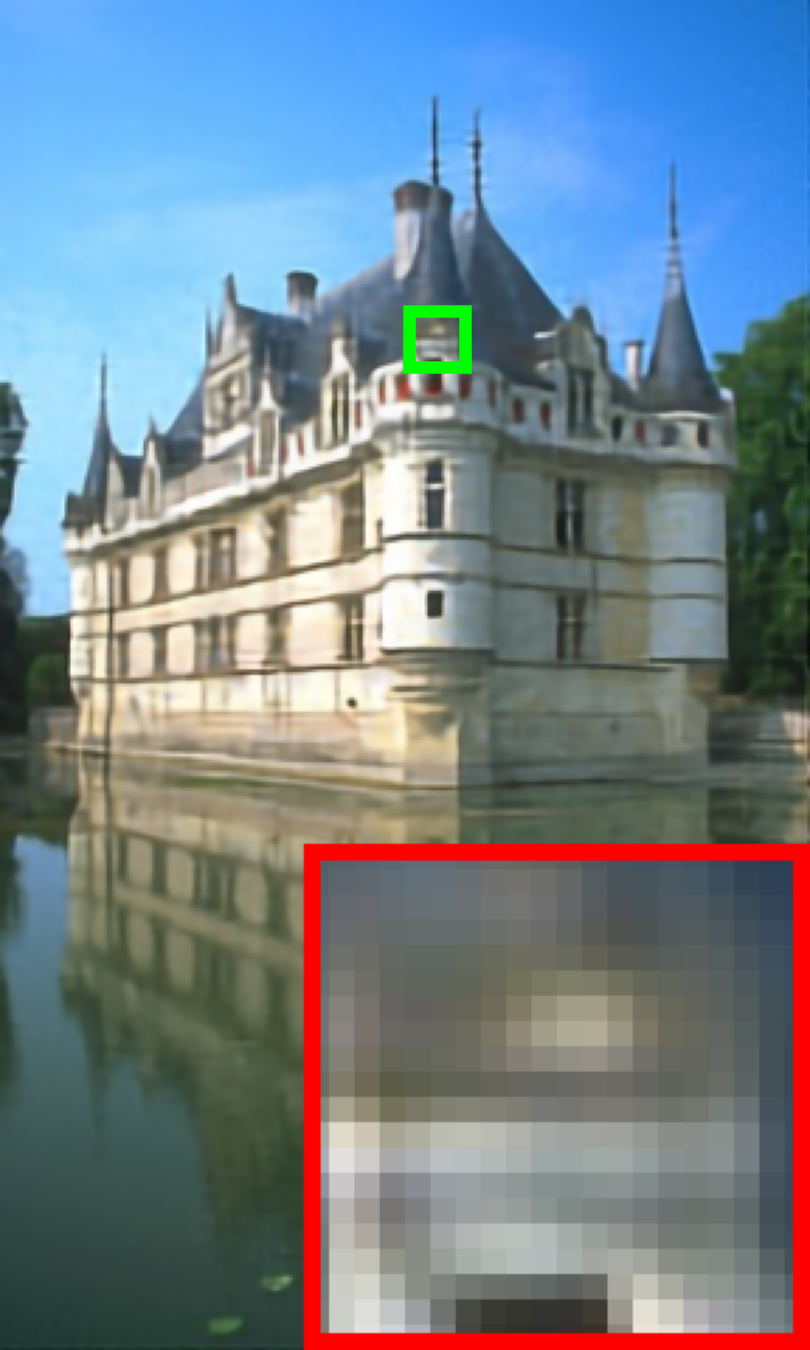}
    &\includegraphics[width=0.07\textwidth]{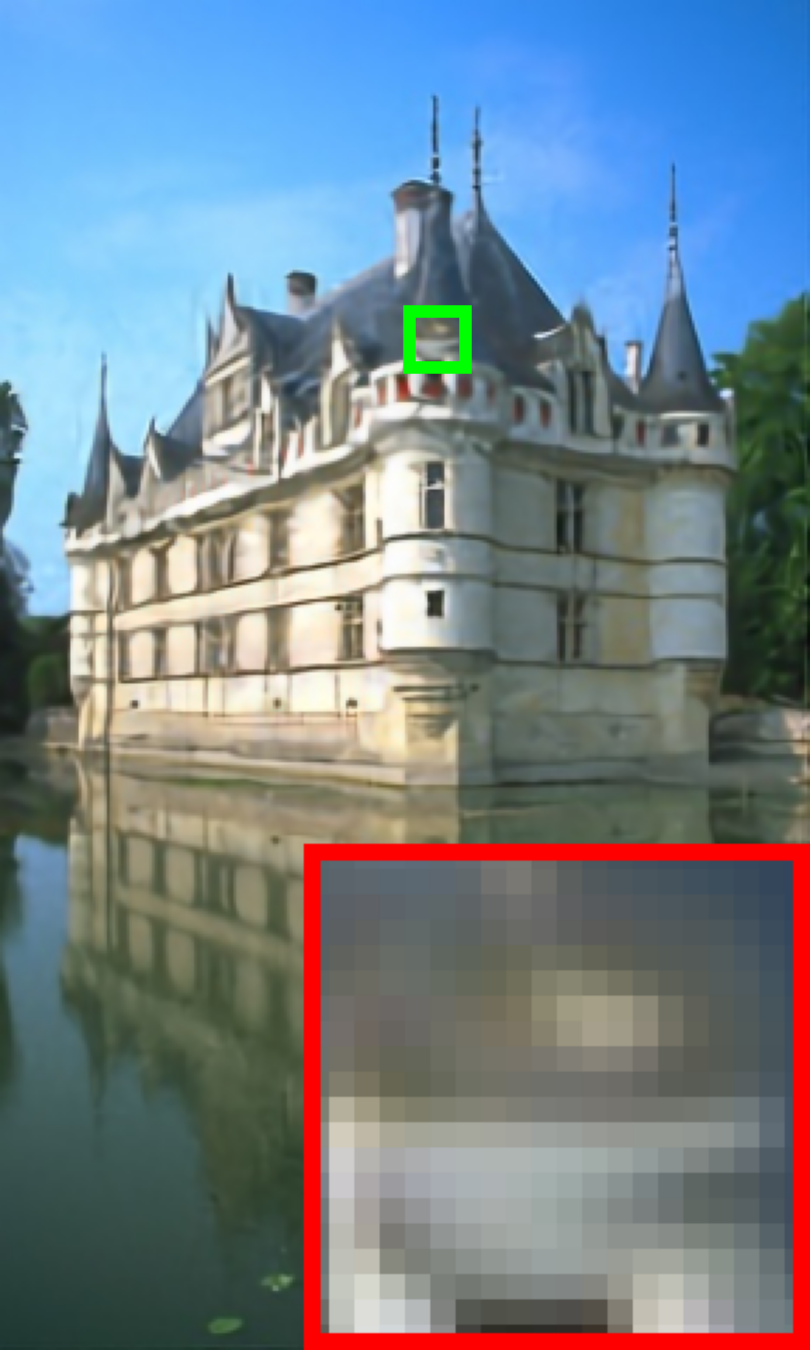}
    &\includegraphics[width=0.07\textwidth]{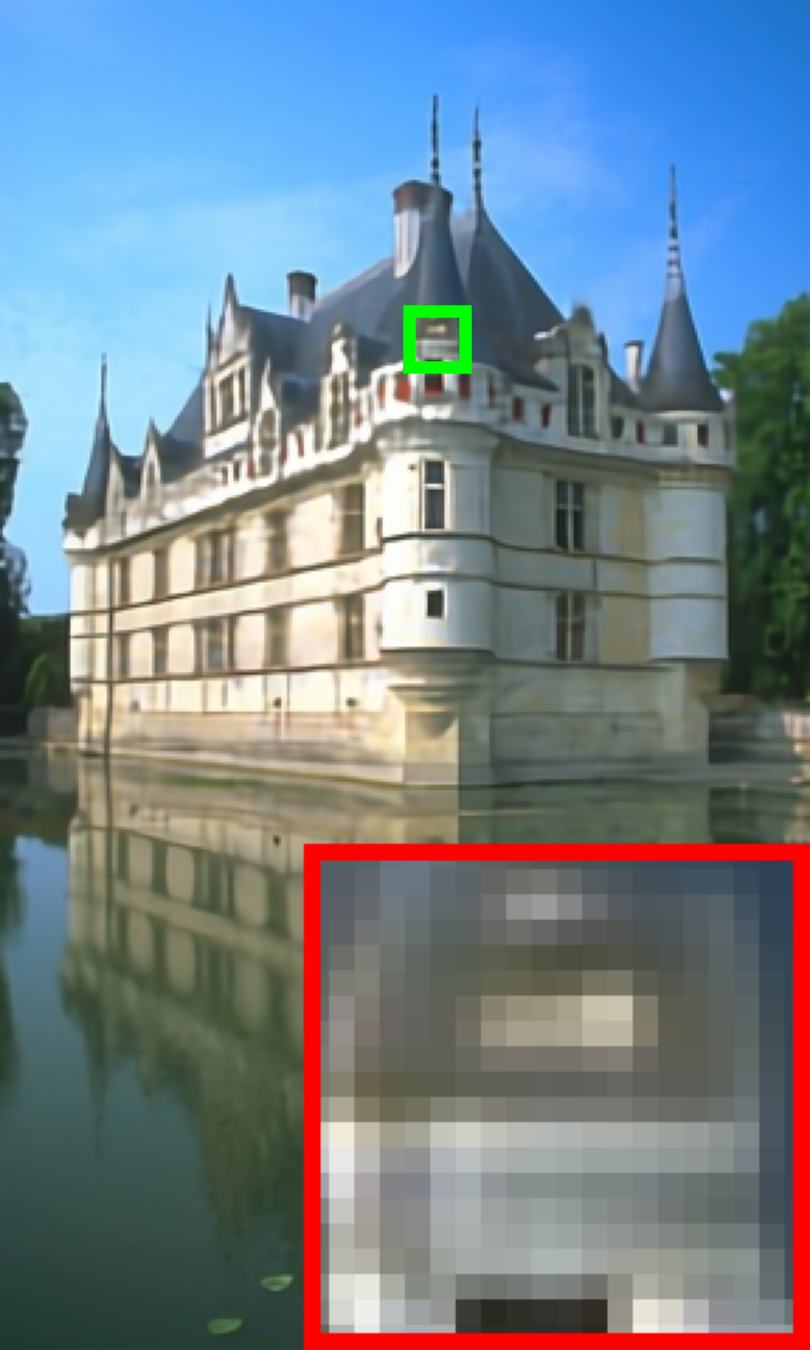}
    &\includegraphics[width=0.07\textwidth]{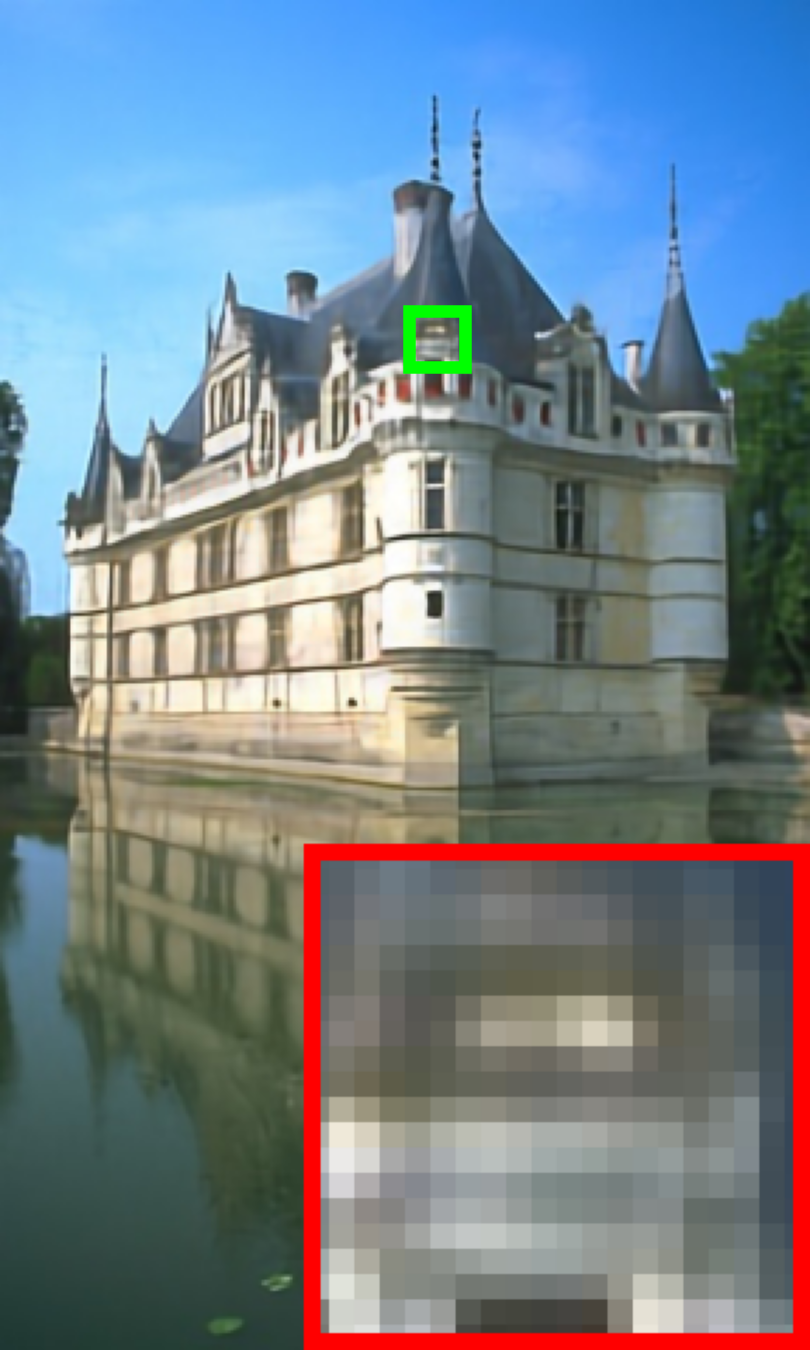}
    &\includegraphics[width=0.07\textwidth]{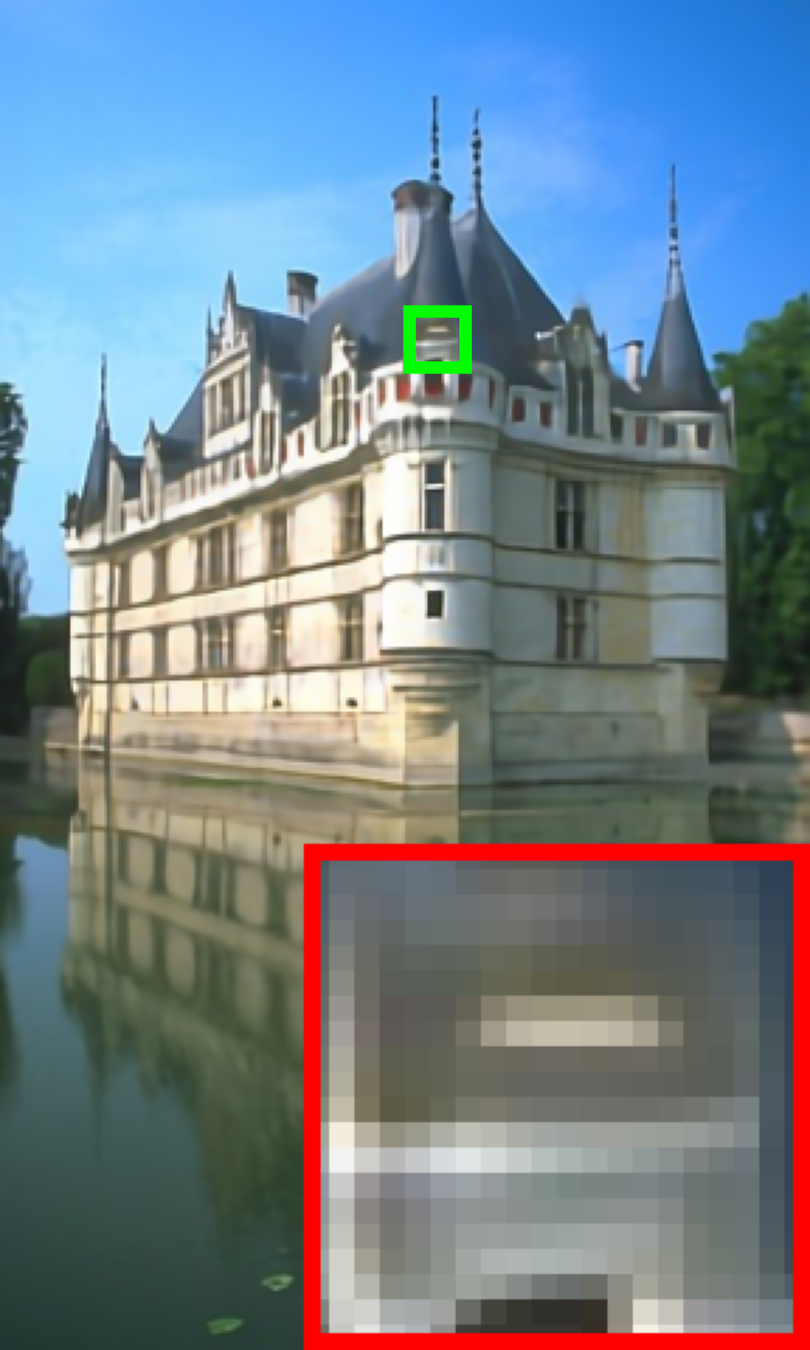}
    &\includegraphics[width=0.07\textwidth]{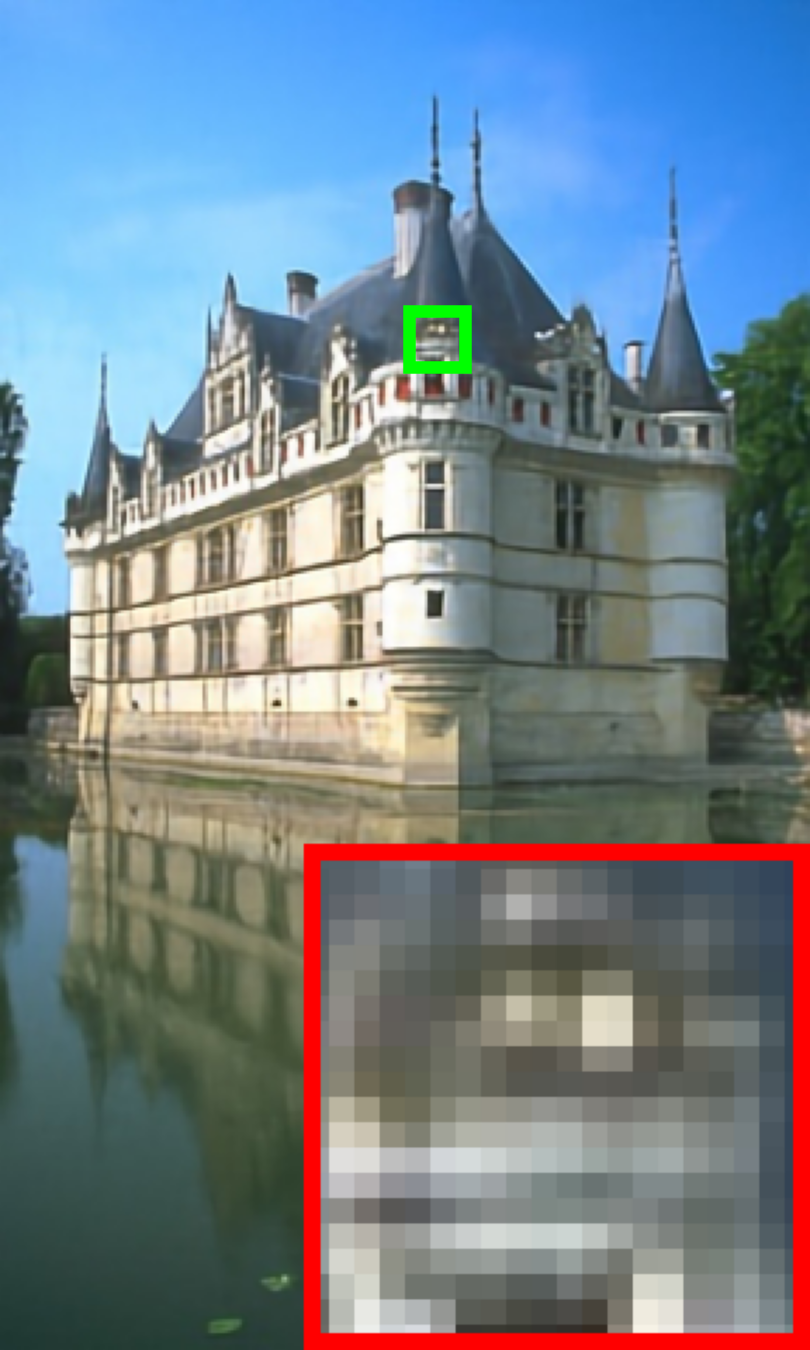}
    &\includegraphics[width=0.07\textwidth]{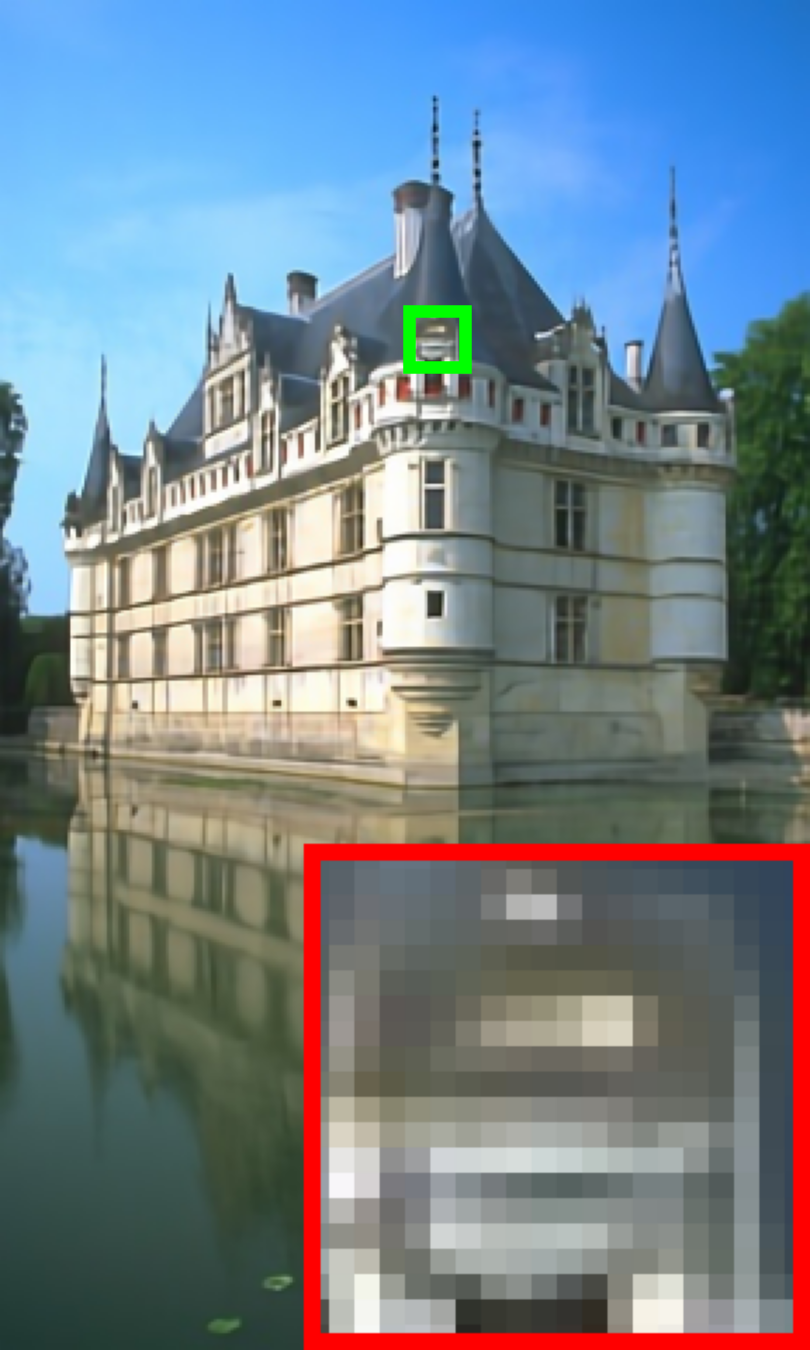}\\
    PSNR/SSIM & 22.97/0.7101 & 23.79/0.7647 & 24.96/0.8084 & 26.09/0.8369 & 26.15/0.8384 & 26.56/0.8544 & 26.73/0.8483 & 26.48/0.8564 & 27.38/0.8747 & 27.47/0.8766 & \textcolor{green}{27.84}/\textcolor{green}{0.8787} & \textcolor{blue}{29.32}/\textcolor{blue}{0.8905} & \textcolor{red}{29.94}/\textcolor{red}{0.8980}\\
    \includegraphics[width=0.07\textwidth]{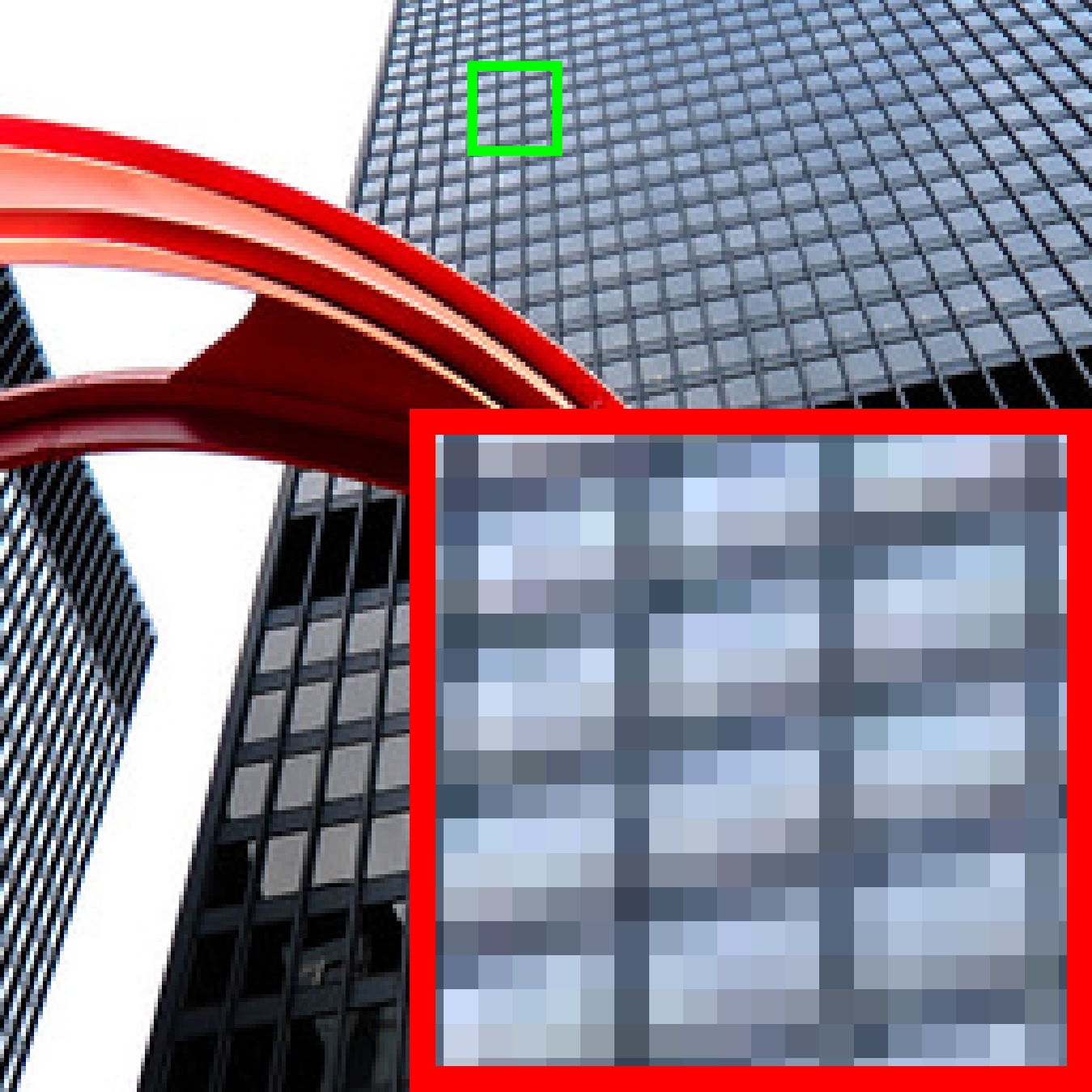}
    &\includegraphics[width=0.07\textwidth]{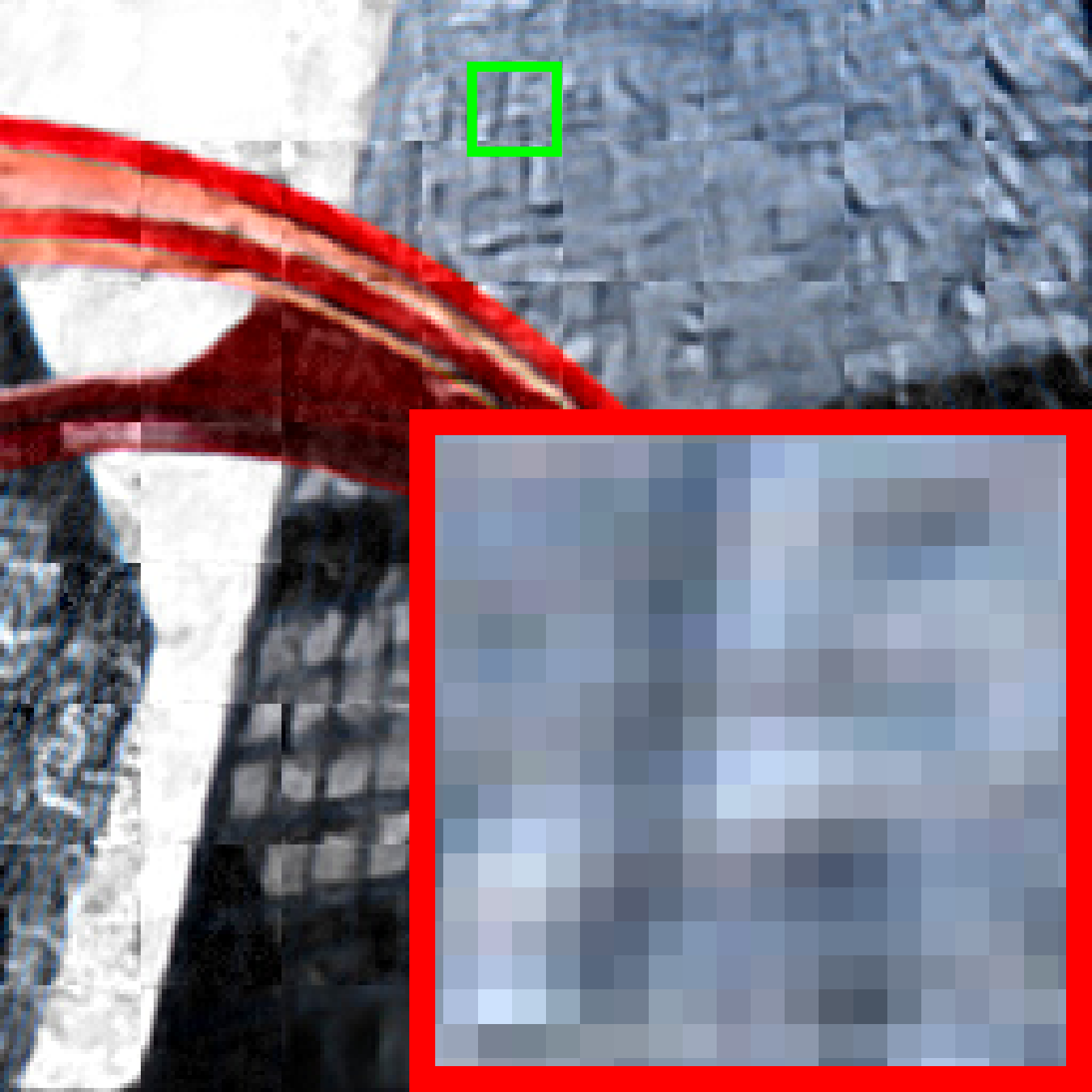}
    &\includegraphics[width=0.07\textwidth]{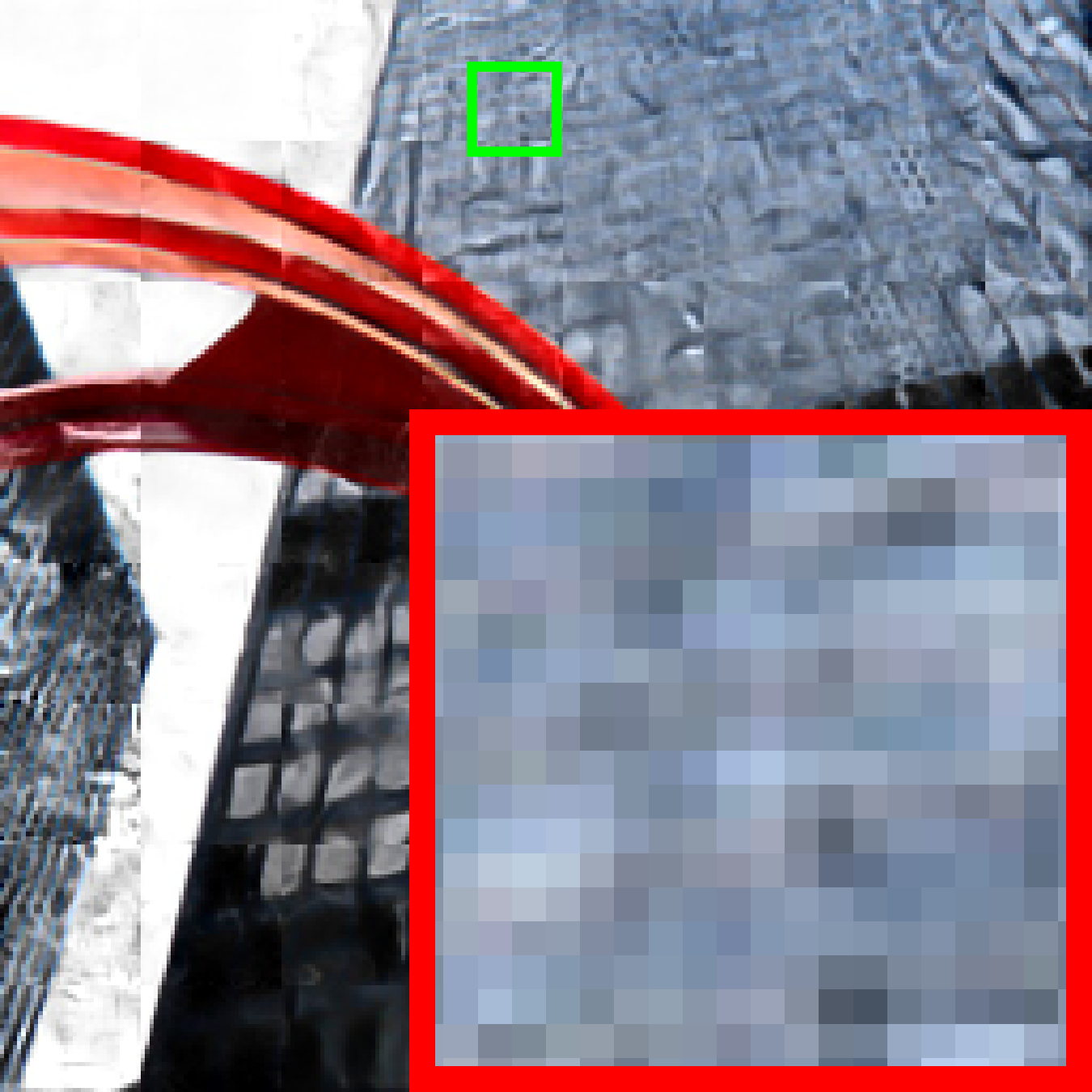}
    &\includegraphics[width=0.07\textwidth]{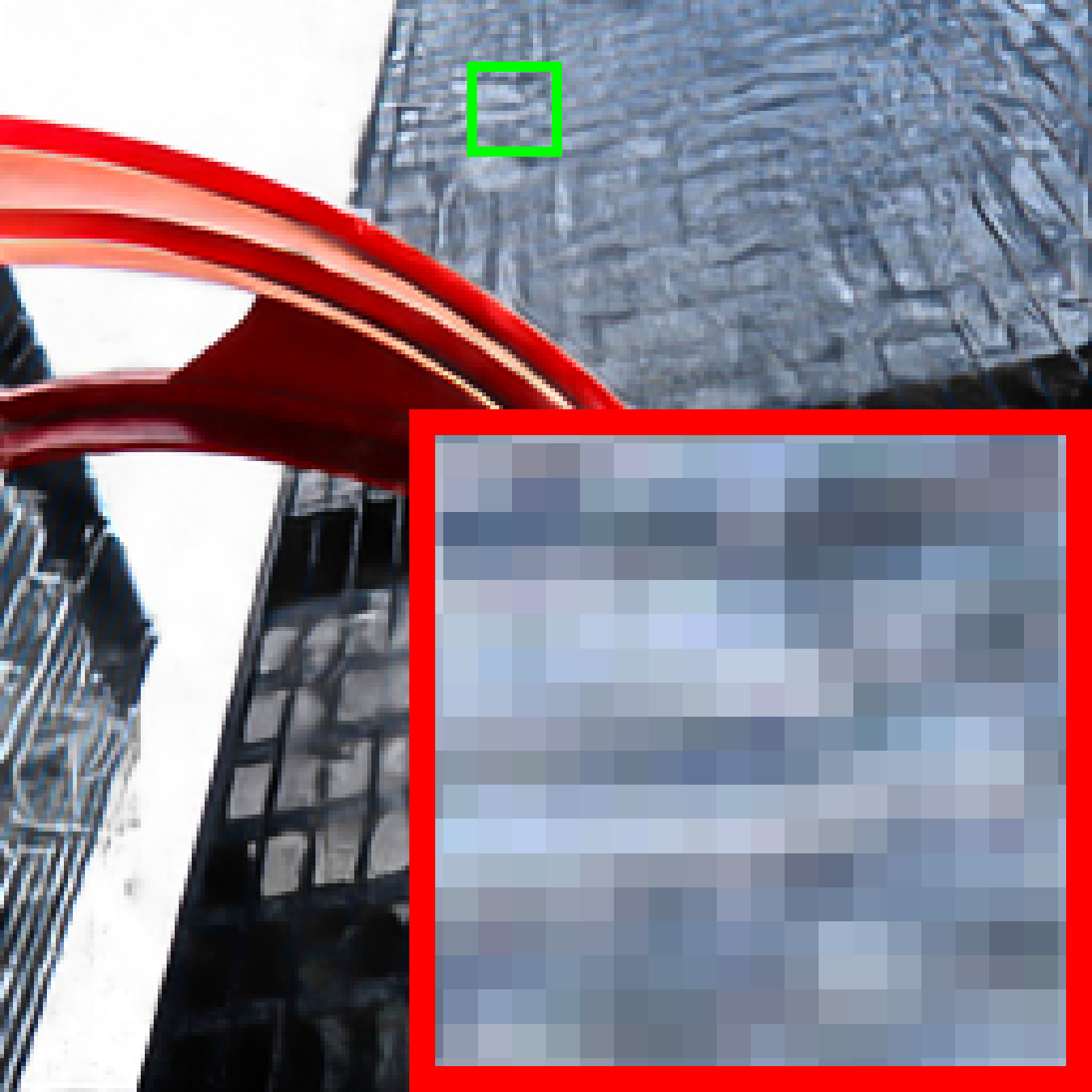}
    &\includegraphics[width=0.07\textwidth]{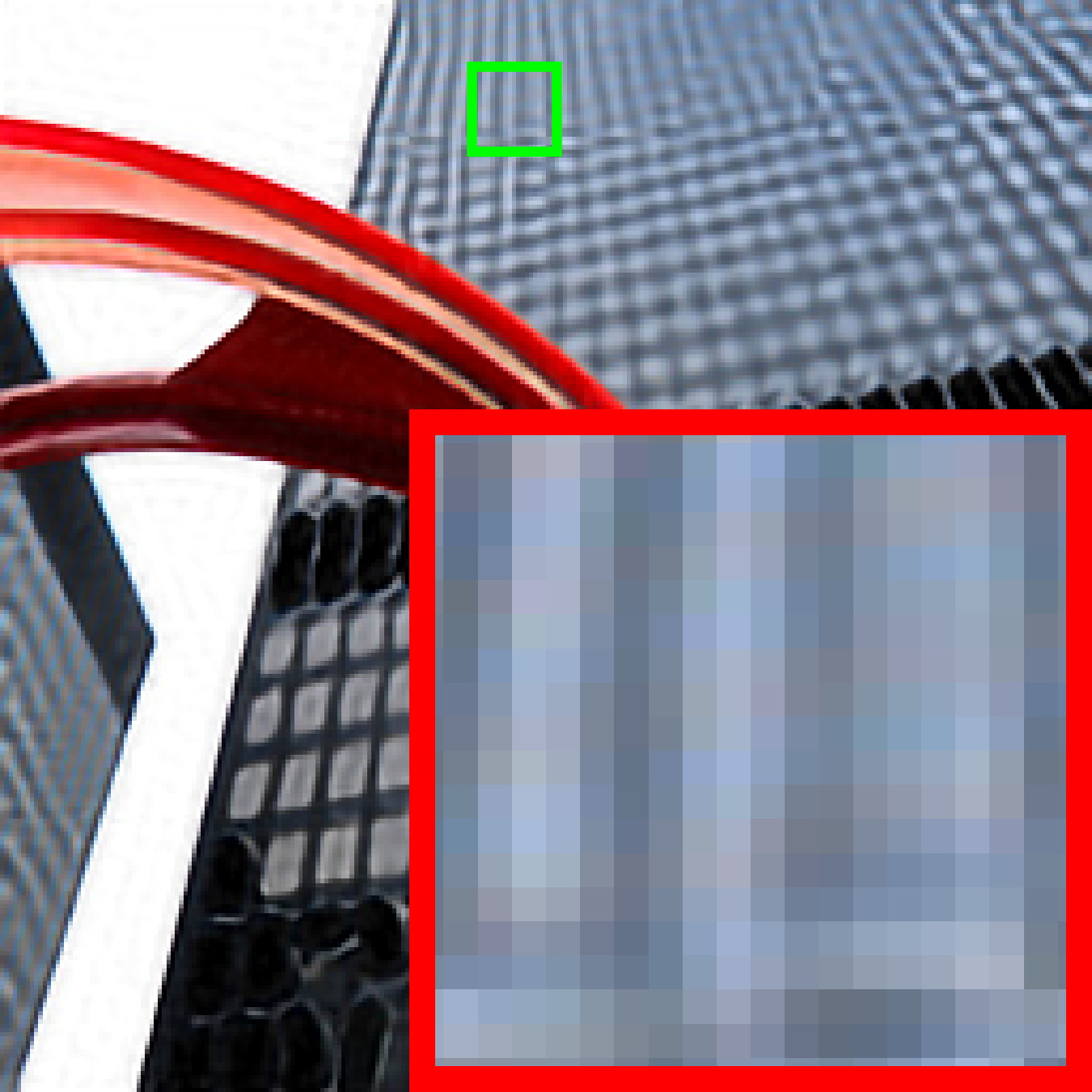}
    &\includegraphics[width=0.07\textwidth]{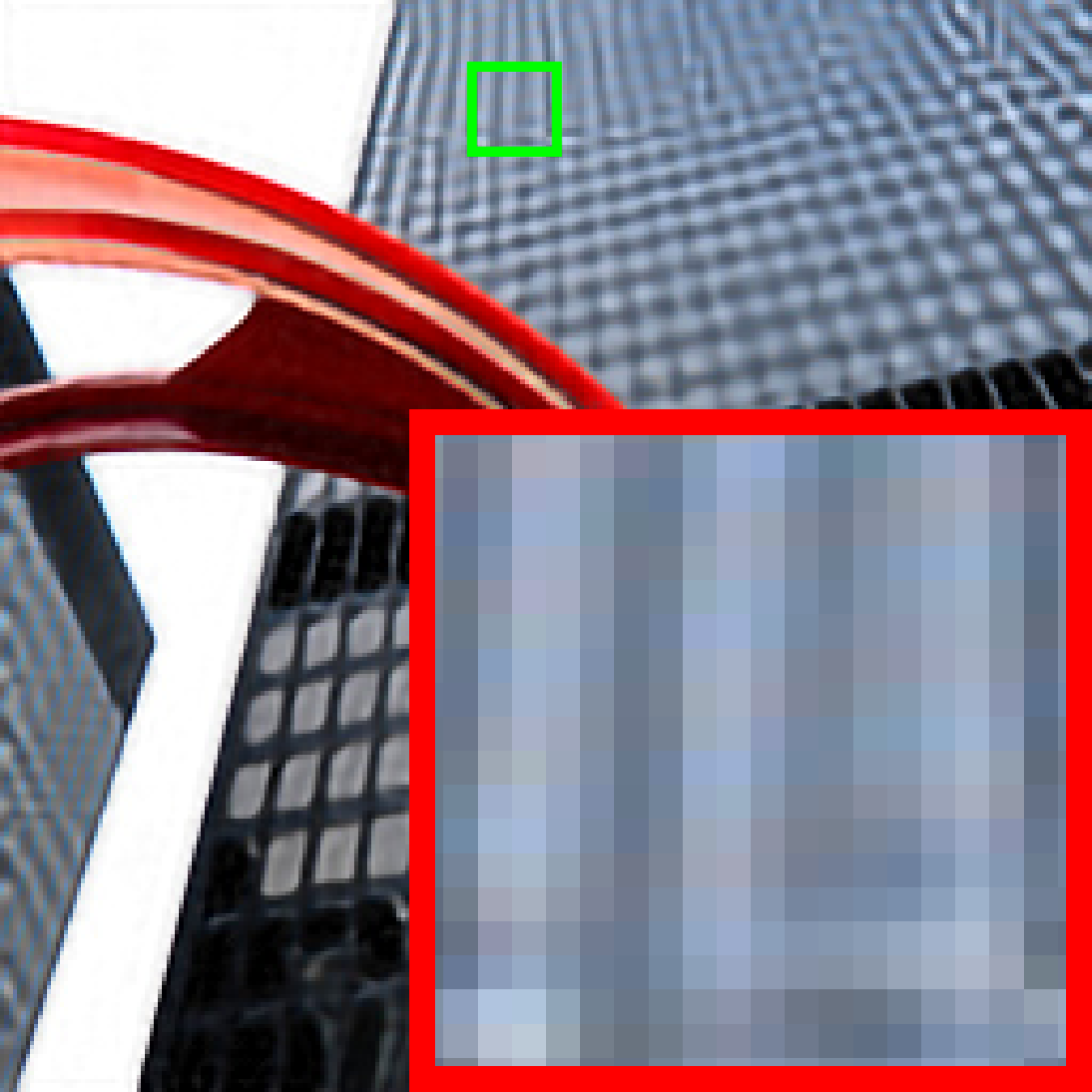}
    &\includegraphics[width=0.07\textwidth]{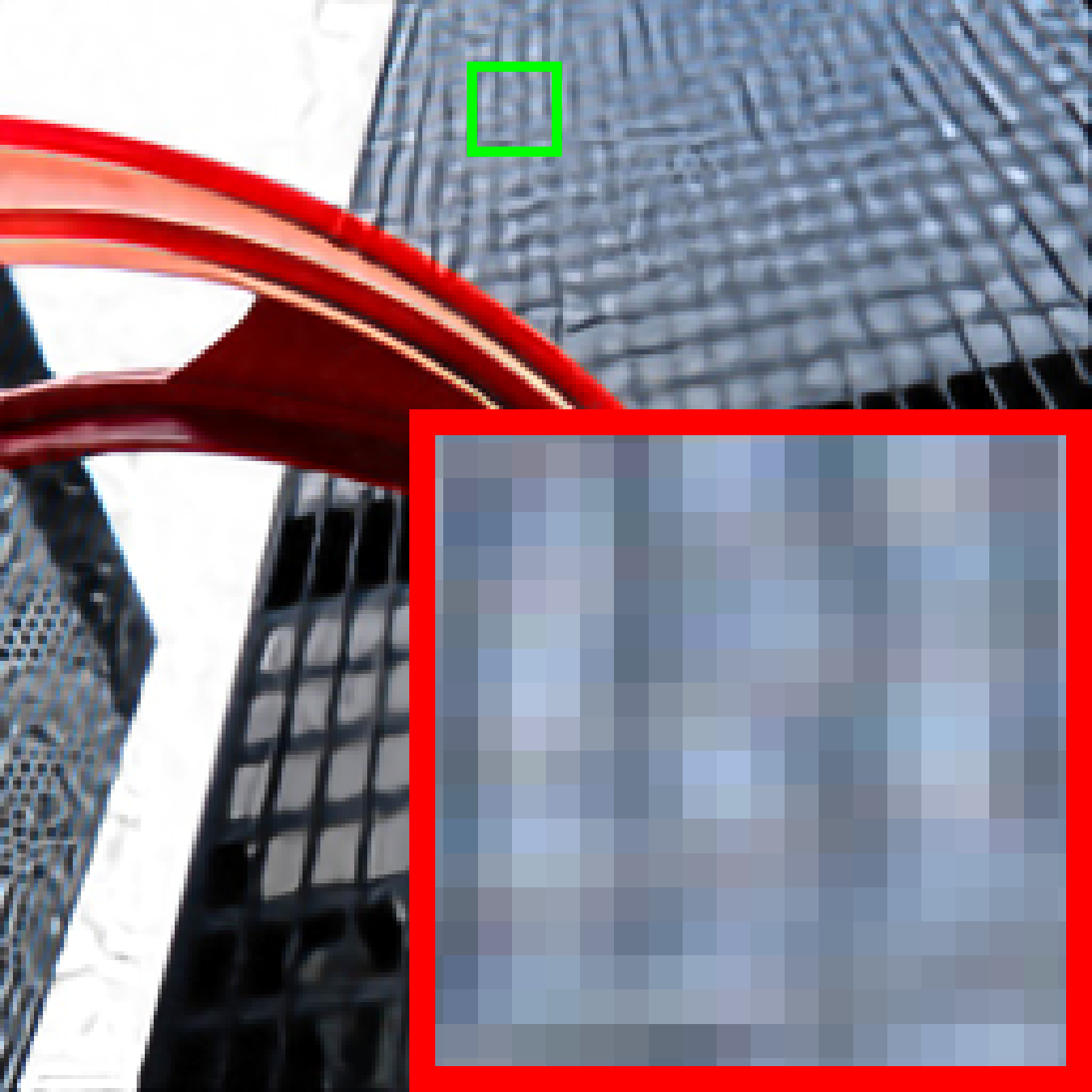}
    &\includegraphics[width=0.07\textwidth]{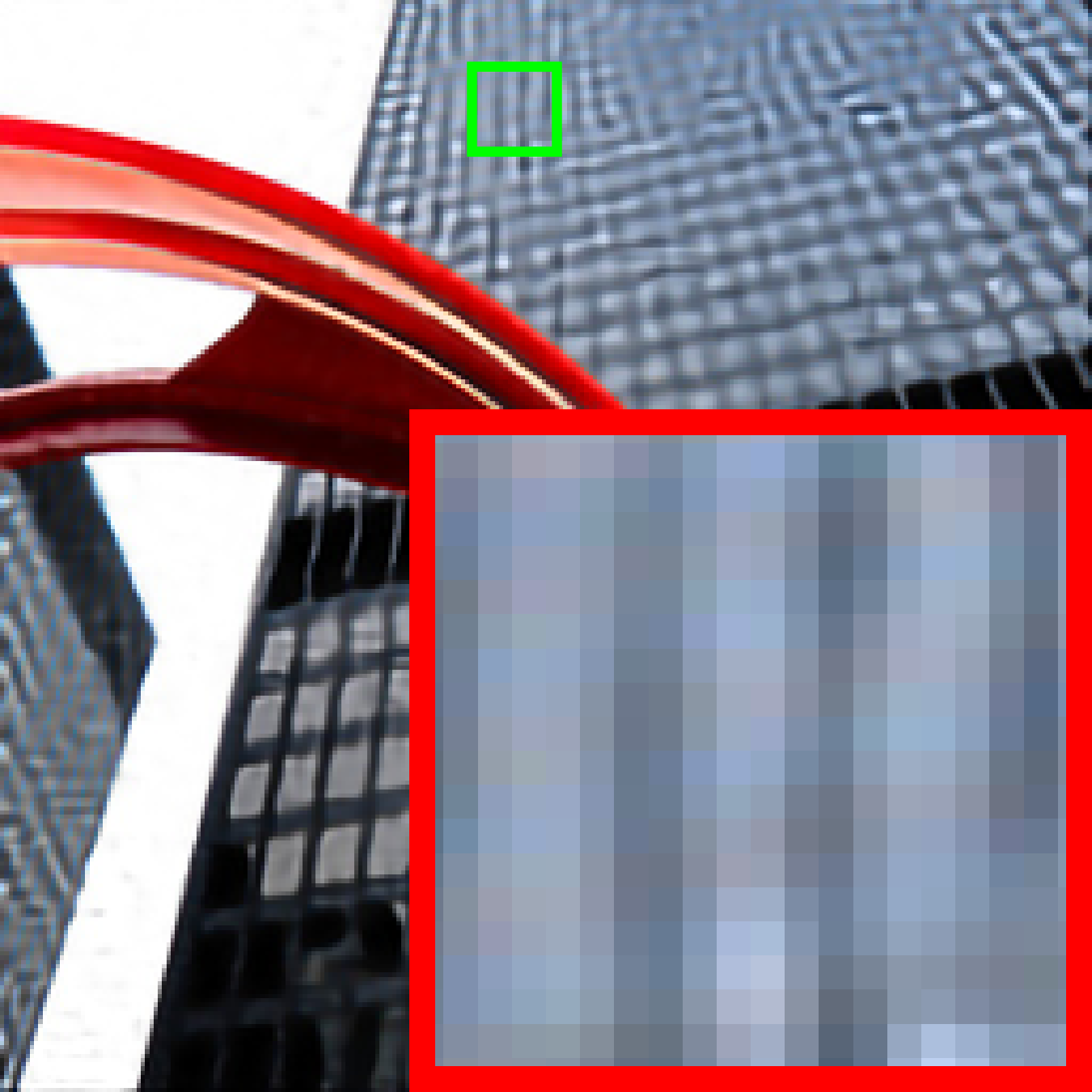}
    &\includegraphics[width=0.07\textwidth]{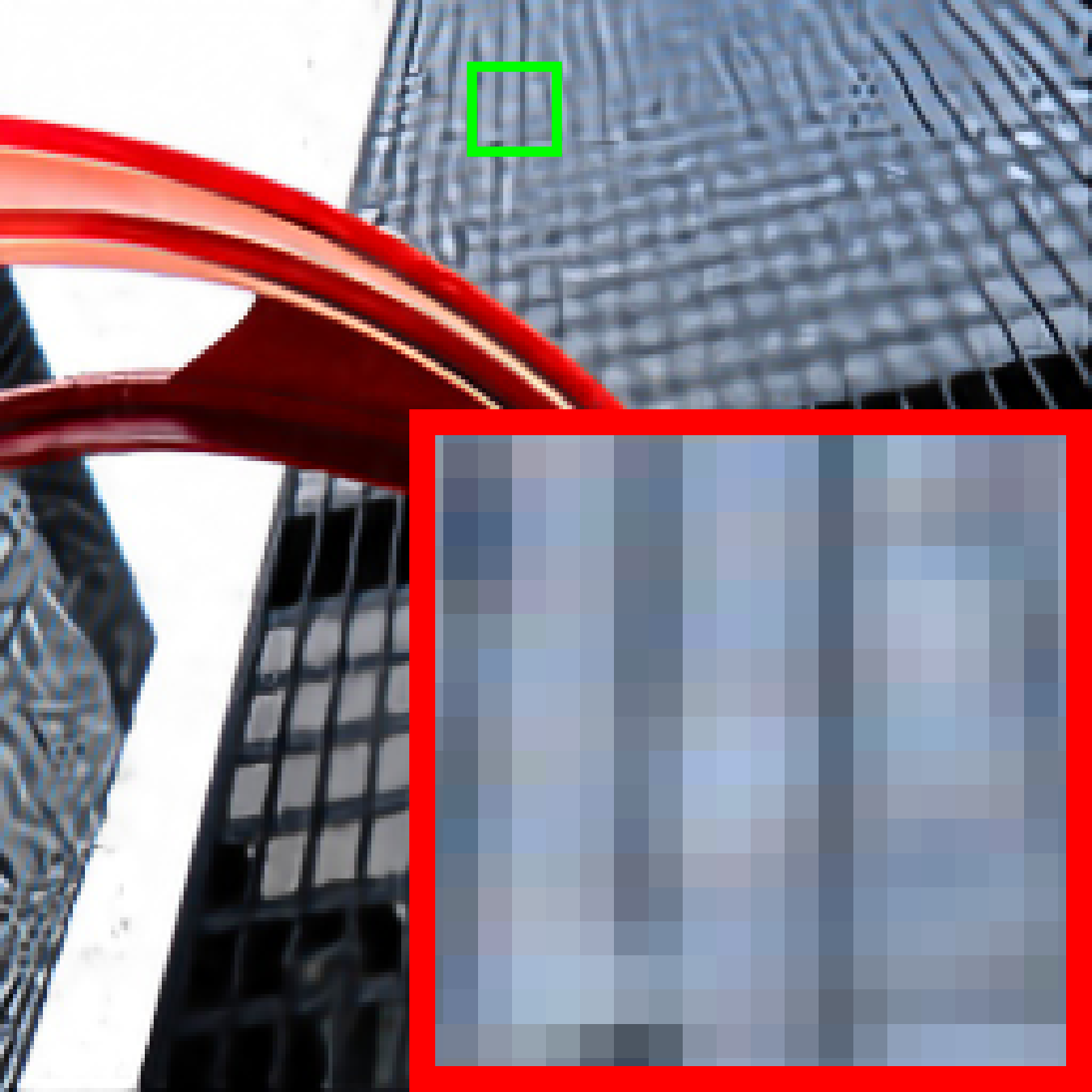}
    &\includegraphics[width=0.07\textwidth]{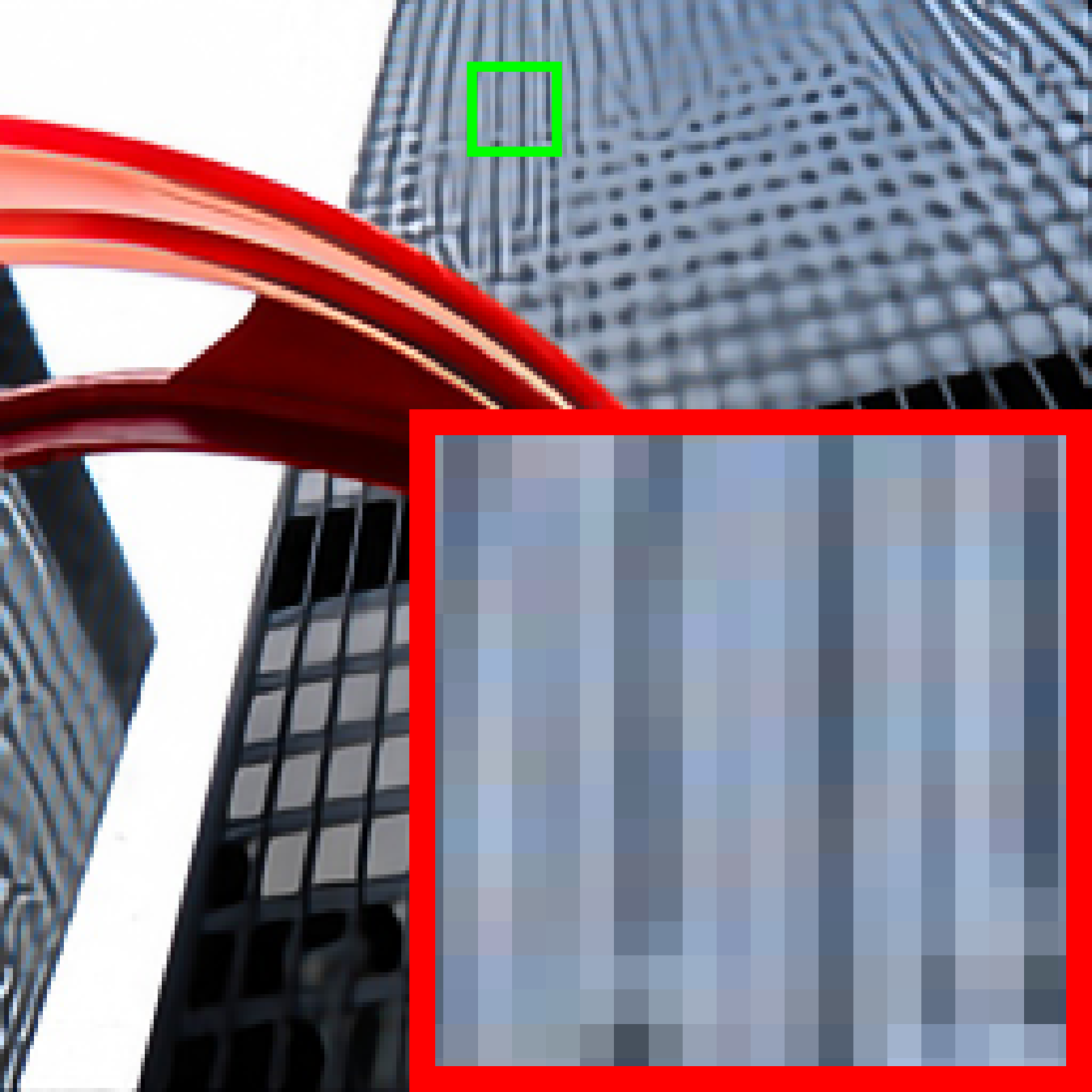}
    &\includegraphics[width=0.07\textwidth]{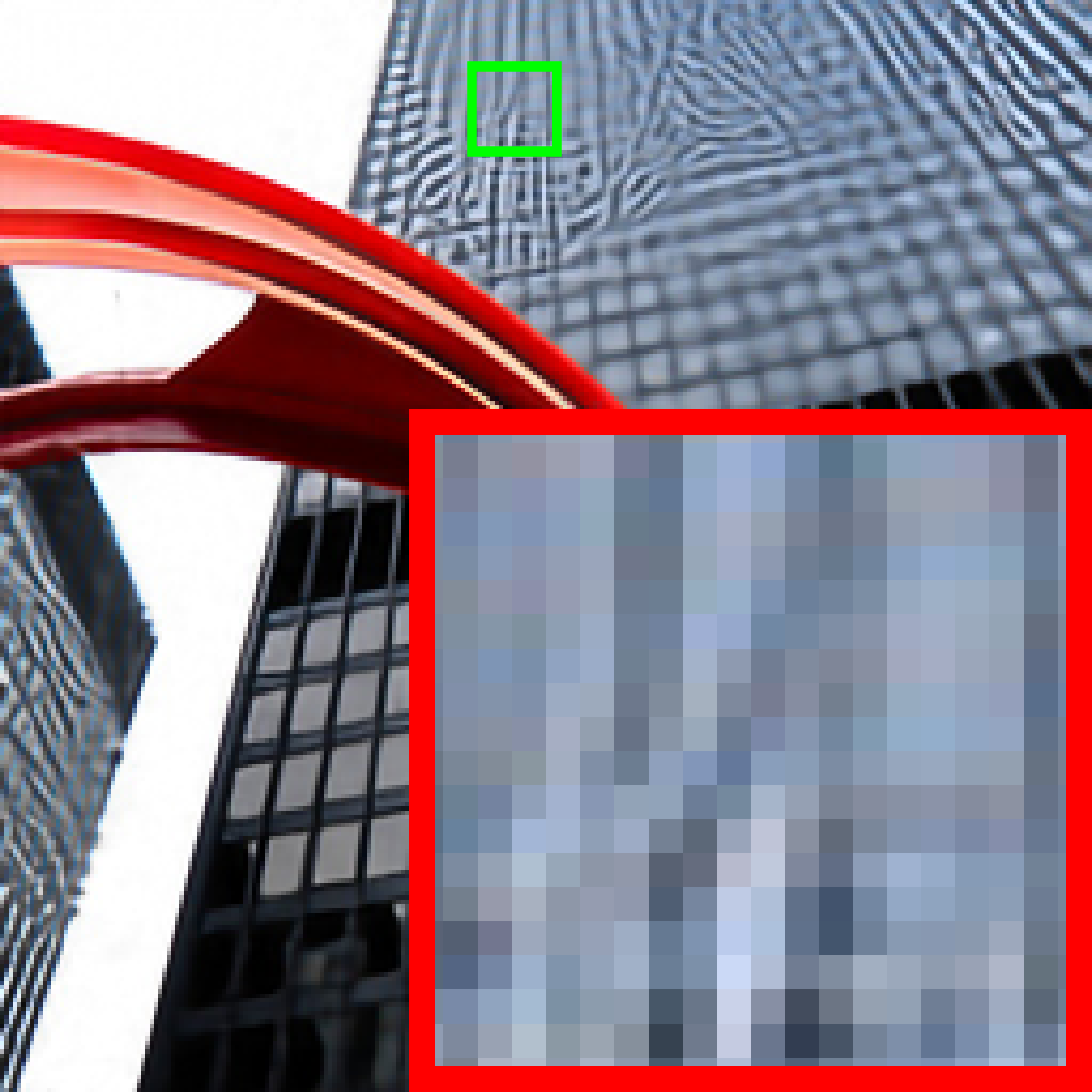}
    &\includegraphics[width=0.07\textwidth]{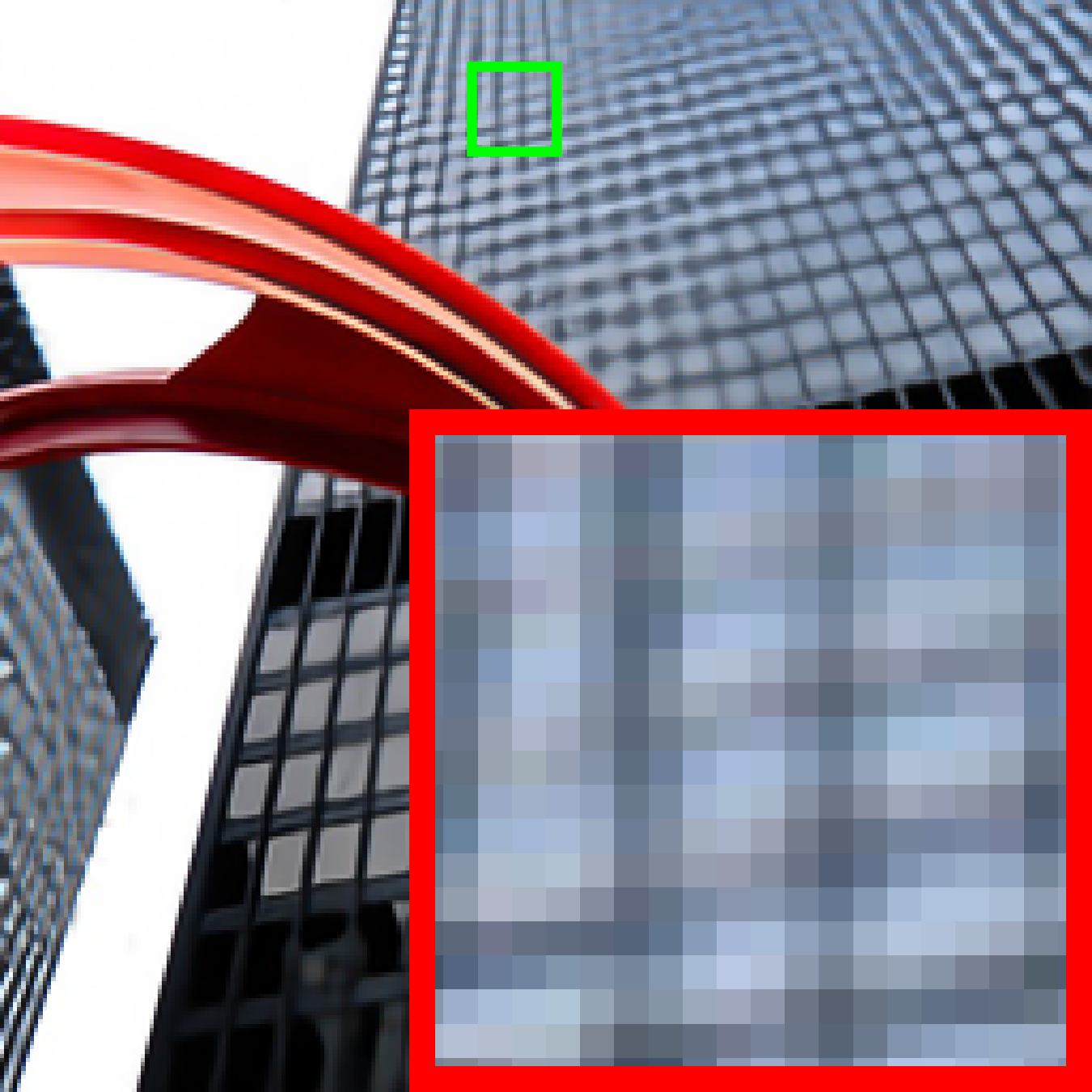}
    &\includegraphics[width=0.07\textwidth]{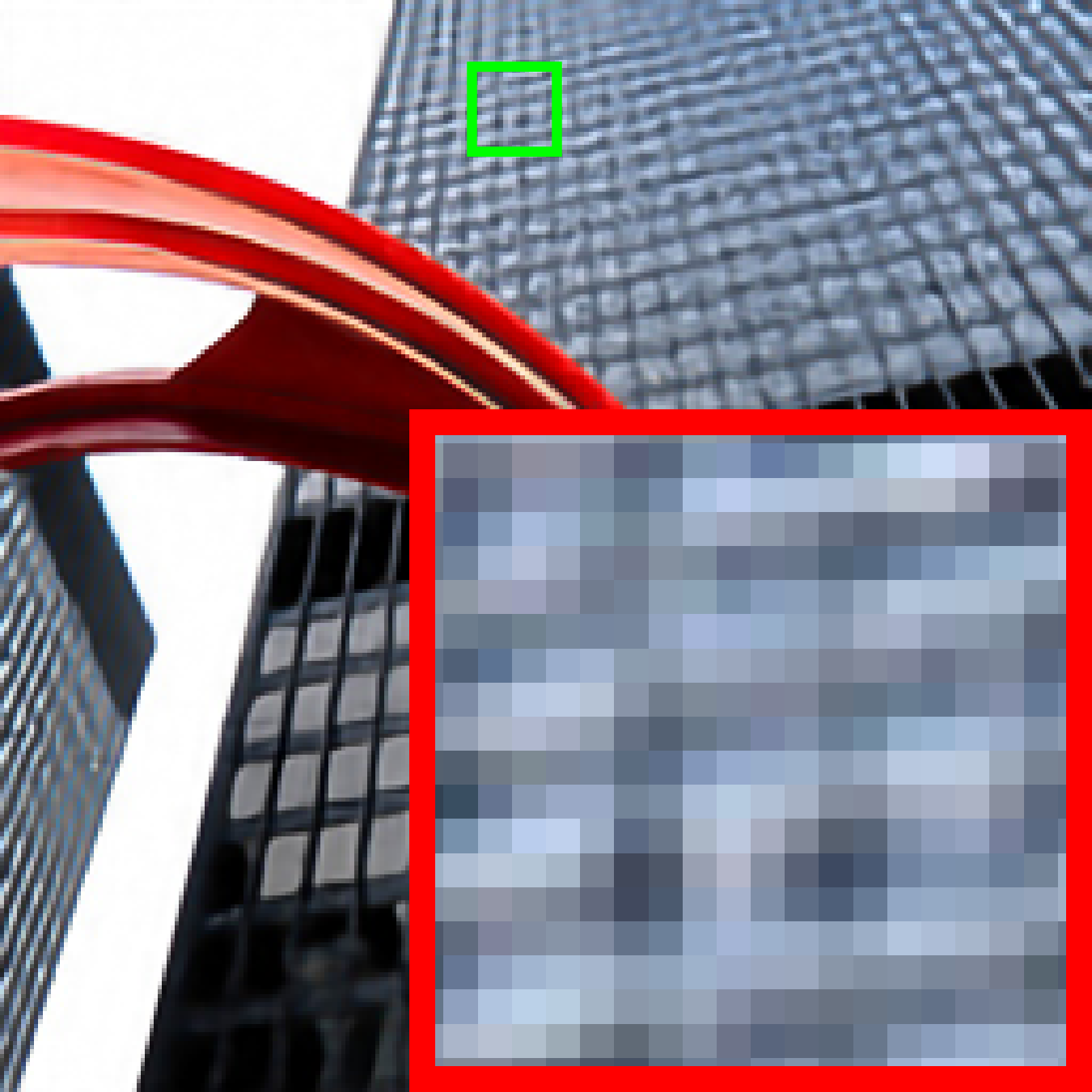}
    &\includegraphics[width=0.07\textwidth]{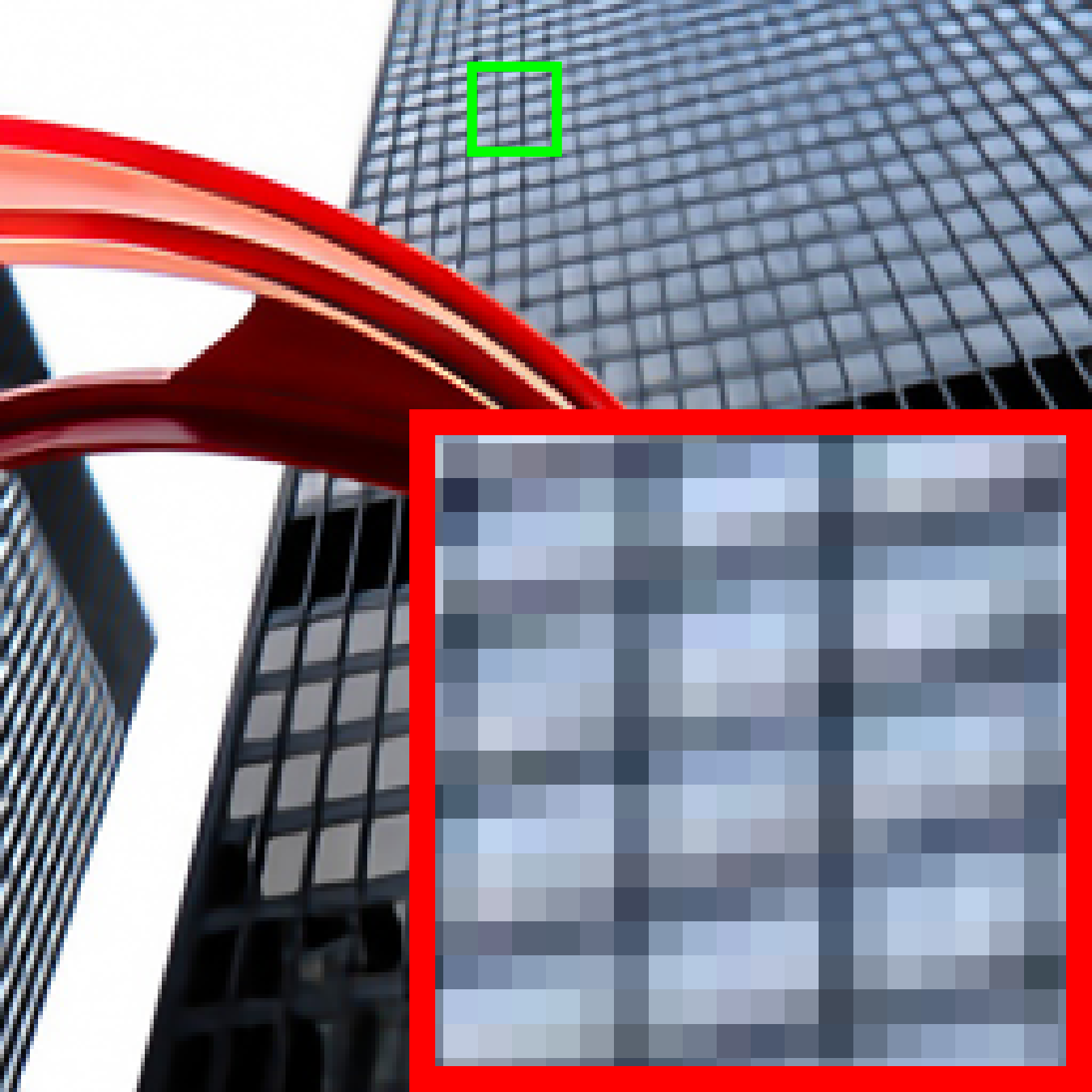}\\
    PSNR/SSIM & 17.09/0.4985 & 18.18/0.5966 & 19.39/0.6898 & 19.54/0.7084 & 19.70/0.7240 & 20.23/0.7743 & 19.97/0.7566 & 20.43/0.7939 & 20.30/0.8132 & 20.34/0.8118 & \textcolor{green}{21.07}/\textcolor{green}{0.8207} & \textcolor{blue}{21.48}/\textcolor{blue}{0.8268} & \textcolor{red}{24.47}/\textcolor{red}{0.9092}\\
    \includegraphics[width=0.07\textwidth]{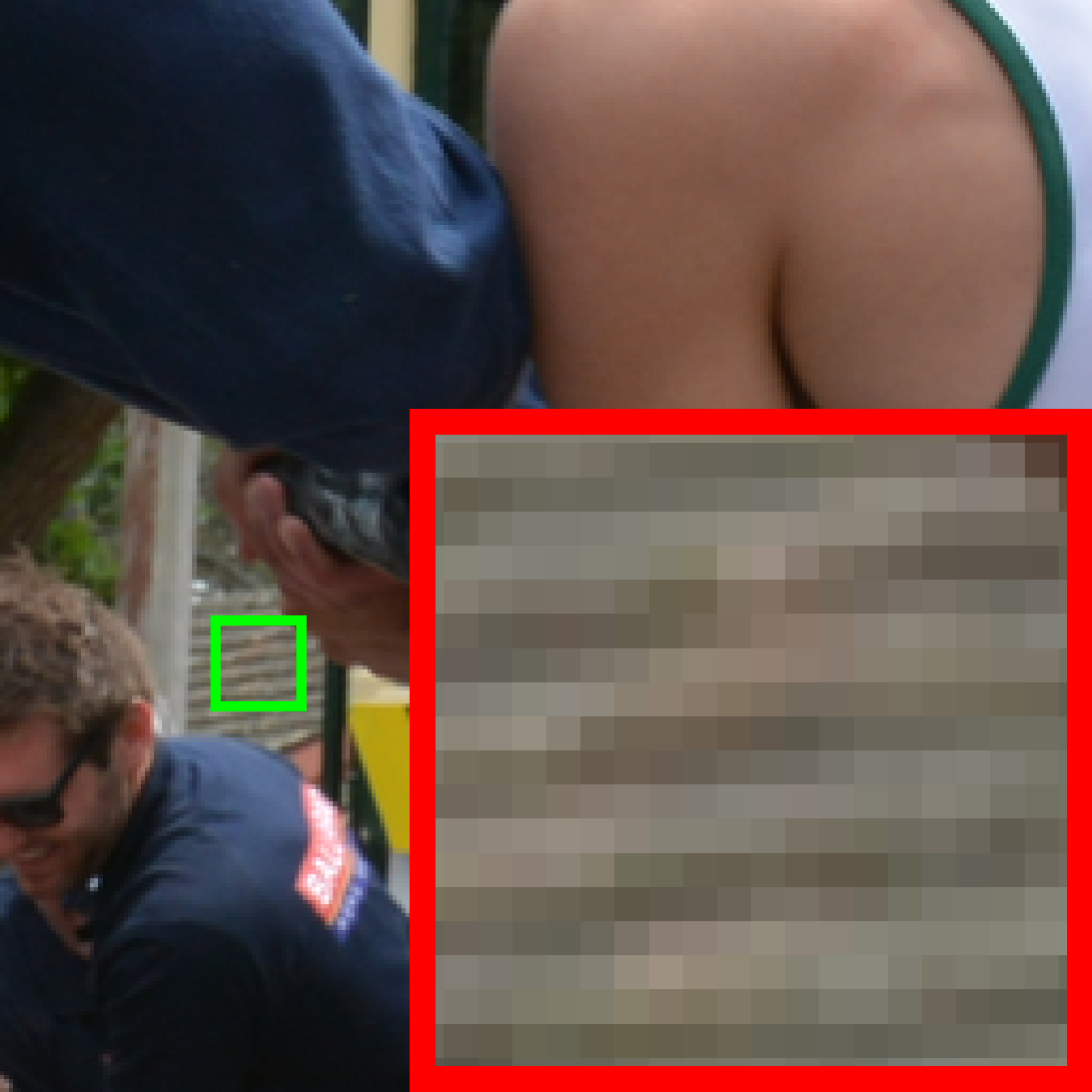}
    &\includegraphics[width=0.07\textwidth]{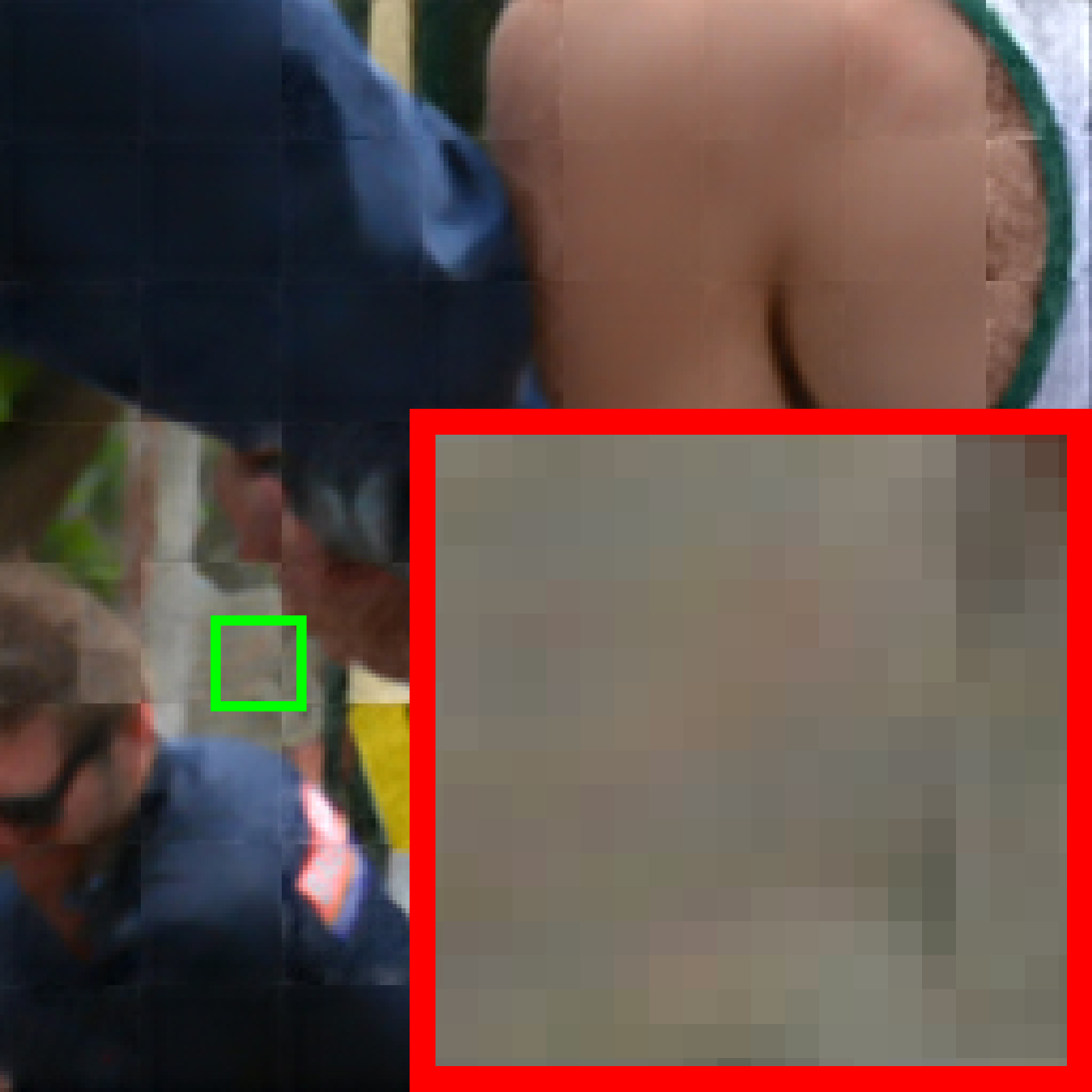}
    &\includegraphics[width=0.07\textwidth]{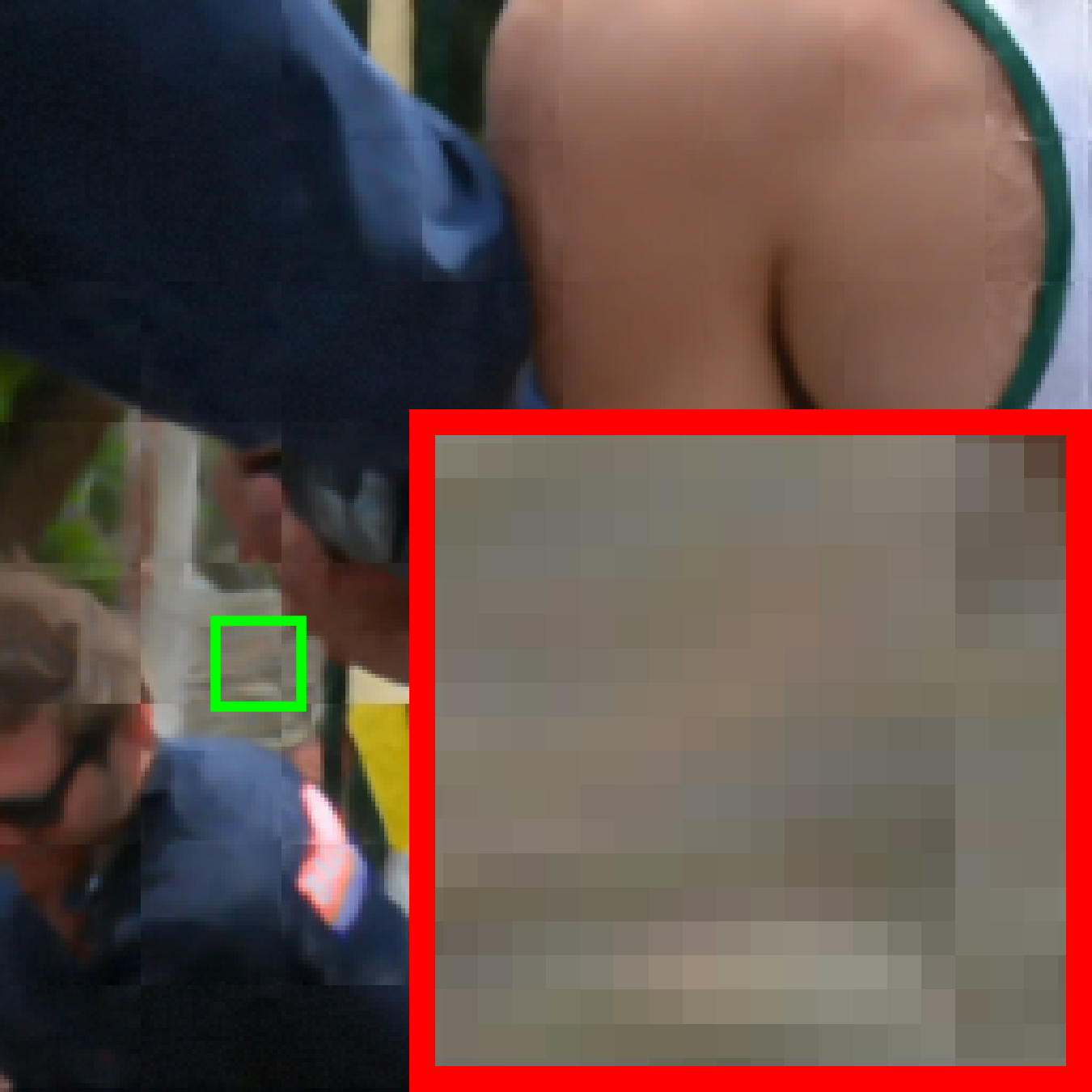}
    &\includegraphics[width=0.07\textwidth]{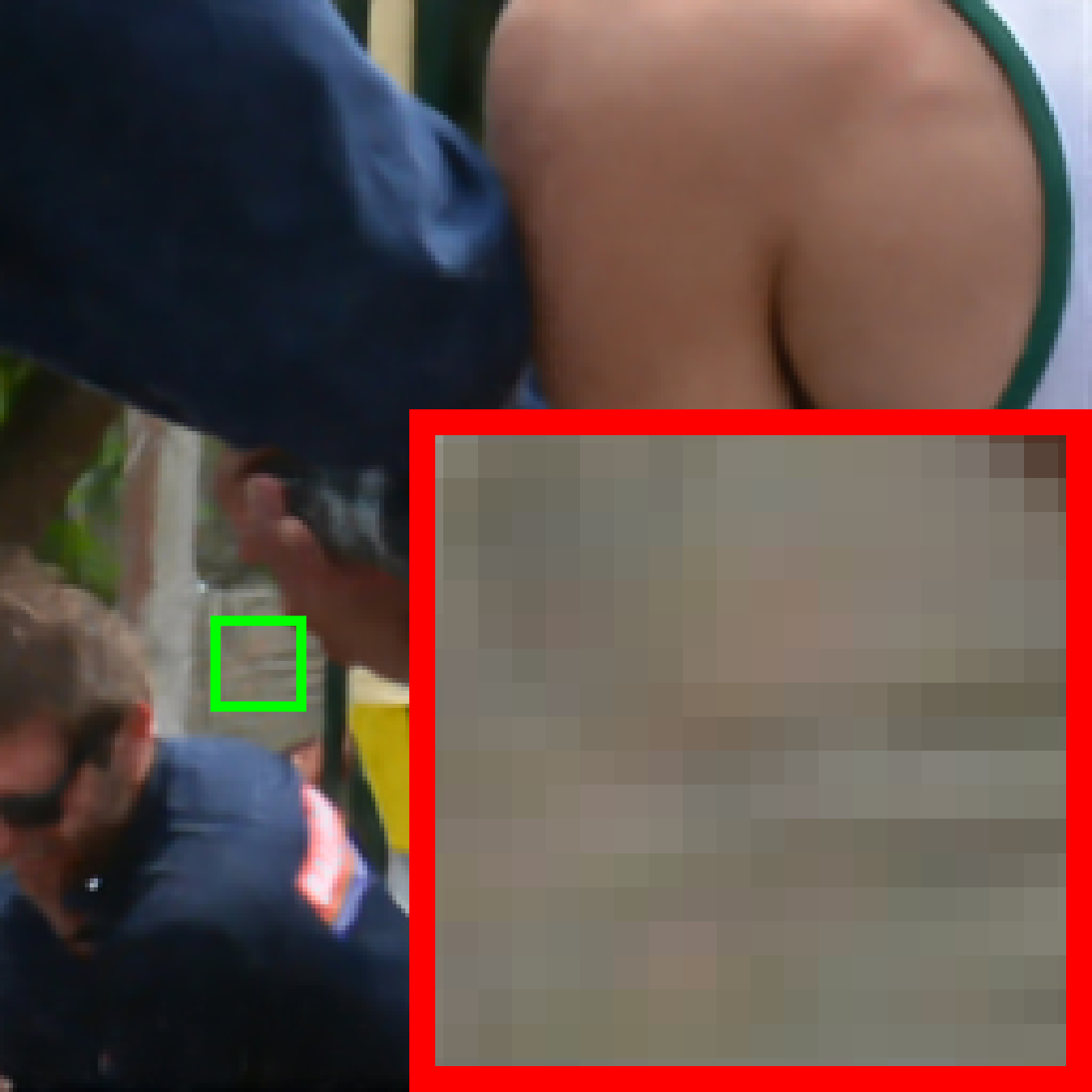}
    &\includegraphics[width=0.07\textwidth]{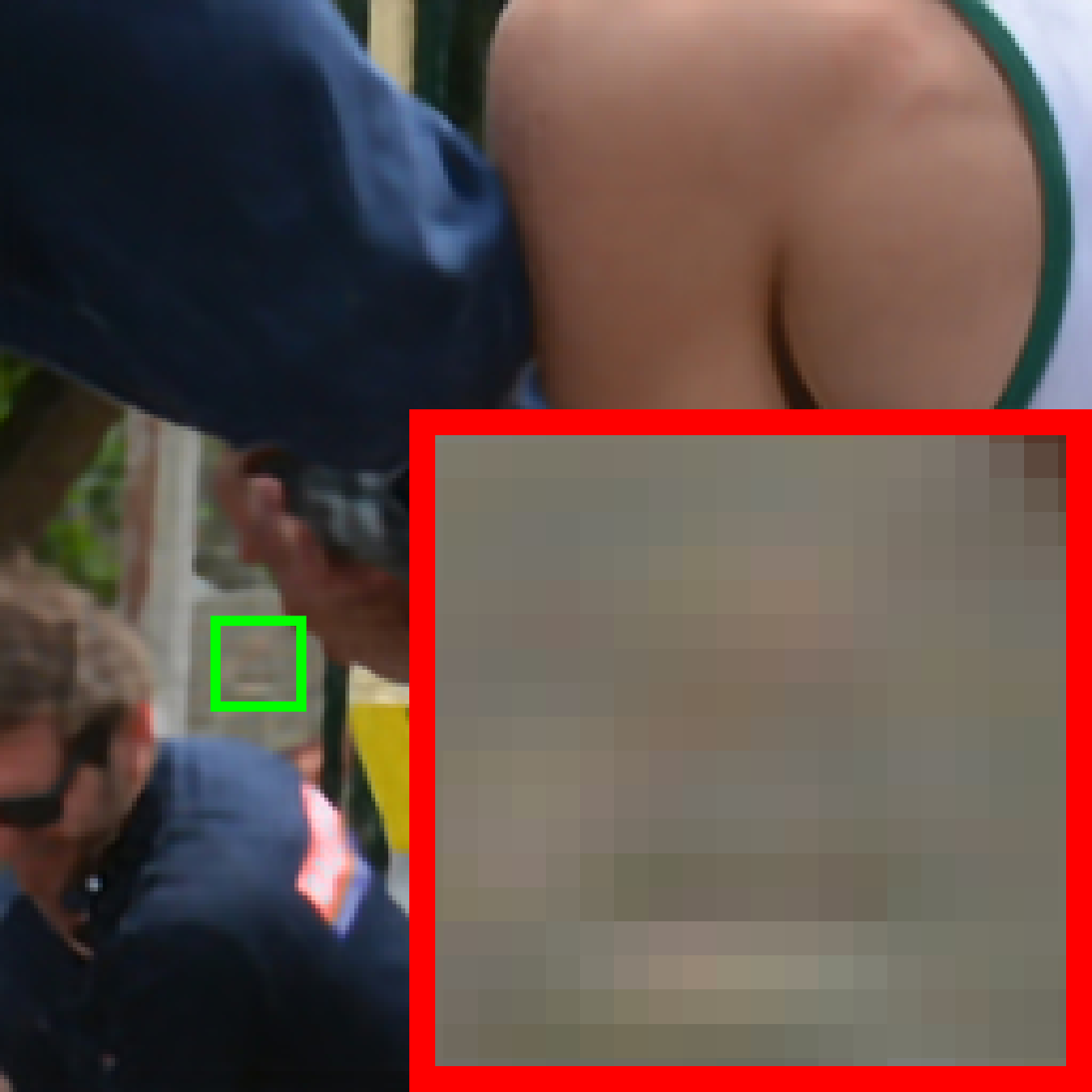}
    &\includegraphics[width=0.07\textwidth]{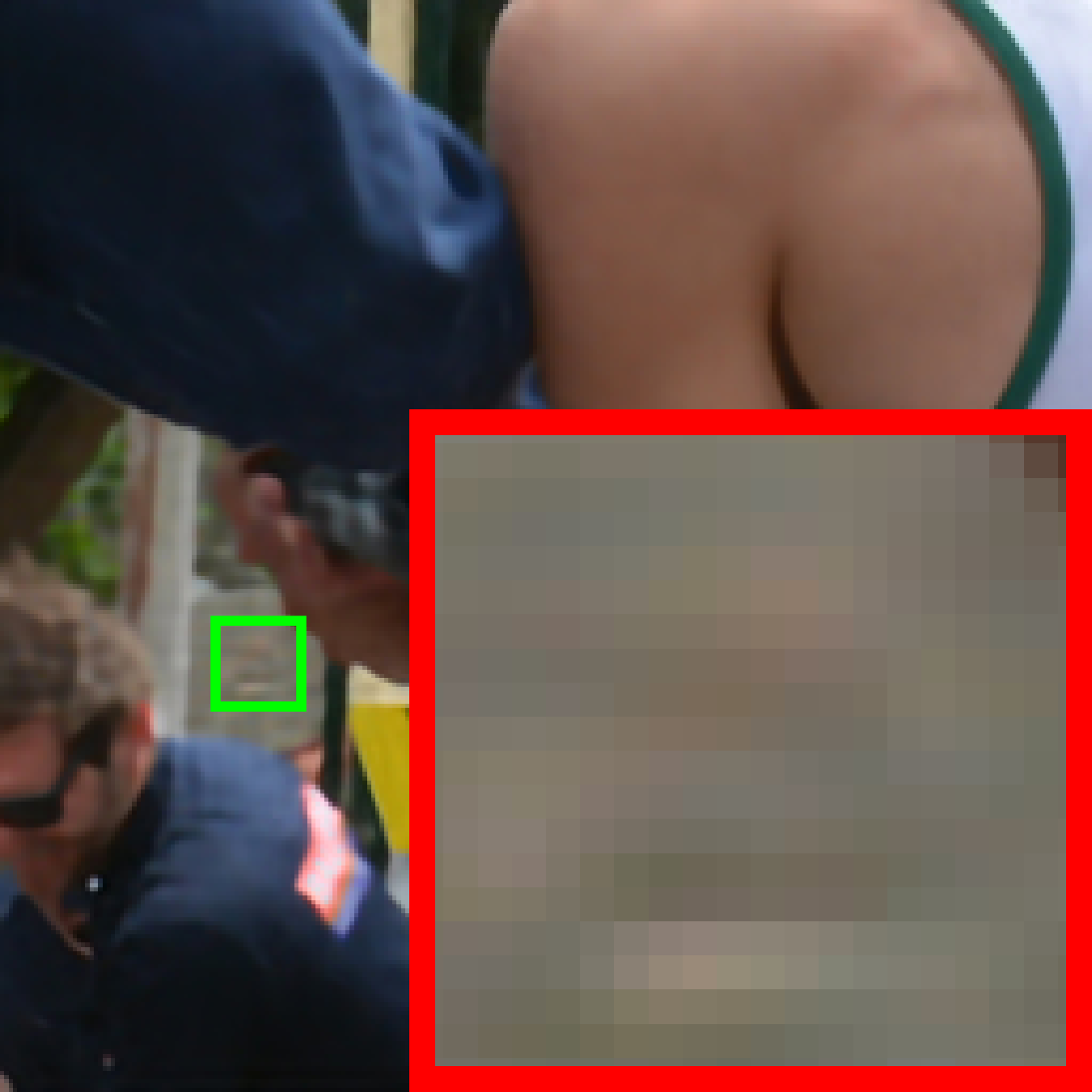}
    &\includegraphics[width=0.07\textwidth]{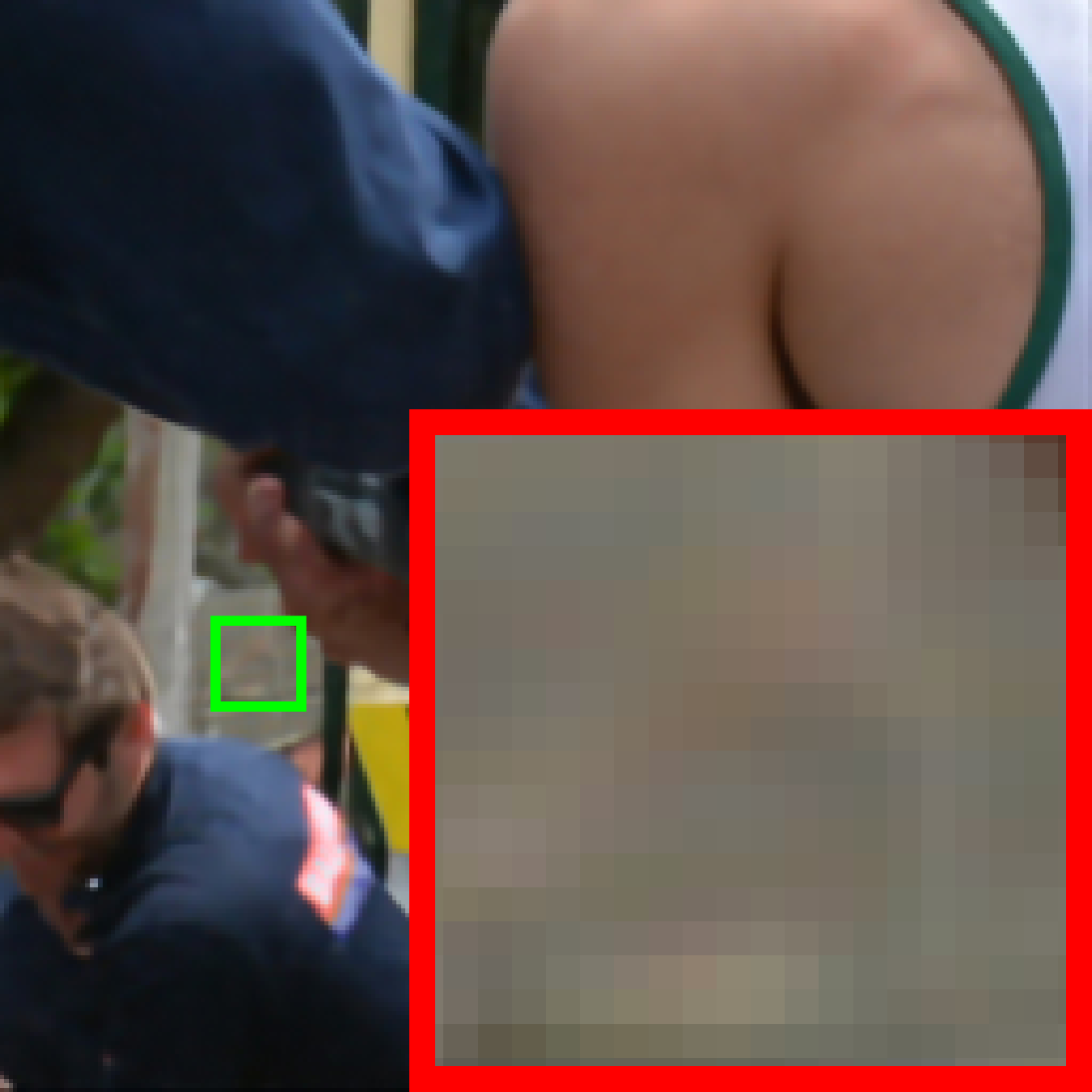}
    &\includegraphics[width=0.07\textwidth]{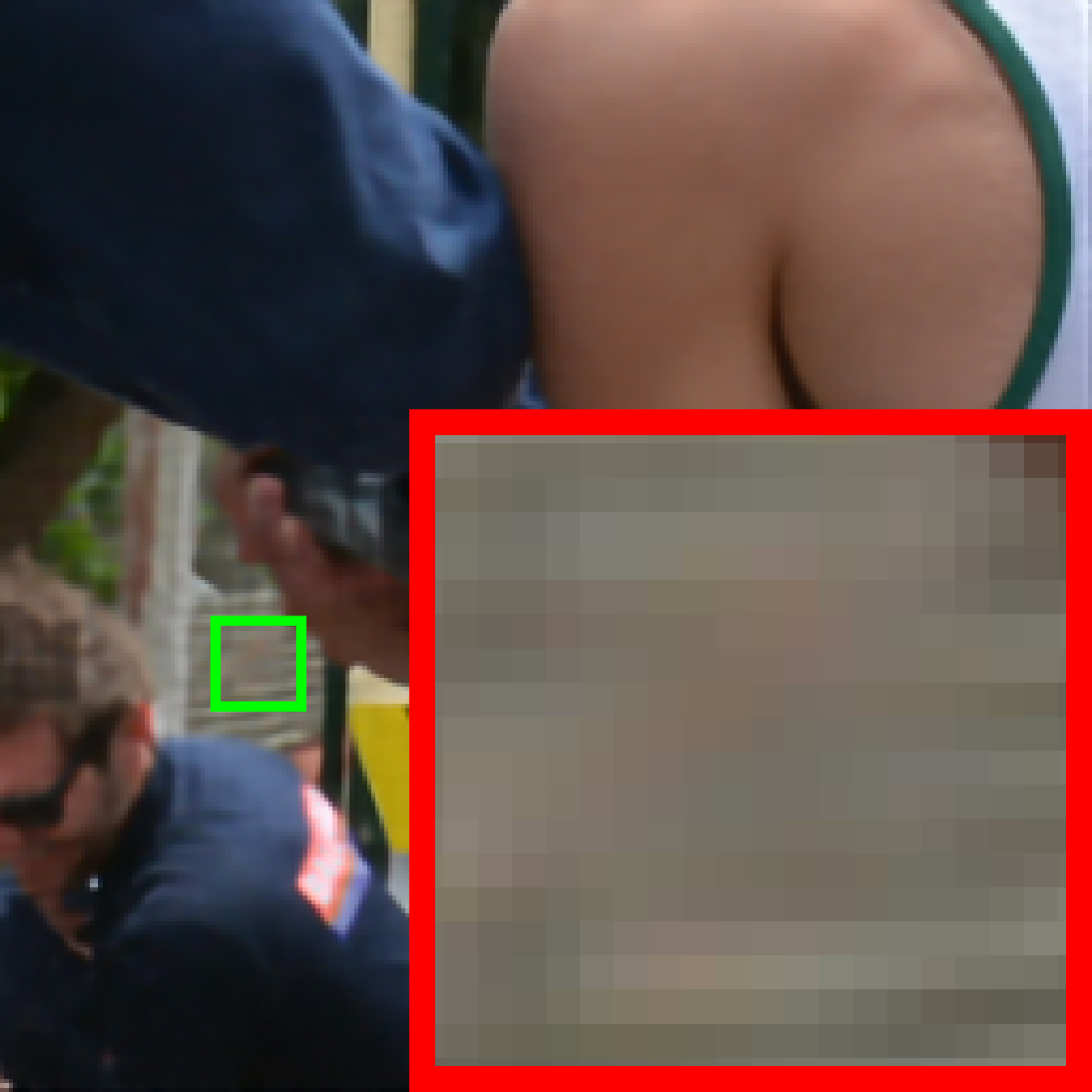}
    &\includegraphics[width=0.07\textwidth]{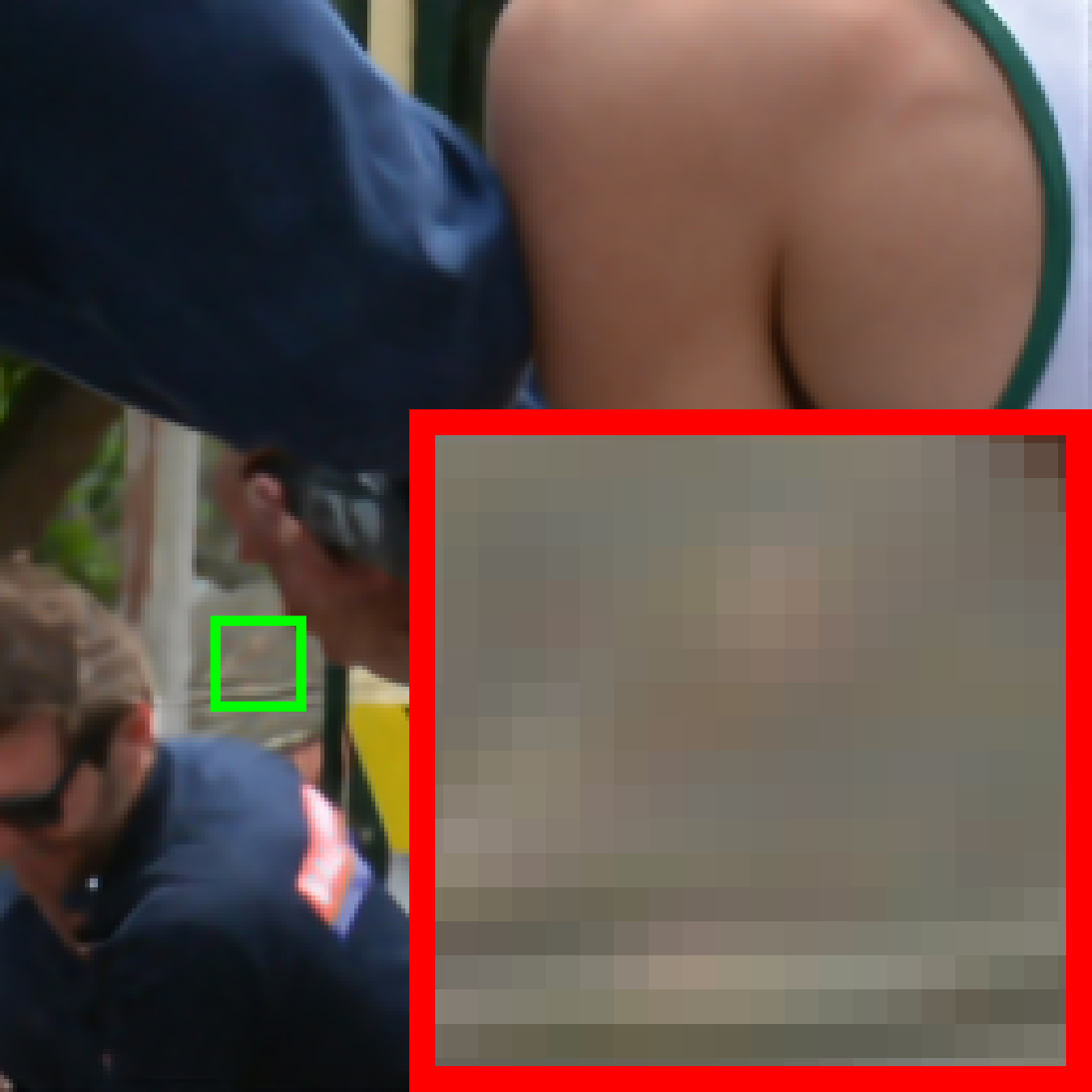}
    &\includegraphics[width=0.07\textwidth]{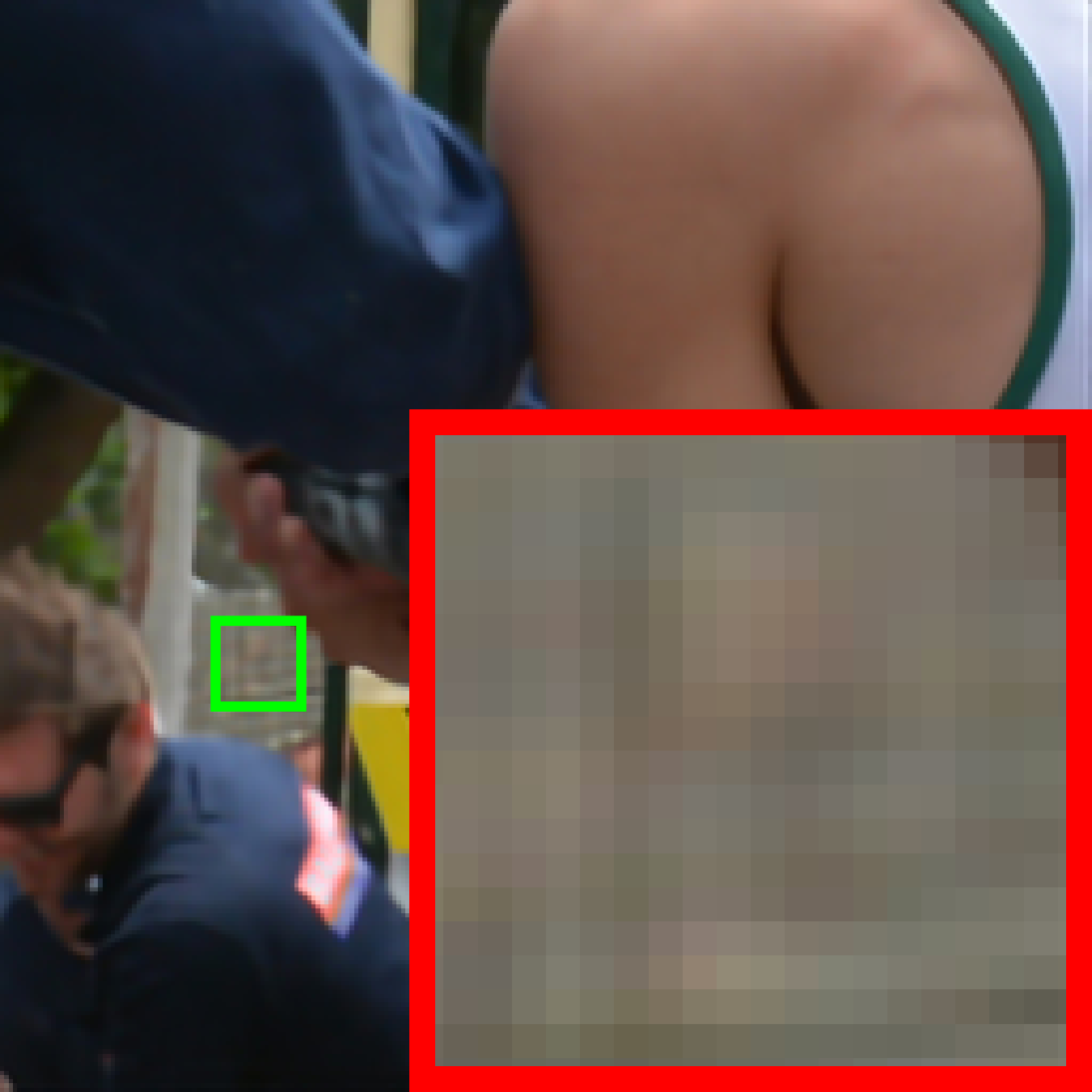}
    &\includegraphics[width=0.07\textwidth]{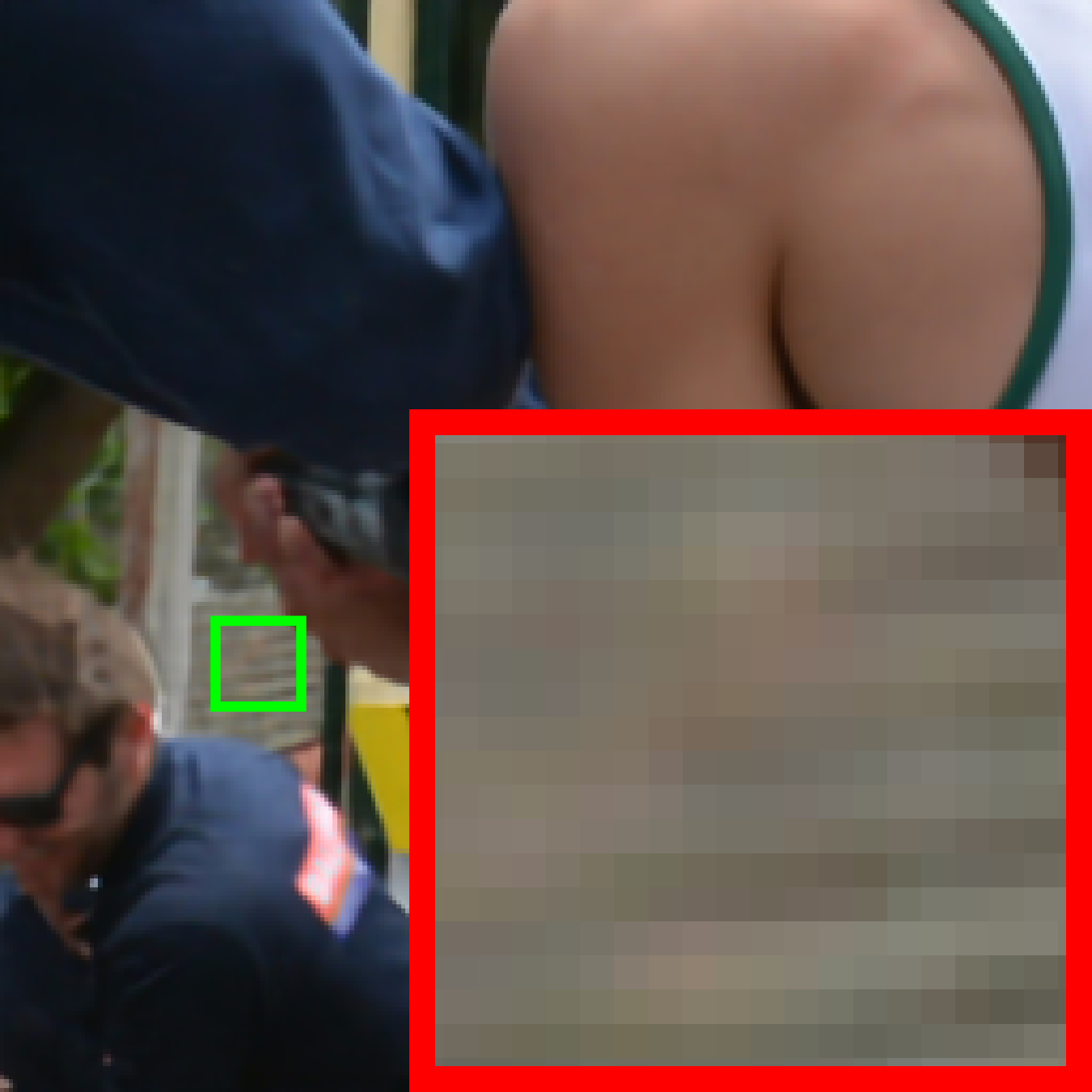}
    &\includegraphics[width=0.07\textwidth]{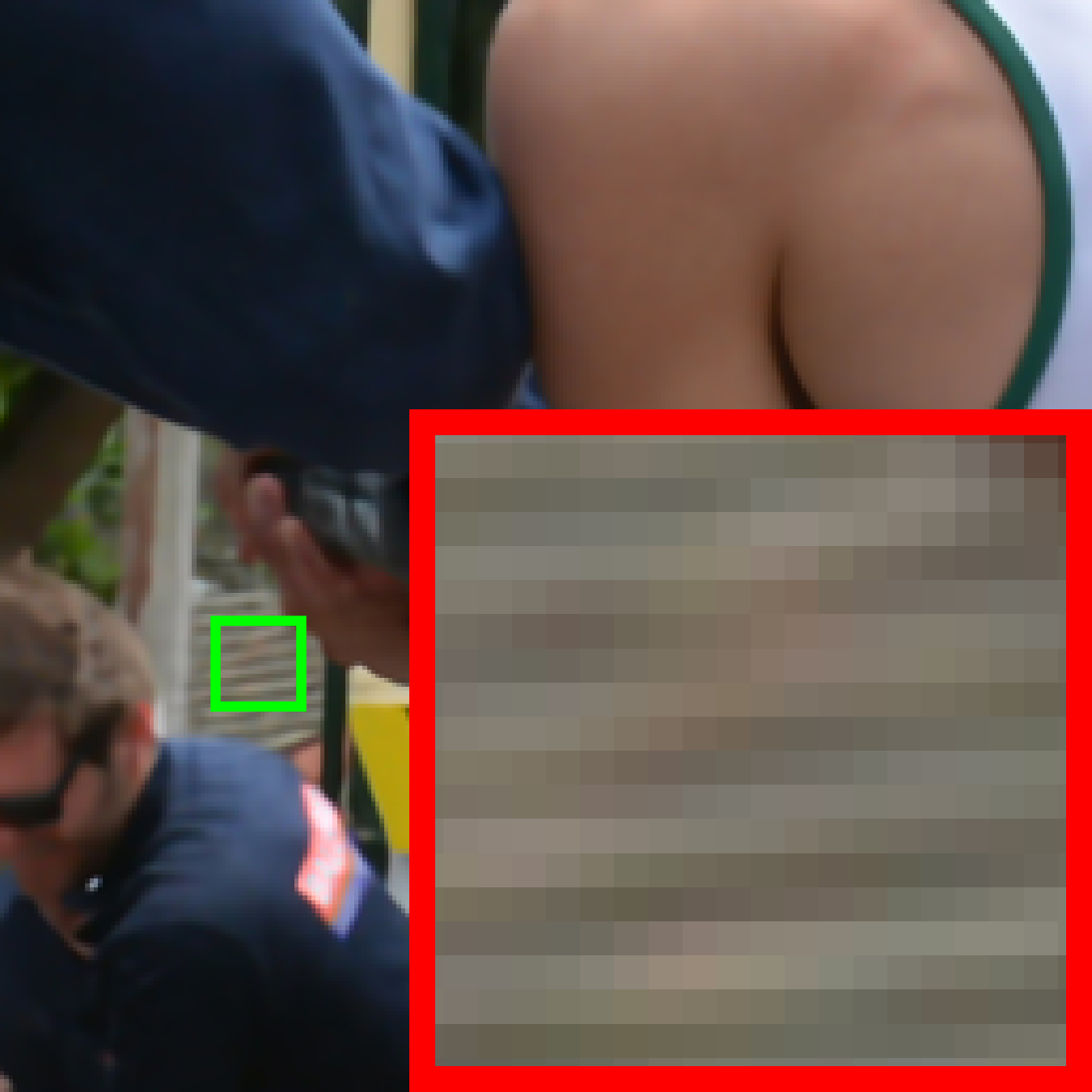}
    &\includegraphics[width=0.07\textwidth]{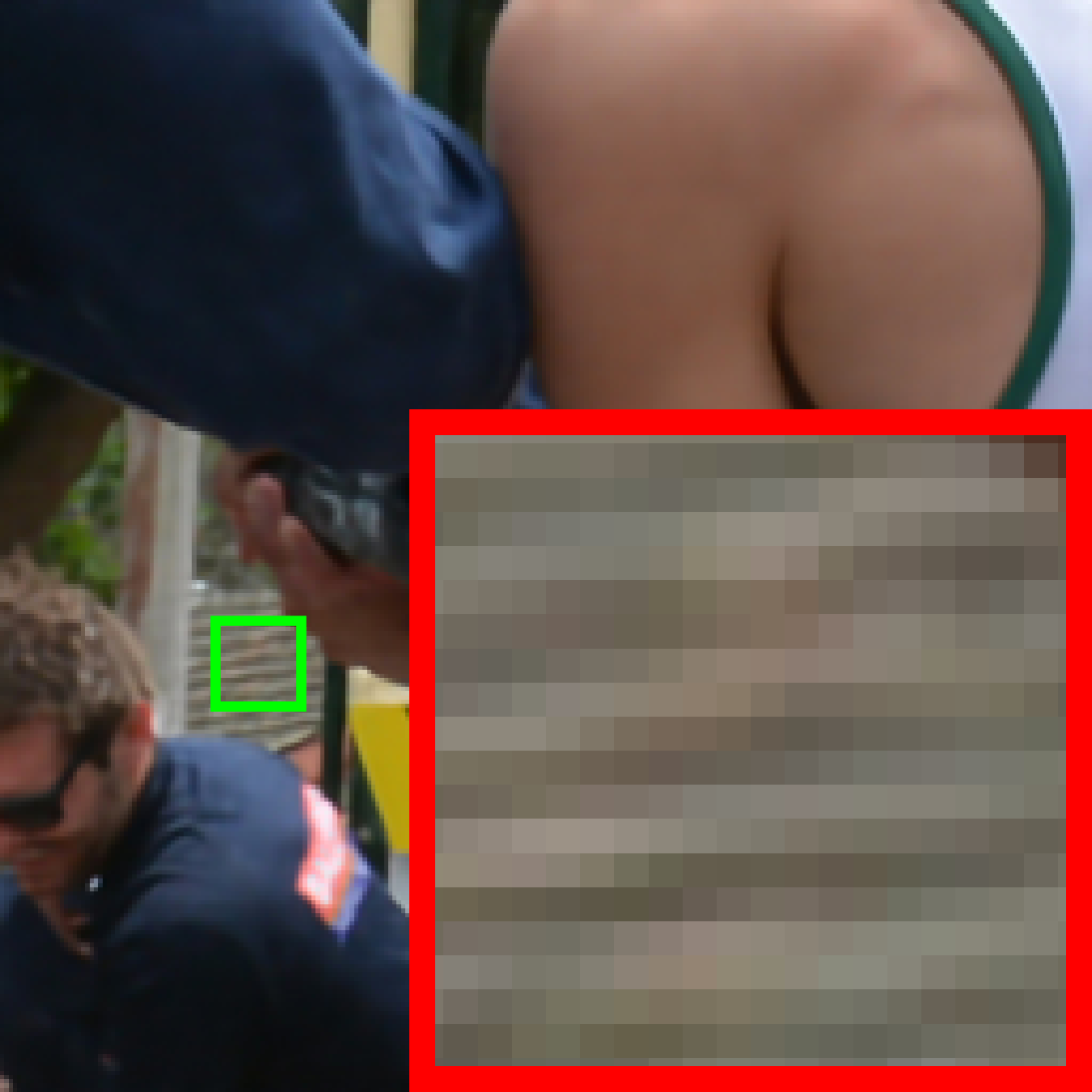}
    &\includegraphics[width=0.07\textwidth]{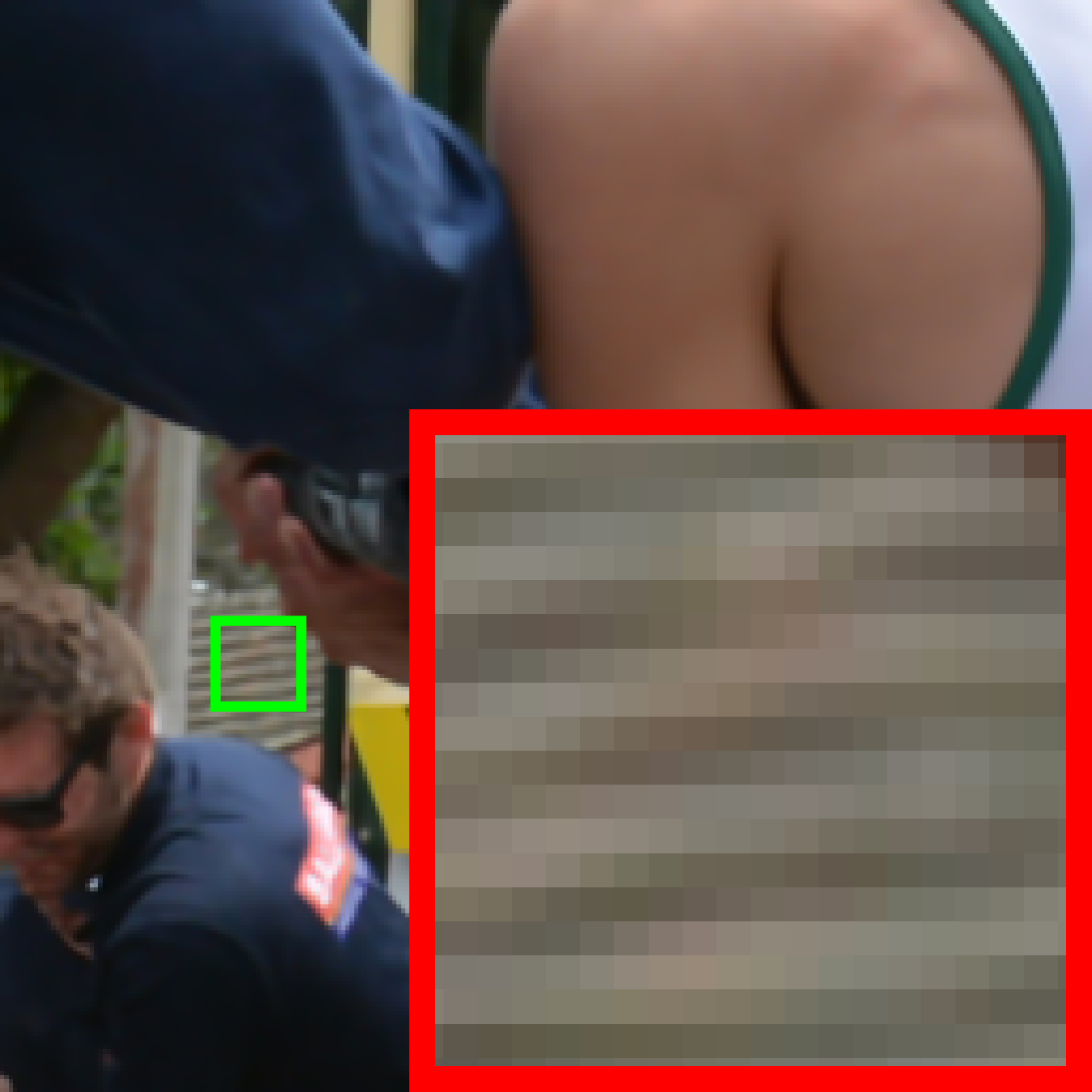}\\
    PSNR/SSIM & 29.29/0.8551 & 32.62/0.9105 & 34.92/0.9441 & 35.71/0.9531 & 35.76/0.9534 & 37.04/0.9646 & 36.78/0.9628 & 37.33/0.9674 & 37.08/0.9679 & 37.89/0.9714 & \textcolor{green}{38.43}/\textcolor{green}{0.9743} & \textcolor{blue}{40.27}/\textcolor{blue}{0.9794} & \textcolor{red}{41.33}/\textcolor{red}{0.9828}
\end{tabular}}
\vspace{-8pt}
\caption{A visual comparison of different methods on four benchmark images named ``Parrots", ``test\_03", ``img\_062" and ``0832" from Set11~\cite{kulkarni2016reconnet} (top), CBSD68~\cite{martin2001database} (upper middle), Urban100~\cite{huang2015single} (lower middle) and DIV2K~\cite{agustsson2017ntire} (bottom), respectively, with the setting of CS ratio $\gamma =10\%$.}
\label{fig:comparison_standard}
\vspace{-10pt}
\end{figure*}

\vspace{-5pt}
\subsection{Proposed PCNet Family}
\label{subsec:pcnet_family}

Based on all the above analyses and results, we first build a convolutional PCNet version, \textbf{PC-CNN}, under the ordinary training setting similar to our baseline and most existing CS networks. As Fig.~\ref{fig:arch} (a) shows, in RS, we use a convolution with $D$ filters of size $2\times 3r\times 3r$, stride $r$ and padding $r$ to extract a shallow feature $\hat{\mathbf{X}}^{(0)}$~$\in$~$\mathbb{R}^{D\times (H/r)\times (W/r)}$ from initialization $\hat{\mathbf{x}}_\text{init}$ of $\mathcal{G}_{\mathbf{A}^\top}$. Then $\hat{\mathbf{X}}^{(0)}$ is sequentially refined by $K$ cascaded deep unrolled stages. The final estimation is generated by another convolution with $r^2$ filters of size $D\times 3\times 3$, stride 1, and padding 1, followed by a PixelShuffle layer for spatial upsampling of ratio $r$. Fig.~\ref{fig:stage} (c) details the design of our $k$-th stage, incorporating PGD-unrolling and high-throughput feature-level refinement. Specifically, in its gradient descent module, we replace the original matrices $\mathbf{A}$, $\mathbf{A}^\top$, and gradient operator $\mathbf{A}^\top(\mathbf{A}(\cdot)-\mathbf{y})$ with our $\mathcal{G}_\mathbf{A}$, $\mathcal{G}_{\mathbf{A}^\top}$ and $\mathcal{H}^{(k)}_\text{grad}(\cdot)$. To enable $\mathcal{H}^{(k)}_\text{grad}$ to enjoy high-throughput transmission, we introduce two skip connections equipped with two zero-initialized learnable factors $s_1^{(k)}$ and $s_2^{(k)}$ for the first and last convolutions of the seven-layer filtering network in our $\mathcal{G}_\mathbf{A}$, as illustrated in Figs.~\ref{fig:sampling_operator} (a) and \ref{fig:stage} (c). These connections link the feature-level input and output of $\mathcal{H}^{(k)}_\text{grad}$ and are jointly optimized as a part of the entire CS network. In implementation, we set $K=20$, $r=1$, and $D=C=32$ by default. Since calculating the equivalent transpose or Moore–Penrose inverse of a matrix form of the deep filtering network is non-trivial, denote measurement $\mathbf{y}=[\mathbf{y}_D,\mathbf{y}_G]$, the initialization operator is defined as $\mathcal{G}_{\mathbf{A}^\top}(\mathbf{y})=[\mathbf{\Phi}_D^\top\mathbf{y}_D,\mathbf{P}^{-1}_G\mathbf{\Phi}_G^\top\mathbf{y}_G]\in\mathbb{R}^{2\times H\times W}$, and the two additional high-throughput connections in $\mathcal{G}_{\mathbf{A}}$ are established only in RS. In the proximal mapping module, we adopt a structure similar to our baseline stage to predict a feature-domain residual, as Fig.~\ref{fig:stage} (a) and (c) exhibit. As shown in Fig.~\ref{fig:roadmap}, the proposed PC-CNN can be viewed as an extension of the baseline equipped with our first three enhancements. In PC-CNN training, we use the ordinary setting and randomly sample permutation $\mathbf{P}_G$, ratios $\gamma_D$ and $\gamma_G$ satisfying $\gamma_D+\gamma_G=\gamma$ for each iteration\footnote{In PC-CNN training, for each batch of 16 patches, we uniformly sample a permutation $\mathbf{P}_G$, 16 ratios $\gamma_i \in[0,1]$ and factors $\alpha_i\in [0,1]$, and compute $\gamma_{D,i}=\alpha_i \gamma_i$ and $\gamma_{D,i}=(1-\alpha_i) \gamma_i$, for $i\in\{1,\cdots,16\}$.}. In evaluation with $\gamma\in[0,0.5]$, we set $\gamma_D=0.4\gamma$ and $\gamma_G=0.6\gamma$ for dual-branch sampling (see Fig.~\ref{fig:sampling_operator}) by default\footnote{To find the optimal dual-branch ratio setting for our evaluation and application, we perform a grid search in Sec.~A of our \textcolor{blue}{\textit{\textbf{supplementary}}}.}.

Our second PCNet model, \textbf{PCT}, is a Transformer NN that incorporates PC-CNN along with seven additional enhancements. The tables at the top left and right of Fig.~\ref{fig:roadmap} detail the main differences among our three networks in terms of training settings and recovery performance, highlighted by red bars. Compared to the baseline, the proposed PC-CNN is enhanced with transmission-augmenting features, and flexibility, and is powered by our proposed COSO ($\mathcal{G}_\mathbf{A}$) as detailed in Fig.~\ref{fig:sampling_operator}, achieving a PSNR lead of 3.01dB. Our PCT builds on PC-CNN, further improving by 0.85dB through advanced SCBs (with $r=2$) and sufficient training.

\hspace{-8pt}
\begin{figure*}[!t]
\setlength{\tabcolsep}{0.5pt}
\resizebox{1.005\textwidth}{!}{
\scriptsize
\begin{tabular}{cccccc}
    Ground Truth & ISTA-Net$^\text{+}$ & ISTA-Net$^\text{++}$ & FSOINet & \textbf{PC-CNN (Ours)} & \textbf{PCT (Ours)}\\
    \includegraphics[width=0.16\textwidth]{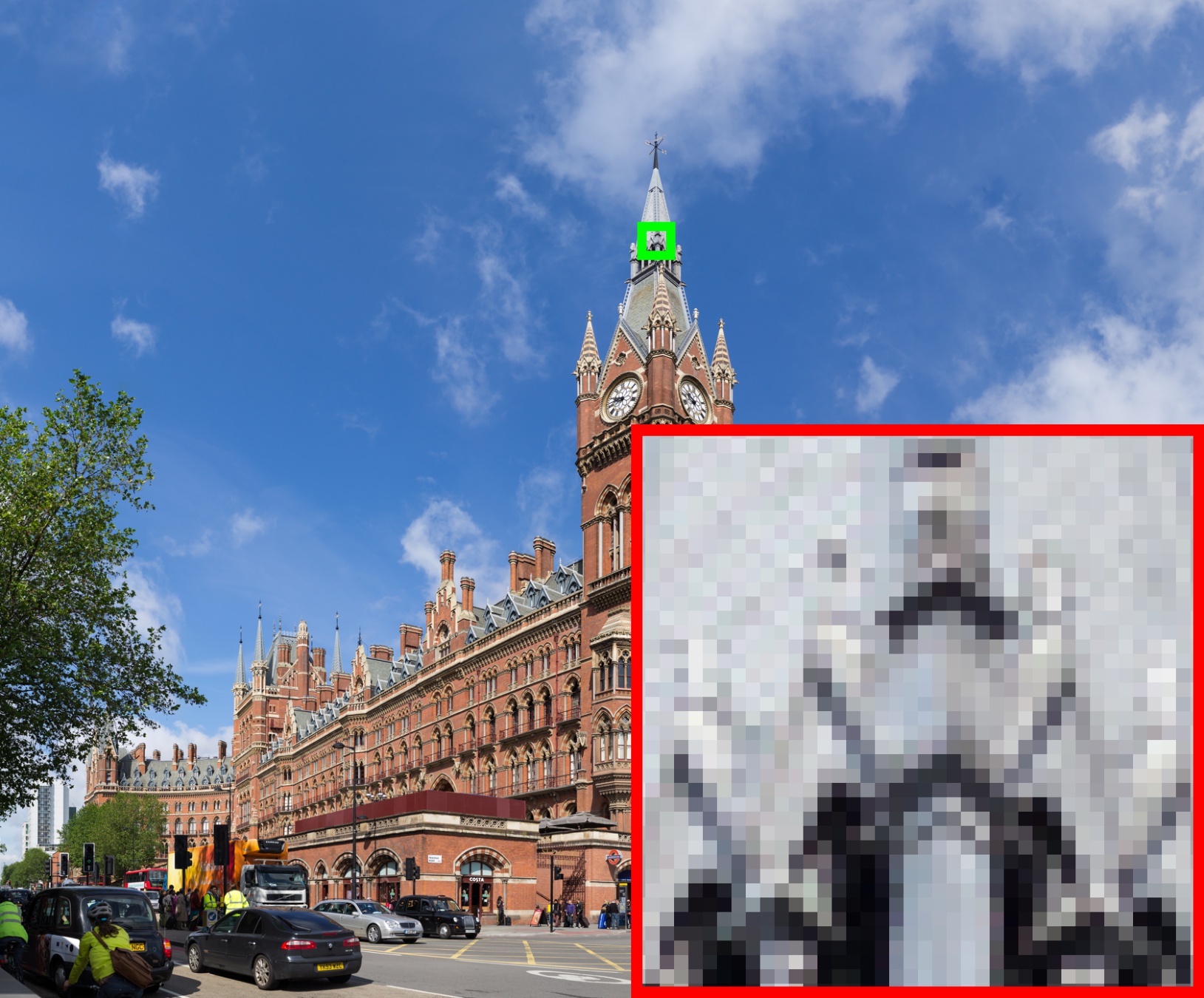}
    &\includegraphics[width=0.16\textwidth]{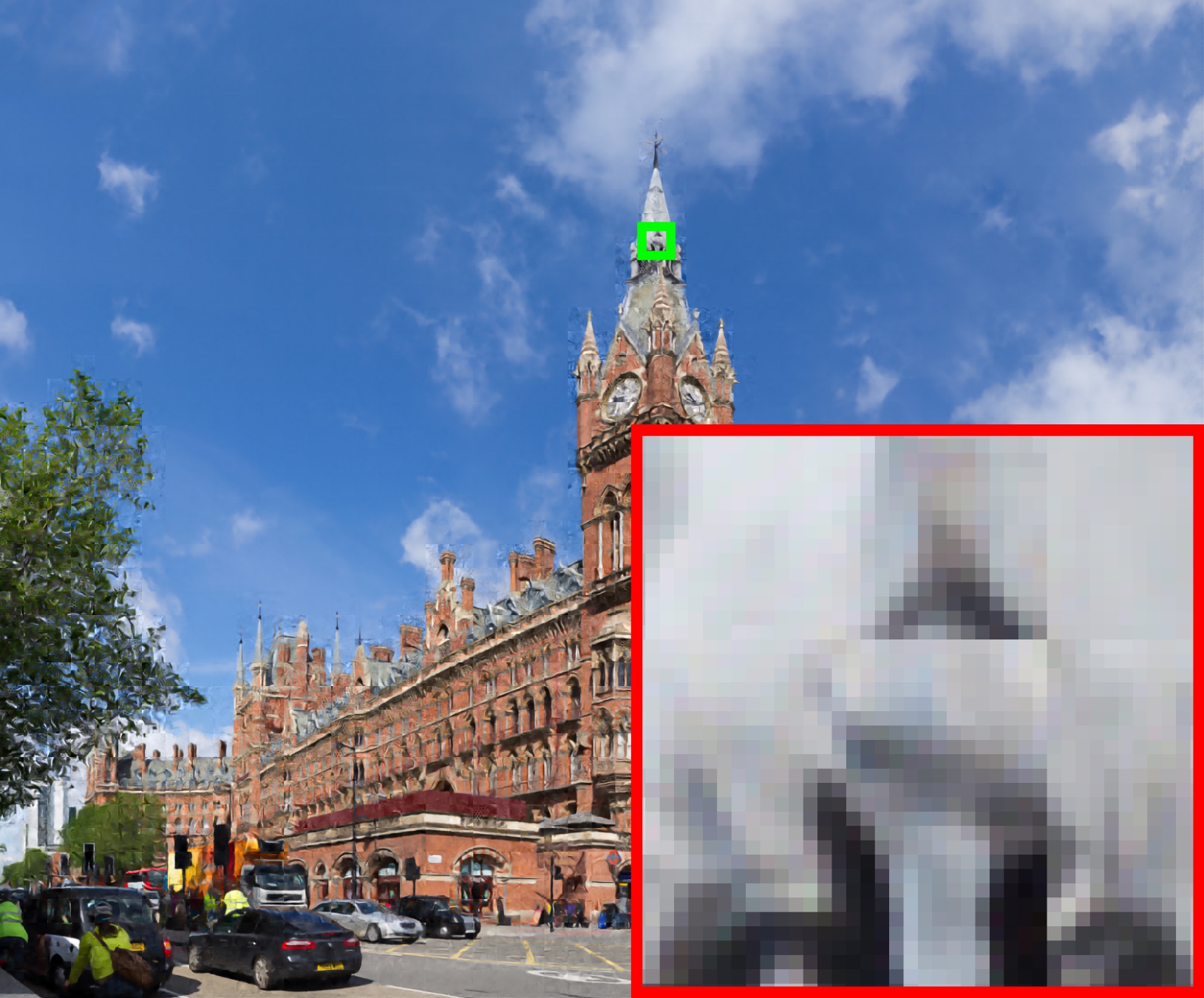}
    &\includegraphics[width=0.16\textwidth]{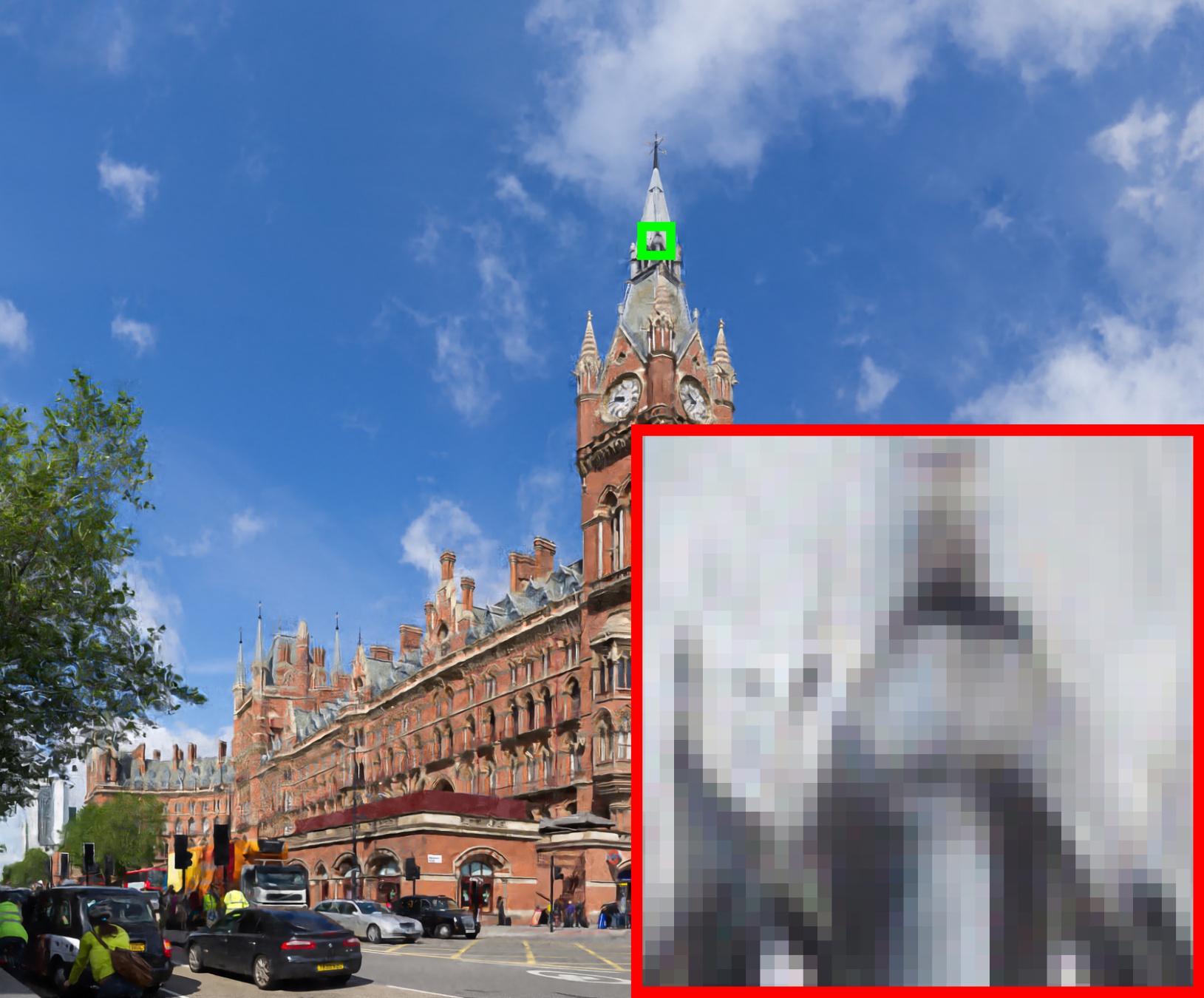}
    &\includegraphics[width=0.16\textwidth]{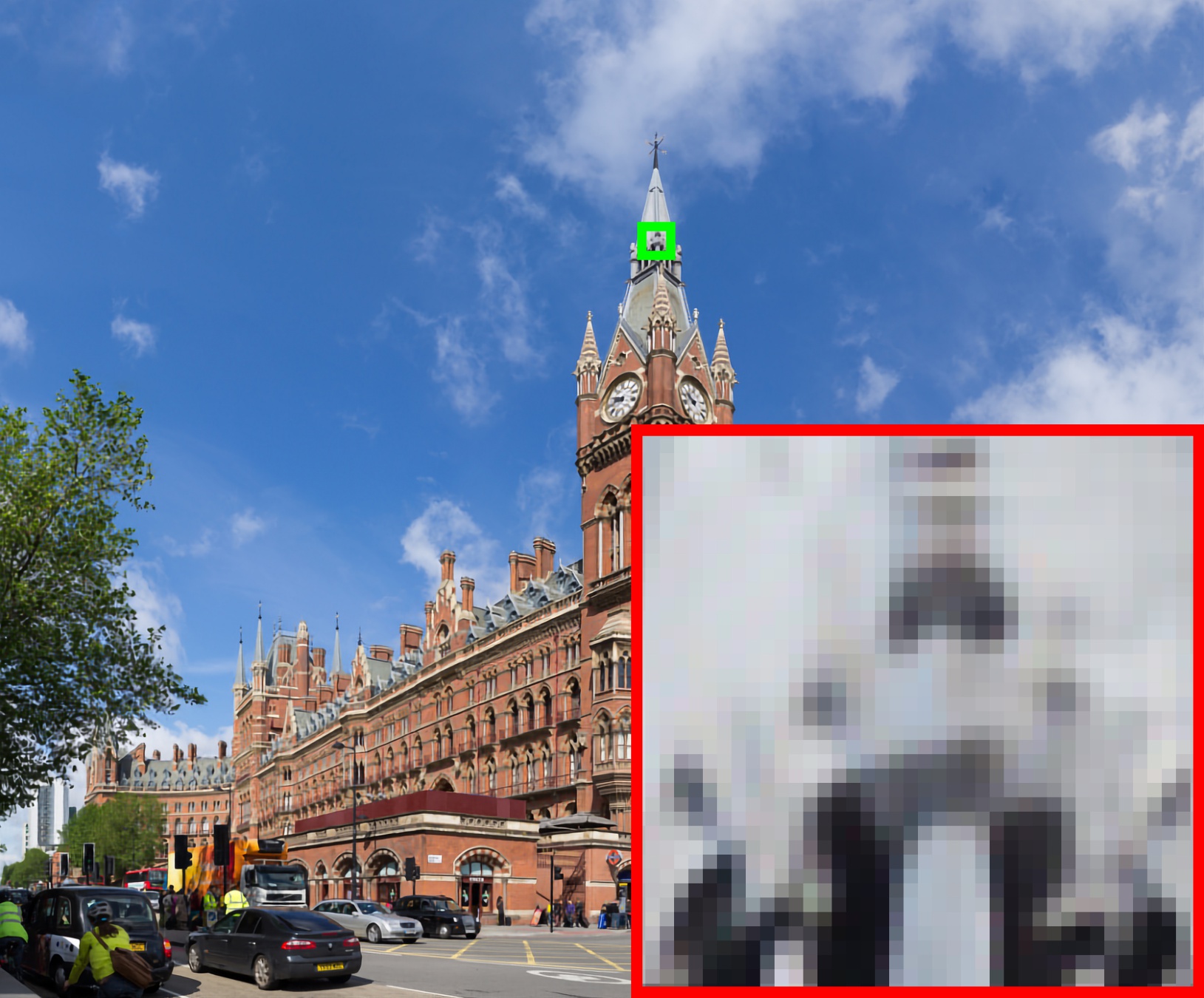}
    &\includegraphics[width=0.16\textwidth]{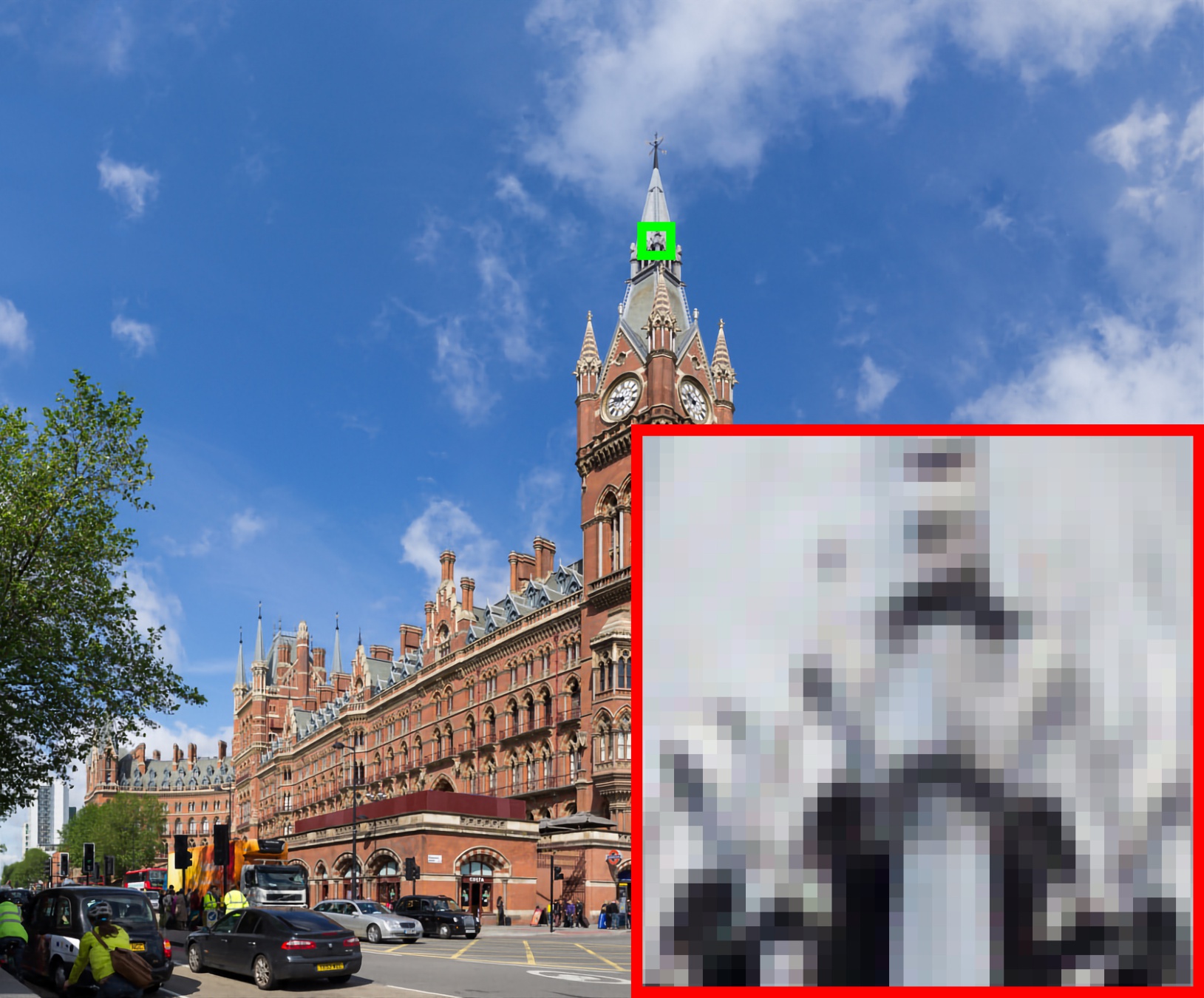}
    &\includegraphics[width=0.16\textwidth]{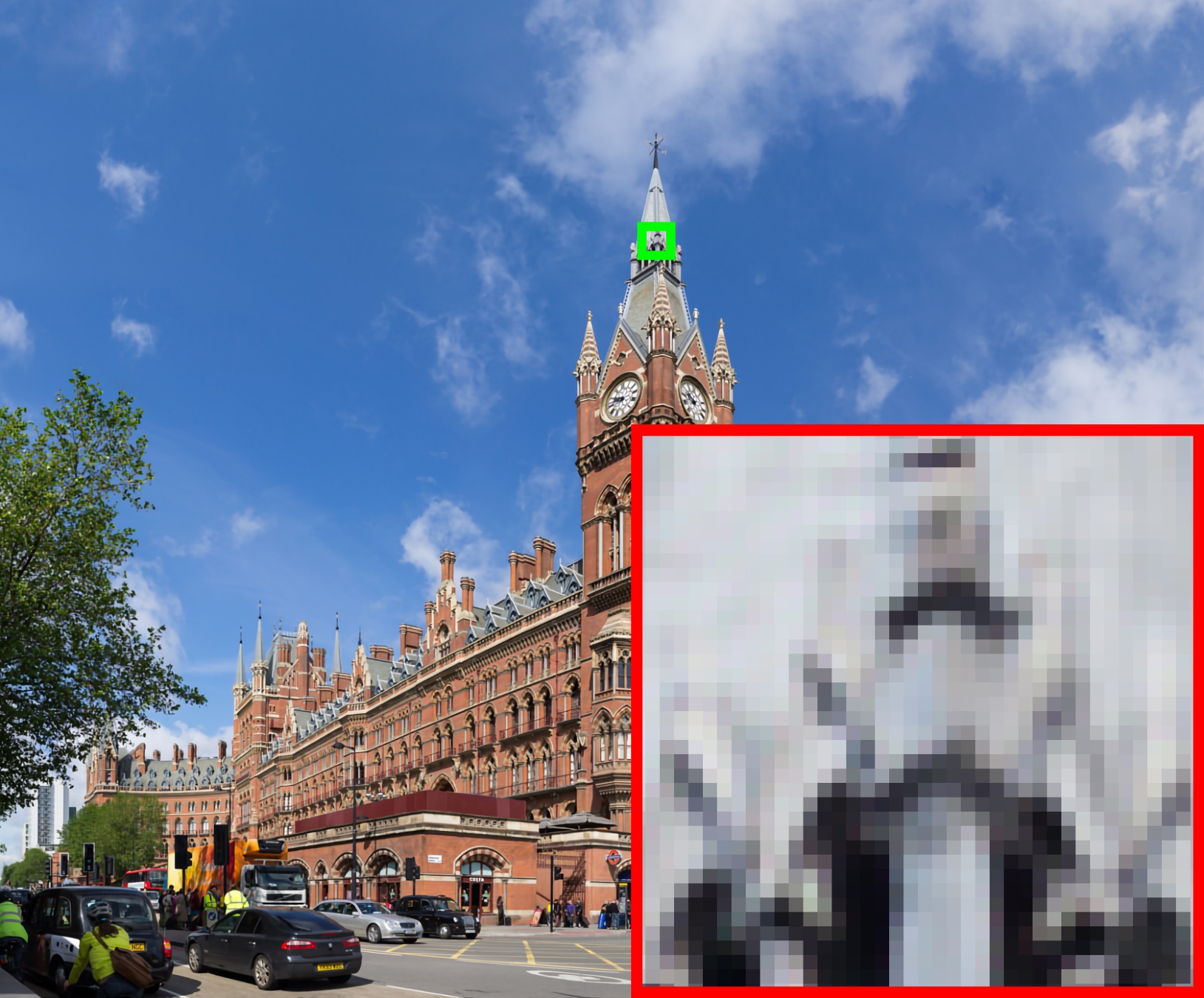}\\
    PSNR/SSIM & 24.05/0.8253 & 24.99/0.8599 & \textcolor{green}{27.12}/\textcolor{green}{0.9145} & \textcolor{blue}{30.00}/\textcolor{blue}{0.9466} & \textcolor{red}{30.54}/\textcolor{red}{0.9538}
\end{tabular}}
\scriptsize
\resizebox{1.005\textwidth}{!}{
\begin{tabular}{ccc}
    \includegraphics[width=0.333\textwidth]{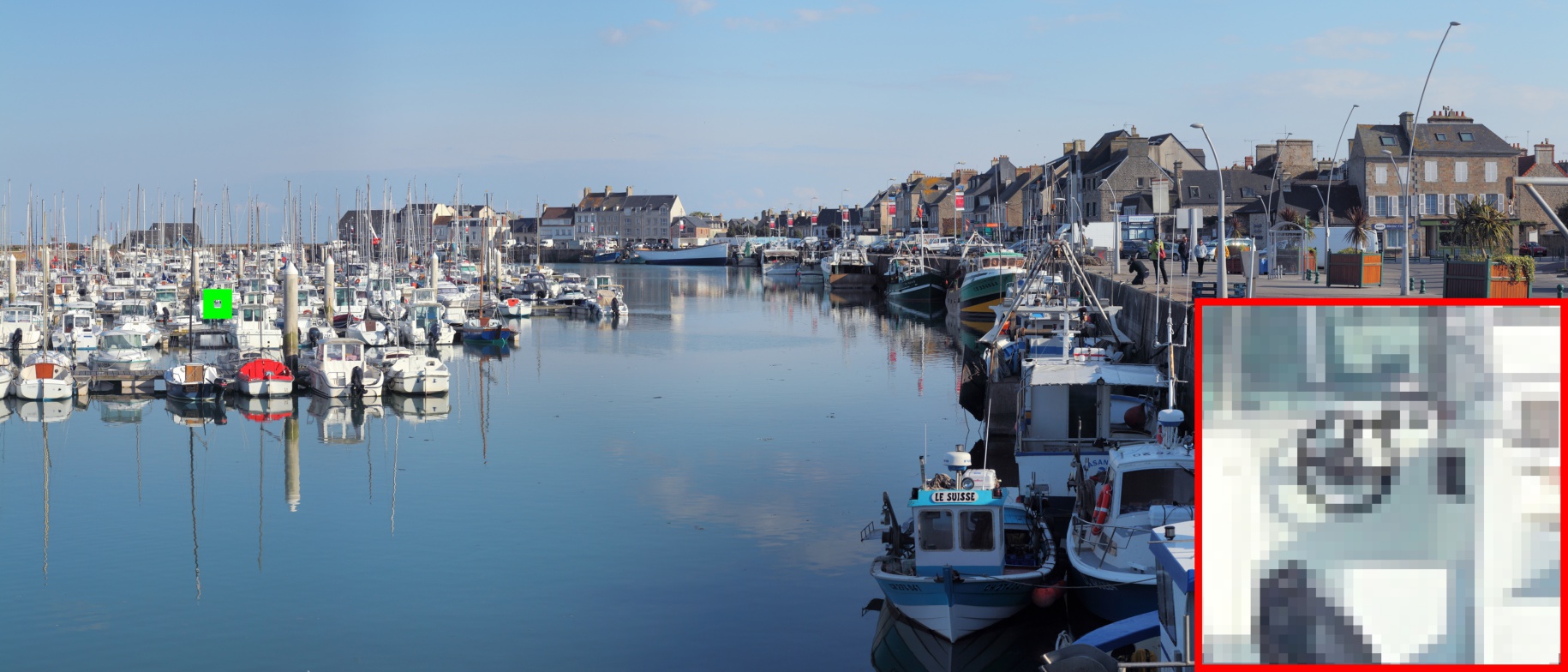}
    &\includegraphics[width=0.333\textwidth]{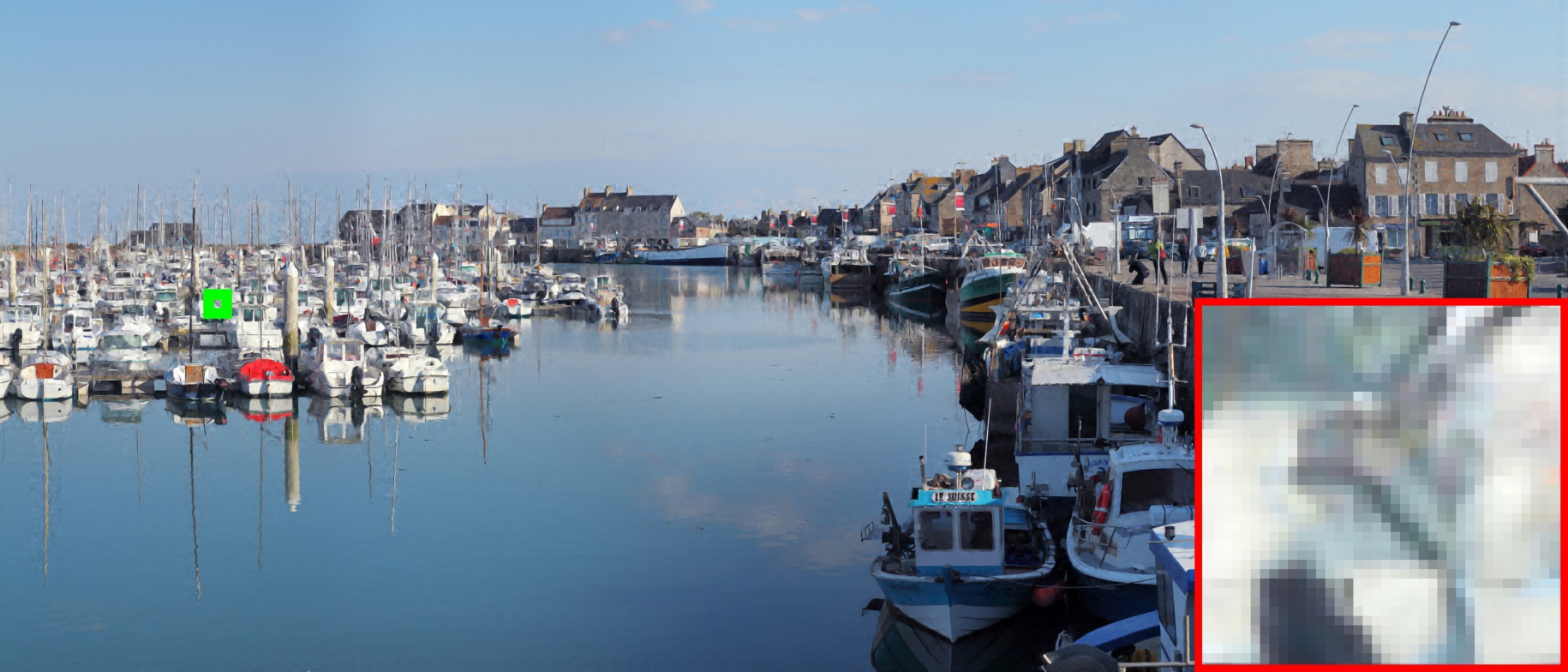}
    &\includegraphics[width=0.333\textwidth]{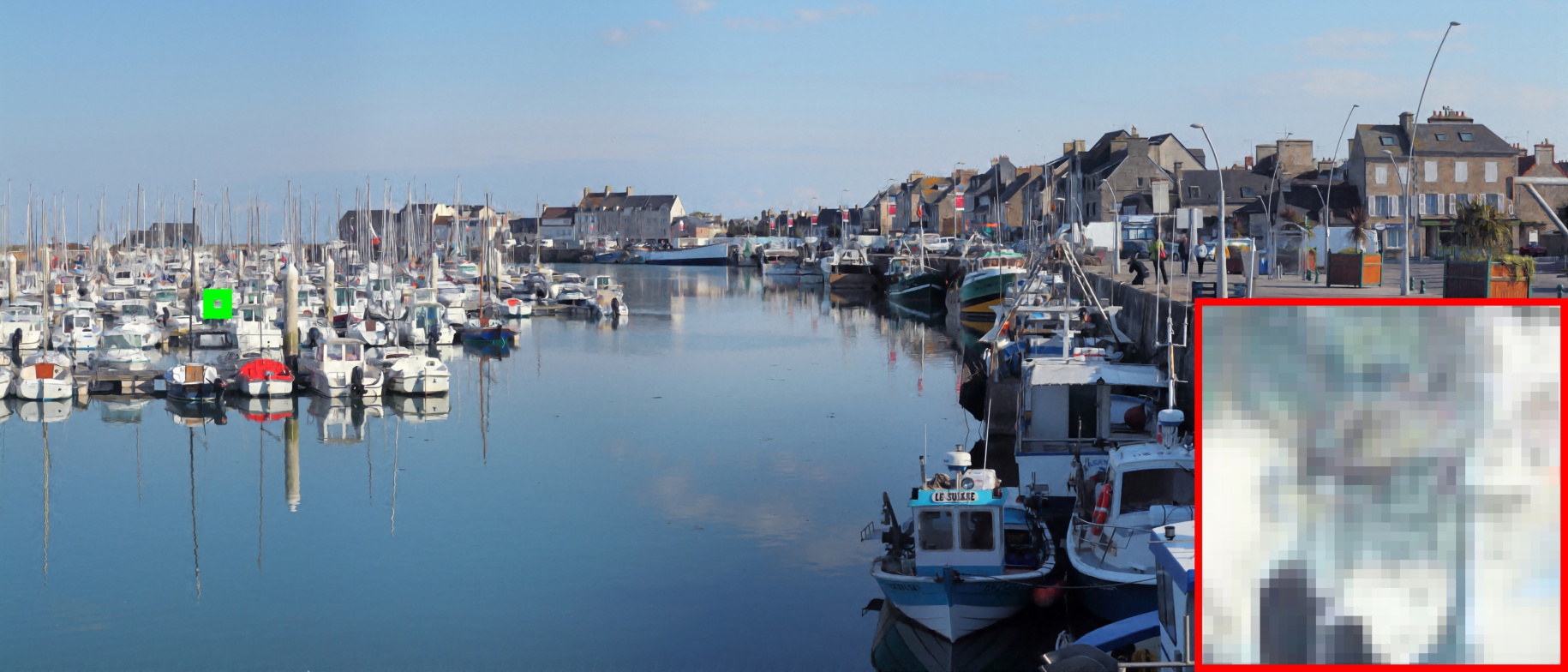}\\
    Ground Truth (PSNR/SSIM) & ISTA-Net$^\text{+}$ (27.01/0.8585) & ISTA-Net$^\text{++}$ (28.34/0.8911) \\
    \includegraphics[width=0.333\textwidth]{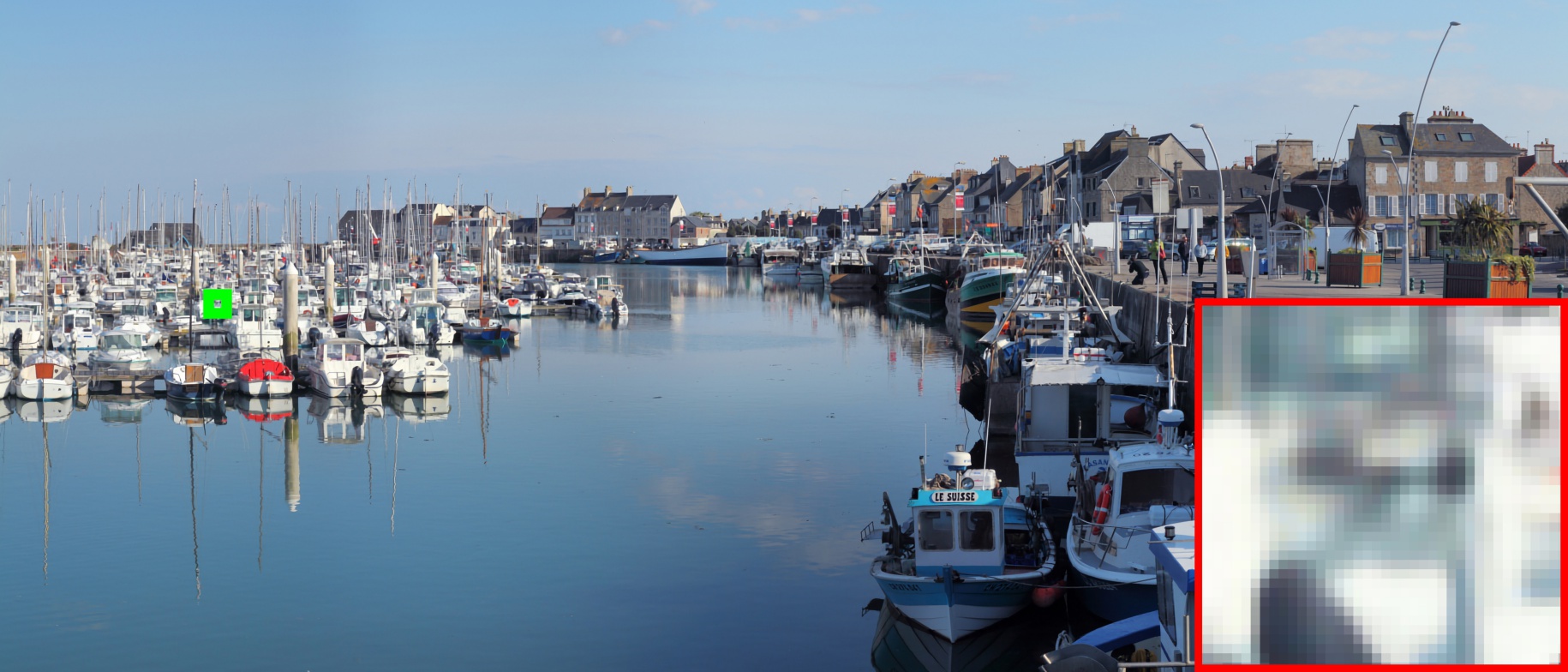}
    &\includegraphics[width=0.333\textwidth]{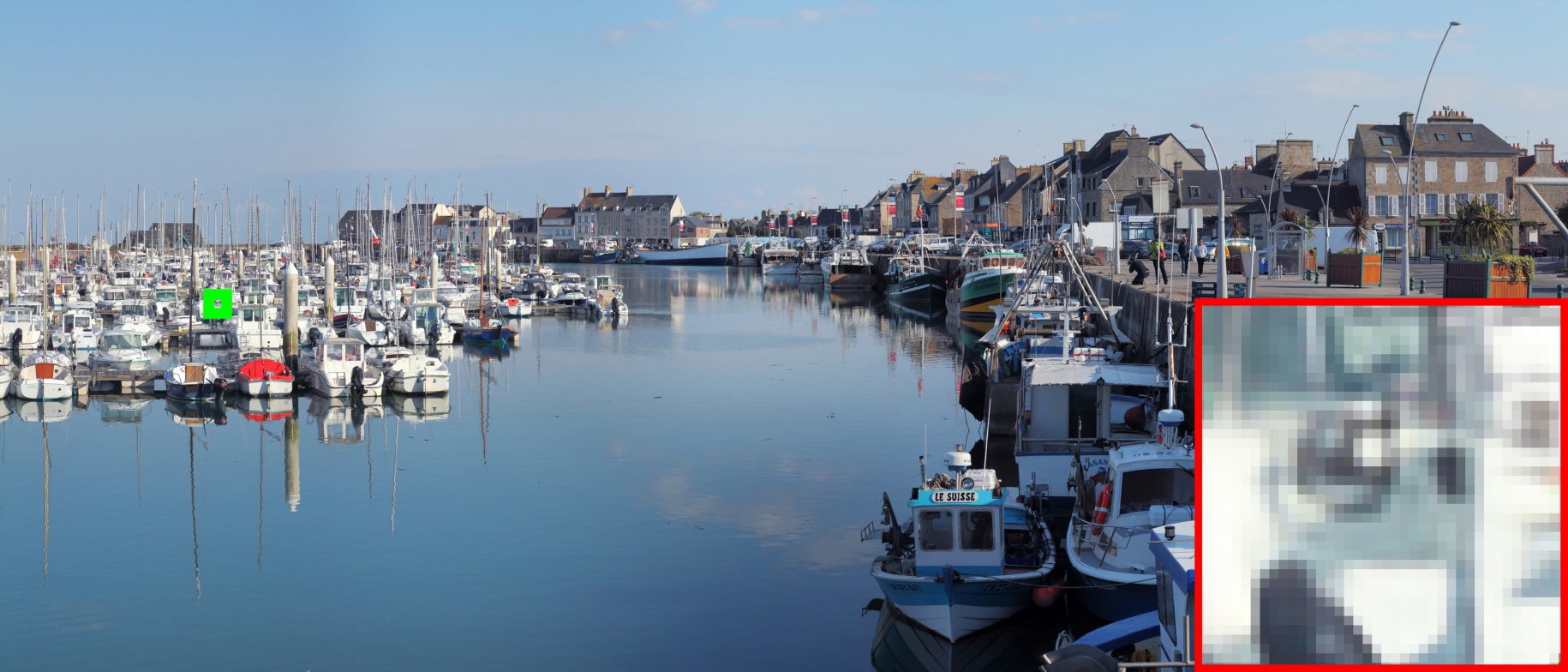}
    &\includegraphics[width=0.333\textwidth]{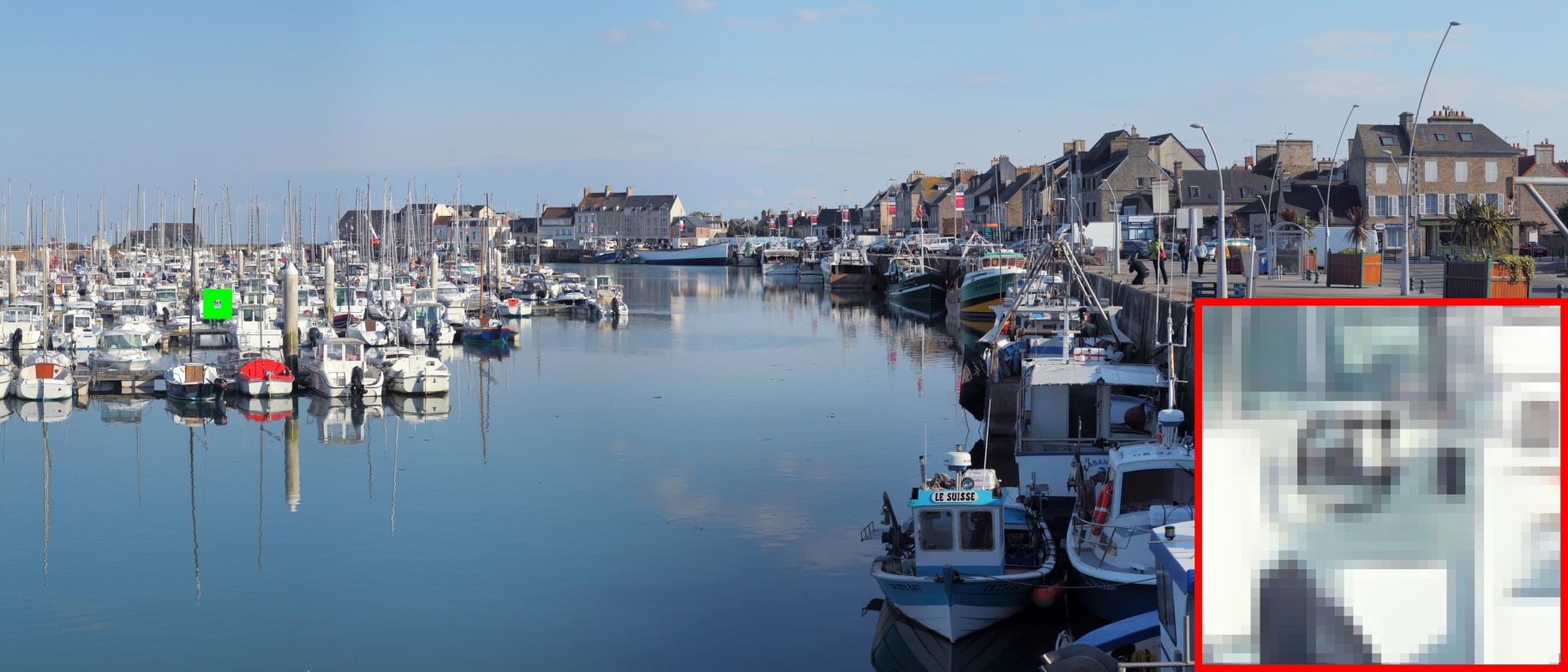}\\
    FOSINet (\textcolor{green}{30.66}/\textcolor{green}{0.9330}) & \textbf{PC-CNN (Ours)} (\textcolor{blue}{33.62}/\textcolor{blue}{0.9506}) & \textbf{PCT (Ours)} (\textcolor{red}{34.64}/\textcolor{red}{0.9593})
\end{tabular}}
\vspace{-10pt}
\hspace{8pt}
\caption{A visual comparison on two images: ``1222" and ``1337" from Test2K (top) and Test4K (bottom) sets \cite{gu2019div8k,kong2021classsr}, respectively, with $\gamma =10\%$.}
\label{fig:comparison_high_resolution}
\vspace{-10pt}
\end{figure*}

\vspace{-5pt}
\section{Experiments}
\label{sec:experiments}

\subsection{Performance Comparisons with State-of-the-Arts}

We compare our PCNets with fifteen representative state-of-the-art end-to-end deep CS NNs: ReconNet \cite{kulkarni2016reconnet}, ISTA-Net$^+$ \cite{zhang2018ista}, DPA-Net \cite{sun2020dual}, MAC-Net \cite{chen2020learning}, ISTA-Net$^{++}$ \cite{you2021ista}, CSNet$^+$ \cite{shi2019image}, SCSNet \cite{shi2019scalable}, OPINE-Net$^+$ \cite{zhang2020optimization}, AMP-Net \cite{zhang2021amp}, COAST \cite{you2021coast}, MADUN \cite{song2021memory}, FSOINet \cite{chen2022fsoinet}, CASNet \cite{chen2022content}, RK-CCSNet \cite{zheng2020sequential}, and MR-CCSNet$^+$ \cite{fan2022global}. The former five, middle eight, and latter two adopt the block-diagonal fixed Gaussian matrix, jointly learned matrix, and stacked linear network by default, respectively. In Tab.~\ref{tab:compare_sota_high_level}, we first conduct a high-level comparison among different approaches. PCNets show their organic integration of multiple merits, enjoying a global sampling, high throughput, interpretable recovery, efficient implementation, and acceptable complexity.

\subsubsection{Comparison on Normal-Sized Benchmark Images}

Tab.~\ref{tab:compare_sota_psnr} and Fig.~\ref{fig:comp_param_PSNR} report the PSNR results of different methods on four widely adopted benchmarks: Set11 \cite{kulkarni2016reconnet}, CBSD68 \cite{martin2001database}, Urban100 \cite{huang2015single}, and the validation set of DIV2K \cite{agustsson2017ntire}. To avoid the memory overflow of some large-capacity networks on a 3090 GPU, all images from Urban100 and DIV2K are 256$\times$256 center-cropped. From our provided quantitative results, we observe that the networks equipped with learnable sampling operators generally achieve higher PSNR values than the random matrix-based ones. Due to the explicit physical information exploitation, most deep unrolled methods attain a large performance gap to the other physics-free ones. By integrating the proposed collaborative sampling operator and high-throughput unrolled architecture, our PC-CNN achieves a significant PSNR lead of about 0.37-1.88dB over the existing best CS approaches. Note that some recent methods: AMP-Net \cite{zhang2021amp}, MADUN \cite{song2021memory}, FSOINet \cite{chen2022fsoinet}, and MR-CCSNet$^+$ \cite{fan2022global} need to train separate models for different rates. Being further enhanced by Transformer blocks and sufficient training, our PCT obtains PSNR gains of 0.31-1.31dB upon PC-CNN. In Fig.~\ref{fig:comparison_standard}, our visual comparison shows that PCNets accurately recover the patterns of stripes and windows. Their recovered images achieve the highest PSNR and SSIM values, correct shapes, vivid textures, and sharper edges. More visual comparison results are provided in Sec.~B of our \textcolor{blue}{\textit{\textbf{supplementary}}}.

\begin{table}[!t]
\vspace{-5pt}
\caption{A PSNR comparison among six CS networks on the 200 full-resolution large-sized images of Test2K and Test4K \cite{kong2021classsr}, generated from the DIV8K \cite{gu2019div8k} dataset, with three settings of $\gamma\in\{10\%,30\%,50\%\}$. }
\vspace{-8pt}
\label{tab:compare_sota_high_resolution}
\centering
\hspace{-4pt}
\resizebox{0.48\textwidth}{!}{
\begin{tabular}{lc|ccc|ccc}
\shline
\rowcolor[HTML]{EFEFEF} 
\multicolumn{1}{l|}{\cellcolor[HTML]{EFEFEF}} &
  \begin{tabular}[c]{@{}c@{}}Test Set\\ (Standard\\ Resolution)\end{tabular} &
  \multicolumn{3}{c|}{\cellcolor[HTML]{EFEFEF}\begin{tabular}[c]{@{}c@{}}Test2K \cite{gu2019div8k,kong2021classsr}\\ ($2048\times 1080$)\end{tabular}} &
  \multicolumn{3}{c}{\cellcolor[HTML]{EFEFEF}\begin{tabular}[c]{@{}c@{}}Test4K \cite{gu2019div8k,kong2021classsr}\\ ($3840\times 2160$)\end{tabular}} \\ \hhline{>{\arrayrulecolor[HTML]{EFEFEF}}->{\arrayrulecolor{black}}|-------} 
\rowcolor[HTML]{EFEFEF} 
\multicolumn{1}{l|}{\multirow{-4}{*}{\cellcolor[HTML]{EFEFEF}Method}} & CS Ratio $\gamma$ & 10\%  & 30\%  & 50\%  & 10\%  & 30\%  & 50\%  \\ \hline \hline
\multicolumn{2}{l|}{ISTA-Net$^+$ \cite{zhang2018ista}}                                                & 25.08 & 29.89 & 33.59 & 26.32 & 31.62 & 35.48 \\
\multicolumn{2}{l|}{ISTA-Net$^{++}$ \cite{you2021ista}}                                             & 26.00 & 30.67 & 34.45 & 27.32 & 32.45 & 36.35 \\
\multicolumn{2}{l|}{FSOINet \cite{chen2022fsoinet}}                                                     & \textcolor{green}{28.08} & \textcolor{green}{33.19} & \textcolor{green}{37.30} & \textcolor{green}{29.56} & \textcolor{green}{35.12} & \textcolor{green}{39.37} \\
\multicolumn{2}{l|}{MR-CCSNet$^{+}$ \cite{fan2022global}}                                             & -     & -     & 35.70 & -     & -     & 37.86 \\ \hline \hline
\multicolumn{2}{l|}{\textbf{PC-CNN (Ours)}}                                               & \textcolor{blue}{29.36} & \textcolor{blue}{35.65} & \textcolor{blue}{40.71} & \textcolor{blue}{31.21} & \textcolor{blue}{37.73} & \textcolor{blue}{42.57}      \\
\multicolumn{2}{l|}{\textbf{PCT (Ours)}}                                                  & \textcolor{red}{29.89} & \textcolor{red}{36.40} & \textcolor{red}{42.07} &  \textcolor{red}{31.78} & \textcolor{red}{38.37} & \textcolor{red}{43.59}\\ \shline
\end{tabular}}
\vspace{-10pt}
\end{table}

\begin{figure}[!t]
\centering
\vspace{-2pt}
\hspace{-2pt}\includegraphics[width=0.48\textwidth]{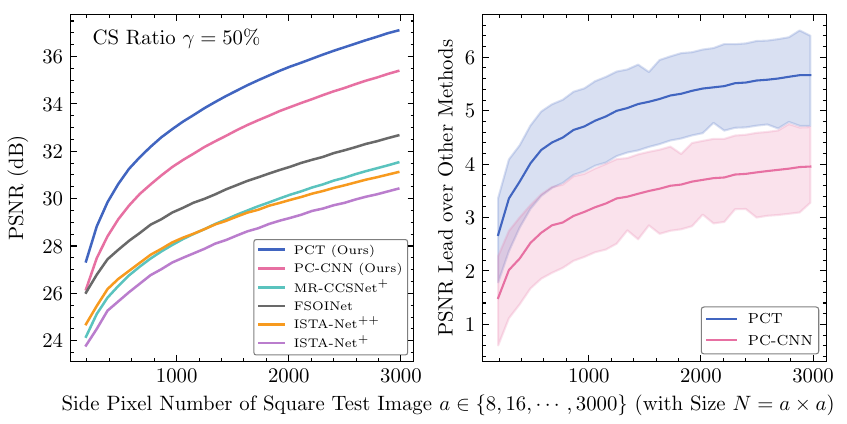}
\vspace{-10pt}
\caption{A comparison of our 375 PSNR evaluations for six CS networks on the 100 downscaled images of size $a\times a$ from Test8K \cite{kong2021classsr}. \textcolor{blue}{\textbf{(1) Left:}} As the side pixel number $a$ increases, the PSNR values of all networks become higher, and the internal differences of PCNets and other methods tend to be stable. \textcolor{blue}{\textbf{(2) Right:}} The PSNR lead of PCNets over the other four networks becomes larger, from about 1-3dB to 3-6dB.}
\label{fig:high_resolution_PSNR_curve}
\vspace{-10pt}
\end{figure}

\begin{figure}[!t]
\centering
\hspace{-2pt}\includegraphics[width=0.48\textwidth]{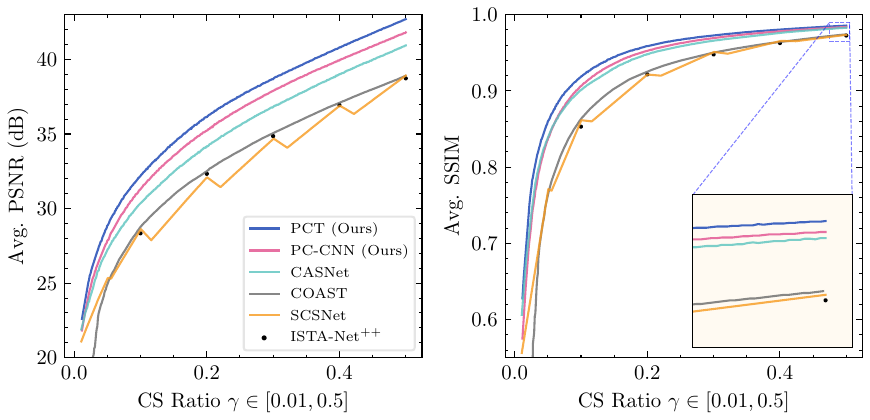}
    \vspace{-10pt}
\caption{A comparison of six scalable deep CS approaches on the Set11 \cite{kulkarni2016reconnet} benchmark, with a widely adopted ratio range of $\gamma\in[0.01,0.5]$.}
\label{fig:scalable_curve}
\vspace{-10pt}
\end{figure}

\begin{table*}[!t]
\vspace{-5pt}
\caption{An ablation study of our structural designs on Set11 \cite{kulkarni2016reconnet}, started from the default PC-CNN version in (1). In (2)-(7), we gradually reduce our deep filtering network to an identity. In (8)-(11), we evaluate four under-sampling designs. In (12)-(13), we train two PC-CNN variants with $K\in\{10,30\}$. For each case, the average PSNR (dB) under $\gamma\in\{10\%,30\%,50\%\}$ and parameter number (M) are provided with impacts highlighted in purple.}
\label{tab:ablation_sampling_operator}
\vspace{-8pt}
\centering
\hspace{-2pt}\resizebox{1.0\textwidth}{!}{
\begin{tabular}{cl|ccc|c}
\shline
\rowcolor[HTML]{EFEFEF} 
\multicolumn{2}{c|}{\cellcolor[HTML]{EFEFEF}} &
  \multicolumn{3}{c|}{\cellcolor[HTML]{EFEFEF}CS Ratio $\gamma$} &
  \cellcolor[HTML]{EFEFEF} \\ \hhline{>{\arrayrulecolor[HTML]{EFEFEF}}-->{\arrayrulecolor{black}}|--->{\arrayrulecolor[HTML]{EFEFEF}}->{\arrayrulecolor{black}}}
\rowcolor[HTML]{EFEFEF} 
\multicolumn{2}{c|}{\multirow{-2}{*}{\cellcolor[HTML]{EFEFEF}Experimental Setting}} &
  10\% &
  30\% &
  50\% &
  \multirow{-2}{*}{\cellcolor[HTML]{EFEFEF}\#Param.} \\ \hline \hline
(1) &
  \textbf{PC-CNN} (Our default convolutional PCNet version) &
  31.39 &
  37.91 &
  41.83 &
  1.16 \\ \hline \hline
(2) &
  Share the two filtered images $\mathbf{X}_D$ and $\mathbf{X}_G$ on (1) &
  31.23~{\scriptsize(\textcolor{purple}{-0.16})} &
  37.79~{\scriptsize(\textcolor{purple}{-0.12})} &
  41.67~{\scriptsize(\textcolor{purple}{-0.16})} &
  1.16~{\scriptsize(\textcolor{purple}{-0.00})} \\
(3) &
  Remove the conditional CS ratio information from (2) &
  31.13~{\scriptsize(\textcolor{purple}{-0.10})} &
  37.50~{\scriptsize(\textcolor{purple}{-0.11})} &
  41.22~{\scriptsize(\textcolor{purple}{-0.45})} &
  1.16~{\scriptsize(\textcolor{purple}{-0.00})} \\
(4) &
  Remove the two high-throughput skip connections in RS of (3) &
  31.06~{\scriptsize(\textcolor{purple}{-0.07})} &
  37.40~{\scriptsize(\textcolor{purple}{-0.10})} &
  41.10~{\scriptsize(\textcolor{purple}{-0.08})} &
  1.16~{\scriptsize(\textcolor{purple}{-0.00})} \\
(5) &
  Reduce the middle convolution layer number of (4) from 5 to 3 &
  30.90~{\scriptsize(\textcolor{purple}{-0.16})} &
  37.26~{\scriptsize(\textcolor{purple}{-0.14})} &
  40.99~{\scriptsize(\textcolor{purple}{-0.11})} &
  1.14~{\scriptsize(\textcolor{purple}{-0.02})} \\
(6) &
  Reduce the middle convolution layer number of (5) from 3 to 1 &
  30.71~{\scriptsize(\textcolor{purple}{-0.19})} &
  37.16~{\scriptsize(\textcolor{purple}{-0.10})} &
  40.87~{\scriptsize(\textcolor{purple}{-0.12})} &
  1.12~{\scriptsize(\textcolor{purple}{-0.02})} \\
(7) &
  Remove the whole filtering network from (6) ($\mathcal{G}_\mathbf{A}(\cdot)=[\mathbf{M}_D\mathbf{\Phi}_D(\cdot),\mathbf{M}_G\mathbf{\Phi}_G\mathbf{P}_G(\cdot)]$) &
  30.42~{\scriptsize(\textcolor{purple}{-0.29})} &
  36.94~{\scriptsize(\textcolor{purple}{-0.22})} &
  40.65~{\scriptsize(\textcolor{purple}{-0.22})} &
  1.11~{\scriptsize(\textcolor{purple}{-0.01})} \\ \hline \hline
(8) &
  Remove the D-branch from (7) ($\mathcal{G}_\mathbf{A}(\cdot)=\mathbf{M}_G\mathbf{\Phi}_G\mathbf{P}_G(\cdot)$) &
  29.97~{\scriptsize(\textcolor{purple}{-0.45})} &
  36.60~{\scriptsize(\textcolor{purple}{-0.34})} &
  40.32~{\scriptsize(\textcolor{purple}{-0.33})} &
  1.11~{\scriptsize(\textcolor{purple}{-0.00})} \\
(9) &
  Remove the random permutation $\mathbf{P}_G$ from (7) ($\mathcal{G}_\mathbf{A}(\cdot)=[\mathbf{M}_D\mathbf{\Phi}_D(\cdot),\mathbf{M}_G\mathbf{\Phi}_G(\cdot)]$) &
  29.31~{\scriptsize(\textcolor{purple}{-1.11})} &
  35.62~{\scriptsize(\textcolor{purple}{-1.32})} &
  39.42~{\scriptsize(\textcolor{purple}{-1.23})} &
  1.11~{\scriptsize(\textcolor{purple}{-0.00})} \\
(10) &
  Remove the G-branch from (7) ($\mathcal{G}_\mathbf{A}(\cdot)=\mathbf{M}_D\mathbf{\Phi}_D(\cdot)$) &
  29.43~{\scriptsize(\textcolor{purple}{-0.99})} &
  35.69~{\scriptsize(\textcolor{purple}{-1.25})} &
  40.05~{\scriptsize(\textcolor{purple}{-0.60})} &
  1.11~{\scriptsize(\textcolor{purple}{-0.00})} \\
(11) &
  Remove the D-branch and random permutation $\mathbf{P}_G$ from (7) ($\mathcal{G}_\mathbf{A}(\cdot)=\mathbf{M}_G\mathbf{\Phi}_G(\cdot)$) &
  28.91~{\scriptsize(\textcolor{purple}{-1.51})} &
  35.28~{\scriptsize(\textcolor{purple}{-1.66})} &
  39.14~{\scriptsize(\textcolor{purple}{-1.51})} &
  1.11~{\scriptsize(\textcolor{purple}{-0.00})} \\ \hline \hline
(12) &
  Reduce the stage number $K$ of (1) from 20 to 10 &
  31.06~{\scriptsize(\textcolor{purple}{-0.33})} &
  37.63~{\scriptsize(\textcolor{purple}{-0.28})} &
  41.56~{\scriptsize(\textcolor{purple}{-0.27})} &
  0.61~{\scriptsize(\textcolor{purple}{-0.55})} \\
(13) &
  Increase the stage number $K$ of (1) from 20 to 30 &
  31.54~{\scriptsize(\textcolor{purple}{+0.15})} &
  38.19~{\scriptsize(\textcolor{purple}{+0.28})} &
  41.97~{\scriptsize(\textcolor{purple}{+0.14})} &
  1.72~{\scriptsize(\textcolor{purple}{+0.56})} \\ \shline
\end{tabular}}
\vspace{-10pt}
\end{table*}

\subsubsection{Comparison on High-Resolution Benchmark Images}

To compare the performance of existing methods and PCNets in recovering large-sized images, we select four typical networks: ISTA-Net$^+$ \cite{zhang2018ista}, ISTA-Net$^{++}$ \cite{you2021ista}, FSOINet \cite{chen2022fsoinet}, and MR-CCSNet$^+$ \cite{fan2022global}, which achieve best results under the settings of block-diagonal matrices and stacked linear network, respectively. Two high-resolution benchmarks are employed for our evaluation: Test2K and Test4K \cite{kong2021classsr}, which contain 200 large images and are generated from the DIV8K \cite{gu2019div8k} dataset. In Tab.~\ref{tab:compare_sota_high_resolution} and Fig.~\ref{fig:comparison_high_resolution}, our comparisons on full-resolution images present that PCNets achieve significant PSNR distances of 1.28-4.77dB to other networks. They can better recover complex structures and fine textures.

In Fig.~\ref{fig:high_resolution_PSNR_curve}, we further compare the above six CS networks on Test8K \cite{gu2019div8k,kong2021classsr} dataset with different test image sizes. Specifically, following \cite{kong2021classsr}, we perform 375 evaluations on the 100 bicubic downscaled image versions of size $a\times a$, with $a\in\{8,16,\cdots,3000\}$. We observe that our designed PCNets consistently outperform the other methods in nearly all cases. As image size grows, the internal PSNR differences within PCNets and within the other four methods remain stable, yet the PSNR lead of PCNet significantly widens (from 1-3dB to 3-6dB), verifying the superiority of our global sampling design for large images.

\subsubsection{Comparison of Scalable Image CS}

We select four scalable networks: ISTA-Net$^{++}$ \cite{you2021ista}, SCSNet \cite{shi2019scalable}, COAST \cite{you2021coast}, and CASNet \cite{chen2022content} for comparison, in which ISTA-Net$^{++}$ supports five ratios $\gamma\in\{10\%,20\%,\cdots,50\%\}$, SCSNet is equipped with a greedy algorithm, to construct the sampling matrix for a specific CS ratio, COAST can handle arbitrary block-based Gaussian matrices, and CASNet learns a complete generating matrix for content-aware CS sampling. Fig.~\ref{fig:scalable_curve} reports the PSNR/SSIM results of six methods with $\gamma$~$\in$~$[0.01, 0.5]$. We observe that PCNets consistently outperform others in nearly all cases. They exhibit a large PSNR lead over the other four methods as the ratio increases. It demonstrates that our ratio-conditioned collaborative sampling operator design is effective in making PCNets scalable, flexible, and robust as $\gamma$ varies.

\begin{figure}[!t]
\centering
\vspace{-5pt}
\hspace{-2pt}\includegraphics[width=0.46\textwidth]{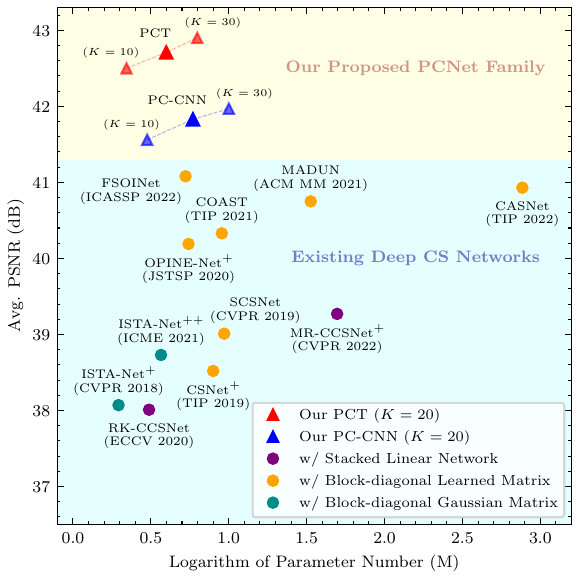}
    \vspace{-10pt}
\caption{A comparison of PSNR and parameter number among our six PCNets with $K\in\{10,20,30\}$ and eleven existing networks on Set11 \cite{kulkarni2016reconnet} under $\gamma=50\%$. The previous methods are classified and highlighted in different colors according to their adopted sampling operators.}
\label{fig:comp_param_PSNR}
\vspace{-10pt}
\end{figure}

\begin{table*}[!t]
\vspace{-5pt}
\caption{Visualizations of the equivalent kernels $\{\mathbf{K}_D,\mathbf{K}_G\}$ and filtered image versions $\{\mathbf{X}_D,\mathbf{X}_G\}$ of our learned deep conditional filtering network in PC-CNN and PCT for D-branch and G-branch, respectively, under five CS ratio settings $\gamma\in\{0.1\%,1\%,10\%,50\%,100\%\}$. The equivalent kernels are extracted by Algo.~\ref{alg:matrix_extraction}, and the filtering evaluation is performed on a benchmark image named ``Butterfly". In kernel visualizations, the darker red/blue colors indicate positive/negative weights with higher absolute values. In image visualizations, the lighter pixels indicate larger values.}
\vspace{-8pt}
\label{tab:visualize_learned_filter}
\centering
\hspace{-4pt}
\setlength{\tabcolsep}{2pt}
\resizebox{1.0\textwidth}{!}{
\begin{tabular}{l|ccccc|ccccc}
\shline
\rowcolor[HTML]{EFEFEF} 
Our Learned PCNet Model           & \multicolumn{5}{c|}{\cellcolor[HTML]{EFEFEF}PC-CNN} & \multicolumn{5}{c}{\cellcolor[HTML]{EFEFEF}PCT} \\ \hline
\rowcolor[HTML]{EFEFEF} 
CS Ratio $\gamma$ (with our default setting $\gamma_D=0.4\gamma$ and $\gamma_G=0.6\gamma$) & 0.1\%        & 1\%       & 10\%   & 50\%     & 100\%       & 0.1\%       & 1\%      & 10\%   & 50\%    & 100\%      \\ \hline \hline
\raisebox{2.5\height}{Deep learned kernel $\mathbf{K}_D$ (2D ``heatmap" visualization)} & \includegraphics[width=0.065\textwidth]{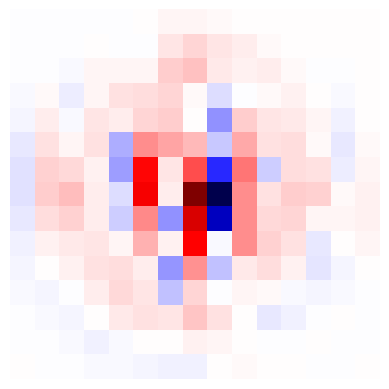} & \includegraphics[width=0.065\textwidth]{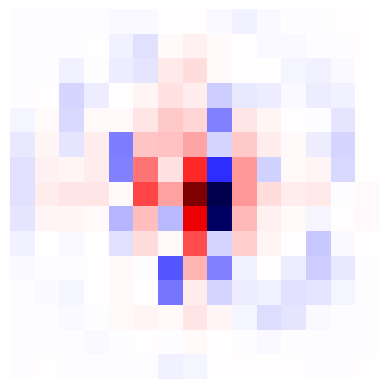} & \includegraphics[width=0.065\textwidth]{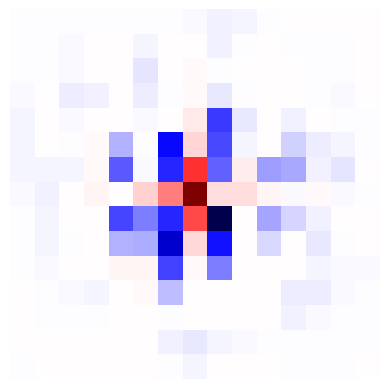} & \includegraphics[width=0.065\textwidth]{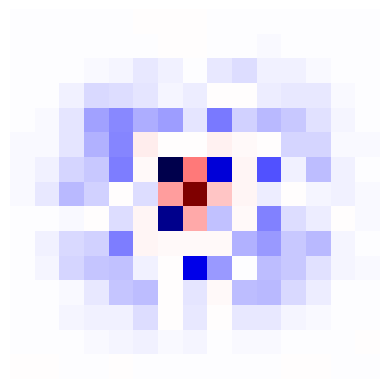} & \includegraphics[width=0.065\textwidth]{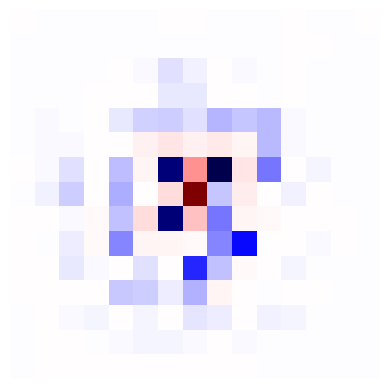} & \includegraphics[width=0.065\textwidth]{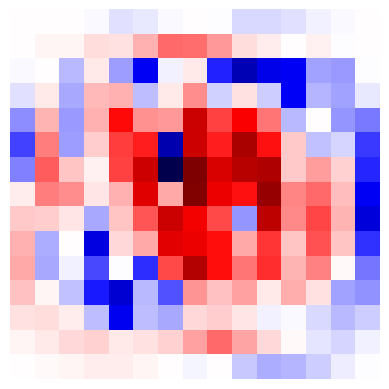} & \includegraphics[width=0.065\textwidth]{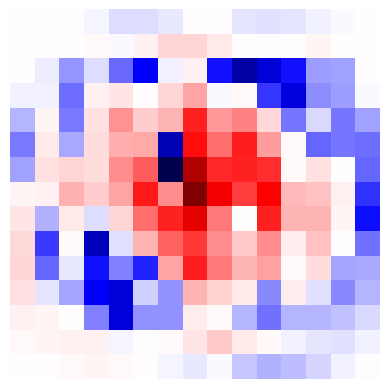} & \includegraphics[width=0.065\textwidth]{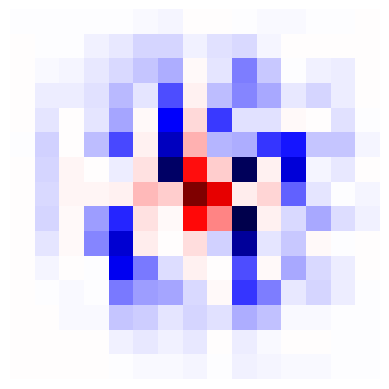} & \includegraphics[width=0.065\textwidth]{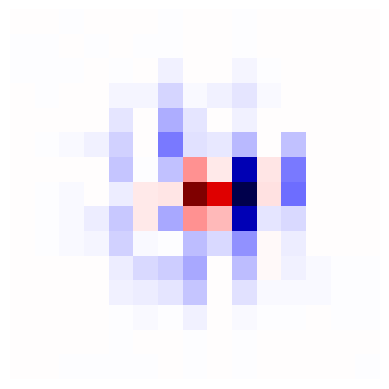} & \includegraphics[width=0.065\textwidth]{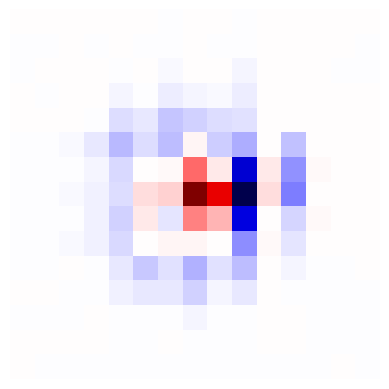} \\
\raisebox{2.5\height}{Deep learned kernel $\mathbf{K}_D$ (3D ``surface" visualization)} & \includegraphics[width=0.065\textwidth]{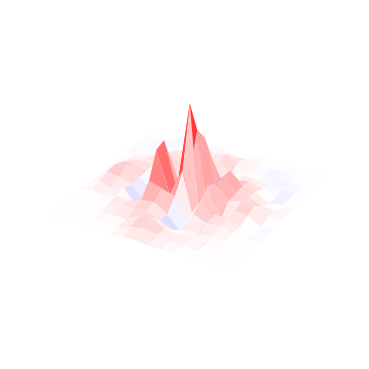} & \includegraphics[width=0.065\textwidth]{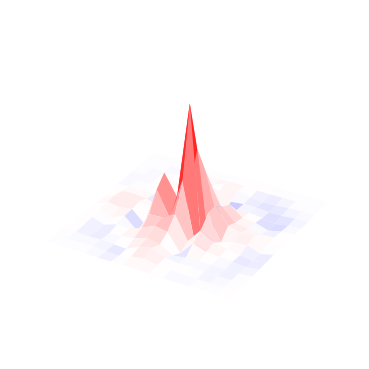} & \includegraphics[width=0.065\textwidth]{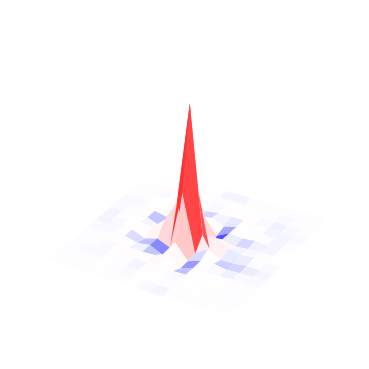} & \includegraphics[width=0.065\textwidth]{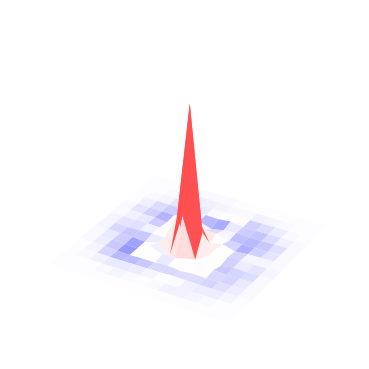} & \includegraphics[width=0.065\textwidth]{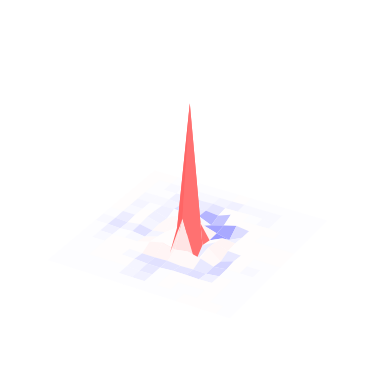} & \includegraphics[width=0.065\textwidth]{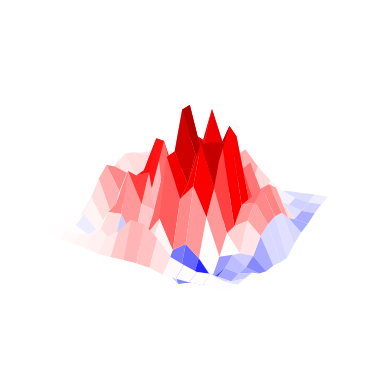} & \includegraphics[width=0.065\textwidth]{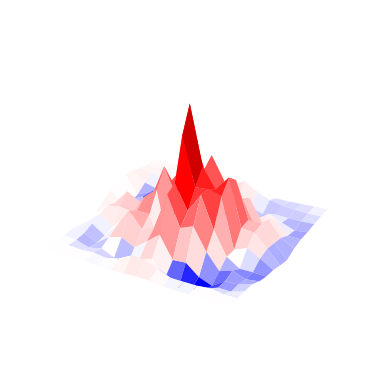} & \includegraphics[width=0.065\textwidth]{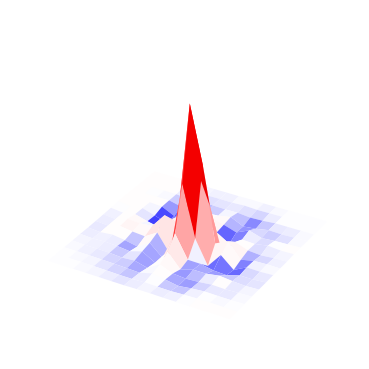} & \includegraphics[width=0.065\textwidth]{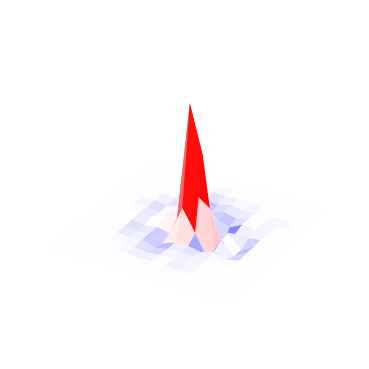} & \includegraphics[width=0.065\textwidth]{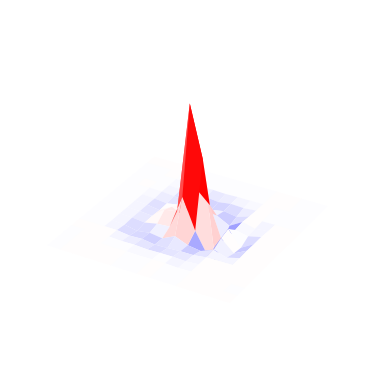} \\ 
\raisebox{2.5\height}{Adaptively filtered image $\mathbf{X}_D$ (Spatial-domain visualization)} & \includegraphics[width=0.065\textwidth]{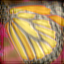} & \includegraphics[width=0.065\textwidth]{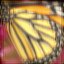} & \includegraphics[width=0.065\textwidth]{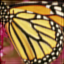} & \includegraphics[width=0.065\textwidth]{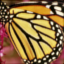} & \includegraphics[width=0.065\textwidth]{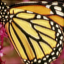} & \includegraphics[width=0.065\textwidth]{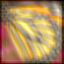} & \includegraphics[width=0.065\textwidth]{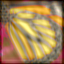} & \includegraphics[width=0.065\textwidth]{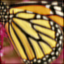} & \includegraphics[width=0.065\textwidth]{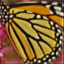} & \includegraphics[width=0.065\textwidth]{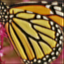} \\
\raisebox{2.5\height}{Adaptively filtered image $\mathbf{X}_D$ (Frequency-domain visualization)} & \includegraphics[width=0.065\textwidth]{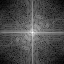} & \includegraphics[width=0.065\textwidth]{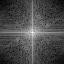} & \includegraphics[width=0.065\textwidth]{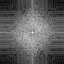} & \includegraphics[width=0.065\textwidth]{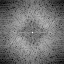} & \includegraphics[width=0.065\textwidth]{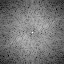} & \includegraphics[width=0.065\textwidth]{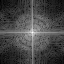} & \includegraphics[width=0.065\textwidth]{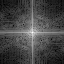} & \includegraphics[width=0.065\textwidth]{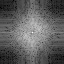} & \includegraphics[width=0.065\textwidth]{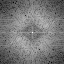} & \includegraphics[width=0.065\textwidth]{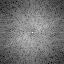} \\ \hline \hline
\raisebox{2.5\height}{Deep learned kernel $\mathbf{K}_G$ (2D ``heatmap" visualization)} & \includegraphics[width=0.065\textwidth]{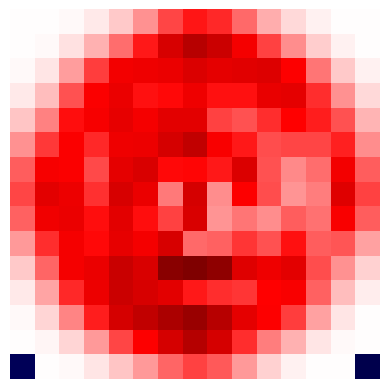} & \includegraphics[width=0.065\textwidth]{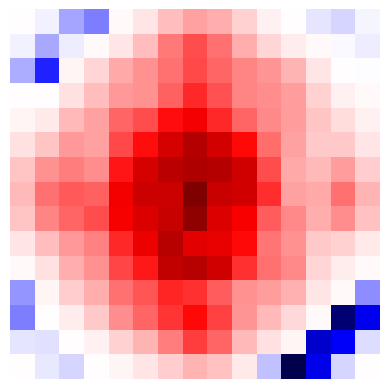} & \includegraphics[width=0.065\textwidth]{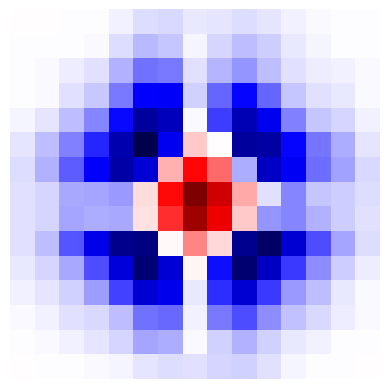} & \includegraphics[width=0.065\textwidth]{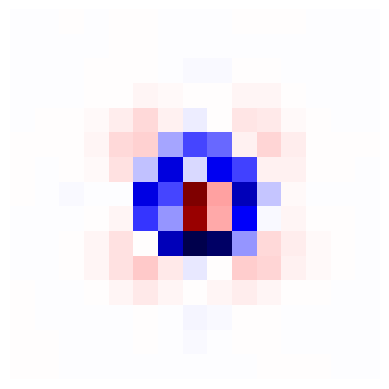} & \includegraphics[width=0.065\textwidth]{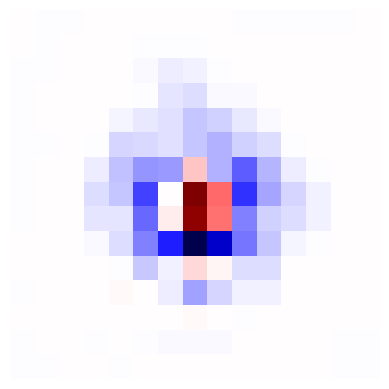} & \includegraphics[width=0.065\textwidth]{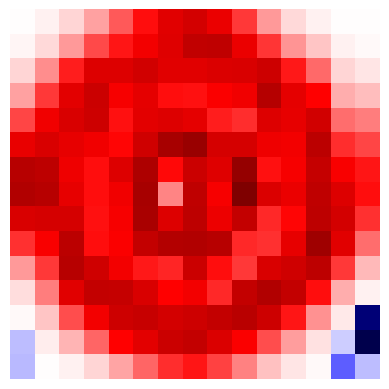} & \includegraphics[width=0.065\textwidth]{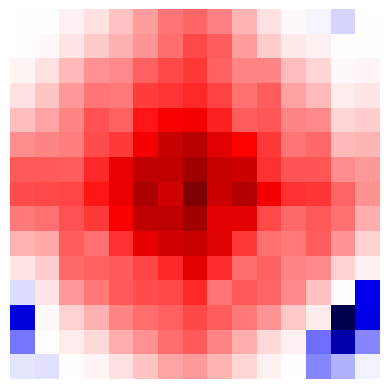} & \includegraphics[width=0.065\textwidth]{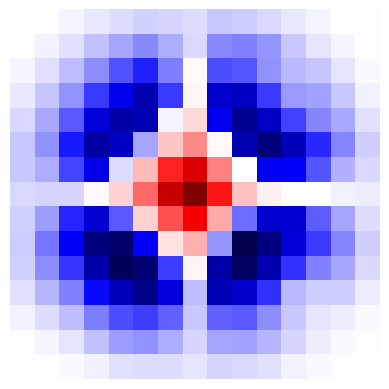} & \includegraphics[width=0.065\textwidth]{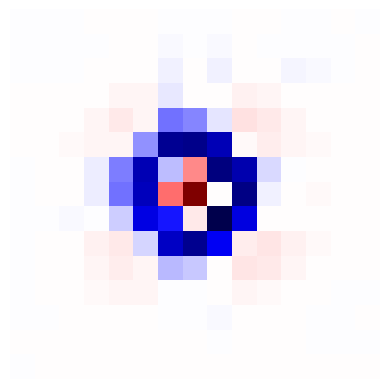} & \includegraphics[width=0.065\textwidth]{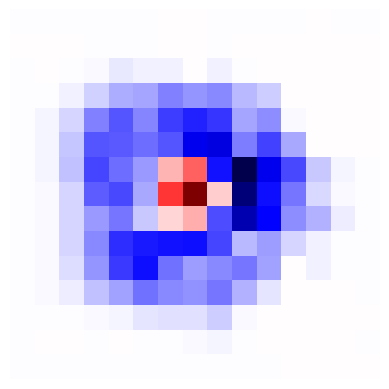} \\
\raisebox{2.5\height}{Deep learned kernel $\mathbf{K}_G$ (3D ``surface" visualization)} & \includegraphics[width=0.065\textwidth]{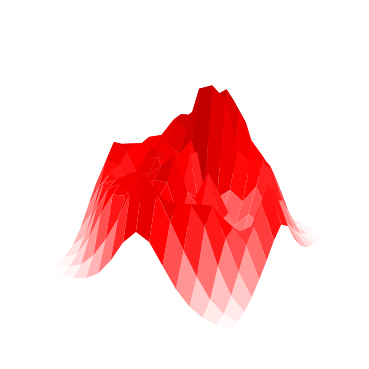} & \includegraphics[width=0.065\textwidth]{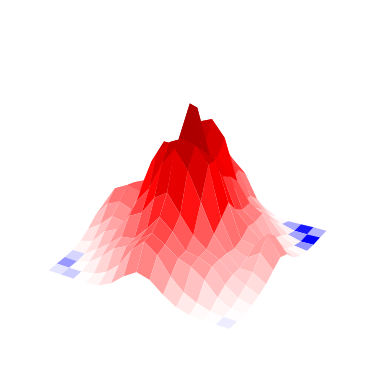} & \includegraphics[width=0.065\textwidth]{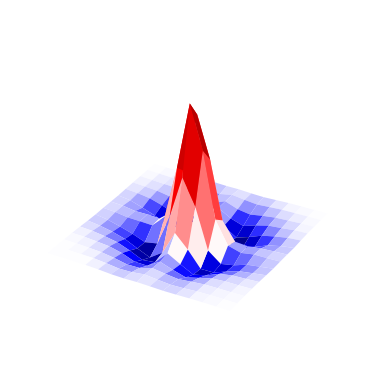} & \includegraphics[width=0.065\textwidth]{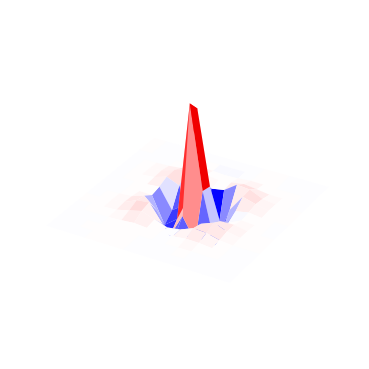} & \includegraphics[width=0.065\textwidth]{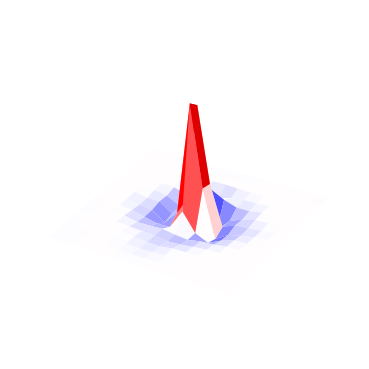} & \includegraphics[width=0.065\textwidth]{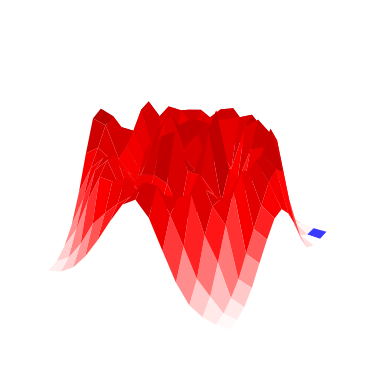} & \includegraphics[width=0.065\textwidth]{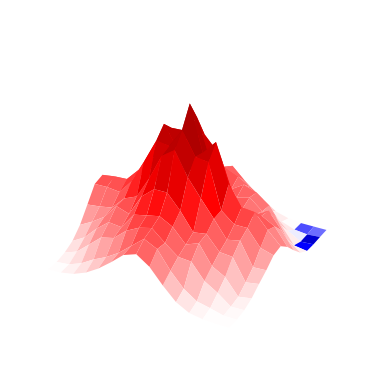} & \includegraphics[width=0.065\textwidth]{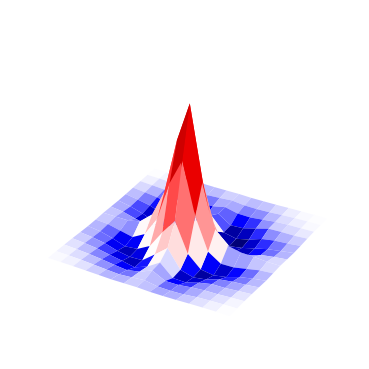} & \includegraphics[width=0.065\textwidth]{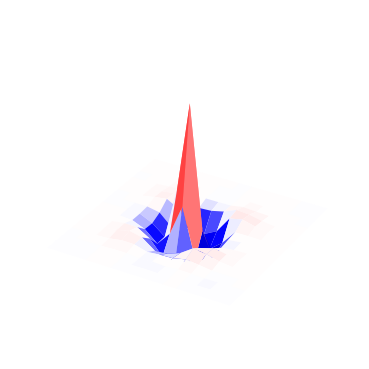} & \includegraphics[width=0.065\textwidth]{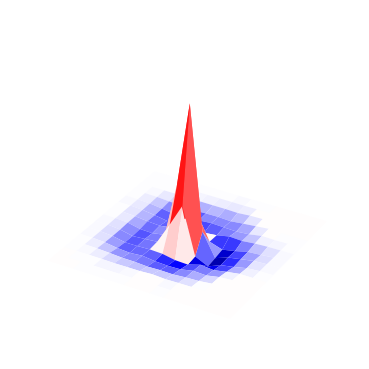} \\ 
\raisebox{2.5\height}{Adaptively filtered image $\mathbf{X}_G$ (Spatial-domain visualization)} & \includegraphics[width=0.065\textwidth]{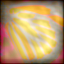} & \includegraphics[width=0.065\textwidth]{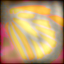} & \includegraphics[width=0.065\textwidth]{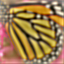} & \includegraphics[width=0.065\textwidth]{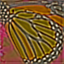} & \includegraphics[width=0.065\textwidth]{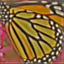} & \includegraphics[width=0.065\textwidth]{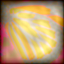} & \includegraphics[width=0.065\textwidth]{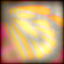} & \includegraphics[width=0.065\textwidth]{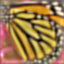} & \includegraphics[width=0.065\textwidth]{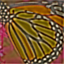} & \includegraphics[width=0.065\textwidth]{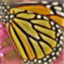} \\
\raisebox{2.5\height}{Adaptively filtered image $\mathbf{X}_G$ (Frequency-domain visualization)} & \includegraphics[width=0.065\textwidth]{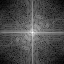} & \includegraphics[width=0.065\textwidth]{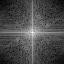} & \includegraphics[width=0.065\textwidth]{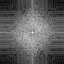} & \includegraphics[width=0.065\textwidth]{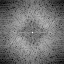} & \includegraphics[width=0.065\textwidth]{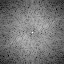} & \includegraphics[width=0.065\textwidth]{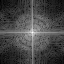} & \includegraphics[width=0.065\textwidth]{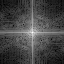} & \includegraphics[width=0.065\textwidth]{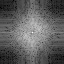} & \includegraphics[width=0.065\textwidth]{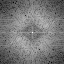} & \includegraphics[width=0.065\textwidth]{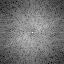} \\ \shline
\end{tabular}}
\vspace{-10pt}
\end{table*}

\subsection{Ablation Studies and Discussions}

\subsubsection{Effect of Our Sampling Operator Designs}

To verify the effect of structural designs in our proposed collaborative sampling operator, we begin with PC-CNN and perform ten fine-grained sequential experiments, progressively simplifying our operator towards the traditional block-based Gaussian matrix. The settings and results are reported in Tab.~\ref{tab:ablation_sampling_operator}. We first share the two filtered images $\{\mathbf{X}_D,\mathbf{X}_G\}$ for dual-branch sampling, eliminate the CS ratio-conditioned modulating layers for the middle five convolutions, and remove our two supplemental feature-level paths for RS (see the red dotted lines in Figs.~\ref{fig:sampling_operator} (a) and \ref{fig:stage} (c)). These changes bring a substantial average PSNR drop of 0.45dB. Subsequently, we remove our deep filtering network, achieving a 0.05M parameter saving at the cost of a further 0.52dB average PSNR drop. The significant impact of our filtering network can be attributed to two main factors. First, it effectively smooths out non-dominant details to enhance the sampling of dominant structural components, thereby increasing the amount of critical information in the measurements, especially at low ratios. Second, it adaptively supplements the learned high-throughput physics information about the forward process and sampling rates, improving recovery performance.

Based on the current PC-CNN variant without a deep filtering step in Tab.~\ref{tab:ablation_sampling_operator} (7), we evaluate the effect of our dual-branch sampling design. Specifically, we build four PC-CNN variants: (8) using only the G-branch (\textit{i.e.}, the scrambled block-diagonal Gaussian matrix), (9) without global permutation, (10) using only the D-branch (\textit{i.e.}, the global DCT basis), and (11) using the traditional block-diagonal Gaussian matrix. The results in Tab.~\ref{tab:ablation_sampling_operator} show significant PSNR drops of approximately 0.37dB, 1.22dB, 0.95dB, and 1.56dB, respectively. These results validate the effectiveness of our dual-branch design in achieving a better balance of low-, mid-, and high-frequency components with global perception. Notably, our PC-CNN with the scrambled matrix in Tab.~\ref{tab:ablation_sampling_operator} (8) achieves a $>$1dB PSNR lead over the variant with the traditional block-diagonal matrix in Tab.~\ref{tab:ablation_sampling_operator} (11).

\begin{table}[!t]
\vspace{-5pt}
\caption{An experiment evaluated on Set11 \cite{kulkarni2016reconnet} with $\gamma =50\%$ of sequentially introducing our ordinary training setting (in Sec.~\ref{subsec:baseline_setup}), proposed deep sampling operator (in Sec.~\ref{subsec:sampling_operator_architecture}) and enhancement strategies (in Sec.~\ref{subsec:training_enhancements}) to boost two existing deep unrolled networks \cite{you2021ista,chen2022fsoinet}.}
\vspace{-8pt}
\label{tab:boost_existing_methods}
\centering
\hspace{-4pt}
\resizebox{0.48\textwidth}{!}{
\begin{tabular}{cl|c|c}
\shline
\rowcolor[HTML]{EFEFEF} 
\multicolumn{2}{c|}{\cellcolor[HTML]{EFEFEF}}                                         & CS Ratio $\gamma$ & \cellcolor[HTML]{EFEFEF}                            \\ \hhline{>{\arrayrulecolor[HTML]{EFEFEF}}-->{\arrayrulecolor{black}}|->{\arrayrulecolor[HTML]{EFEFEF}}->{\arrayrulecolor{black}}}
\rowcolor[HTML]{EFEFEF} 
\multicolumn{2}{c|}{\multirow{-2}{*}{\cellcolor[HTML]{EFEFEF}Experimental Setting}}   & 50\%              & \multirow{-2}{*}{\cellcolor[HTML]{EFEFEF}\#Param.} \\ \hline \hline
(1) & ISTA-Net$^{++}$ \cite{you2021ista} (The default version) & 38.73 & 0.76 \\
(2) & Train (1) under our ordinary setting            & 38.80~{\scriptsize(\textcolor{purple}{+0.07})} & 0.76~{\scriptsize(\textcolor{purple}{-0.00})} \\
(3) & Introduce our $\{\mathcal{G}_\mathbf{A},\mathcal{G}_{\mathbf{A}^\top}\}$ to (2) & 41.05~{\scriptsize(\textcolor{purple}{+1.25})}             & 0.83~{\scriptsize(\textcolor{purple}{+0.07})}                                                \\
(4) & Train (3) under our enhanced setting            & 41.22~{\scriptsize(\textcolor{purple}{+0.17})} & 0.83~{\scriptsize(\textcolor{purple}{-0.00})} \\ \hline \hline
(5) & FSOINet \cite{chen2022fsoinet} (The default version)         & 41.08 & 1.06 \\
(6) & Train (5) under our ordinary setting            & 41.11~{\scriptsize(\textcolor{purple}{+0.03})} & 1.06~{\scriptsize(\textcolor{purple}{-0.00})} \\
(7) & Introduce our $\{\mathcal{G}_\mathbf{A},\mathcal{G}_{\mathbf{A}^\top}\}$ to (6)                         & 41.37~{\scriptsize(\textcolor{purple}{+0.26})} & 0.55~{\scriptsize(\textcolor{purple}{-0.51})} \\
(8) & Train (7) under our enhanced setting            & 41.49~{\scriptsize(\textcolor{purple}{+0.12})} & 0.55~{\scriptsize(\textcolor{purple}{-0.00})} \\ \shline
\end{tabular}}
\vspace{-10pt}
\end{table}

\begin{table*}[!t]
\vspace{-5pt}
\caption{Visualizations of the equivalent global matrix forms (denoted by $\mathbf{A}$s, see Fig.~\ref{fig:sampling_operator_comparison} (a)) of eight sampling operators \cite{you2021ista,chen2022fsoinet,fan2022global}: global dense (G. D.) Gaussian/DCT/Hadamard matrices, scrambled (S.) block-diagonal (B.) Gaussian/learned matrices, learned stacked linear (S. L.) network, and our proposed COSO learned by PC-CNN. The image (DMD) size is set to $96\times 96$. \textcolor{blue}{\textbf{(1) Top:}} We uniformly extract five rows of $\mathbf{A}$s and reshape them to the size of $96\times 96$ with target ratio $\gamma^\prime=10\%$. All their spatial/frequency representations are provided. \textcolor{blue}{\textbf{(2) Middle:}} We find that all the $\mathbf{A}$s have full rank and can approximately satisfy the diagonal form $\mathbf{AA}^\top\approx diag([q_1,\cdots,q_M])$, but they exhibit different element distributions. \textcolor{blue}{\textbf{(3) Bottom:}} The effective receptive field (ERF) is evaluated on an image named ``House" by computing the partial derivative of a measurement to the input.} \vspace{-8pt}
\label{tab:visualize_equivalent_matrix}
\centering
\hspace{-4pt}
\setlength{\tabcolsep}{1pt}
\resizebox{1.0\textwidth}{!}{
\begin{tabular}{l|cc|cc|cc|cc|cc|cc|cc|cc}
\shline
\rowcolor[HTML]{EFEFEF} 
\cellcolor[HTML]{EFEFEF} &
  \multicolumn{2}{c|}{\cellcolor[HTML]{EFEFEF}G. D. Gaussian} &
  \multicolumn{2}{c|}{\cellcolor[HTML]{EFEFEF}G. D. DCT} &
  \multicolumn{2}{c|}{\cellcolor[HTML]{EFEFEF}G. D. Hadamard} &
  \multicolumn{2}{c|}{\cellcolor[HTML]{EFEFEF}B. Gaussian} &
  \multicolumn{2}{c|}{\cellcolor[HTML]{EFEFEF}B. Leanred} &
  \multicolumn{2}{c|}{\cellcolor[HTML]{EFEFEF}S. B. Gaussian} &
  \multicolumn{2}{c|}{\cellcolor[HTML]{EFEFEF}S. L. Network} &
  \multicolumn{2}{c}{\cellcolor[HTML]{EFEFEF}\textbf{Our COSO}} \\ \hhline{>{\arrayrulecolor[HTML]{EFEFEF}}->{\arrayrulecolor{black}}|---------------->{\arrayrulecolor[HTML]{EFEFEF}}>{\arrayrulecolor{black}}} 
\rowcolor[HTML]{EFEFEF} 
\multirow{-2}{*}{\cellcolor[HTML]{EFEFEF}Sampling Matrix (Operator)} &
 {\scriptsize Spatial} &
 {\scriptsize Frequency} &
{\scriptsize  Spatial }&
 {\scriptsize Frequency }&
 {\scriptsize Spatial }&
 {\scriptsize Frequency }&
 {\scriptsize Spatial }&
 {\scriptsize Frequency }&
 {\scriptsize Spatial }&
 {\scriptsize Frequency }&
 {\scriptsize Spatial }&
 {\scriptsize Frequency }&
 {\scriptsize Spatial }&
 {\scriptsize Frequency }&
 {\scriptsize Spatial }&
 {\scriptsize Frequency }\\ \hline \hline
\raisebox{2\height}{Row of $\mathbf{A}$ at $\gamma=(1/5)\gamma^\prime$} &
 \includegraphics[width=0.05\textwidth]{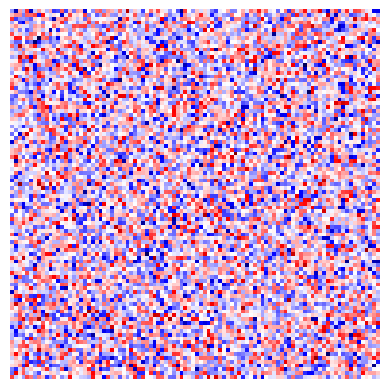}  &
 \includegraphics[width=0.05\textwidth]{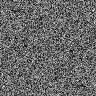}  &
\includegraphics[width=0.05\textwidth]{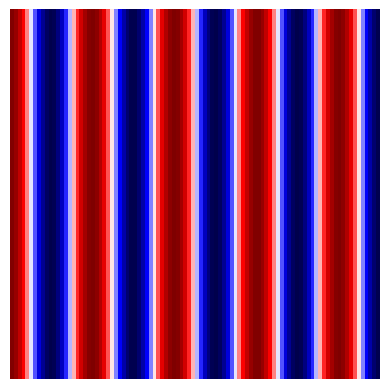}  &
 \includegraphics[width=0.05\textwidth]{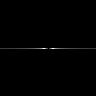}  &
\includegraphics[width=0.05\textwidth]{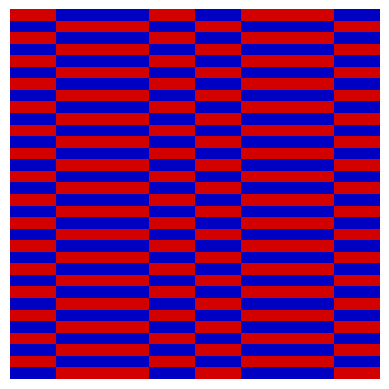}  &
 \includegraphics[width=0.05\textwidth]{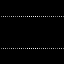}  &
\includegraphics[width=0.05\textwidth]{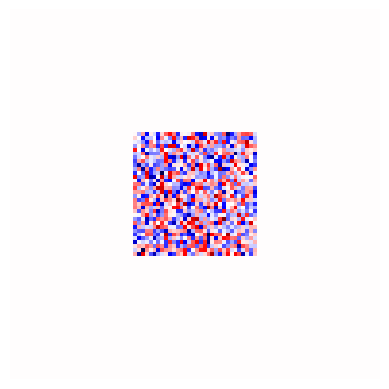}  &
 \includegraphics[width=0.05\textwidth]{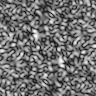}  &
\includegraphics[width=0.05\textwidth]{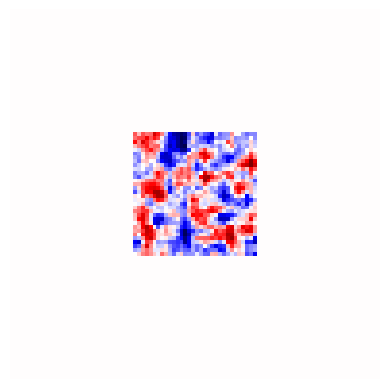}  &
 \includegraphics[width=0.05\textwidth]{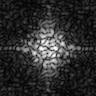}  &
\includegraphics[width=0.05\textwidth]{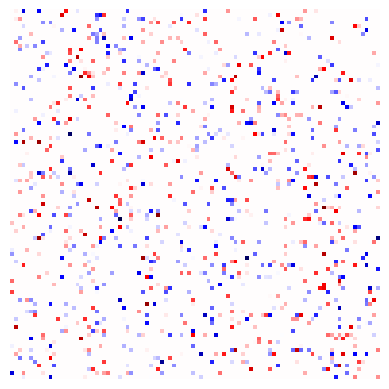}  &
 \includegraphics[width=0.05\textwidth]{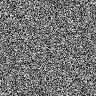}  &
\includegraphics[width=0.05\textwidth]{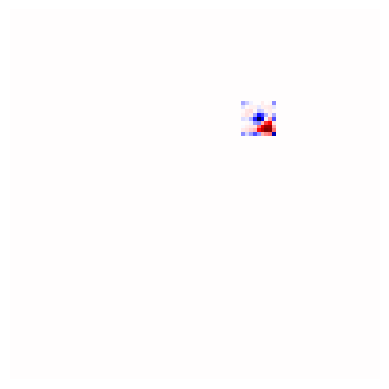}  &
 \includegraphics[width=0.05\textwidth]{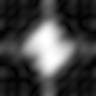}  &
\includegraphics[width=0.05\textwidth]{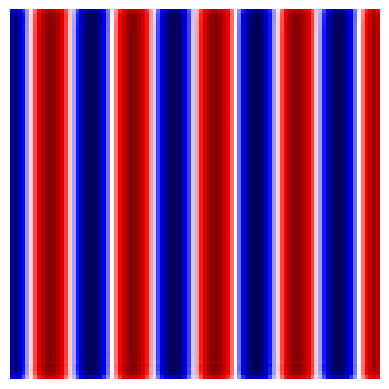}  &
 \includegraphics[width=0.05\textwidth]{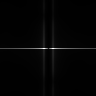} \\
\raisebox{2\height}{Row of $\mathbf{A}$ at $\gamma=(2/5)\gamma^\prime$} &
 \includegraphics[width=0.05\textwidth]{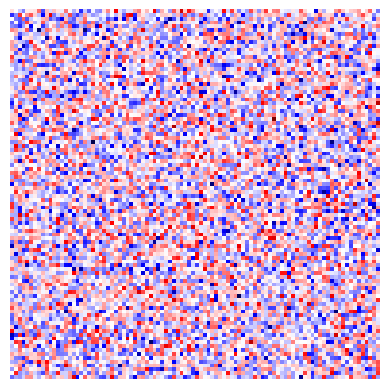}  &
 \includegraphics[width=0.05\textwidth]{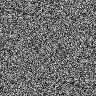}  &
\includegraphics[width=0.05\textwidth]{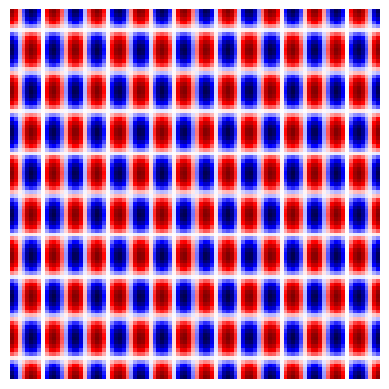}  &
 \includegraphics[width=0.05\textwidth]{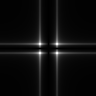}  &
\includegraphics[width=0.05\textwidth]{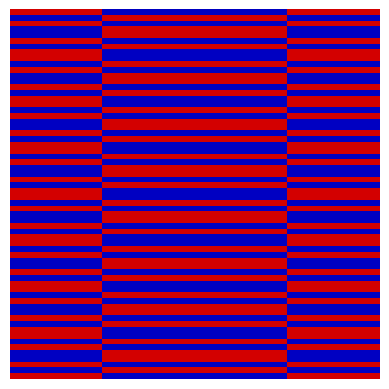}  &
 \includegraphics[width=0.05\textwidth]{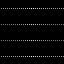}  &
\includegraphics[width=0.05\textwidth]{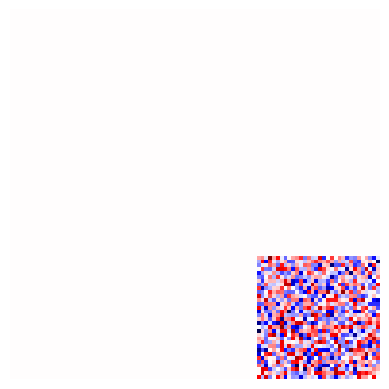}  &
 \includegraphics[width=0.05\textwidth]{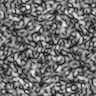}  &
\includegraphics[width=0.05\textwidth]{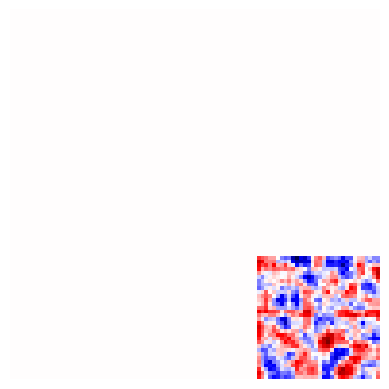}  &
 \includegraphics[width=0.05\textwidth]{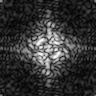}  &
\includegraphics[width=0.05\textwidth]{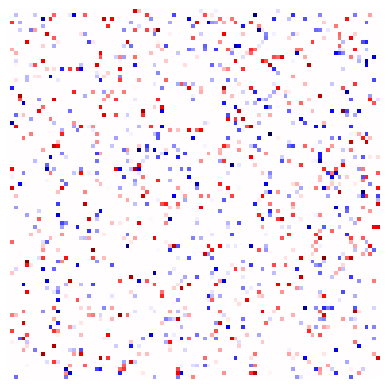}  &
 \includegraphics[width=0.05\textwidth]{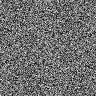}  &
\includegraphics[width=0.05\textwidth]{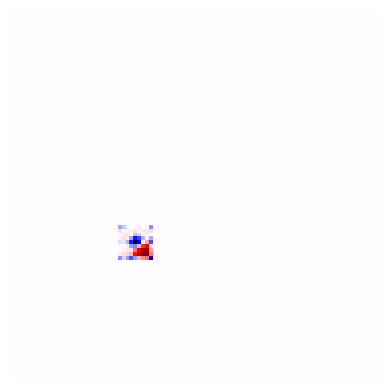}  &
 \includegraphics[width=0.05\textwidth]{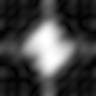}  &
\includegraphics[width=0.05\textwidth]{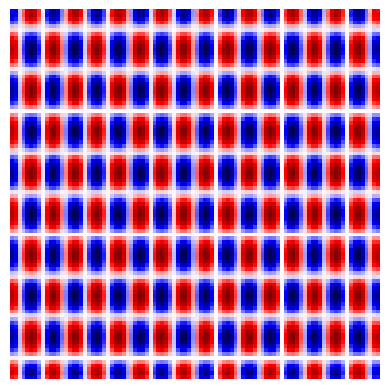}  &
 \includegraphics[width=0.05\textwidth]{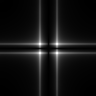} \\
\raisebox{2\height}{Row of $\mathbf{A}$ at $\gamma=(3/5)\gamma^\prime$} &
 \includegraphics[width=0.05\textwidth]{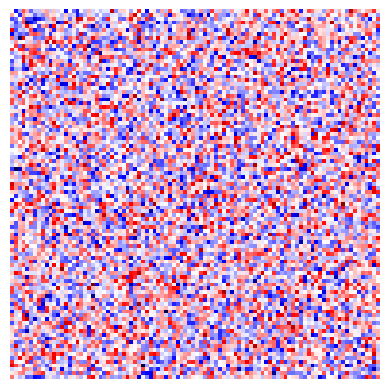}  &
 \includegraphics[width=0.05\textwidth]{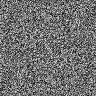}  &
\includegraphics[width=0.05\textwidth]{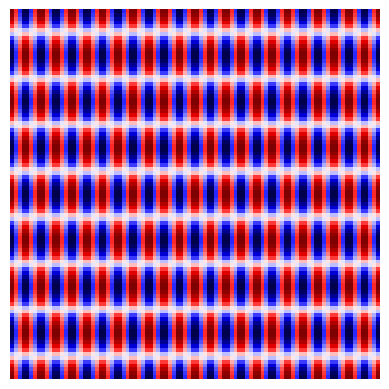}  &
 \includegraphics[width=0.05\textwidth]{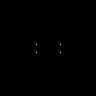}  &
\includegraphics[width=0.05\textwidth]{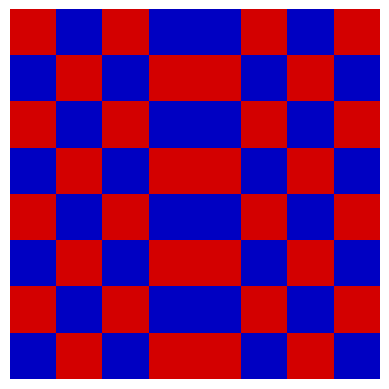}  &
 \includegraphics[width=0.05\textwidth]{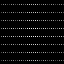}  &
\includegraphics[width=0.05\textwidth]{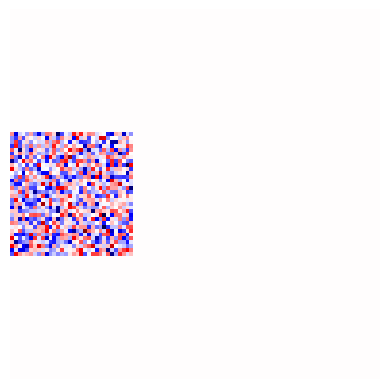}  &
 \includegraphics[width=0.05\textwidth]{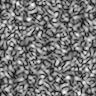}  &
\includegraphics[width=0.05\textwidth]{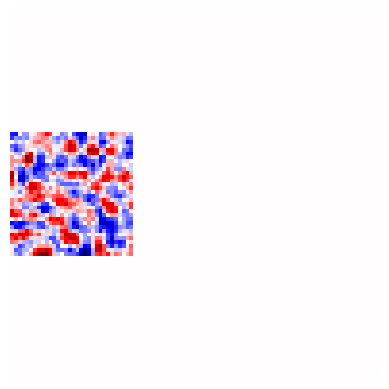}  &
 \includegraphics[width=0.05\textwidth]{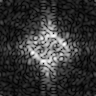}  &
\includegraphics[width=0.05\textwidth]{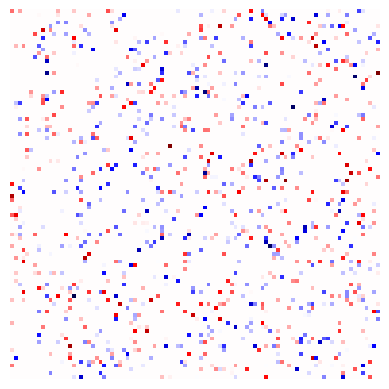}  &
 \includegraphics[width=0.05\textwidth]{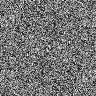}  &
\includegraphics[width=0.05\textwidth]{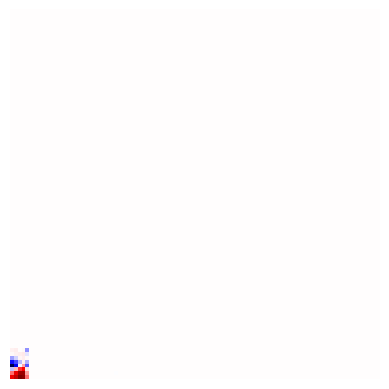}  &
 \includegraphics[width=0.05\textwidth]{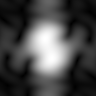}  &
\includegraphics[width=0.05\textwidth]{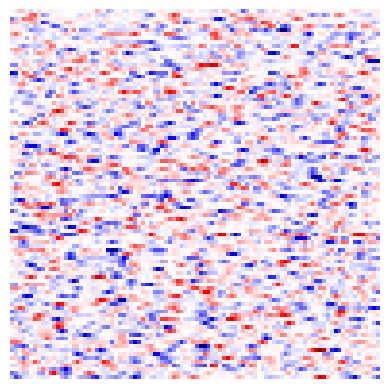}  &
 \includegraphics[width=0.05\textwidth]{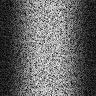} \\
\raisebox{2\height}{Row of $\mathbf{A}$ at $\gamma=(4/5)\gamma^\prime$} &
 \includegraphics[width=0.05\textwidth]{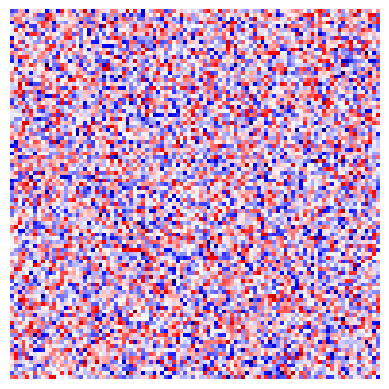}  &
 \includegraphics[width=0.05\textwidth]{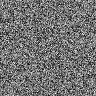}  &
\includegraphics[width=0.05\textwidth]{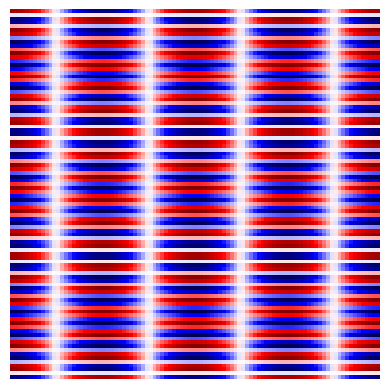}  &
 \includegraphics[width=0.05\textwidth]{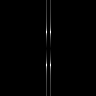}  &
\includegraphics[width=0.05\textwidth]{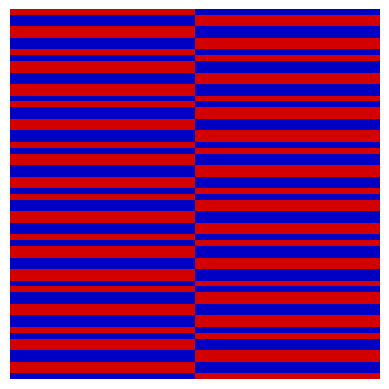}  &
 \includegraphics[width=0.05\textwidth]{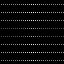}  &
\includegraphics[width=0.05\textwidth]{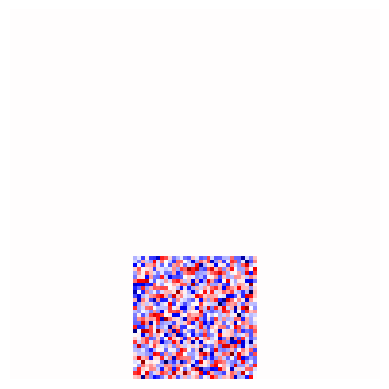}  &
 \includegraphics[width=0.05\textwidth]{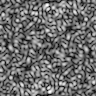}  &
\includegraphics[width=0.05\textwidth]{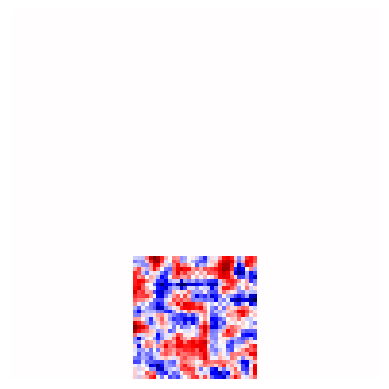}  &
 \includegraphics[width=0.05\textwidth]{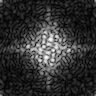}  &
\includegraphics[width=0.05\textwidth]{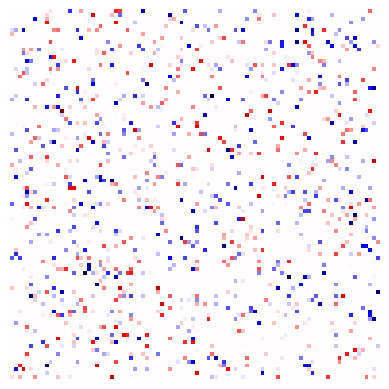}  &
 \includegraphics[width=0.05\textwidth]{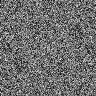}  &
\includegraphics[width=0.05\textwidth]{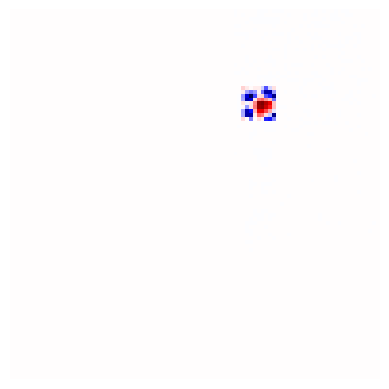}  &
 \includegraphics[width=0.05\textwidth]{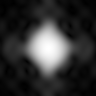}  &
\includegraphics[width=0.05\textwidth]{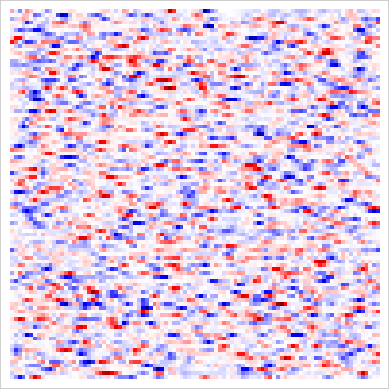}  &
 \includegraphics[width=0.05\textwidth]{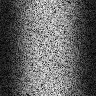} \\
\raisebox{2\height}{Row of $\mathbf{A}$ at $\gamma=(5/5)\gamma^\prime$} &
 \includegraphics[width=0.05\textwidth]{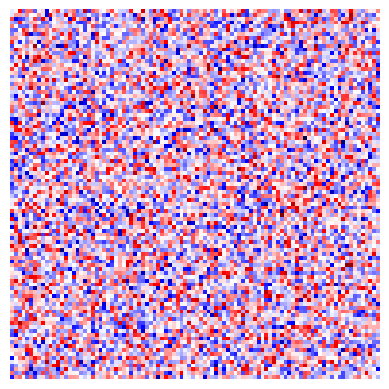}  &
 \includegraphics[width=0.05\textwidth]{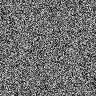}  &
\includegraphics[width=0.05\textwidth]{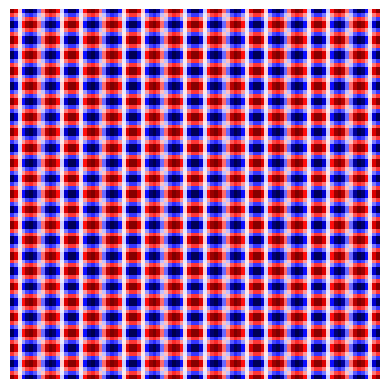}  &
 \includegraphics[width=0.05\textwidth]{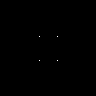}  &
\includegraphics[width=0.05\textwidth]{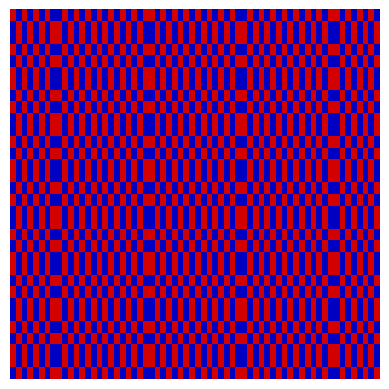}  &
 \includegraphics[width=0.05\textwidth]{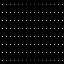}  &
\includegraphics[width=0.05\textwidth]{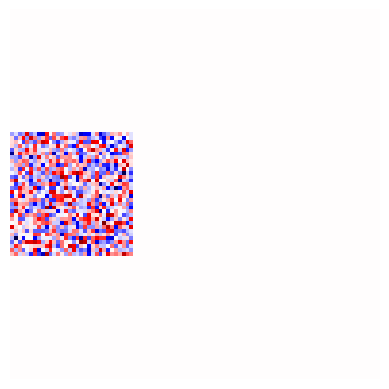}  &
 \includegraphics[width=0.05\textwidth]{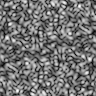}  &
\includegraphics[width=0.05\textwidth]{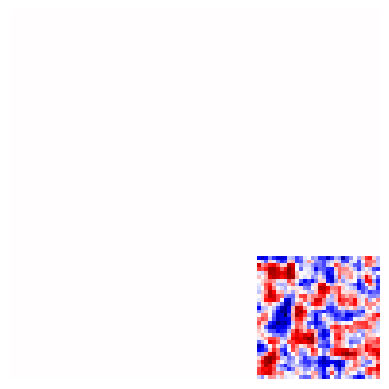}  &
 \includegraphics[width=0.05\textwidth]{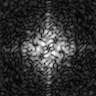}  &
\includegraphics[width=0.05\textwidth]{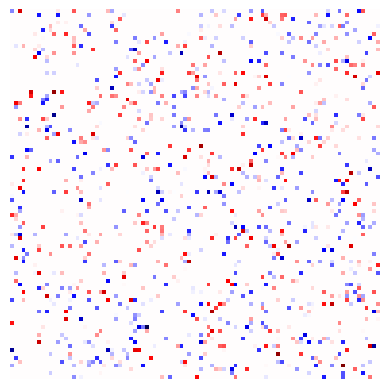}  &
 \includegraphics[width=0.05\textwidth]{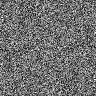}  &
\includegraphics[width=0.05\textwidth]{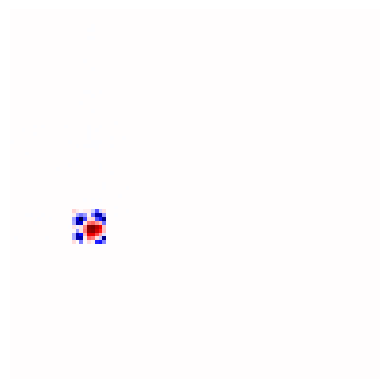}  &
 \includegraphics[width=0.05\textwidth]{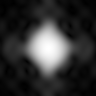}  &
\includegraphics[width=0.05\textwidth]{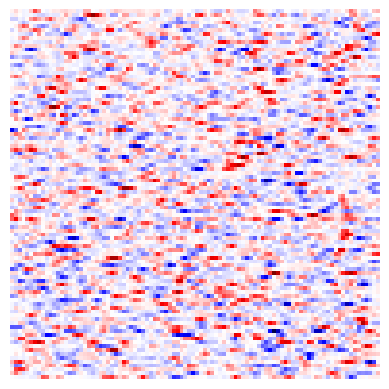}  &
 \includegraphics[width=0.05\textwidth]{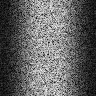} \\ \hline \hline
 \raisebox{1.75\height}{Visualization of $\mathbf{A}\in\mathbb{R}^{M\times N}$} &
  \multicolumn{2}{c|}{\includegraphics[width=0.1\textwidth]{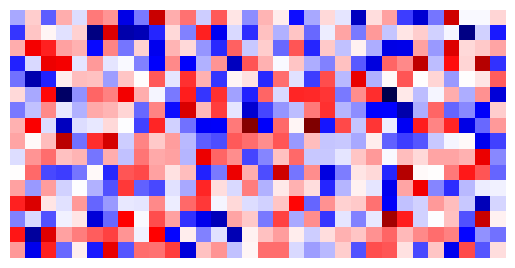}} &
  \multicolumn{2}{c|}{\includegraphics[width=0.1\textwidth]{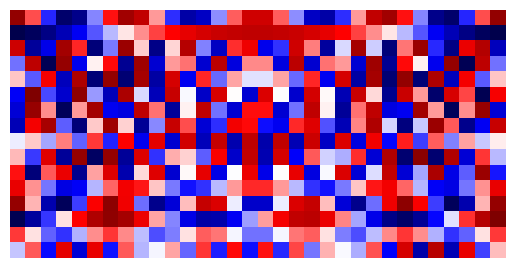}} &
  \multicolumn{2}{c|}{\includegraphics[width=0.1\textwidth]{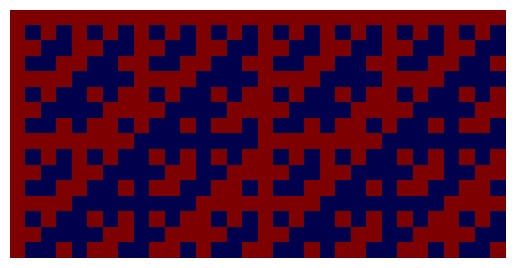}} &
  \multicolumn{2}{c|}{\includegraphics[width=0.1\textwidth]{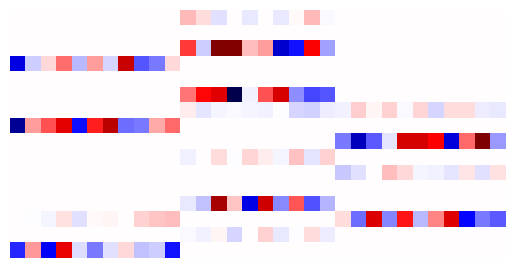}} &
  \multicolumn{2}{c|}{\includegraphics[width=0.1\textwidth]{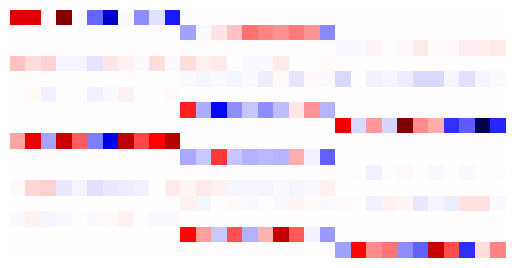}} &
  \multicolumn{2}{c|}{\includegraphics[width=0.1\textwidth]{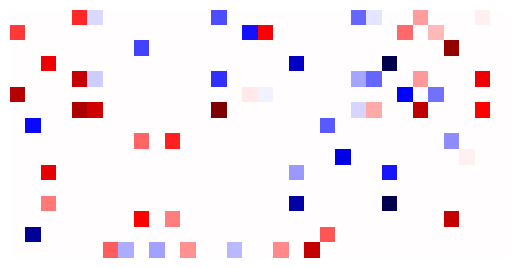}} &
  \multicolumn{2}{c|}{\includegraphics[width=0.1\textwidth]{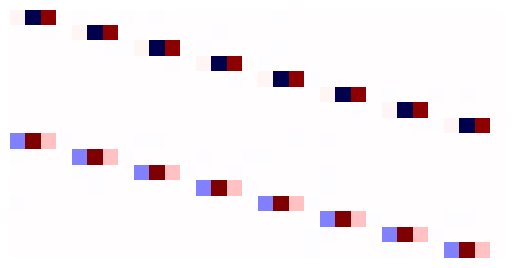}} &
  \multicolumn{2}{c}{\includegraphics[width=0.1\textwidth]{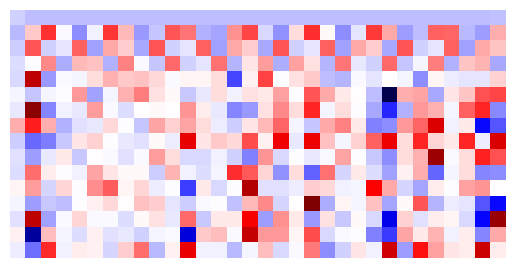}} \\
\raisebox{3.5\height}{Visualization of $\mathbf{AA}^\top\in\mathbb{R}^{M\times M}$} &
  \multicolumn{2}{c|}{\includegraphics[width=0.1\textwidth]{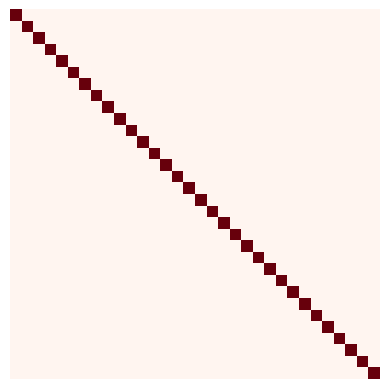}} &
  \multicolumn{2}{c|}{\includegraphics[width=0.1\textwidth]{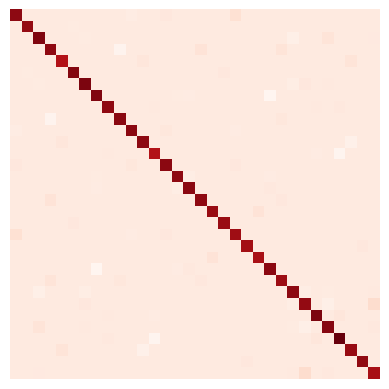}} &
  \multicolumn{2}{c|}{\includegraphics[width=0.1\textwidth]{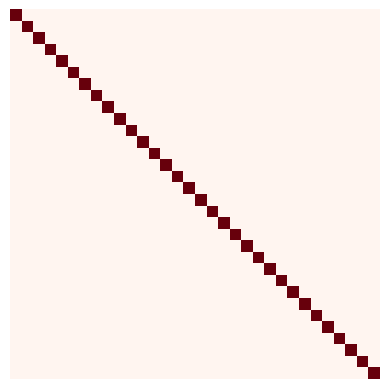}} &
  \multicolumn{2}{c|}{\includegraphics[width=0.1\textwidth]{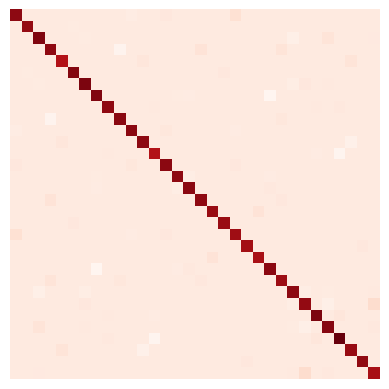}} &
  \multicolumn{2}{c|}{\includegraphics[width=0.1\textwidth]{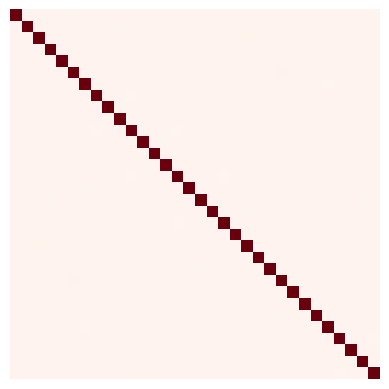}} &
  \multicolumn{2}{c|}{\includegraphics[width=0.1\textwidth]{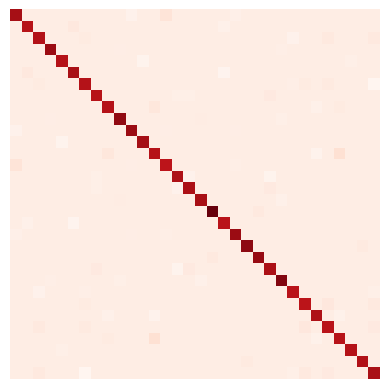}} &
  \multicolumn{2}{c|}{\includegraphics[width=0.1\textwidth]{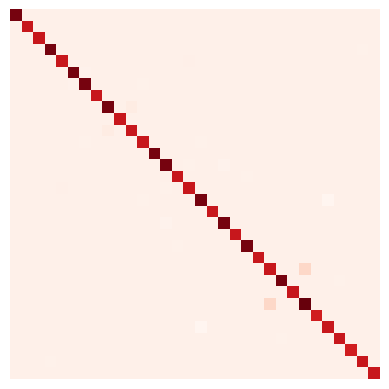}} &
  \multicolumn{2}{c}{\includegraphics[width=0.1\textwidth]{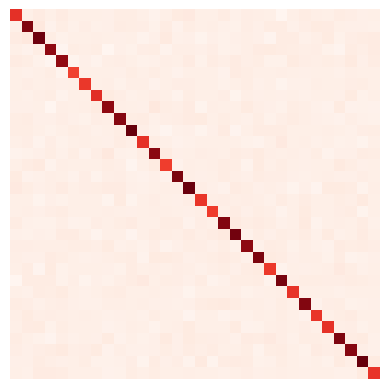}} \\ \raisebox{4\height}{Element Distribution of $\mathbf{A}$} &
  \multicolumn{2}{c|}{\includegraphics[width=0.1\textwidth]{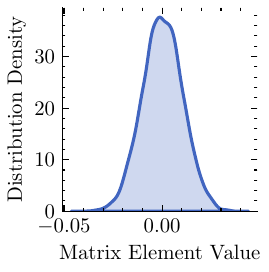}} &
  \multicolumn{2}{c|}{\includegraphics[width=0.1\textwidth]{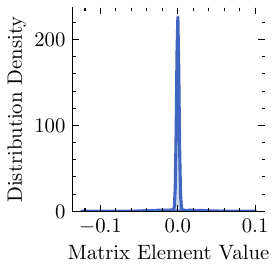}} &
  \multicolumn{2}{c|}{\includegraphics[width=0.1\textwidth]{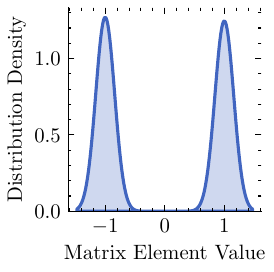}} &
  \multicolumn{2}{c|}{\includegraphics[width=0.1\textwidth]{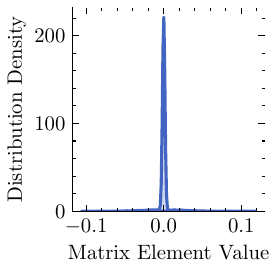}} &
  \multicolumn{2}{c|}{\includegraphics[width=0.1\textwidth]{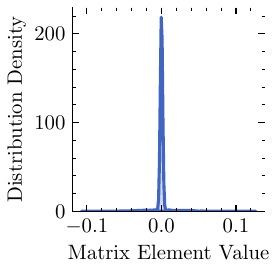}} &
  \multicolumn{2}{c|}{\includegraphics[width=0.1\textwidth]{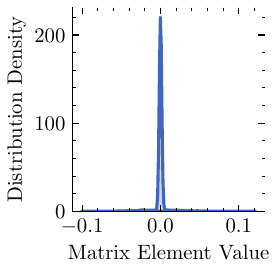}} &
  \multicolumn{2}{c|}{\includegraphics[width=0.1\textwidth]{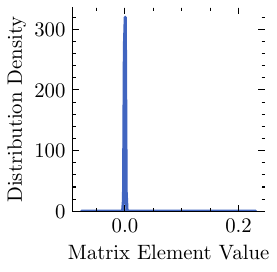}} &
  \multicolumn{2}{c}{\includegraphics[width=0.1\textwidth]{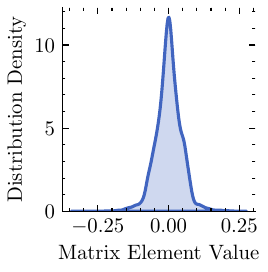}} \\
  \hline \hline
  \raisebox{2\height}{Visualization of $\dfrac{\partial \mathbf{y}_i}{\partial \mathbf{x}}$} &
  \multicolumn{2}{c|}{\includegraphics[width=0.1\textwidth]{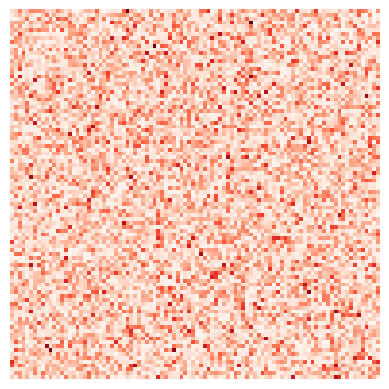}} &
  \multicolumn{2}{c|}{\includegraphics[width=0.1\textwidth]{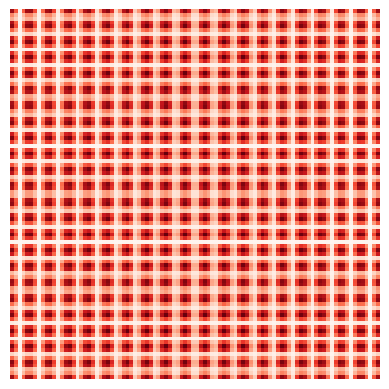}} &
  \multicolumn{2}{c|}{\includegraphics[width=0.1\textwidth]{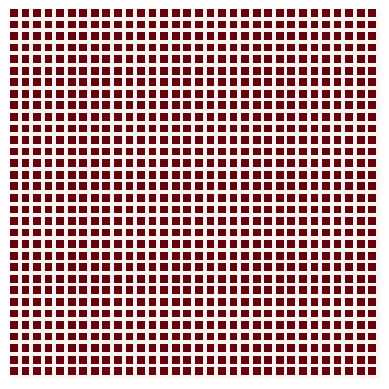}} &
  \multicolumn{2}{c|}{\includegraphics[width=0.1\textwidth]{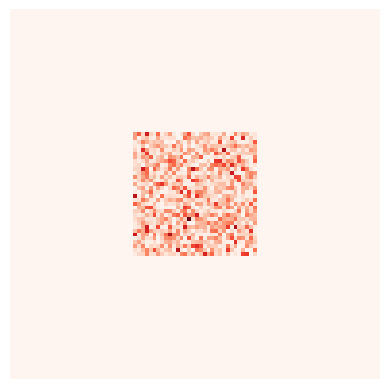}} &
  \multicolumn{2}{c|}{\includegraphics[width=0.1\textwidth]{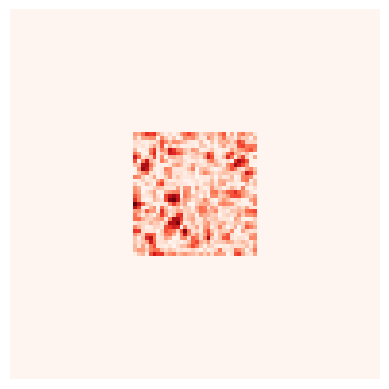}} &
  \multicolumn{2}{c|}{\includegraphics[width=0.1\textwidth]{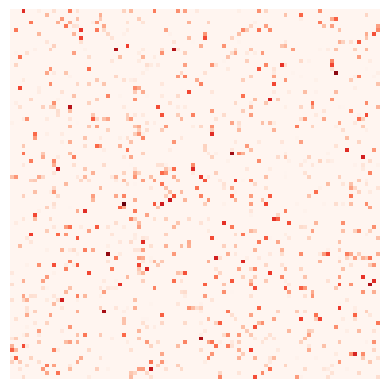}} &
  \multicolumn{2}{c|}{\includegraphics[width=0.1\textwidth]{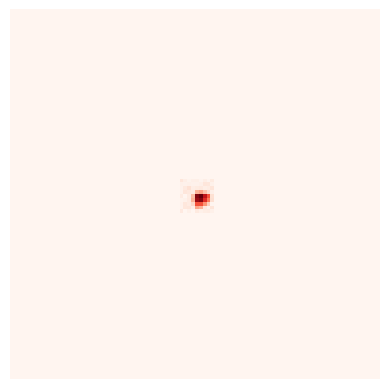}} &
  \multicolumn{2}{c}{\includegraphics[width=0.1\textwidth]{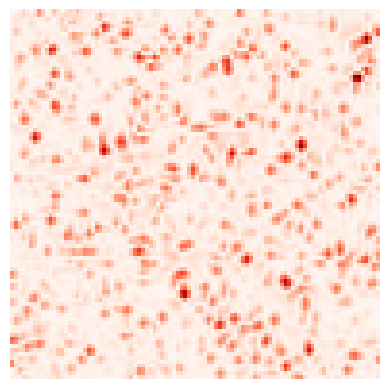}} \\ \shline
\end{tabular}}
\vspace{-10pt}
\end{table*}

\begin{table}[!t]
\vspace{-5pt}
\caption{A visual comparison of recovering a benchmark image named ``House" among four CS NNs \cite{chen2022fsoinet,fan2022global} at $\gamma =10\%$. We provide their recoveries \textcolor{blue}{\textbf{(top)}} and visualizations of the effective receptive field (ERF) showing the utilization of informative regions in initialized results \textcolor{blue}{\textbf{(bottom)}}.}
\vspace{-8pt}
\label{tab:compare_erf_network}
\centering
\hspace{-4pt}
\setlength{\tabcolsep}{1.5pt}
\resizebox{0.48\textwidth}{!}{
\begin{tabular}{l|cccc}
\shline
\rowcolor[HTML]{EFEFEF} 
Deep CS Network           & FSOINet & MR-CCSNet$^+$ & \textbf{PC-CNN} & \textbf{PCT} \\ \hline \hline
\raisebox{3\height}{Recovered Result $\hat{\mathbf{x}}$} &  \includegraphics[width=0.08\textwidth]{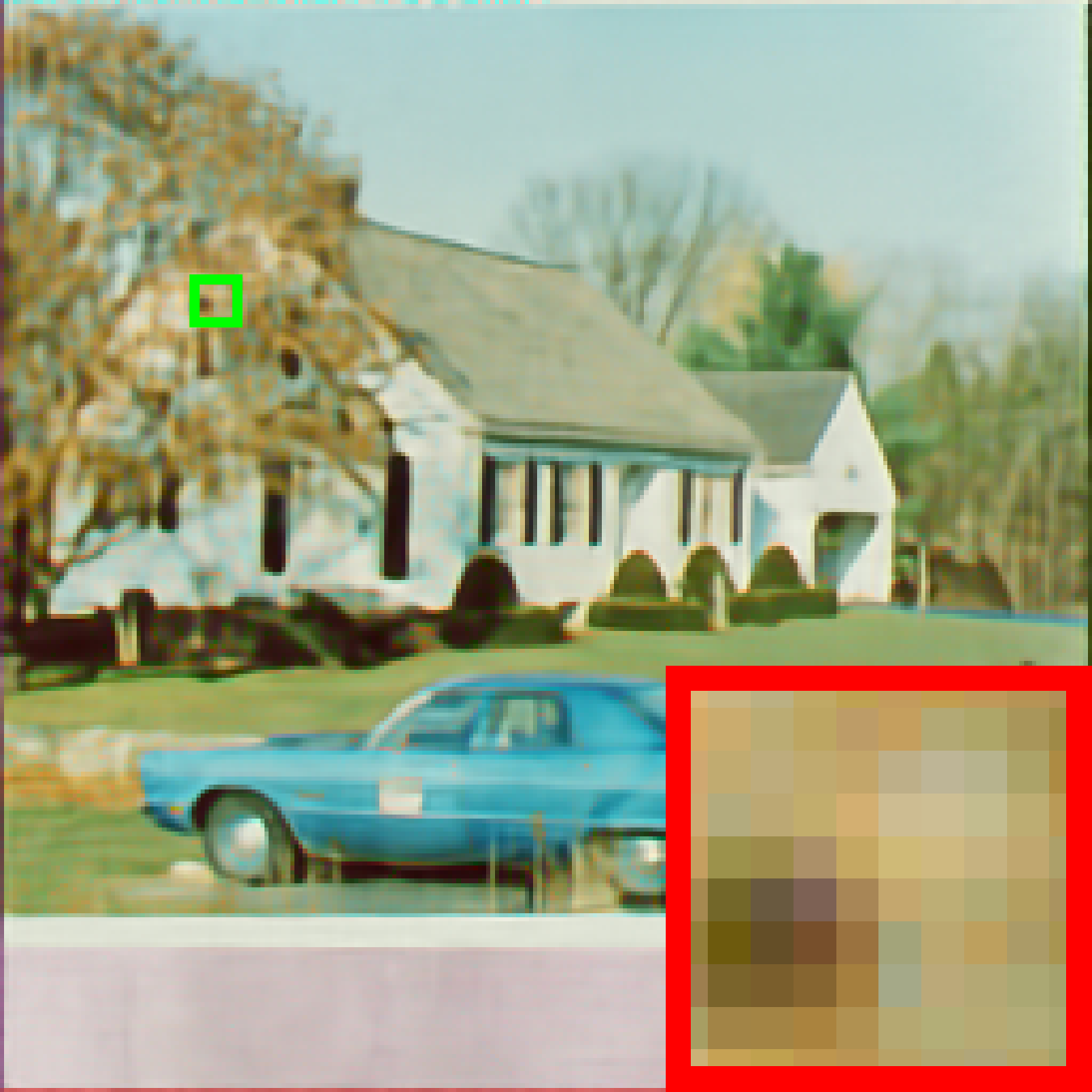}  &  \includegraphics[width=0.08\textwidth]{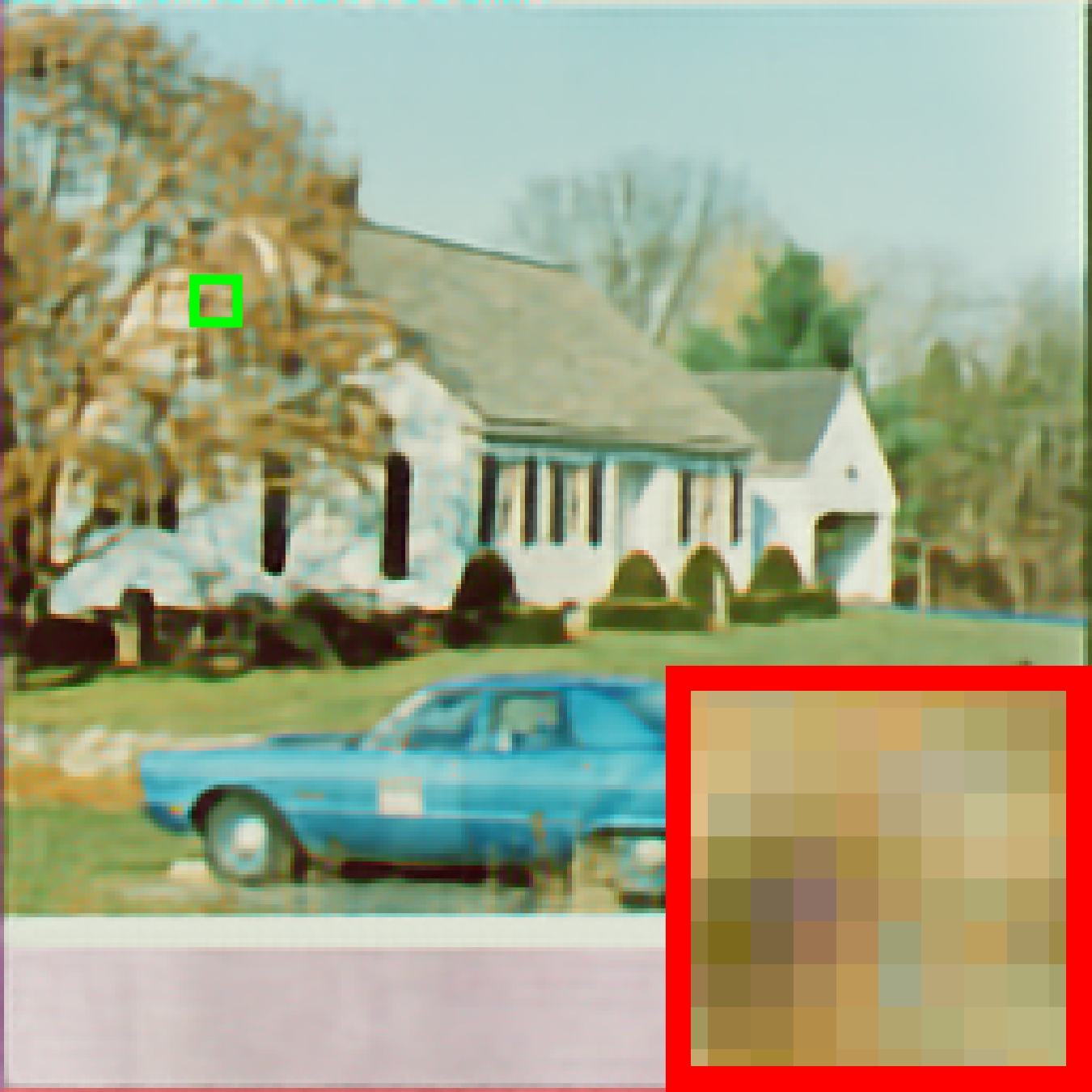}             & \includegraphics[width=0.08\textwidth]{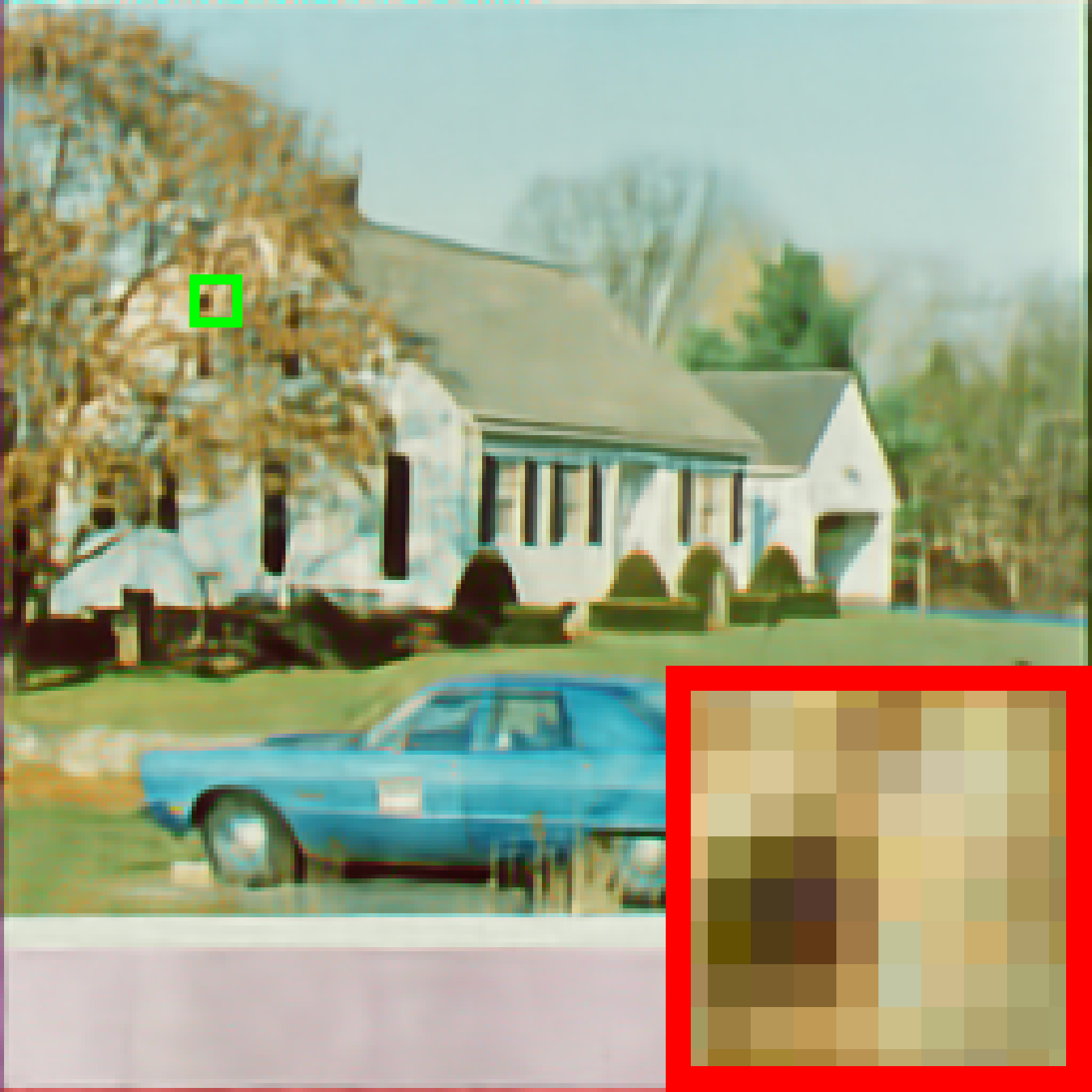}&\includegraphics[width=0.08\textwidth]{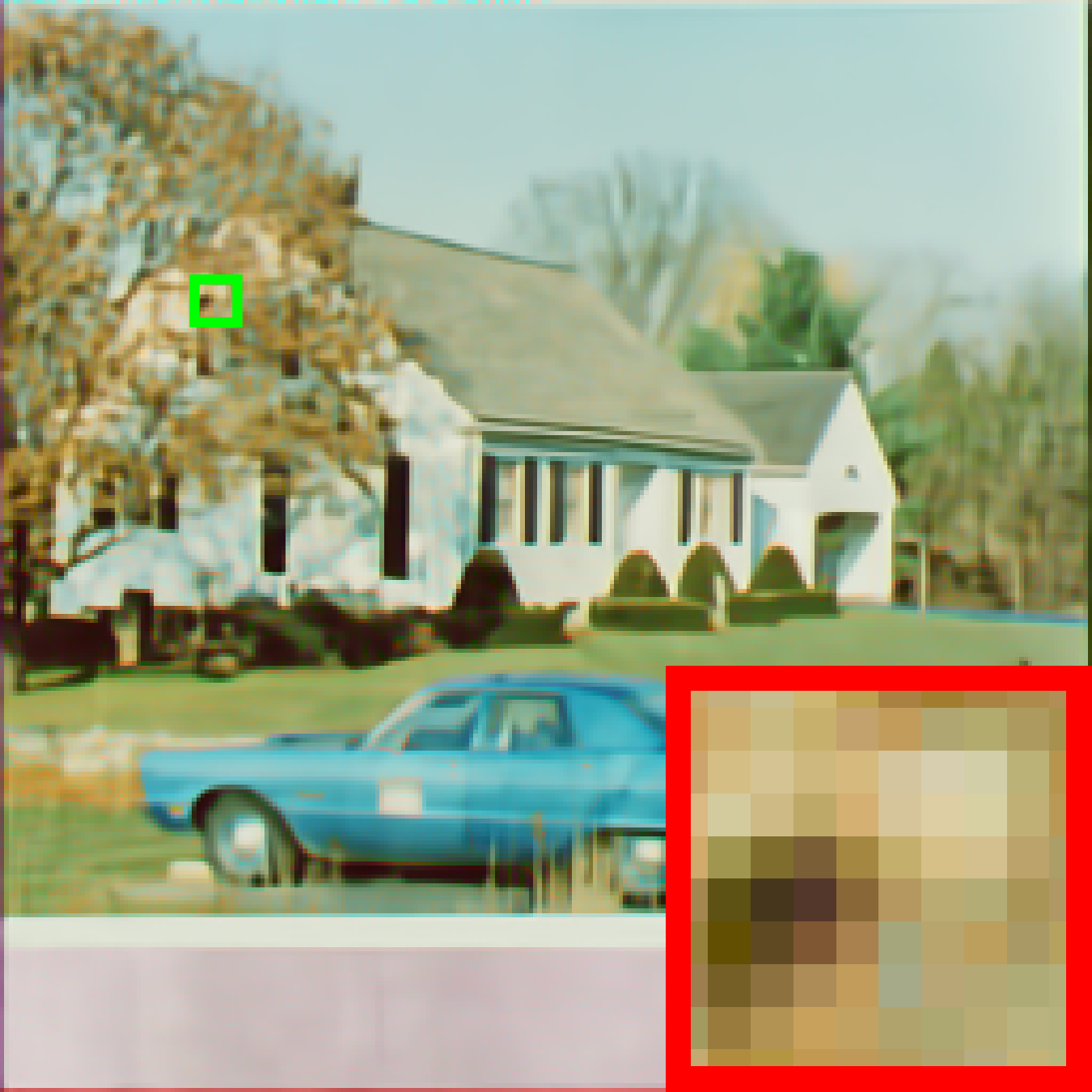}\\
PSNR/SSIM & \textcolor{green}{25.61}/0.8080 & 25.40/\textcolor{green}{0.8139} & \textcolor{blue}{26.25}/\textcolor{blue}{0.8209} & \textcolor{red}{26.73}/\textcolor{red}{0.8345}\\ \hline \hline
\raisebox{1.5\height}{Visualization of $\dfrac{\partial \hat{\mathbf{x}}_i}{\partial \hat{\mathbf{x}}_\text{init}}$} & \includegraphics[width=0.08\textwidth]{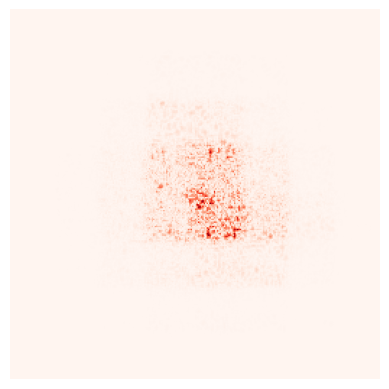} & \includegraphics[width=0.08\textwidth]{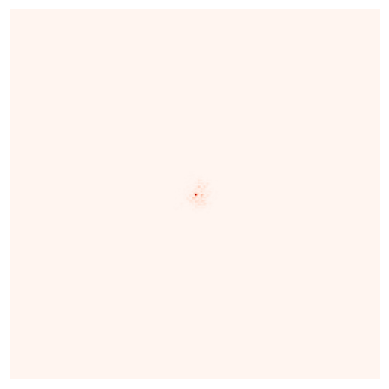} & \includegraphics[width=0.08\textwidth]{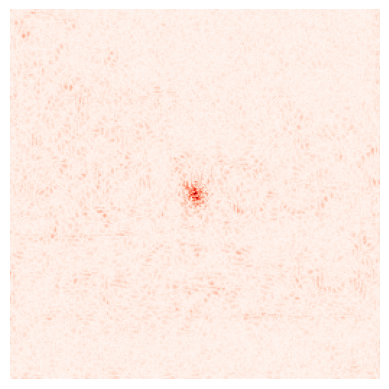} & \includegraphics[width=0.08\textwidth]{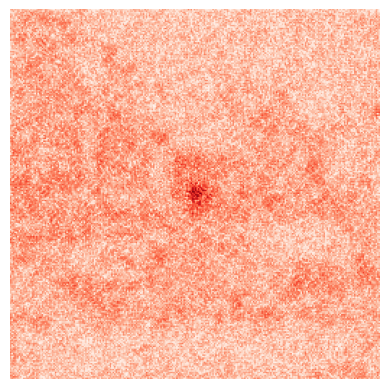}\\ \shline
\end{tabular}}
\vspace{-10pt}
\end{table}

\subsubsection{Extensibility and Generalization Ability of Our Method}
In addition to our default PCNets, we train the other four PC-CNN and PCT variants with $K\in\{10,30\}$. The results on Set11 \cite{kulkarni2016reconnet} are presented in Tab.~\ref{tab:ablation_sampling_operator} (12)-(13) and Fig.~\ref{fig:comp_param_PSNR}. In the ratio setting of $\gamma =50\%$, our PC-CNN and PCT networks with $K\in\{10,20,30\}$ achieve the PSNR values of 41.56/41.83/41.97dB and 42.52/42.71/42.90dB, respectively. This indicates that first, even with much fewer parameters, our PCNets of $K=10$ hold a large performance leading to other methods. Second, there still exists improvement room as capacity increases. We note that this can not be achieved by adding parameters to other unrolled networks \cite{zhang2018ista,you2021coast,zhang2021amp,song2021memory,chen2022content} since their performance reaches saturation.

To verify the generality of our method, we sequentially apply our ordinary setting (in Sec.~\ref{subsec:baseline_setup}), proposed COSO ($\mathcal{G}_\mathbf{A}$ in Sec.~\ref{subsec:sampling_operator_architecture}), and enhanced training strategies (in Sec.~\ref{subsec:training_enhancements}) to the two representative existing deep unfolding methods: ISTA-Net$^{++}$ \cite{you2021ista} and FSOINet \cite{chen2022fsoinet}. The experimental results are reported in Tab.~\ref{tab:boost_existing_methods}. Specifically, we observe that even at a high sampling rate $\gamma =50\%$, the introduction of basic setting and our $\{\mathcal{G}_\mathbf{A},\mathcal{G}_{\mathbf{A}^\top}\}$ brings a significant PSNR gain of 0.81dB and a parameter reduction of 0.22M on average. Our training enhancements further enhance the methods with 0.15dB improvement. Note that our strategies in Tab.~\ref{tab:boost_existing_methods} (1)-(8) are actually extensible to many existing approaches, not limited to deep unrolled networks. They can make each model flexible to handle any CS ratio $\gamma\in[0,1]$ once trained.

\subsubsection{Analysis of Learned Sampling Operator Weights}

To study what is learned in our deep conditional filtering network, we extract its equivalent weights for five $\gamma$s by Algo.~\ref{alg:matrix_extraction}. Since all the spatial sizes of our convolution kernels are $3\times 3$, without considering zero-paddings and biases, the seven layers can be merged into a $15\times 15$ filter $\mathbf{K}$. Tab.~\ref{tab:visualize_learned_filter} visualizes our extracted filters $\{\mathbf{K}_D,\mathbf{K}_G\}$ and corresponding filtered results $\{\mathbf{X}_D,\mathbf{X}_G\}$ of an image named ``Butterfly" for D- and G-branch, respectively. We observe that without any specific constraints in the loss function, the filters adaptively learn to perform smoothing and sharpening on the input at low and high CS ratios, respectively, especially for G-branch.

In the top seven rows of Tab.~\ref{tab:visualize_equivalent_matrix}, we present the equivalent matrices of eight sampling operators for a 96$\times$96 image size. We note that traditional fixed global dense matrices incur high computation and memory costs and lack a data-driven learned image prior. Block-diagonal matrices and stacked networks, which perform sparse pixel aggregations into measurements (characterized by many zero weights), prove to be insufficiently effective in our comparison experiments. In contrast, our proposed collaborative sampling operator can capture low-, mid-, and high-frequency components with adaptively modulated weights for different CS ratios. This approach results in a superior balance of efficiency, complexity, trainability, interpretability, and effectiveness.

\subsubsection{Analysis of Effective Receptive Field}

To further investigate how much image information is preserved in measurements and utilized in reconstruction, we conduct a visual comparison of the effective receptive field (ERF) \cite{luo2016understanding} among various methods. Specifically, following \cite{ding2022scaling}, in the last row of Tab.~\ref{tab:visualize_equivalent_matrix}, we evaluate the ERFs of sampling operators by computing the partial derivative $\partial \mathbf{y}_i/ \partial \mathbf{x}$ for the last measurement value $\mathbf{y}_i$ relative to the input image named ``House'' $\mathbf{x}$. Our approach, which combines deep conditional filtering and dual-branch sampling, demonstrates both local and global characteristics.

In Tab.~\ref{tab:compare_erf_network}, we compare the ERFs of four CS networks by computing $\partial \hat{\mathbf{x}}_{i}/ \partial \hat{\mathbf{x}}_\text{init}$ for the centrally recovered pixel $\hat{\mathbf{x}}_{i}$ relative to the initialization $\hat{\mathbf{x}}_{\text{init}}$. We observe that the existing networks \cite{chen2022content,fan2022global} display limited, unbalanced, and blocky content perceptions. In contrast, our proposed PC-CNN, by integrating our global sampling operator with the unrolled PGD algorithm framework, achieves a balanced integration of both local and global features from the initialization. Additionally, our PCT enhances the utilization of informative image regions using shifted window-based SCBs, effectively increasing the overall performance.

\subsection{Applications}

\subsubsection{Quantized Compressed Sensing}
In practice, each measurement is often discretized to a finite number of $q$ bits, leading to quantized compressed sensing (QCS) model $\mathbf{y}=\mathcal{Q}(\mathbf{Ax+n})$, where $\mathcal{Q}(\cdot)$ is a quantizer of level $q$. Following the setup in \cite{jung2021quantized}, we apply our PCNets and four existing networks to the QCS task. Tab.~\ref{tab:apply_to_qcs} shows PSNR lead 0.75/1.13/5.88dB of PCNet with $q=32/8/1$, thus manifesting the consistent performance and robustness superiority, as well as generalization ability of our method.

\subsubsection{Self-Supervised Compressed Sensing}
We further inject our sampling operator into five deep self-supervised CS methods: DIP \cite{ulyanov2018deep}, BCNN \cite{pang2020self}, EI \cite{chen2021equivariant}, REI \cite{chen2022robust}, and DDSSL \cite{quan2022dual}. The former two methods fit estimations $\hat{\mathbf{x}}$ by NN from only a given measurement $\mathbf{y}$, while the latter three train a recovery network from an external measurement set $\{\mathbf{y}_i\}$. None of these methods see any ground truth image $\mathbf{x}$ and are all learned from block-diagonal Gaussian measurements as the default setting in \cite{quan2022dual}. In Tab.~\ref{tab:apply_to_sscs}, we employ our learned and frozen $\{\mathcal{G}_\mathbf{A},\mathcal{G}_{\mathbf{A}^\top}\}$ from PC-CNN to them and obtain a PSNR boost of 1.96/2.00dB with observation noise level $\sigma=0/10$, verifying that our sampling operator effectively preserves more information than the traditional block-based sampling matrix.

\begin{table}[!t]
\caption{PSNR results tested on Set11 \cite{kulkarni2016reconnet} with $\gamma =50\%$ of applying PCNets and four CS networks to the quantized CS task with level $q\in\{1,8,32\}$.}
\vspace{-8pt}
\label{tab:apply_to_qcs}
\centering
\hspace{-4pt}
\resizebox{0.48\textwidth}{!}{
\begin{tabular}{l|ccc}
\shline
\rowcolor[HTML]{EFEFEF} 
\cellcolor[HTML]{EFEFEF}                         & \multicolumn{3}{c}{\cellcolor[HTML]{EFEFEF}Quantization Level $q$} \\ \hhline{>{\arrayrulecolor[HTML]{EFEFEF}}->{\arrayrulecolor{black}}|--->{\arrayrulecolor[HTML]{EFEFEF}}>{\arrayrulecolor{black}}} 
\rowcolor[HTML]{EFEFEF} 
\multirow{-2}{*}{\cellcolor[HTML]{EFEFEF}Method} & 32 (Full-Precision)               & 8              & 1              \\ \hline \hline
ISTA-Net$^+$ \cite{zhang2018ista}    & 38.07 & 36.63 {\scriptsize(\textcolor{purple}{-1.44})} & 15.09 {\scriptsize(\textcolor{purple}{-21.54})} \\
ISTA-Net$^{++}$ \cite{you2021ista} & 38.73 & 37.31 {\scriptsize(\textcolor{purple}{-1.42})} & 17.45 {\scriptsize(\textcolor{purple}{-19.86})} \\
FSOINet \cite{chen2022fsoinet}         & \textcolor{green}{41.08} & \textcolor{green}{39.13} {\scriptsize(\textcolor{purple}{-1.95})} & \textcolor{green}{20.77} {\scriptsize(\textcolor{purple}{-18.36})} \\
MR-CCSNet$^+$ \cite{fan2022global}                 & 39.27                & 36.76 {\scriptsize(\textcolor{purple}{-2.51})}              & 19.12 {\scriptsize(\textcolor{purple}{-17.64})}              \\ \hline \hline
\textbf{PC-CNN (Ours)}          & \textcolor{blue}{41.83} & \textcolor{blue}{40.26} {\scriptsize(\textcolor{purple}{-1.57})} & \textcolor{blue}{26.65} {\scriptsize(\textcolor{purple}{-13.61})} \\
\textbf{PCT (Ours)}             & \textcolor{red}{42.71} & \textcolor{red}{41.44} {\scriptsize(\textcolor{purple}{-1.27})} & \textcolor{red}{27.34} {\scriptsize(\textcolor{purple}{-14.10})} \\ \shline
\end{tabular}}
\vspace{-10pt}
\end{table}

\begin{table}[!t]
\caption{PSNR results tested on Set11 \cite{kulkarni2016reconnet} with $\gamma =10\%$ of applying our learned and frozen operators $\{\mathcal{G}_\mathbf{A},\mathcal{G}_{\mathbf{A}^\top}\}$ to five self-supervised CS methods. We follow \cite{quan2022dual} to obtain their results with a block-diagonal matrix and then evaluate their corresponding enhanced versions.}
\vspace{-8pt}
\label{tab:apply_to_sscs}
\centering
\hspace{-4pt}
\resizebox{0.48\textwidth}{!}{
\begin{tabular}{l|cc}
\shline
\rowcolor[HTML]{EFEFEF} 
\cellcolor[HTML]{EFEFEF}                         & \multicolumn{2}{c}{\cellcolor[HTML]{EFEFEF}Observation Noise Level $\sigma$} \\ \hhline{>{\arrayrulecolor[HTML]{EFEFEF}}->{\arrayrulecolor{black}}|-->{\arrayrulecolor[HTML]{EFEFEF}}>{\arrayrulecolor{black}}} 
\rowcolor[HTML]{EFEFEF} 
\multirow{-2}{*}{\cellcolor[HTML]{EFEFEF}Deep Self-Supervised CS Method} & 0                               & 10                             \\ \hline \hline
DIP (CVPR 2018) \cite{ulyanov2018deep}    & 26.02 & 23.81 \\
\textbf{PC-DIP}~{\scriptsize(Our sampling-enhanced version)} & 27.87~{\scriptsize(\textcolor{purple}{+1.85})}  & 24.94~{\scriptsize(\textcolor{purple}{+1.13})} \\ \hline \hline
BCNN (ECCV 2020) \cite{pang2020self}   & 27.49 & 25.15 \\
\textbf{PC-BCNN}~{\scriptsize(Our sampling-enhanced version)}            & 29.16~{\scriptsize(\textcolor{purple}{+1.67})} & 27.17~{\scriptsize(\textcolor{purple}{+2.02})} \\ \hline \hline
EI (ICCV 2021) \cite{chen2021equivariant}/REI (CVPR 2022) \cite{chen2022robust}                   & 22.65                           & 22.26                          \\
\textbf{PC-EI/-REI}~{\scriptsize(Our sampling-enhanced versions)}         & 25.03~{\scriptsize(\textcolor{purple}{+2.38})} & 24.90~{\scriptsize(\textcolor{purple}{+2.64})} \\ \hline \hline
DDSSL (ECCV 2022) \cite{quan2022dual} & 26.82 & 25.17 \\
\textbf{PC-DDSSL}~{\scriptsize(Our sampling-enhanced version)}          & 28.77~{\scriptsize(\textcolor{purple}{+1.95})} & 27.38~{\scriptsize(\textcolor{purple}{+2.21})} \\ \shline
\end{tabular}}
\vspace{-10pt}
\end{table}

\vspace{-5pt}
\section{Conclusion}
\label{sec:conclusion}

We propose a practical and compact network dubbed PCNet for image compressed sensing (CS) by designing a novel collaborative sampling operator (COSO) and modernizing a reconstruction backbone. In PCNet, the linear sampling process of our COSO is decomposed into a \textit{deep conditional filtering} step and a \textit{dual-branch fast sampling} step. From three perspectives—RIP \cite{candes2005decoding}, spatial/frequency domains, and ERF \cite{luo2016understanding}—we demonstrate the characteristics of eight sampling operators and the effectiveness of our design. Additionally, we introduce ten enhancement strategies to further improve reconstruction performance. PCNet learns a compact, implicit representation of image priors for a global sampling matrix within several convolutional layers and is flexible to handle arbitrary ratios once trained. It enjoys an interpretable and robust structure comprising a two-step disentangled sampling and an optimization-inspired reconstruction, and exhibits a significant performance lead and extension potential beyond existing methods, especially for high-resolution images. Considering the application of pre-trained sampling operators, we further provide a deployment-oriented matrix extraction scheme for spatial light modulators (SLMs) like digital micro-mirror devices (DMDs) with arbitrary resolutions and verify the consistent superiority of PCNet in quantized CS (QCS) and self-supervised CS tasks. Our future work involves applying this comprehensive solution to real single-pixel imaging systems, and extending our structural reparameterization \cite{ding2022scaling}-inspired sampling learning and recovery enhancements to other tasks like MRI \cite{sun2016deep}, sparse-view CT \cite{szczykutowicz2010dual}, and SCI \cite{cheng2022recurrent}.

% \section*{Acknowledgments}

\bibliographystyle{IEEEtran}
\bibliography{ref}

\vspace{-1.4cm}
\begin{IEEEbiography}[\vspace{-0.8cm}{\includegraphics[width=1in,height=1.25in,clip,keepaspectratio]{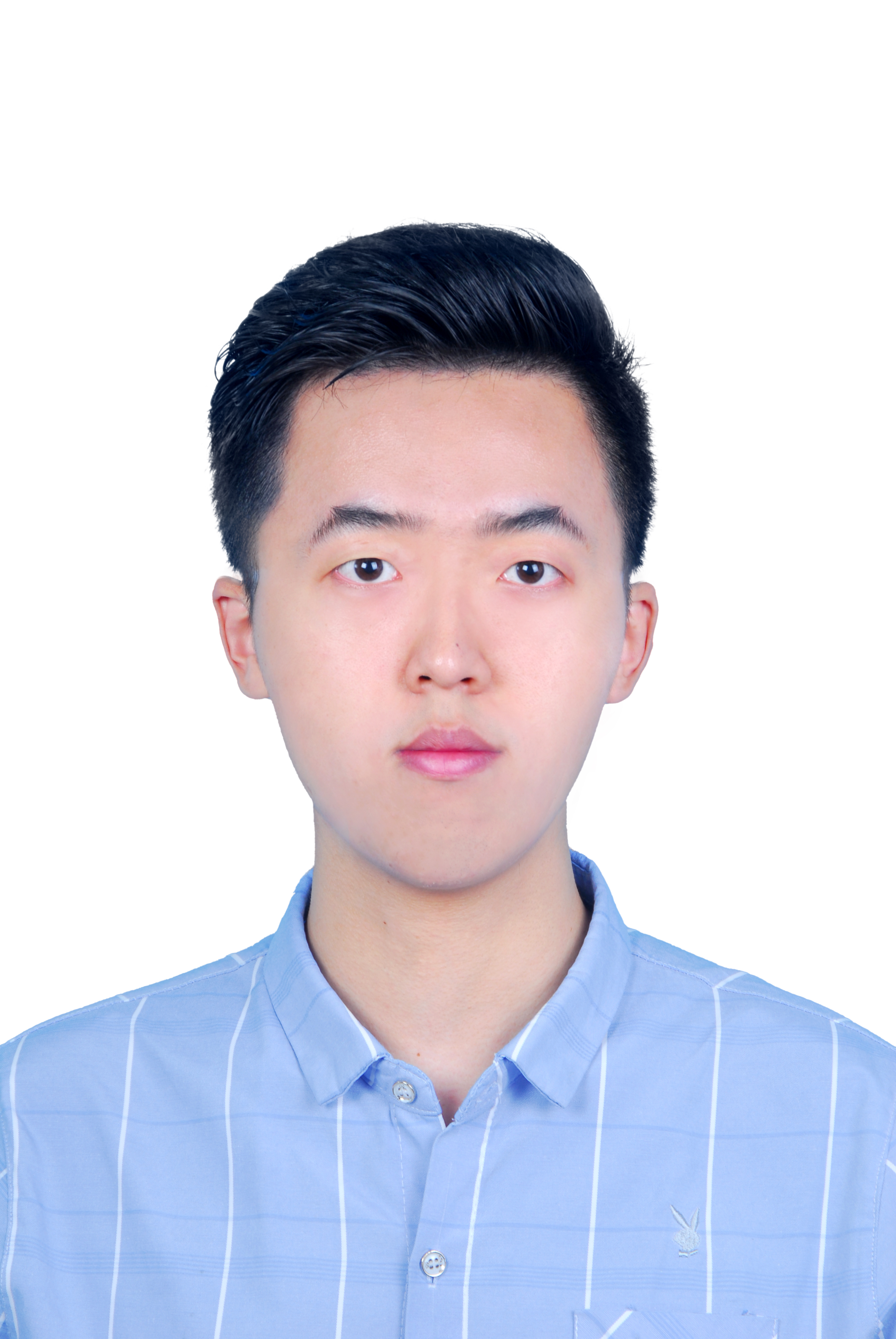}}]{Bin Chen} received the B.S. degree in the School of Computer Science, Beijing University of Posts and Telecommunications, Beijing, China, in 2021. He is currently working toward the Ph.D. degree in computer applications technology at the School of Electronic and Computer Engineering, Peking University, Shenzhen, China. His research interests include image compressive sensing and super-resolution. \end{IEEEbiography}

\vspace{-1.4cm}
\begin{IEEEbiography}[{\includegraphics[width=1in,height=1.25in,clip,keepaspectratio]{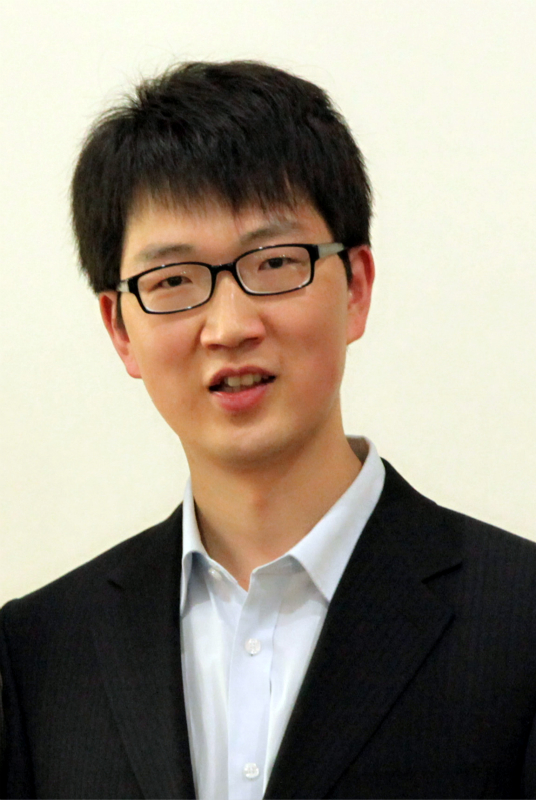}}]{Jian Zhang} (M'14) received Ph.D. degree from the
School of Computer Science and Technology, Harbin Institute of Technology (HIT), Harbin, China, in 2014. He is currently an Assistant Professor and heads the Visual-Information Intelligent Learning LAB (VILLA) at the School of Electronic and Computer Engineering, Peking University (PKU), Shenzhen, China. His research interest focuses on intelligent controllable image generation, encompassing three pivotal areas: efficient image reconstruction, controllable image generation, and precise image editing. He has published over 100 technical articles in refereed international journals and proceedings and has received over 9700 citations. He received several Best Paper Awards at international journals/conferences. He serves as an Associate Editor for the Journal of Visual Communication and Image Representation.
\end{IEEEbiography}

\end{document}